%% file: neurips_data_2024.tex
\documentclass{article}
\PassOptionsToPackage{sort&compress,numbers}{natbib}
\usepackage[preprint]{neurips_data_2024}

\usepackage[utf8]{inputenc} 
\usepackage[T1]{fontenc}    
\usepackage{hyperref}       
\usepackage{url}            
\usepackage{booktabs}       
\usepackage{amsfonts}       
\usepackage{nicefrac}       
\usepackage{microtype}      
\usepackage{xcolor}         
\usepackage{amsmath}
\usepackage{nicefrac}
\usepackage{xcolor}
\usepackage{bm}
\usepackage{relsize}
\usepackage{fix-cm}
\usepackage{longtable}

\usepackage{hyperref}
\usepackage{url}
\usepackage{xspace}
\usepackage{array}
\usepackage{graphicx}
\usepackage{booktabs}
\usepackage{multirow,array}
\usepackage{multicol}
\usepackage{subcaption}
\hypersetup{
	colorlinks=true,%
	citecolor={blue!50!black},
	linkcolor={red!50!black},
	urlcolor={green!50!black}
}
\usepackage{makecell}
\usepackage{csquotes}
\usepackage{tcolorbox}
\usepackage{wrapfig}
\usepackage[font=small]{caption}
\usepackage[T1]{fontenc}
\usepackage{enumitem}
\usepackage{listings}
\usepackage{tabularx}
\usepackage{caption}
\usepackage{pifont}
\usepackage{textcomp}
\usepackage{fontawesome}

\usepackage[linesnumbered,ruled,vlined]{algorithm2e}
\usepackage{adjustbox}

\usepackage{appendix}

\usepackage{tocloft}
\usepackage{minitoc}

\newcommand{\appendixtableofcontents}{
    \begingroup
    \let\clearpage\relax
    \tableofcontents
    \endgroup
}

\definecolor{codegreen}{rgb}{0,0.6,0}
\definecolor{codegray}{rgb}{0.5,0.5,0.5}
\definecolor{codepink}{RGB}{252, 142, 172}
\definecolor{codepurple}{rgb}{0.58,0,0.82}
\definecolor{backcolour}{RGB}{245,245,245}
\lstdefinestyle{mystyle}{
    backgroundcolor=\color{backcolour},   
    commentstyle=\color{magenta},
    keywordstyle=\color{blue},
    numberstyle=\tiny\color{codegray},
    stringstyle=\color{codepurple},
    basicstyle=\fontfamily{\ttdefault}\footnotesize,
    breakatwhitespace=false,        
    breaklines=true,                
    keepspaces=true,    
    frame=single,
    numbersep=5pt,                  
    showspaces=false,              
    showstringspaces=false,
    showtabs=false,               
    tabsize=2,
    classoffset=1, 
    keywordstyle=\color{violet},
    classoffset=0,
}
\newcommand{\projname}{\textbf{\textsc{Cradle}}\xspace}

\newcommand{\ie}{\emph{i.e., }}
\newcommand{\eg}{\emph{e.g., }}

\newcommand{\nonumfootnote}[1]{%
  \begingroup
  \renewcommand{\thefootnote}{}
  \footnotetext{#1}
  \endgroup
}

\DeclareRobustCommand{\Colon}{{%
  \ooalign{%
    \hidewidth\raisebox{0.2ex}{/}\kern0.1em\hidewidth\cr
    C\cr
    \hidewidth\kern0.1em\raisebox{0.2ex}{/}\hidewidth\cr
  }%
}}

\usepackage{tikz}
\usetikzlibrary{shadows}
\usepackage{calc}

\title{\projname: Empowering Foundation Agents Towards General Computer Control}

\author{
    Weihao Tan\textsuperscript{\rm 3}\footnotemark[1]\textbf{,} 
    Wentao Zhang\textsuperscript{\rm 3}\footnotemark[1]\textbf{,} 
    Xinrun Xu\textsuperscript{\rm 5}\footnotemark[1]\textbf{,} 
    Haochong Xia\textsuperscript{\rm 3}\footnotemark[2]\textbf{,} 
    Ziluo Ding\textsuperscript{\rm 2}\footnotemark[2]\textbf{,}  \\
    \textbf{Boyu Li}\textsuperscript{\rm 2}\footnotemark[2]\textbf{,}  
    \textbf{Bohan Zhou}\textsuperscript{\rm 4}\footnotemark[2]\textbf{,}  
    \textbf{Junpeng Yue}\textsuperscript{\rm 4}\footnotemark[2]\textbf{,} 
    \textbf{Jiechuan Jiang}\textsuperscript{\rm 4}\footnotemark[2]\textbf{,}  
    \textbf{Yewen Li}\textsuperscript{\rm 3}\footnotemark[2]\textbf{,}  
    \textbf{Ruyi An}\textsuperscript{\rm 3}\footnotemark[2]\textbf{,} \\
    \textbf{Molei Qin}\textsuperscript{\rm 3}\footnotemark[2]\textbf{,} 
    \textbf{Chuqiao Zong}\textsuperscript{\rm 3}\footnotemark[2]\textbf{,} 
    \textbf{Longtao Zheng}\textsuperscript{\rm 3}\footnotemark[2]\textbf{,} 
    \textbf{Yujie Wu}\textsuperscript{\rm 1}\footnotemark[2]\textbf{,}  
    \textbf{Xiaoqiang Chai}\textsuperscript{\rm 1}\footnotemark[2]\textbf{,} \\
    \textbf{Yifei Bi}\textsuperscript{\rm 2}\textbf{,} 
    \textbf{Tianbao Xie}\textsuperscript{\rm 6}\textbf{,}  
    \textbf{Pengjie Gu}\textsuperscript{\rm 3}\textbf{,}
    \textbf{Xiyun Li}\textsuperscript{\rm 2}\textbf{,} 
    \textbf{Ceyao Zhang}\textsuperscript{\rm 7}\textbf{,} \\
    \textbf{Long Tian}\textsuperscript{\rm 1}\textbf{,} 
    \textbf{Chaojie Wang}\textsuperscript{\rm 1}\textbf{,} 
    \textbf{Xinrun Wang}\textsuperscript{\rm 3}\footnotemark[3]\textbf{,} 
    \textbf{Börje F. Karlsson}\textsuperscript{\rm 2}\footnotemark[3]\footnotemark[2]\textbf{,} \\
    \textbf{Bo An}\textsuperscript{\rm 3, 1}\footnotemark[4]~\textbf{,} 
    \textbf{Shuicheng Yan}\textsuperscript{\rm 1}\footnotemark[4]~\textbf{,} 
    \textbf{Zongqing Lu}\textsuperscript{\rm 4, 2}\footnotemark[4]   \\
    \textsuperscript{\rm 1} Skywork AI \textsuperscript{\rm 2} Beijing Academy of Artificial Intelligence \\
    \textsuperscript{\rm 3} Nanyang Technological University, Singapore 
   \textsuperscript{\rm 4} Peking University \\
   \textsuperscript{\rm 5} Institute of Software, Chinese Academy of Sciences \\
   \textsuperscript{\rm 6} The University of Hong Kong
   \textsuperscript{\rm 7} The Chinese University of Hong Kong, Shenzhen\\    \texttt{weihao001@ntu.edu.sg}\quad\texttt{boan@ntu.edu.sg}\quad\texttt{zongqing.lu@pku.edu.cn} \\ 
    Project website: \url{https://baai-agents.github.io/Cradle/} \\
}

\begin{document}
\doparttoc 
\faketableofcontents 
\maketitle

\nonumfootnote{$^*$Equal contribution \quad $^{\dagger}$Core contribution \quad $^{\ddagger}$Equal advising \quad $^{\S}$Corresponding authors}

\nonumfootnote{Weihao Tan's work was conducted during his internships at Skywork AI and BAAI. Longtao Zheng is also an intern at Skywork AI. Xinrun Xu, Bohan Zhou, and Junpeng Yue are interns at BAAI.}

\begin{figure}[ht]
\vspace{-0.75cm}
    \centering
    \includegraphics[width=.95\linewidth]{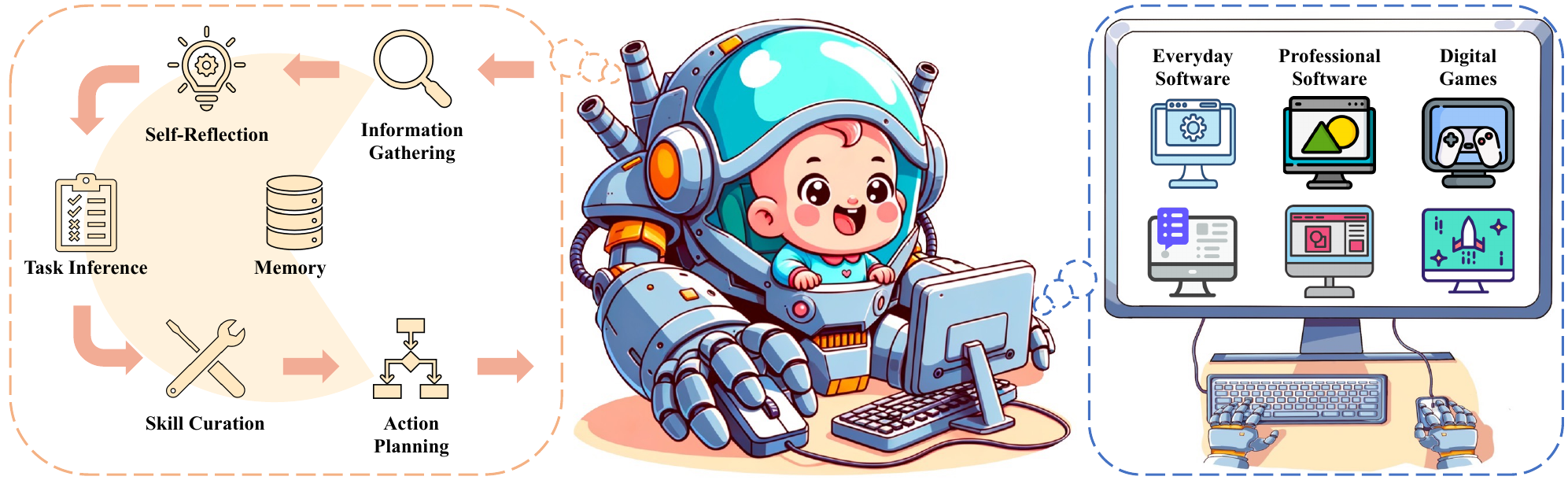}
    \caption{The \projname framework empowers nascent foundation models to perform complex computer tasks via the same unified interface humans use, \ie screenshots as input and keyboard \& mouse operations as output.
}
    \label{fig:main-figure}
    \vspace{-0.3cm}
\end{figure}

\setcounter{footnote}{0}

\begin{abstract}
\label{sec:abstract}
Despite the success in specific scenarios, existing foundation agents still struggle to generalize across various virtual scenarios, mainly due to the dramatically different encapsulations of environments with manually designed observation and action spaces.
To handle this issue, we propose the \textbf{General Computer Control} (GCC) setting to restrict foundation agents to interact with software through the most unified and standardized interface, \ie using screenshots as input and keyboard and mouse actions as output. 
We introduce \projname, a modular and flexible LMM-powered framework, as a preliminary attempt towards GCC. Enhanced by six key modules: Information Gathering, Self-Reflection, Task Inference, Skill Curation, Action Planning, and Memory, \projname is able to understand input screenshots and output executable code for low-level keyboard and mouse control after high-level planning, so that \projname can interact with any software and complete long-horizon complex tasks without relying on any built-in APIs.
Experimental results show that \projname exhibits remarkable generalizability and impressive performance across four previously unexplored commercial video games, five software applications, and a comprehensive benchmark, OSWorld. To our best knowledge, \projname is the first to enable foundation agents to follow the main storyline and complete 40-minute-long real missions in the complex AAA game Red Dead Redemption 2 (RDR2). 
\projname can also create a city of a thousand people in Cities:~Skylines, farm and harvest parsnips in Stardew Valley, and trade and bargain with a maximal weekly total profit of 87\% in Dealer's Life 2. \projname can not only operate daily software, like Chrome, Outlook, and Feishu, but also edit images and videos using Meitu and CapCut. 
With a unified interface to interact with any software, \projname greatly extends the reach of foundation agents by enabling the easy conversion of any software, especially complex games, into benchmarks to evaluate agents' various abilities and facilitate further data collection, thus paving the way for generalist agents.

\end{abstract}

\section{Introduction}
\vspace{-5pt}
\label{sec:introduction}
Artificial General Intelligence (AGI) has long been a north-star goal for the AI community~\citep{morris2023levels}. The recent success of foundation agents, \ie agents empowered by large multimodal models (LMMs) and advanced tools, in various environments, \eg  web browsing~\citep{zhou2023webarena, deng2023mind2web, gur2023real, zheng2023synapse, zheng2024gpt, he2024webvoyager}, operating mobile applications~\citep{yang2023appagent, wang2024mobile} and desktop software~\citep{zhang2024ufo, wu2024copilot}, crafting and exploration in Minecraft~\citep{wang2023describe, wang2023voyager, wang2023jarvis}, and some robotics scenarios~\citep{huang2022inner, brohan2023can, driess2023palm, brohan2023rt}, have shown promise. However, current foundation agents still struggle to generalize across different scenarios, primarily due to the dramatic differences in the encapsulation of environments with human-designed observation and action space. Therefore, developing foundation agents applicable to various environments remains extremely challenging. 

Computers, as the most important and universal interface that connects humans and the increasingly digital world, provide countless rich software, including applications and realistic video games for agents to interact with, while avoiding the challenges of robots in reality, such as hardware requirements, constraints of practicability, and possible catastrophic failures~\citep{raad2024scaling}. Mastering these virtual environments is a promising path for foundation agents to achieve generalizability. Therefore, we propose the \textbf{General Computer Control} (GCC) setting:
\vspace{-2pt}
\begin{tcolorbox}[left=0.1cm, right=0.1cm, bottom=0cm, top=0cm, colback=gray!5, colframe=gray!30, boxrule=1.5pt]
\emph{Building \textbf{foundation agents} that can master \textbf{ANY} computer task via the universal human-style interface by receiving input from screens and audio and outputting keyboard and mouse actions.}
\end{tcolorbox}
\vspace{-2pt}

There are many challenges to achieving GCC: i) good alignment across multi-modalities for better understanding and decision-making;
ii) precise control of keyboard and mouse to interact with the computer, which has a large, hybrid action space, including not only which key to press and where the mouse to move, but also the duration of the press and the speed of the mouse movement; 
iii) long-horizontal reasoning due to the partial observability of complex GCC tasks, which also leads to the demand for long-term memory to maintain past useful experiences; and 
iv) efficient exploration in a structured manner to discover better strategies and solutions autonomously, \ie self-improving, which can allow agents to generalize across the myriad tasks in the digital world. 

As shown in Figure~\ref{fig:main-figure}, we introduce {\projname}, a novel modular LMM-powered framework that empowers foundation agents towards GCC. \projname consists of six key modules: 1) information gathering, to extract the relevant information from multimodal observations; 2) self-reflection, to rethink past experiences about whether the actions and tasks are successfully completed and reasons for possible failures; 3) task inference, to determine whether to continue current tasks or propose a new task given the current situation; 4) skill curation, for generating, updating, and retrieving useful skills for the current task; 5) action planning, to generate specific executable operations for keyboard and mouse control via skills; and 6) memory, for storage, summary, and retrieval of past experiences. 

\begin{figure}[t]
    \vspace{-15pt}
    \centering
    \includegraphics[width=\textwidth]{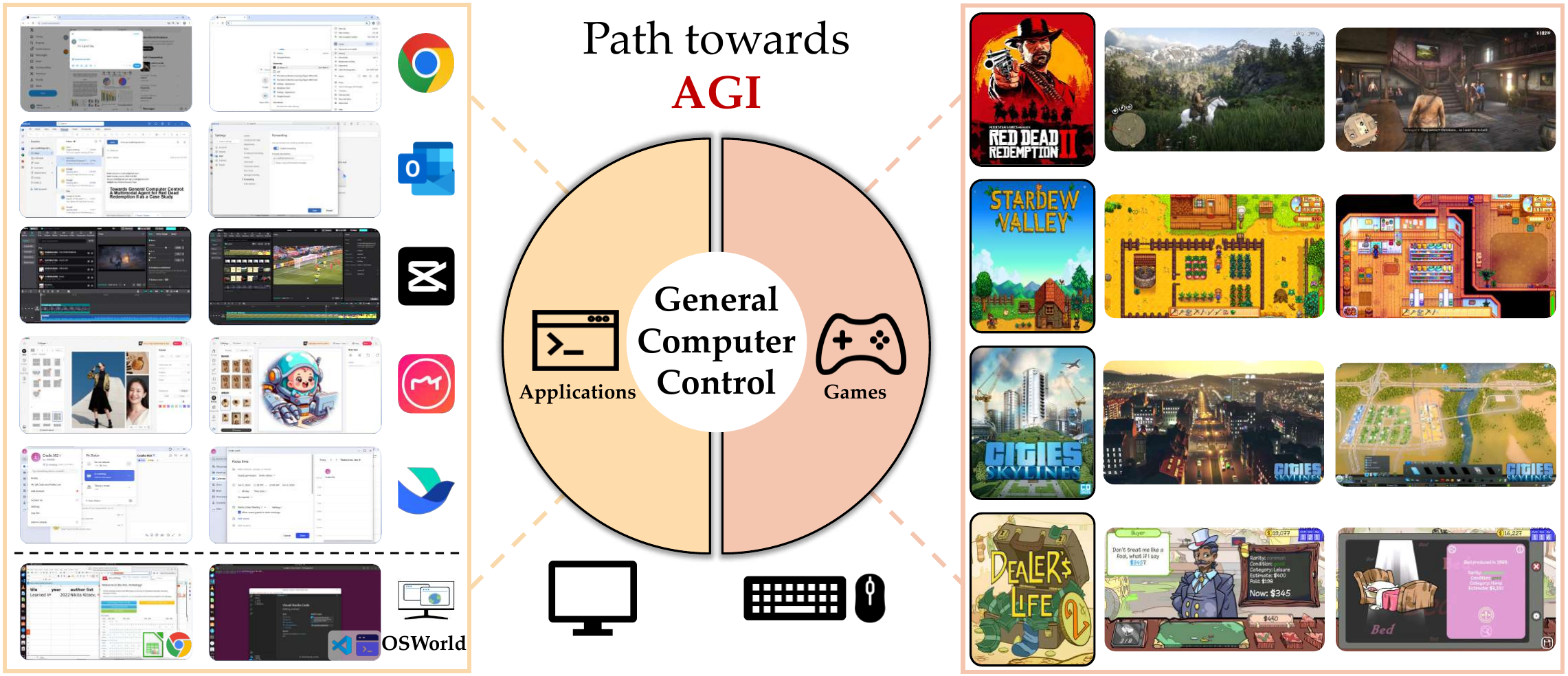}
    \caption{Taxonomy of GCC and the games and software investigated in this work.}
    \label{fig:agi_gcc_games}
    \vspace{-20pt}
\end{figure}

\begin{figure}[b]
    \vspace{-15pt}
\centering
\begin{subfigure}{0.47\textwidth}
\centering
\includegraphics[width=1\textwidth]{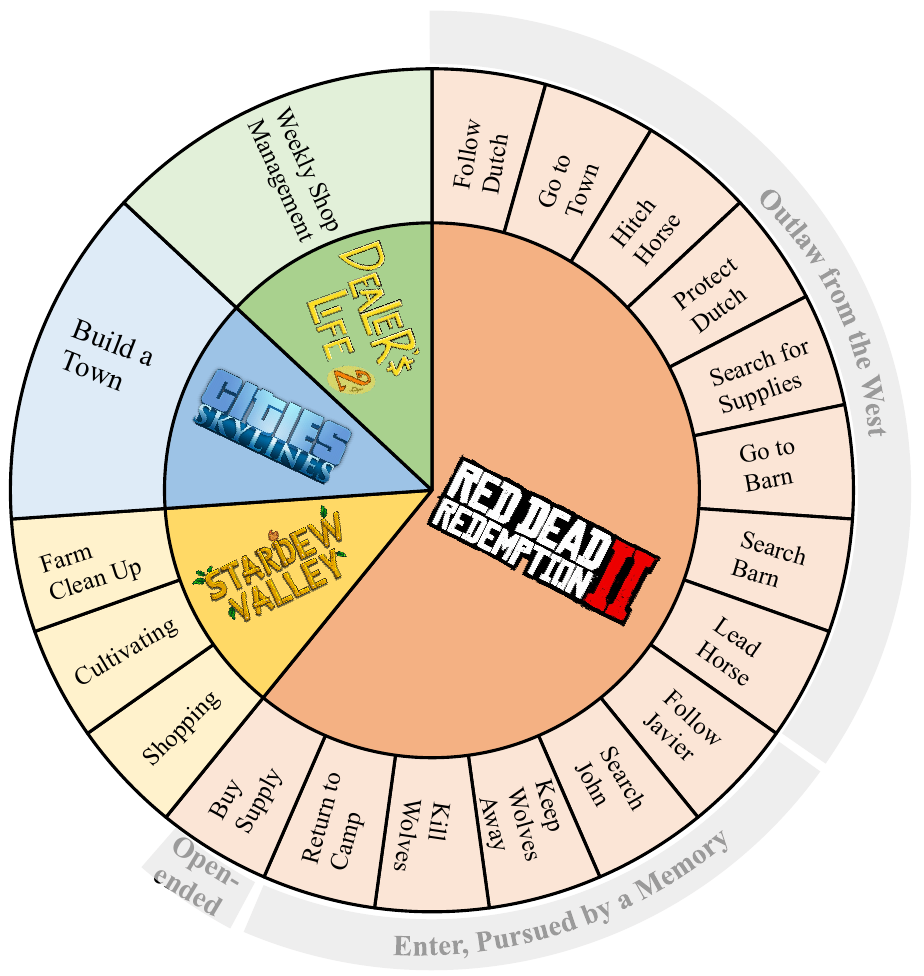}
\end{subfigure}
\hfill
\begin{subfigure}{0.485\textwidth}
\centering
\includegraphics[width=1\textwidth]{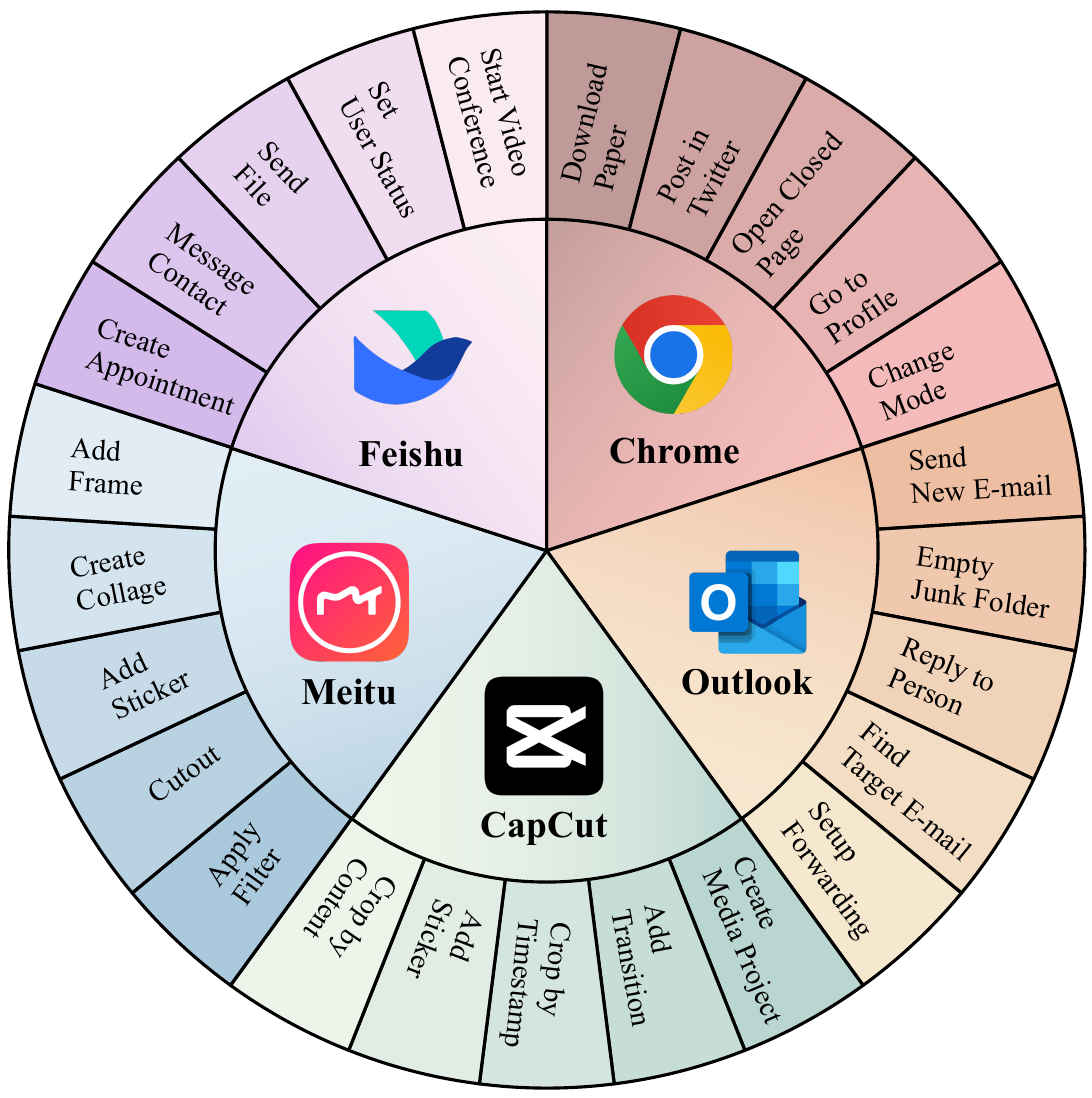}
\end{subfigure}
\caption{Overview of all game tasks (left) in RDR2, Stardew Valley, Cities: Skylines, and Dealer's Life 2 and application tasks (right) in Chrome, Outlook, CapCut, Meitu, and Feishu.}
\label{fig:tasks}
    \vspace{-15pt}
\end{figure}

As illustrated in Figure~\ref{fig:agi_gcc_games}, tasks in GCC can be broadly divided into two categories: video game playing and software application manipulation. 
video games offer the most challenging tasks in GCC due to several key factors. First, the complexity of game environments requires sophisticated problem-solving and adaptive strategies. Second, long-term reasoning is essential to navigate and succeed in these intricate virtual worlds. Third, understanding and mastering new, complex mechanics within games demand rapid learning and cognitive flexibility. Finally, video games test a player's ability to react quickly and perform precise control and operations, which together create a unique and demanding computational challenge. In addition to the typical embodied control, classical UI manipulation, like menu use, is also common during gameplay, which is similar to the other software applications~\citep{raad2024scaling}. Therefore, games provide rich comprehensive and challenging testbeds to evaluate and improve agents' various abilities. 
In this work, we conduct extensive experiments to demonstrate the generalizability of \projname in such complex environments, while also mastering diverse everyday software applications in distinct domains. We managed to prove that commercial software is out-of-box testbeds under our framework.
The four selected representative games are: 
\begin{itemize}[leftmargin=*, topsep=0pt, partopsep=0pt, parsep=0pt, itemsep=0pt]
\item \textbf{Red Dead Redemption 2} 
(RDR2), an epic AAA 3D role-playing game (RPG) with rich storylines, realistic scenes, and an immersive open-ended world; where players can complete missions by following the instructions, freely explore the world, interact with non-player characters (NPCs) and engage in a variety of activities such as hunting and fishing, in a first- or third-person perspective. This game offers great challenges in 3D embodied navigation and interaction.
\item \textbf{Stardew Valley}, a 2D pixel-art farming simulation game where players can restore and expand a farm through carefully planned activities such as planting crops, mining, fishing, and crafting. Players can build relationships with the villagers, participate in seasonal events, and uncover the mysteries of the valley. The game encourages strategic planning and time management, as each day brings new opportunities and challenges. Players have to balance their energy and resources to maximize their farm's productivity and profitability.
\item \textbf{Dealer's Life 2}, a simulation game where players manage a pawn shop. They must assess the value of items, haggle with customers, and make strategic decisions to grow their business. The game offers a dynamic market influenced by trends, customer preferences, and random events, requiring players to adapt and refine their negotiation tactics.
\item \textbf{Cities:~Skylines}, a 3D, top-down view, city-building game where players take on the role of a city mayor, tasked with the development and management of a thriving metropolis, engaging in urban planning by controlling zoning, road placement, taxation, public services, and public transportation in an area. They must balance the needs and desires of the population with the city's budget, addressing issues such as traffic congestion, pollution, and citizen satisfaction. The game provides a sandbox environment where creativity and strategic thinking are key to building efficient and aesthetically pleasing urban landscapes. It also requires highly precise mouse control.

\end{itemize}
The target set of diverse software applications for evaluation includes: \textbf{Chrome}, \textbf{Outlook}, \textbf{CapCut}, \textbf{Meitu}, and \textbf{Feishu}, as well as one comprehensive software benchmark, \textbf{OSWorld}~\citep{xie2024osworld}.

As shown in Figure~\ref{fig:tasks}, for each game and software application, representative tasks are designed to measure the various abilities of the agent comprehensively. Experimental results show that \projname exhibits remarkable generalization ability and impressive performance across the four previously unexplored commercial video games, the five target software applications, and the comprehensive contemporaneous OSWorld benchmark. To our best knowledge, \projname is the first to enable LMM-based agents to follow the main storyline and complete 40-minute-long real missions in a complex AAA game, RDR2. 
\projname also manages to create a city of a thousand people in Cities:~Skylines, farm and harvest parsnips in Stardew Valley, trade and bargain with a maximal weekly total profit of 87\% in Dealer's Life 2. Besides, \projname can not only operate daily software, like Chrome and Outlook, but also edit images and videos using Meitu and CapCut, and perform office tasks in Feishu. 
Able to interact with software in a unified manner, \projname greatly extends the reach of AI agents by making it easy to convert any software, especially complex games, into benchmarks to evaluate agents' various abilities and facilitate further data collection, paving the way for generalism. We hope the open-source \projname framework and its holistic evaluation protocol, \ie tasks and metrics, on various environments, can accelerate the development of more powerful foundation agents, thereby advancing the path towards AGI.

\vspace{-10pt}
\section{Related Work}
\vspace{-5pt}

\subsection{Environments and Benchmarks for Computer Control}
\textbf{Environments and Benchmarks on Software Applications}. Simulated environments on computers have been popular benchmarks and testbeds for the research community. Earlier computer control environments primarily focused on web navigation tasks~\citep{shi2017world, liu2018reinforcement, yao2022webshop, deng2023mind2web, zhou2023webarena, koh2024visualwebarena}. Recent benchmarks start to include various common software~\citep {kapoor2024omniact, xie2024osworld}, aiming to develop a generalist agent in the digital world. However, none of them takes video games into consideration, missing a key component of computer control. 

\textbf{Environments and Benchmarks on Video Games}.
On the other side, many research environments are built on top of video games, significantly advancing the study of decision-making, especially, reinforcement learning (RL). Examples include but are not limited to Atari games~\citep{bellemare2013arcade}, 
Super Mario Bros~\citep{gym-super-mario-bros}, Google Research Football~\citep{kurach2020google},  
Minecraft~\citep{johnson2016malmo, guss2019minerl, fan2022minedojo}, Dota II~\citep{berner2019dota}, StarCraft II~\citep{vinyals2019alphastar, samvelyan2019starcraft, ellis2023smacv2}, Quake III~\citep{jaderberg2019human}, Gran Turismo~\citep{wurman2022outracing},  Diplomacy~\citep{meta2022human} and Civilization~\citep{qi2024civrealm}.
Additionally, many custom-built environments, especially grid world and embodied scenarios, are created from scratch in a game-like manner to facilitate agent development, such as BabyAI~\citep{babyai_iclr19}, Melting Pot~\citep{leibo2021scalable}, Overcooked~\citep{carroll2019utility, wu_wang2021too, xiao2022asynchronous}, 
VRKitchen~\citep{gao2019vrkitchen},
VirtualHome~\citep{puig2018virtualhome}, 
iGibson~\citep{shen2021igibson, li2021igibson},
ProcTHOR~\citep{deitke2022️}, Habitat~\citep{habitat19iccv, szot2021habitat, puig2023habitat}, and Generative agents~\citep{park2023generative}.

Each of these environments highly relies on the accessibility of the open-source code or provided built-in APIs. Significant human efforts are required for implementation and encapsulation, enabling agent interaction. Therefore, despite the abundance of software and games available for human use, only a limited number are accessible to agents, especially for commercial closed-source games and software applications.
Additionally, the lack of consensus on environment standards further complicates the interaction, as each environment has specific observation and action spaces, tailored to its unique requirements. This variation exacerbates the challenge of enabling agents to interact with diverse environments and collect data with a consistent level of fine-grained semantics to improve the agent's capabilities. Few agents can complete tasks across multiple environments so far. 

Similar to OpenAI Universe~\citep{openaiUniverse} and SIMA~\citep{raad2024scaling}, our goal is to explore a unified way that allows agents to interact for measuring and training agents' abilities across a wide range of games, websites, and other applications without heavy human efforts needed. This approach aims to prove that diverse software applications and games can serve as out-of-the-box environments for AI development.

\subsection{LMM-based Agents for Computer Tasks}
\textbf{Agents for Software Manipulation.} Agents for software applications are developed to complete tasks such as web navigation~\citep{zhou2023webarena, deng2023mind2web, mialon2023gaia} and software application control~\citep{rawles2023android, yang2023appagent,  kapoor2024omniact}. While previous LLM-based web agents~\citep{deng2023mind2web, zhou2023webarena, gur2023real, zheng2023synapse} show some promising results in effectively interacting with content on webpages, they usually use raw HTML code and DOM tree as input and interact with the available element IDs, ignoring the rich visual patterns with key information, like icons, images, and spatial relations. Recently, multimodal web agents~\citep{yan2023gpt, gao2023assistgui,he2024webvoyager, zheng2024gpt, niu2024screenagent, zhang2024ufo, wu2024copilot} and mobile app agents~\citep{yang2023appagent,wang2024mobile} have been explored. Though using screenshots as input, they still rely on built-in APIs and advanced tools to get internal information, like available interactive element IDs, to execute corresponding actions, which greatly limits their applicability. Other train-based agents~\citep{hong2023cogagent, furuta2023multimodal, cheng2024seeclick} also suffer from generalizing to unseen software and tasks. Moreover, all of these works primarily focus on static websites and software, which greatly reduces the need for timeliness and simplifies the setting by ignoring the dynamics between adjacent screenshots, \ie animations, and incomplete action space without considering the duration of the key press and different mouse mode. It results in the failure of deployment to the tasks with rapid graphics changes, \eg game playing.

\textbf{Agents for Game Playing.} Several attempts try to develop foundation agents for complex video games, such as Minecraft~\citep{wang2023describe, wang2023jarvis, wang2023voyager}, Starcraft II~\citep{ma2023large} and Civilization-like game~\citep{qi2024civrealm} with textual observations obtained from internal APIs and pre-defined semantic actions. Although JARVIS-1~\citep{wang2023jarvis} claims to interact with the environment in a human-like manner with the screenshots as input and mouse and keyboard for control, its action space is predefined as a hybrid space composed of keyboard, mouse, and API. The game-specific observation and action spaces prohibit the generalization of them to other novel games. Pre-trained with videos with action labels, VPT~\citep{baker2022video} manages to output mouse and keyboard control with raw screenshots as input without any additional information. However, collecting videos with action labels is time-consuming and costly, which is difficult to generalize to multiple environments. Another concurrent work, SIMA~\citep{raad2024scaling} trained embodied agents to complete 10-second-long tasks over ten 3D video games. Though their results are promising to scale up, they focus on behavior cloning with gameplay data from human experts, resulting in a high expense. 

In both targeting complex video games and diverse software applications, \projname attempts to explore a new way to efficiently interact with different complex environments in a unified manner and facilitate further data collection. In a nutshell, to our best knowledge, there are currently no agents under the GCC setting, reported to show superior performance and generalization in complex video games and across computer tasks. In this work, we make a preliminary attempt to explore and benchmark diverse environments in this setting, applying our framework to diverse challenging environments under GCC and proposing an approach where any software can be used to benchmark agentic capabilities in it.

\section{The \projname Framework}
\begin{figure}[t]
\centering
\vspace{-15pt}
\includegraphics[width=\linewidth]{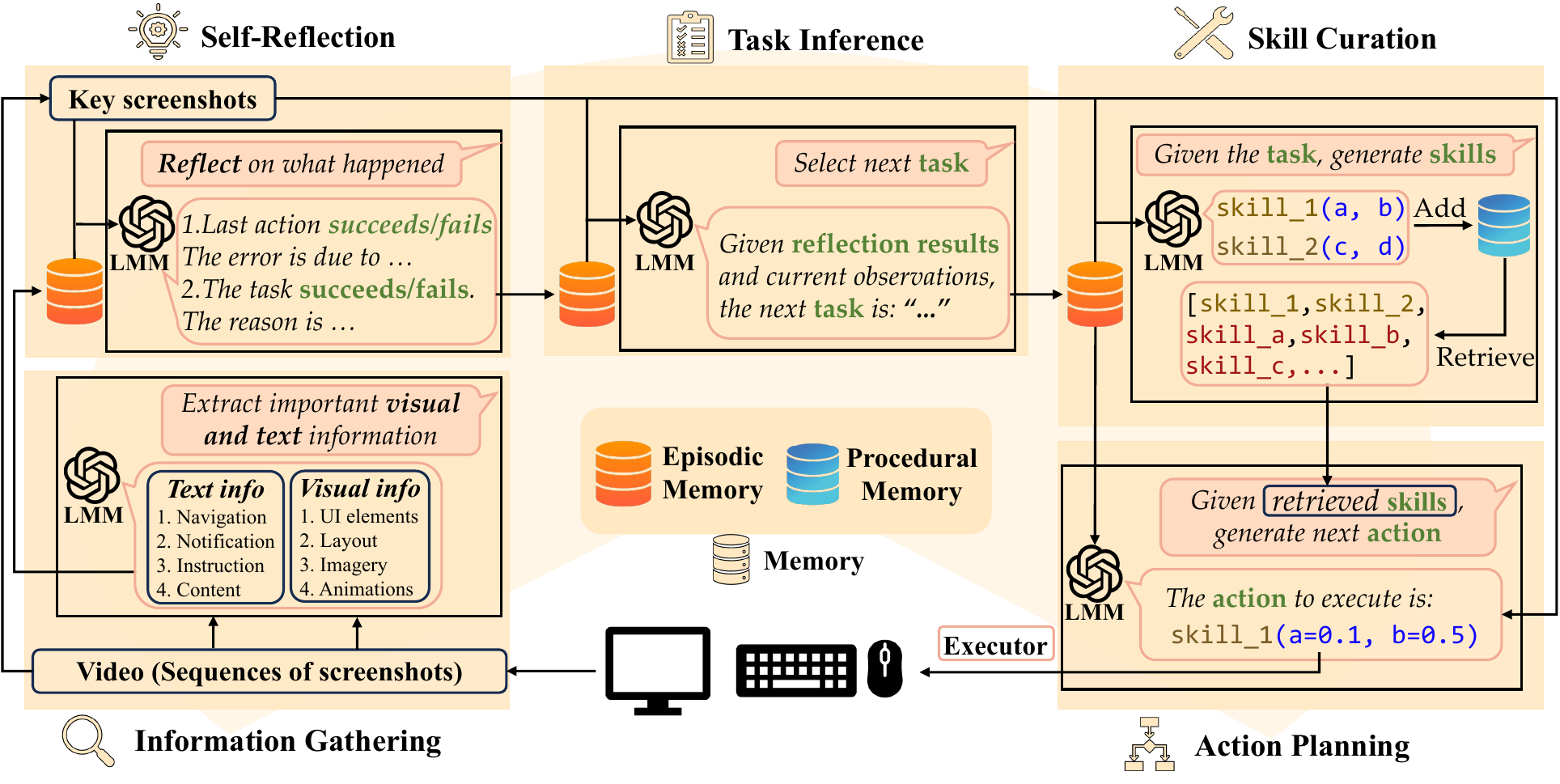}
\caption{An overview of the \projname framework. \projname takes video from the computer screen as input and outputs computer keyboard and mouse control determined through inner reasoning.}
\label{fig:system}
\vspace{-15pt}
\end{figure}

\vspace{-5pt}
To pursue GCC, we propose \projname, illustrated in Figure~\ref{fig:system}, a modular and flexible LMM-powered framework that can properly handle the challenges GCC presents.
The framework should have the ability to understand and interpret computer screens and dynamic changes between consecutive frames from arbitrary software and be able to generate reasonable computer control actions to be executed precisely. This suggests that a multimodal model with powerful vision and reasoning capabilities, in addition to rich knowledge of computer UI and control, is a requirement. In this work, we leverage GPT-4o~\citep{openaigpt4o} as the framework's backbone model.

\begin{figure}[b]
\centering
\vspace{-15pt}
\includegraphics[width=\linewidth]{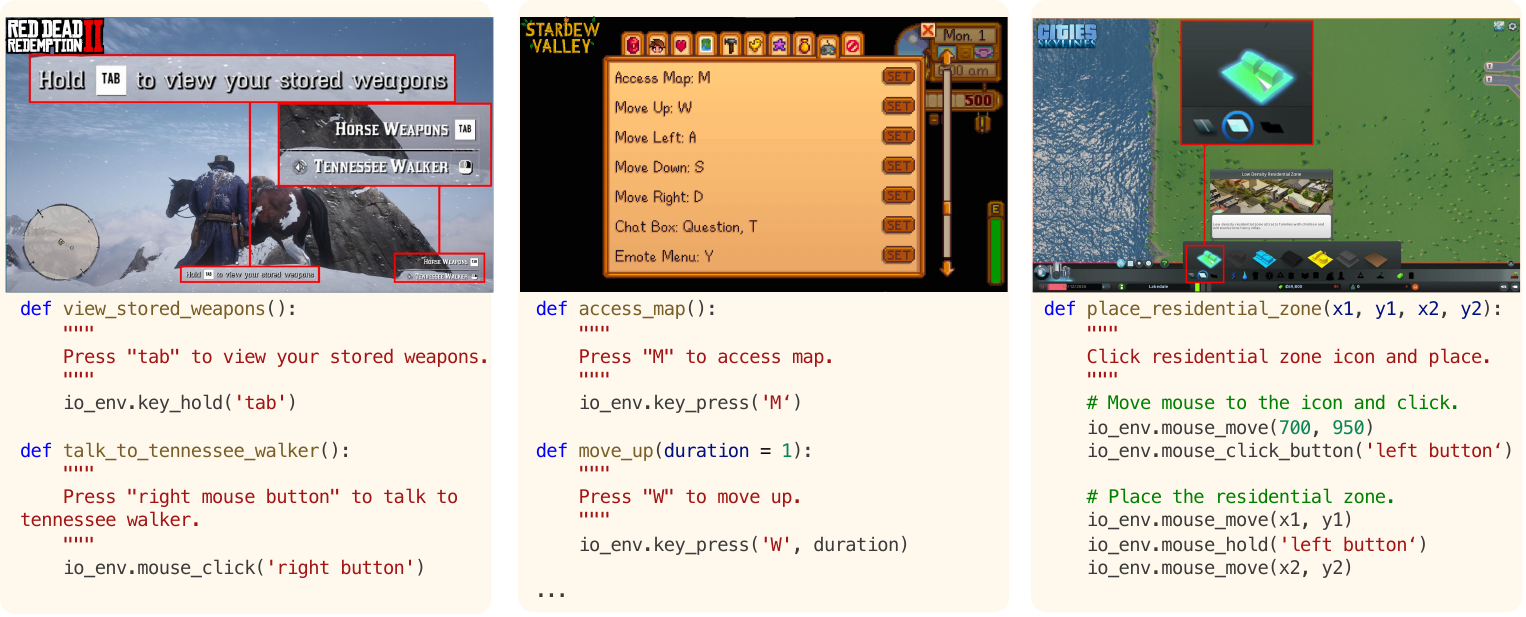}    
\caption{Examples for skill generation according to in-game guidance in RDR2 (left), in-game manual in Stardew Valley (middle), self-exploration in Cities: Skylines (right). Code and comments are shown in brevity.
}
\label{fig:skill_gen}
\vspace{-18pt}
\end{figure}

\vspace{-10pt}
\subsection{Environment IO}
\vspace{-5pt}

\textbf{Observation and Action Space.} \projname only takes a video clip, recording the execution of the last action, as input and outputs keyboard and mouse operations to interact with environments. The observation space is made up of complete screen videos with different lengths. For the action space, it includes all possible keyboard and mouse operations, including \texttt{key\_press}\footnote{Keys include both keyboard keys and mouse buttons.}, \texttt{key\_hold}, \texttt{key\_release}, \texttt{mouse\_move}, and \texttt{wheel\_scroll}. These operations can be combined in various ways to form combos and shortcuts, execute rapid key sequences, or coordinate timings. We choose to use Python code to simulate these operations and encapsulate them into an \texttt{io\_env} class. Note that seldom previous work takes \texttt{key\_hold}, \texttt{key\_release} and mouse moving speed into consideration, which are critical in games (\eg RDR2, for opening weapon wheel and changing view). The asynchronization brought by these temporal-extended actions introduces additional challenges for the control.

\textbf{Information Gathering.} 
Provided with a video clip as input, it is critical for \projname to capture and extract all useful visual and textual information to understand the recent situation and perform further reasoning.
Visual information includes layout, imagery, animations, and UI elements 
which pose high spatial perception and visual understanding requirements for LMM models.
Moreover, we depend on their OCR capabilities to extract textual information in images, which usually includes content (headings and paragraphs), navigation labels (menus and links), notifications, and instructions to convey messages and guide users. Moreover, for each environment, we enhance LMMs' abilities with different tools such as template matching~\citep{brunelli2009template}, Grounding DINO~\citep{liu2023grounding}, and SAM~\citep{kirillov2023segment} to provide additional grounding for object detection and localization.

\textbf{Skill and Action Generation} As shown in Figure~\ref{fig:skill_gen}, to bridge the gap between semantic actions generated by LMMs and OS-level executable actions, \projname uses LMMs to generate code functions as semantic-level skills, which encapsulate lower-level keyboard and mouse control. Similarly to how humans improve while playing, these skills can be developed from scratch according to in-game tutorials and guidance, game manuals and settings, or through self-exploration as the game progresses. These skills can also be pre-defined or composited to solve more complex tasks. An action usually consists of a single or multiple skills instantiated with any necessary parametric aspects, such as duration, position, and speed.

\textbf{Action Execution.} After \projname generates actions and decides to execute them in the environment, an \textbf{\textit{Executor}} is then triggered to map these semantic actions to the OS-level keyboard and mouse commands to interact with the environment.

\begin{figure}[t]
\centering
\vspace{-15pt}
\includegraphics[width=\linewidth]{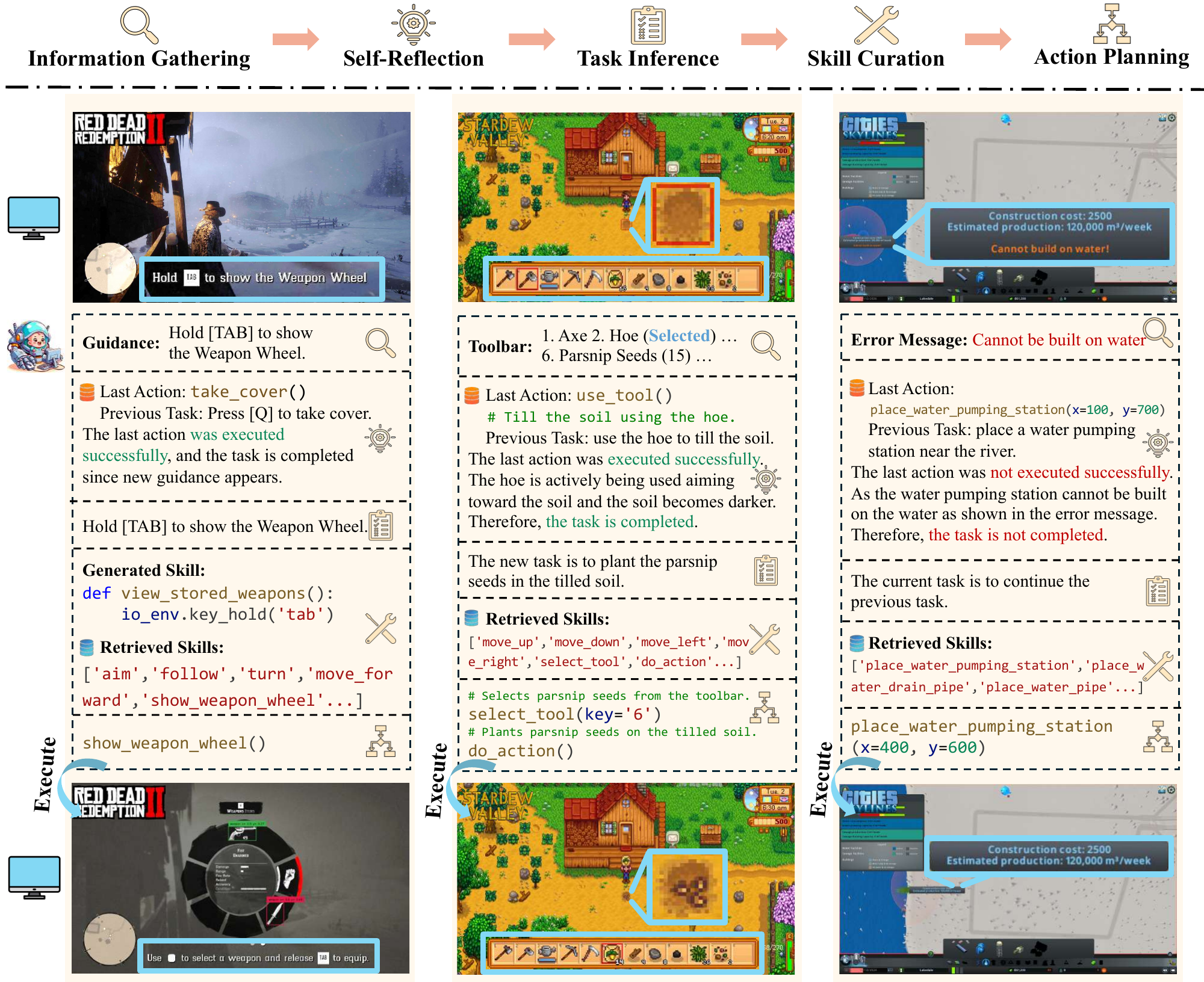}
\caption{Illustrative examples of \projname's complete workflow in RDR2 (left), Stardew Valley (middle) and Cities: Skylines (right). Prompts are shown partially for brevity.}
\label{fig:cradle_examples}
\vspace{-18pt}
\end{figure}

\vspace{-5pt}
\subsection{Memory}
\projname stores and maintains all the useful information from the environment or outputted by each module through a memory mechanism, consisting of episodic memory and procedural memory. 
 
\textbf{Episodic Memory.} Episodic memory is used to maintain current and past experiences, including key screenshots from each video observation, and everything useful outputted by LMMs and advanced tools, \eg textual and visual information, actions, tasks, and reasoning from each module. To facilitate retrieval and storage, periodical summarization is conducted to abstract recently added multimodal information into long-term summaries. The incorporation of episodic memory enables \projname to effectively retain crucial information over extended periods.

\textbf{Procedural Memory.} This memory is specific to storing and retrieving skills in code form, which can be learned from scratch, as shown in Figure~\ref{fig:skill_gen}, or pre-defined in procedural memory.  Skills can be added, updated, or composed in the procedural memory. The most relevant skills for a given task and situation will be retrieved to support action planning, therefore, as \projname continuously acquires new skills during interactions, it is critical that this memory can effectively calculate skill relevance.

\vspace{-5pt}
\subsection{Reasoning}
\vspace{-5pt}
Based on the extracted information from the video observations and relevant information from its memory, \projname needs to do high-level reasoning and then make the next decision. This process is analogous to ``\textbf{reflect on the past, summarize the present, and plan for the future}'', which is broken down into the following modules.

\textbf{Self-Reflection.} The reflection module initially evaluates whether the last executed action was successfully carried out and whether the task was completed. Sequential key screenshots from the last video observation, along with the previous context for action planning and task inference are fed to the LMM for reasoning. Additionally, we also request the LMM to provide an analysis of any failure. This valuable information enables \projname to try and remedy inappropriate decisions or less-than-ideal actions. Furthermore, reflection can also be leveraged to inform re-planning of the task and bring the agent closer to target task completion, better understand the factors that led to previous successes, or suggest how to update or improve specific skills.

\textbf{Task Inference}. After reflecting on the outcome of the last executed action, \projname needs to analyze the current situation to infer the most suitable task for the current moment. We let LMMs estimate the highest priority task to perform and when to stop an ongoing task and start a new one.

\textbf{Skill Curation.}
As the task is determined, \projname needs to prepare the tactics to accomplish it, by retrieving useful skills from the procedural memory, updating skills, or generating new ones. The new skill will be stored in the procedural memory for future utilization.
\textbf{Action Planning.}
\projname needs to select the appropriate skills from the curated skill set and instantiate these skills into a sequence of executable actions by specifying any necessary parametric aspects (\eg duration, position, and target) according to the current task and history information. The generated action is then fed to the \textit{\textbf{Executor}} for interaction with the environment. 

Through these six modules, the input video is processed in stages: first into reasoning, then into semantic skills and actions, and finally into low-level keyboard and mouse operations. This comprehensive conversion covers all essential interactive data needed for both high-level planning and low-level control, enabling a unified approach to data collection for further self-improvement.

\vspace{-10pt}
\section{Empirical Studies}
\vspace{-5pt}
\label{sec:empirical_studies}
In this section, we report empirical results of applying \projname in various challenging environments representative of GCC setting to demonstrate \projname's capabilities in decision-making, UI understanding, and manipulation\footnote{Result tables present two main indicator formats: $\mu \pm \sigma$, representing mean and standard deviation of five runs; and $(\mathlarger{\mathlarger{\nicefrac{s}{t}}})$ representing $s$ successful runs out of a total of $t$ runs.}. To facilitate reproducibility, we also provide setup and load scripts (\eg game saves and checkpoints) in the code repository to reset state for task execution where appropriate.

\subsection{Experimental Settings}
Here we provide a brief introduction to our experimental settings. More implementation details, environment and task descriptions, and prompts used can be found in Appendices~\ref{app:general_implementation} to~\ref{appx:prompts}.

\textbf{Implementation Details.} 
To lower the frequency of interaction with the backbone model, video observation is recorded at 2 fps, which proves sufficient for information gathering without missing any important information in most situations. If not specifically mentioned, all experiments are conducted in five runs under a maximum step limit, using OpenAI's latest model, \emph{gpt-4o-2024-05-13}~\citep{openaigpt4o}. 
Same as Voyager~\citep{wang2023voyager}, we use OpenAI's \emph{text-embedding-ada-002 model}~\citep{openaiada} to generate embeddings for each skill, stored in the procedural memory and retrieved according to the similarities. 
It is important to note that, due to the dynamism of the RDR2 and Stardew Valley and the LMM inference and communication latency, we must pause those game environments while waiting for backbone model responses. Other environments execute continuously. 
All software and games can be run on regular Windows 10 machines, except for RDR2, which is tested on two machines with an NVIDIA RTX-3060 GPU and RTX-4090 GPU separately. We observe a slight performance gain due to the stability of RTX-4090 GPU during the gameplay.

\textbf{Evaluation Methods.} Unlike conventional research benchmarks, which usually provide grounding signals for evaluation, it is difficult to have a unified and general method to determine whether a task is completed automatically in diverse software, especially in video games. Similarly to SIMA~\citep{raad2024scaling}, we apply human evaluation to all tasks across application software and games. Moreover, to provide more quantitative results and a comparison baseline, we provide results for the OSWorld~\citep{xie2024osworld} benchmark, a contemporaneous benchmark that provides evaluation scripts for at least one solution per task.

\textbf{Task Introduction.}
For \textbf{RDR2}, we mainly focus on evaluating agents on the first two missions of the main storyline in Chapter I, which can be divided into 13 tasks according to the in-game checkpoints, which include but are not limited to NPC following, house exploration, and combat. A novice player typically takes about 40 minutes to complete these missions. Few studies tackle such long-duration tasks and rich semantic environments. It is an ideal scenario to emulate a new human player learning to play the game from scratch according to the rich in-game tutorials and hints. Despite the missions in the main storyline, we also designed an open-ended task, \emph{Buy Supply}, in the open-ended world, Chapter II, where the agent is instructed to go to the General Store in Valentine town from the camp for supplementary supply. For this open-ended task, seldom in-game guidance will appear. The agent needs to analyze and propose feasible solutions to complete the mission. For \textbf{Stardew Valley}, we propose three essential tasks at the stage of the game, \ie \emph{Farm Clearup}: Clear the obstacles on the farm, such as weeds, stones, and trees, as much as possible to prepare for farming; 
2) \emph{Cultivation}: Plant the parsnip seed, water every day and harvest at least one parsnip; 
3) \emph{Shopping}: Go to the general store in the town, which is out of the scope of the current map, to buy more seeds and return home. For \textbf{Dealer's Life}, the agent is tasked with managing a shop for a week, appraising item values and haggling with the customers to secure deals. For \textbf{Cities: Skylines}, the task is to build a reasonable city ending in as much population as possible, with the initial starting funds of \textcolonmonetary{}70,000, and basic road, water and power facilities. Moreover, we define five representative domain-specific tasks for each of the five \textbf{Software Applications} in our diverse target set. 
\vspace{-10pt}
\subsection{Performance across Tasks and Environments}

\begin{figure}[ht]
\centering
\vspace{-10pt}
\centering
\includegraphics[width=\textwidth]{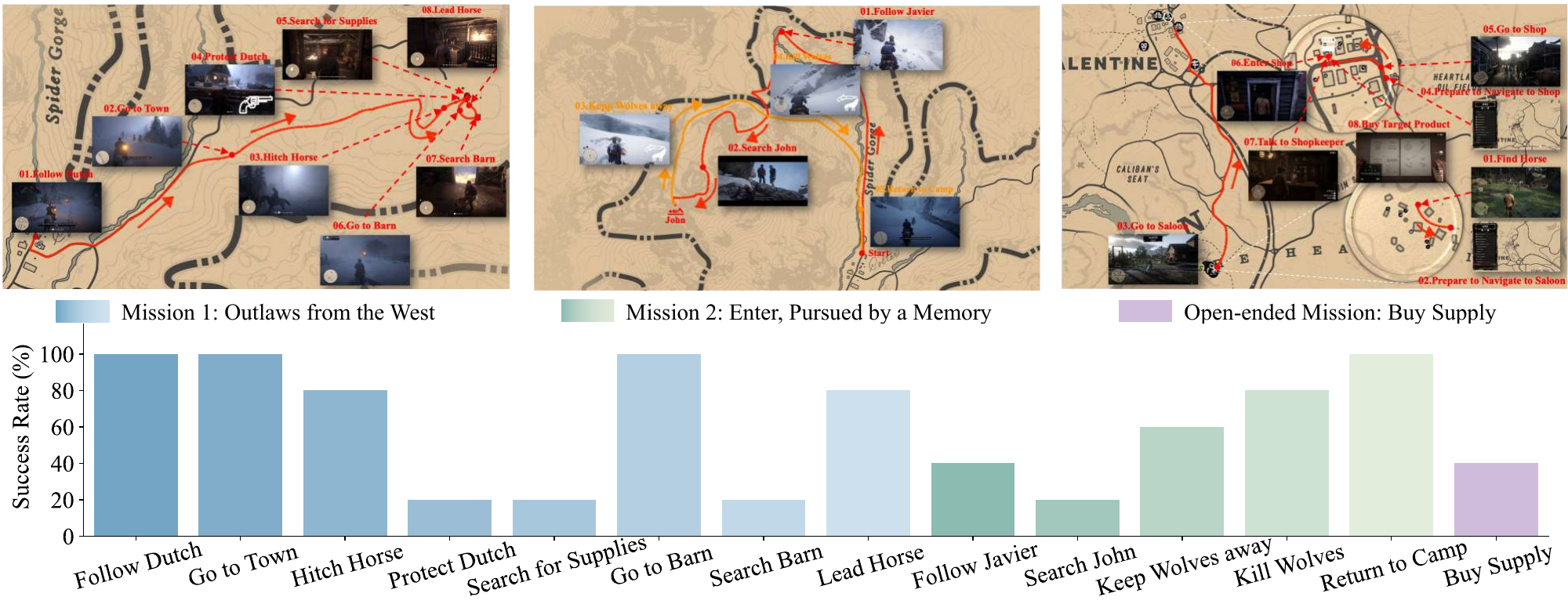}
\caption{Trajectory and success rates of 13 main storyline tasks and 1 open-world task in RDR2. Each task is run with a maximum of 500 steps. Agents can retry the checkpoint if the task fails or the character dies.}
\label{fig:main_mission_success_rates}
\vspace{-10pt}
\end{figure}

\textbf{Red Dead Red Redemption 2.} 
As shown in Figure~\ref{fig:main_mission_success_rates}, \projname can achieve a high success rate in simple tasks like following an NPC or going to specific locations on the ground (\eg \emph{Follow Dutch} and \emph{Go to Barn}). Another following task, \emph{Follow Javier}, and the searching task, \emph{Search John}, are dangerous for the rugged and winding path up to the snow mountain with cliffs. In addition, GPT-4o struggles with real-time combat tasks and searching tasks due to its inability to accurately locate enemies or objects and precisely time decisions. Even equipped with additional detection tools, like Grounding DINO~\citep{liu2023grounding}, the success rate drops significantly to 20\% in the task of \emph{Protect Dutch}, which requires nighttime combat.
Additionally, indoor tasks like \emph{Search for Supplies} and \emph{Search Barn} are also challenging due to GPT4-o's poor spatial perception, which finds it difficult to locate target objects and ends up circling aimlessly. The open-ended task, \emph{Buy Supply}, shows that even without in-game guidance, \projname still manages to complete the task with its superior reasoning ability.

\begin{figure}[ht]
\centering
\begin{minipage}{0.66\textwidth}
\begin{subfigure}[b]{0.5\textwidth}
\centering
\includegraphics[width=\textwidth]{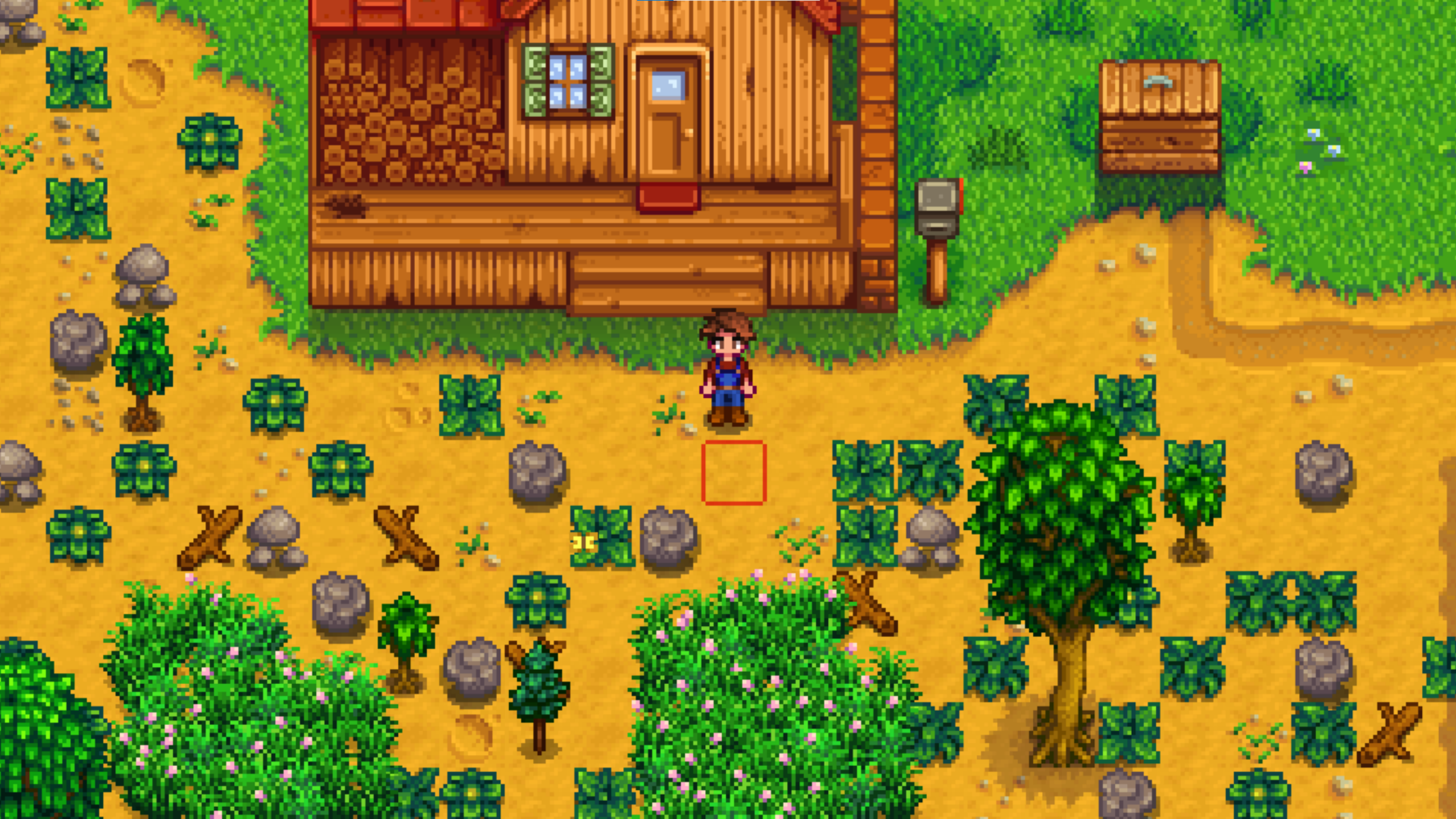}
\caption{Before \emph{Farm Clear-up}}
\label{fig:stardew_clearup_start}
\end{subfigure}
\begin{subfigure}[b]{0.5\textwidth}
\centering
\includegraphics[width=\textwidth]{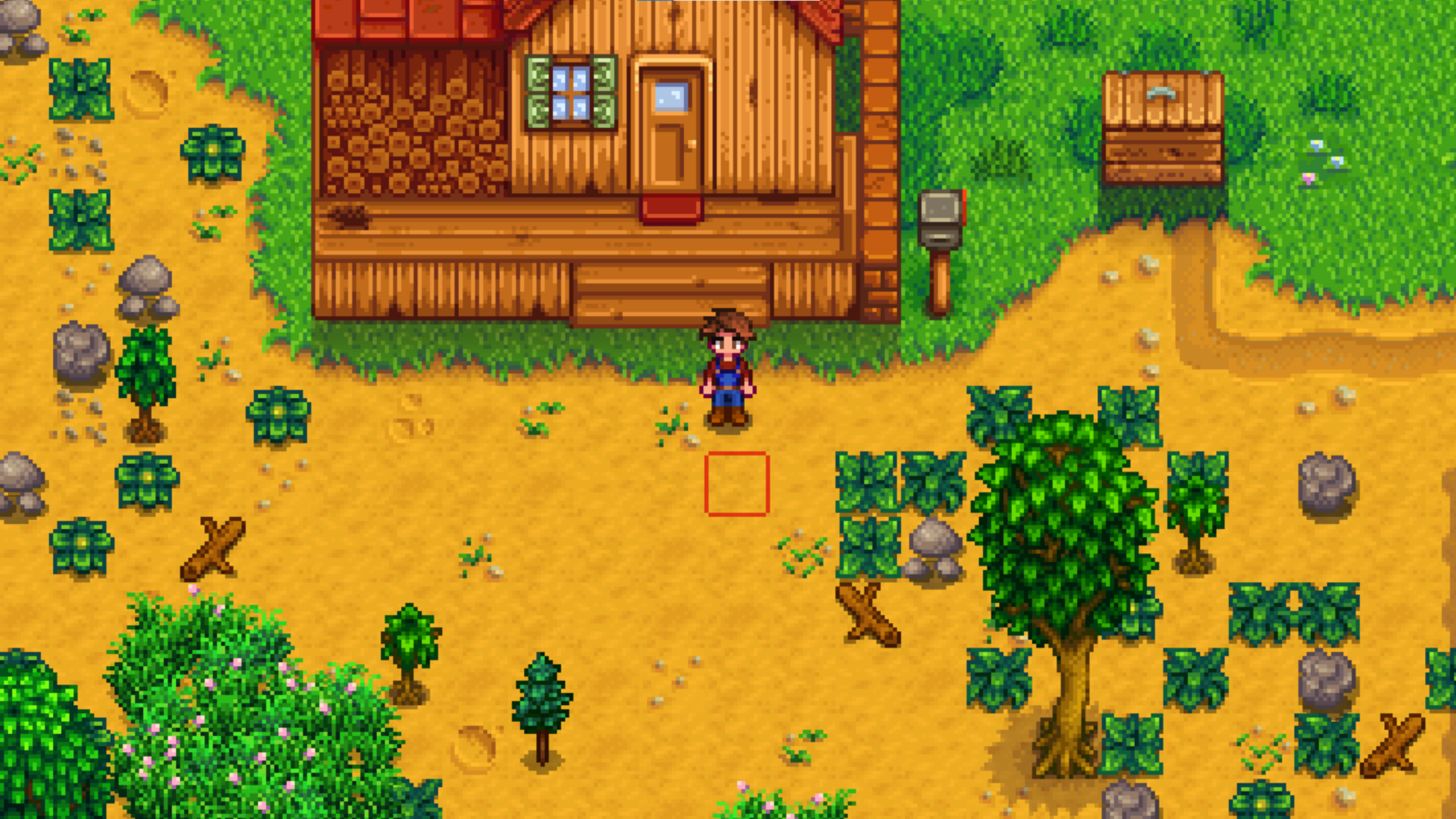}
\caption{After \emph{Farm Clear-up}}
\label{fig:stardew_clearup_end}
\end{subfigure}
\begin{subfigure}[b]{0.5\textwidth}
\centering
\includegraphics[width=\textwidth]{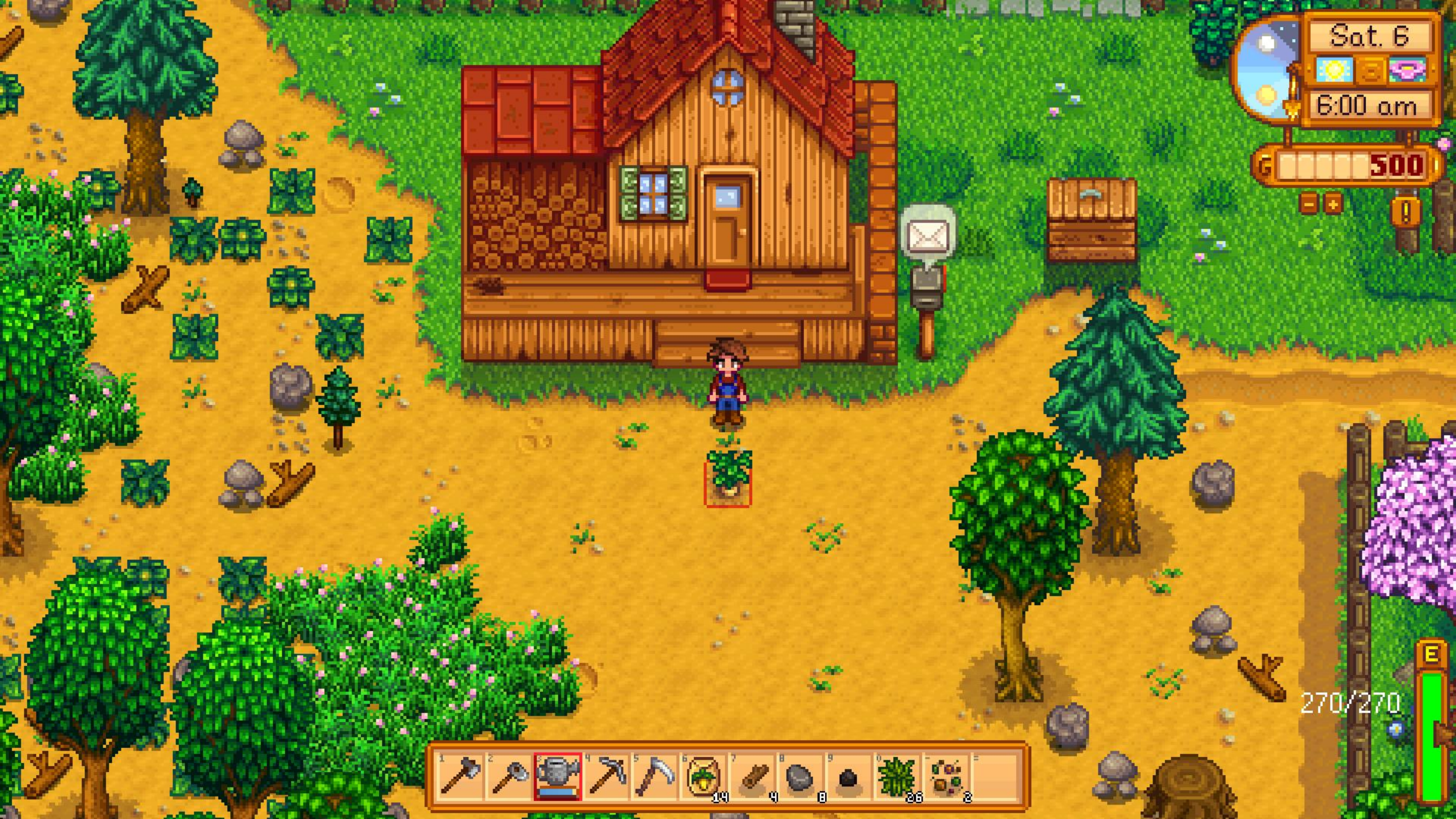}
\caption{\emph{Cultivation}}
\label{fig:stardew_cultivation}
\end{subfigure}
\begin{subfigure}[b]{0.5\textwidth}
\centering
\includegraphics[width=\textwidth]{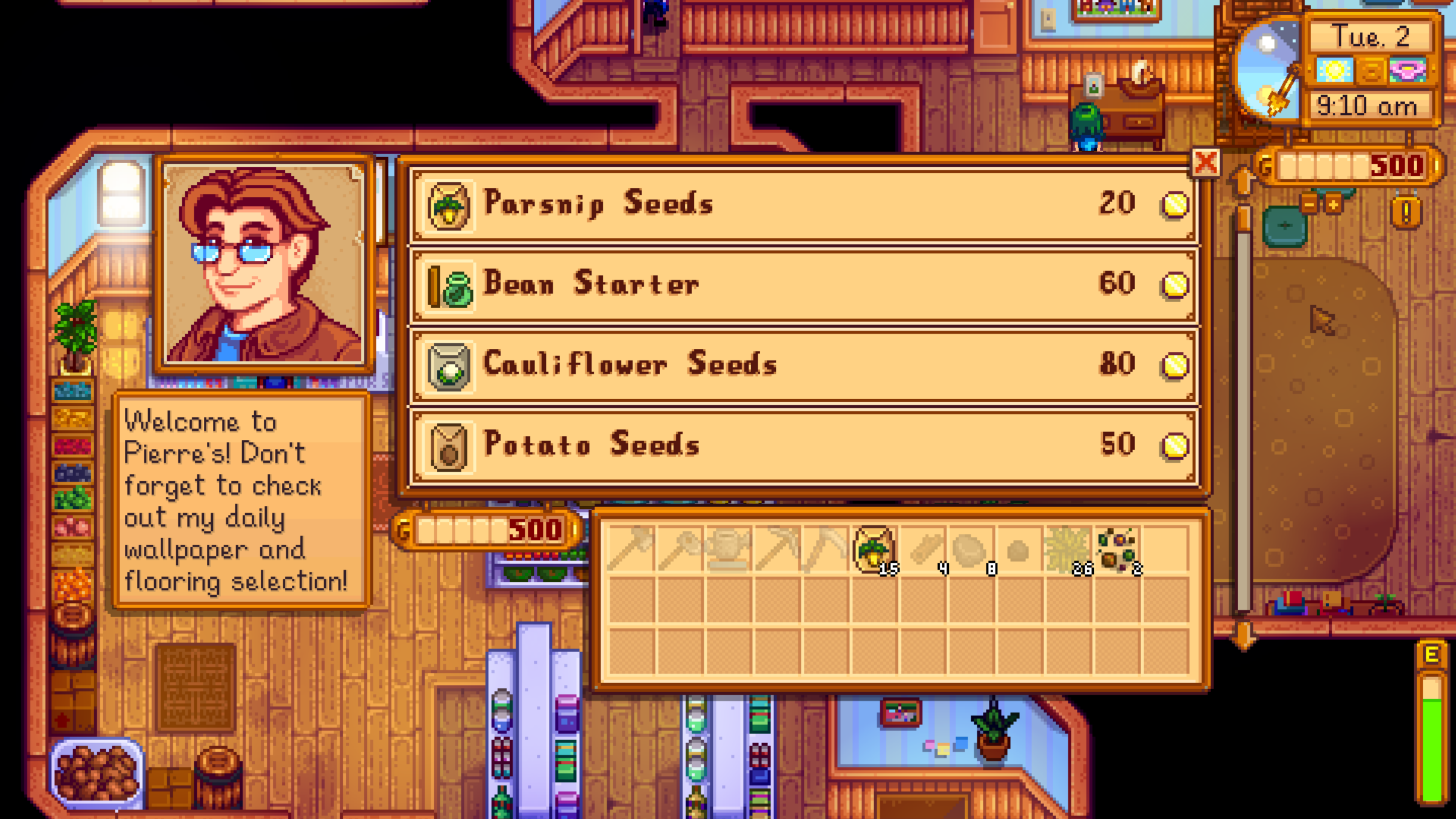}
\caption{\emph{Shopping}}
\label{fig:stardew_shopping}
\end{subfigure}
\caption{Three tasks in Stardew Valley. Each task is run with a maximum of 100 steps.}
\label{fig:stardew_tasks}
\end{minipage}
\begin{minipage}{0.33\textwidth}
\begin{subfigure}[b]{\textwidth}
\centering
\includegraphics[width=\textwidth]{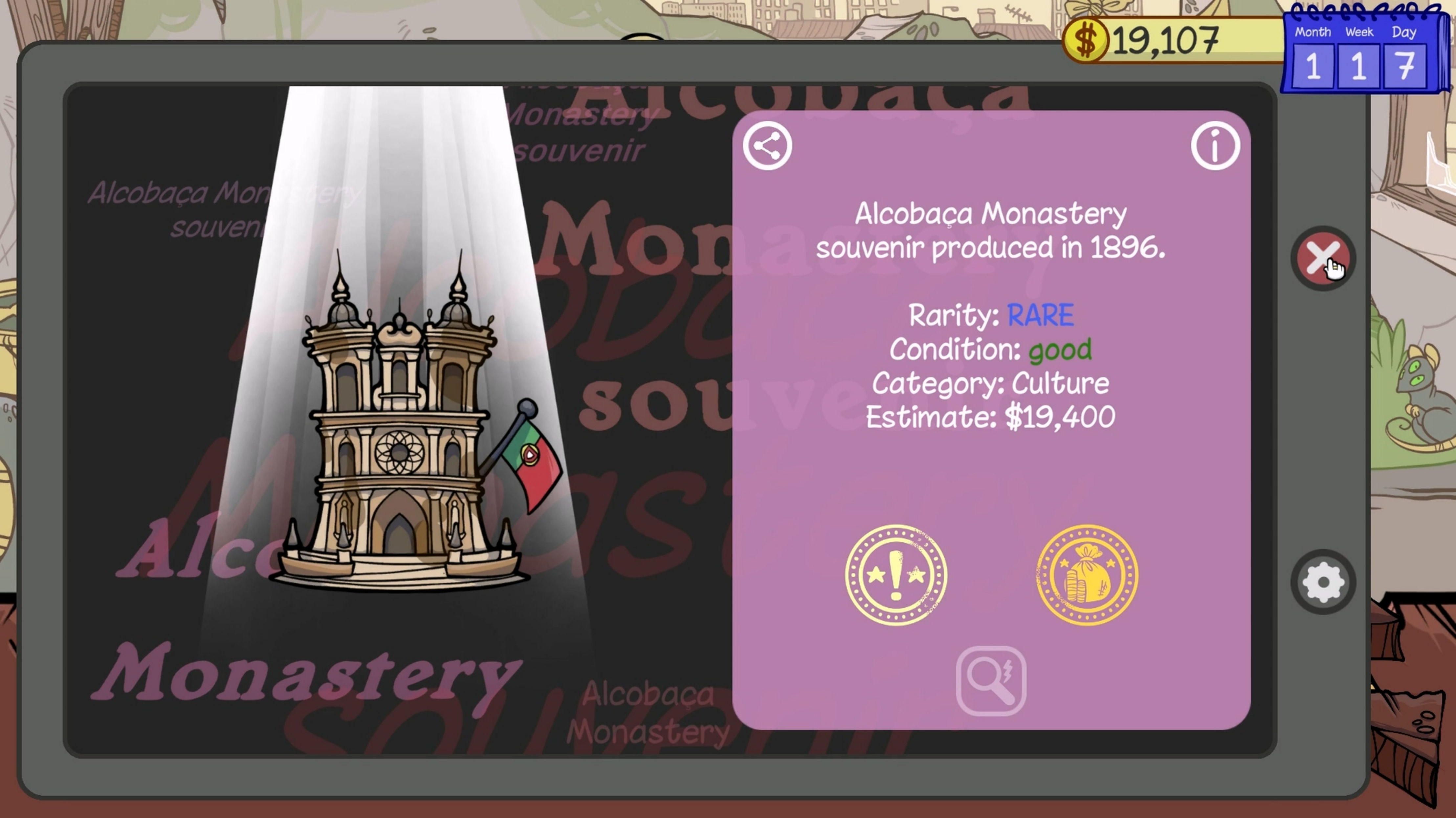}
\caption{Evaluate an item on offer.}
\label{fig:dealers_main_figure_1}
\end{subfigure}
\begin{subfigure}[b]{\textwidth}
\centering
\includegraphics[width=\textwidth]{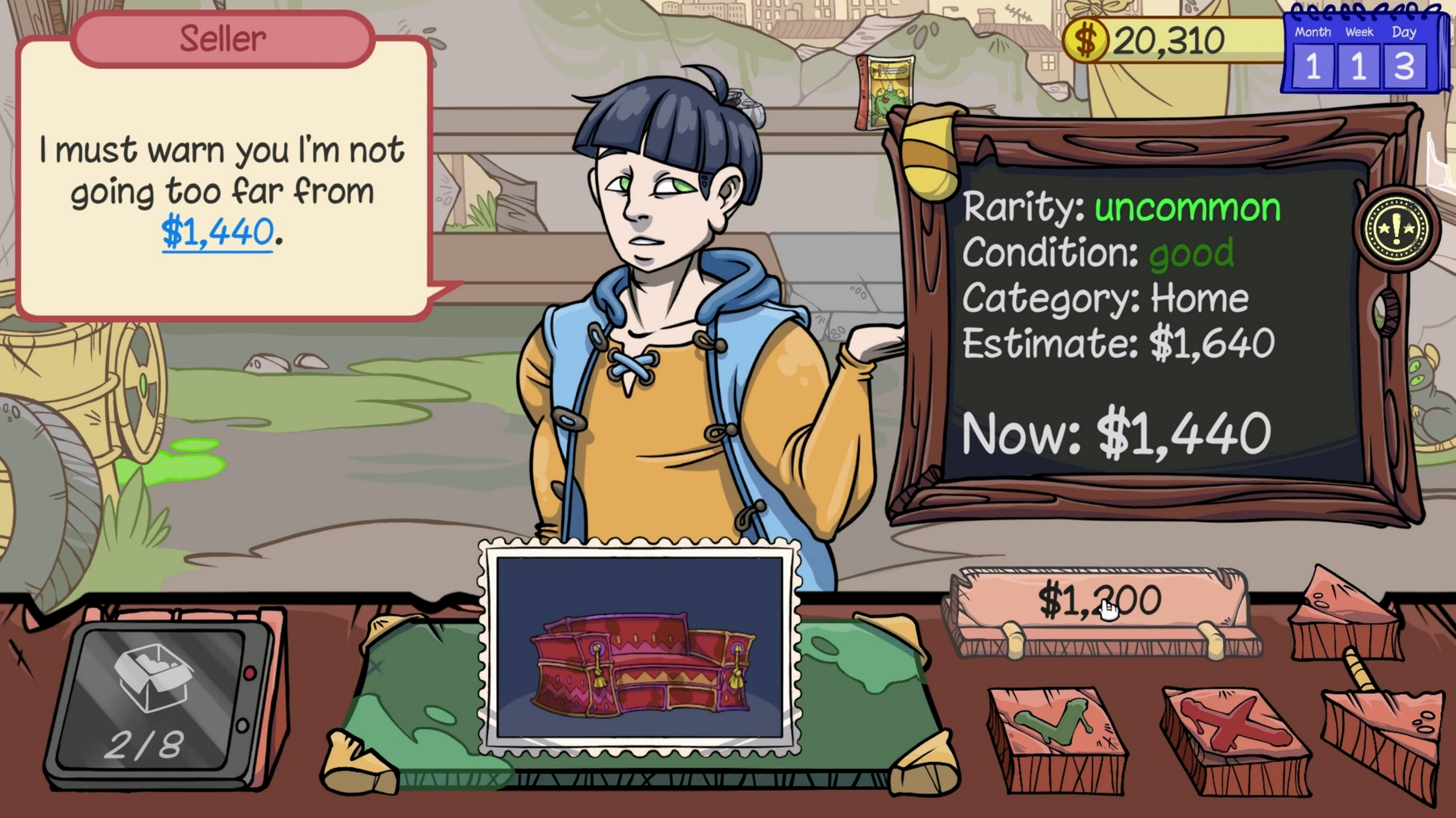}
\caption{Haggle by proposing price.}
\label{fig:dealers_main_figure_2}
\end{subfigure}
\caption{
The task in Dealer's Life 2 of running the shop for a week.
}
\end{minipage}
\end{figure}

\begin{wraptable}{r}{0.33\textwidth}
\vspace{-16pt}
\centering
\setlength{\tabcolsep}{3pt}
\caption{\projname in Stardew Valley.}
\vspace{-6pt}
\label{tab:stardew_results}
\begin{small}
\vspace{-2pt}

\begin{tabular}{c|c}
\toprule
\begin{tabular}[c]{@{}c@{}} Task \end{tabular}
& \begin{tabular}[c]{@{}c@{}} Results\end{tabular} \\ 
\midrule
\begin{tabular}[c]{@{}c@{}} Farm Clearup \\ (Grids Num.)\end{tabular}
& \begin{tabular}[c]{@{}c@{}} $14.8 \pm 5.0$ \end{tabular} \\
\begin{tabular}[c]{@{}c@{}} Cultivation  \end{tabular}
& \begin{tabular}[c]{@{}c@{}} $\mathlarger{\mathlarger{\nicefrac{4}{5}}}$ \end{tabular}  \\
\begin{tabular}[c]{@{}c@{}} Shopping  \end{tabular}
& \begin{tabular}[c]{@{}c@{}} $\mathlarger{\mathlarger{\nicefrac{1}{5}}}$ \end{tabular} \\
\bottomrule
\end{tabular}
\end{small}
\vspace{-12pt}
\end{wraptable}
\textbf{Stardew Valley.} As shown in Table~\ref{tab:stardew_results}, we surprisingly find that GPT-4o  struggles with accurately recognizing and locating objects near the player in this 2D game. This leads to difficulties for the agent to interact with objects or people, as it requires the player to stand precisely in front of them in the grid (\eg when entering doors, using a pickaxe to break stones). It explains the inefficiency in the farming task though the agent manages to clear up most of the obstacles in front of the house within 100 steps (as shown in Figure~\ref{fig:stardew_clearup_start} and~\ref{fig:stardew_clearup_end}) and poor performance in the shopping task. On the other hand, relying on episodic summarization and task inference, \projname manages to obtain the parsnip by watering the seed for four days and harvesting.

\begin{wraptable}{r}{0.33\textwidth}
\vspace{-12pt}
\centering
\setlength{\tabcolsep}{2pt}
\caption{\projname in Dealer's Life 2.}
\vspace{-4pt}
\label{tab:dealers_main_results}
\vspace{-2pt}
\begin{small}

\begin{tabular}{c|c}
\toprule
\begin{tabular}[c]{@{}c@{}} Metrics \end{tabular} &
\begin{tabular}[c]{@{}c@{}} Results \end{tabular} \\ 
\midrule
\begin{tabular}[c]{@{}c@{}} Avg. Haggling Count \end{tabular} 
& \begin{tabular}[c]{@{}c@{}} 1.95 $\pm$ 0.43 \end{tabular} \\
\begin{tabular}[c]{@{}c@{}} Turnover Rate (\%)  \end{tabular} 
& \begin{tabular}[c]{@{}c@{}} 93.6 $\pm$ 6.9{\phantom 0} \end{tabular} \\
\begin{tabular}[c]{@{}c@{}} Item Profit Rate (\%) \end{tabular} 
& \begin{tabular}[c]{@{}c@{}} 37.8 $\pm$ 19.1 \end{tabular} \\
\begin{tabular}[c]{@{}c@{}} Total Profit Rate (\%) \end{tabular} 
& \begin{tabular}[c]{@{}c@{}} 39.6 $\pm$ 27.3 \end{tabular} \\
\bottomrule
\end{tabular}
\end{small}
\vspace{-18pt}
\end{wraptable}
\textbf{Dealer's Life 2.}
Table~\ref{tab:dealers_main_results} shows that 
\projname demonstrates robust performance and efficient profit-making on the \emph{Weekly Shop Management} task, successfully finalizing 93.6\% of potential transactions, with an average of 2 negotiation rounds per customer, and generally aiming for a profit rate of over 50\% at the initial offer. It consistently generates profit across all runs, maintaining a total profit rate of +39.6\%, peaking at +87.4\% in a single run.

\begin{figure}[ht]
\vspace{-5pt}
\begin{subfigure}[b]{0.49\textwidth}
\centering
\includegraphics[width=\textwidth]{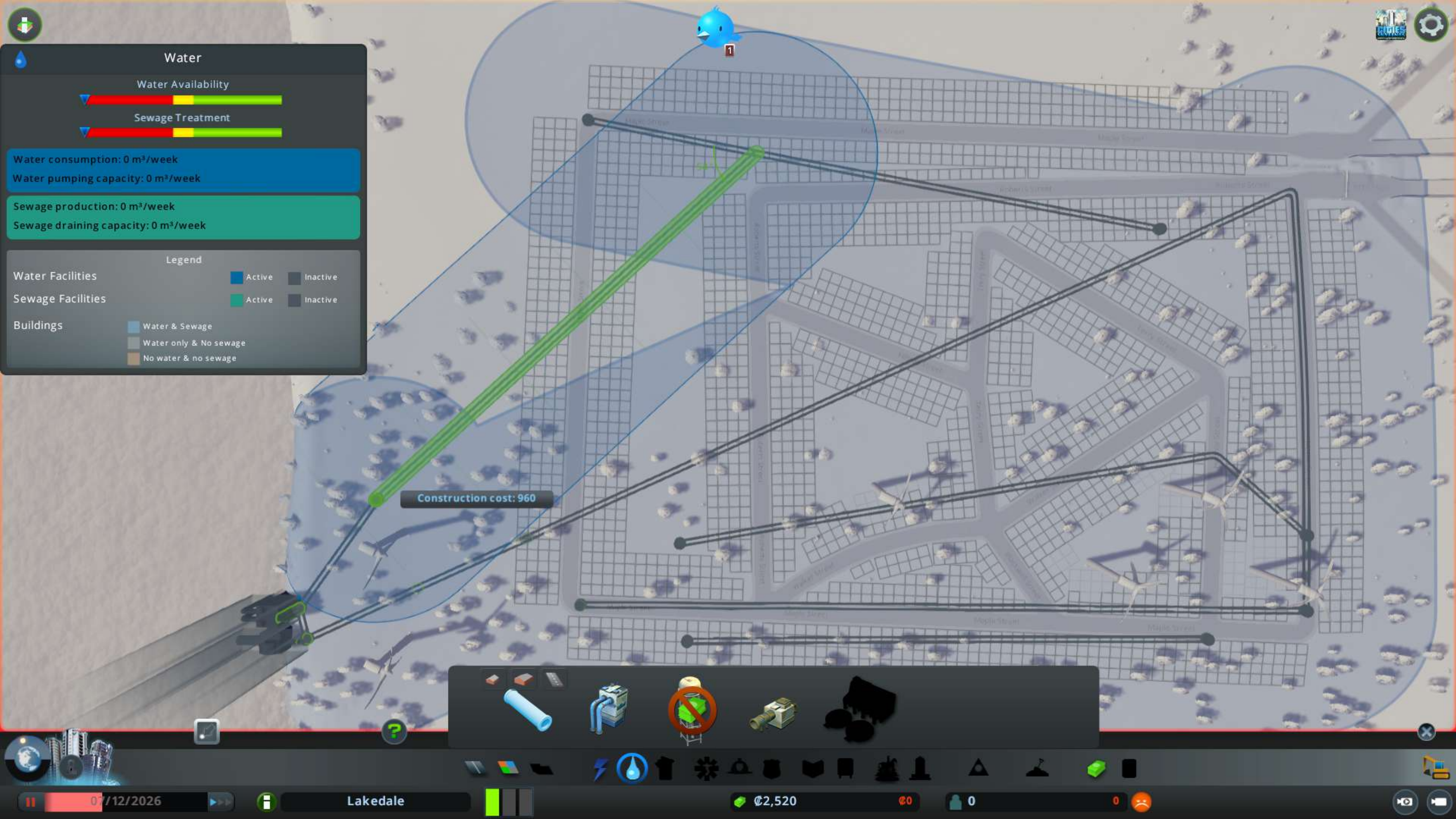}
\caption{\projname fails to connect all water pipes and cover all zones. Missing pipe is shown in green.}
\label{fig:skylines_water_failure}
\end{subfigure}
\centering
\begin{subfigure}[b]{0.49\textwidth}
\centering
\includegraphics[width=\textwidth]{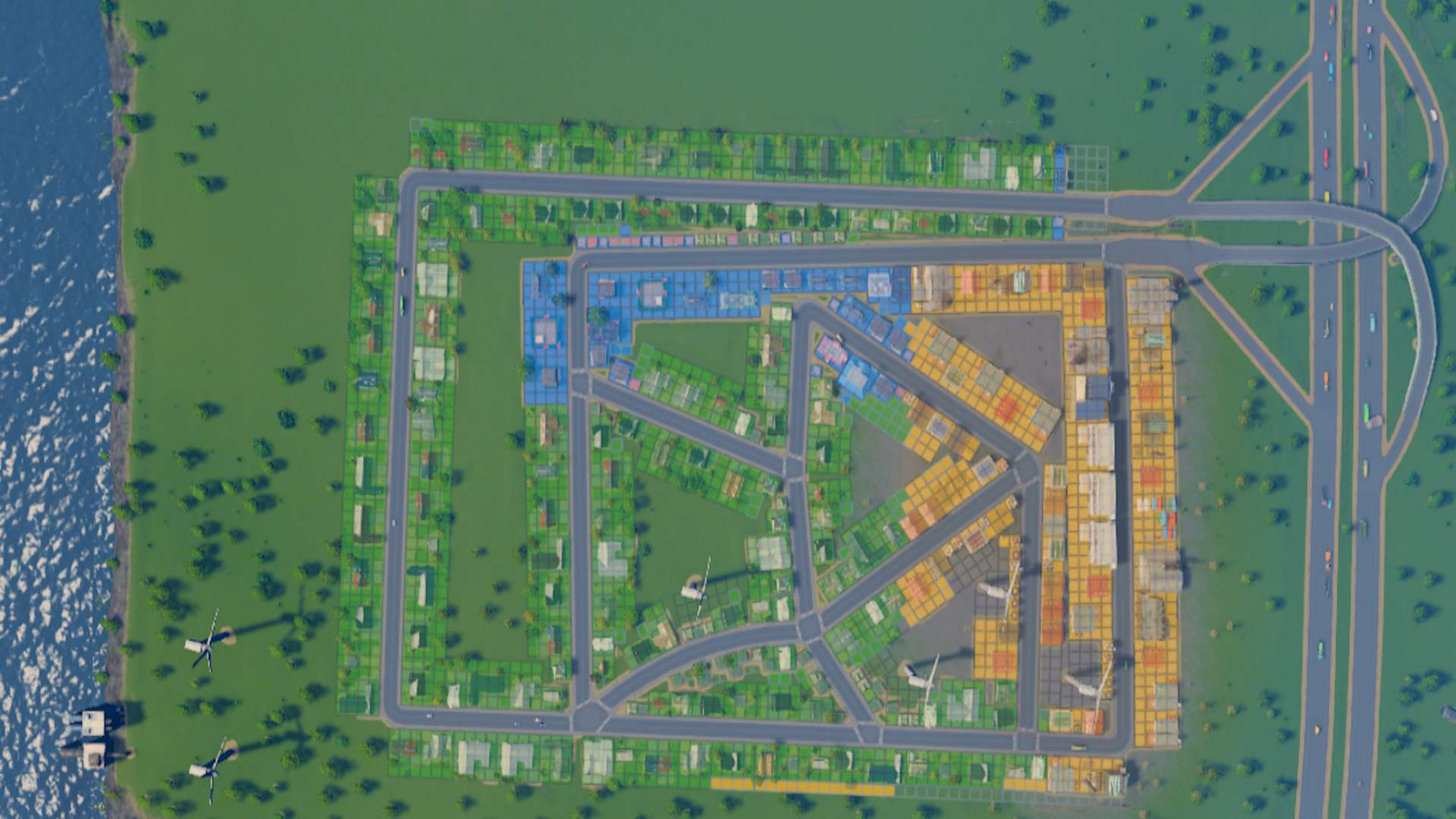}
\caption{The city built by \projname of 1100+ people after human assistance to get water pipes connected.}
\label{fig:skylines_result}
\end{subfigure}
\caption{\projname's craft work in Cities: Skylines, shown in water view (left) and zoning view (right) with the same city. After fixing the error shown on the left, the city can finally be built into what is shown on the right. Each task is run with a maximum of 1000 steps or the budget is used up.
}
\vspace{-12pt}
\end{figure}

\begin{wraptable}{r}{0.4\textwidth}
\centering
\setlength{\tabcolsep}{2pt}
\caption{\projname in Cities: Skylines.}
\vspace{-4pt}
\label{tab:skylines_results}
\begin{small}
\begin{tabular}{c|p{1.5cm}<{\centering}}
\toprule
\begin{tabular}[c]{@{}l@{}} Task \& Metrics \end{tabular} 
& \begin{tabular}[c]{@{}c@{}} Results \end{tabular} \\ 
\midrule
\begin{tabular}[c]{@{}c@{}} Roads in Closed Loop  \end{tabular} 
& \begin{tabular}[c]{@{}c@{}} $\mathlarger{\mathlarger{\nicefrac{4}{5}}}$ \end{tabular}  \\
\begin{tabular}[c]{@{}c@{}} Sufficient Water Supply  \end{tabular} 
& \begin{tabular}[c]{@{}c@{}} $\mathlarger{\mathlarger{\nicefrac{1}{5}}}$ \end{tabular}  \\
\begin{tabular}[c]{@{}c@{}} Sufficient Electricity Supply \end{tabular} 
& \begin{tabular}[c]{@{}c@{}} $\mathlarger{\mathlarger{\nicefrac{5}{5}}}$ \end{tabular}  \\
\begin{tabular}[c]{@{}c@{}} Zones Area $\geq$ $90\%$  \end{tabular} 
& \begin{tabular}[c]{@{}c@{}} $\mathlarger{\mathlarger{\nicefrac{4}{5}}}$ \end{tabular}  \\
\midrule
\begin{tabular}[c]{@{}c@{}} Maximal Population  \end{tabular} 
& \begin{tabular}[c]{@{}c@{}} $450\pm224$ \end{tabular}  \\
\begin{tabular}[c]{@{}c@{}} Maximal Population-w-HA  \end{tabular} 
& \begin{tabular}[c]{@{}c@{}} $850\pm142$ \end{tabular}  \\
\bottomrule
\end{tabular}
\end{small}
\vspace{-8pt}
\end{wraptable}
\textbf{Cities: Skylines}. 
Table~\ref{tab:skylines_results} shows that while \projname manages to build the roads in a closed loop to ensure smooth traffic flow, place multiple wind turbines to provide sufficient electricity supply and cover more than 90\% of available area with residential, commercial and industrial zones, it fails to provide sufficient water supply reliably. As shown in Figure~\ref{fig:skylines_water_failure}, the most common failure case is that water pipes are not connected with each other, resulting in localized water shortages in the city, and preventing new residents from moving in. As shown in Figure~\ref{fig:skylines_result}, with human assistance (-w-HA) to correct the mistakes within three unit operations (building or removing a road/facility/a place of zones is counted as one unit operation), the city built by \projname can eventually reach a population of more than one thousand. Table~\ref{tab:skylines_results} also indicates that \projname nearly completes the city design, albeit with a few mistakes or omissions. With human assistance to fix these small issues, the final population can be doubled.

\begin{figure}[ht]
\vspace{-10pt}
  \centering
    \includegraphics[width=\textwidth]{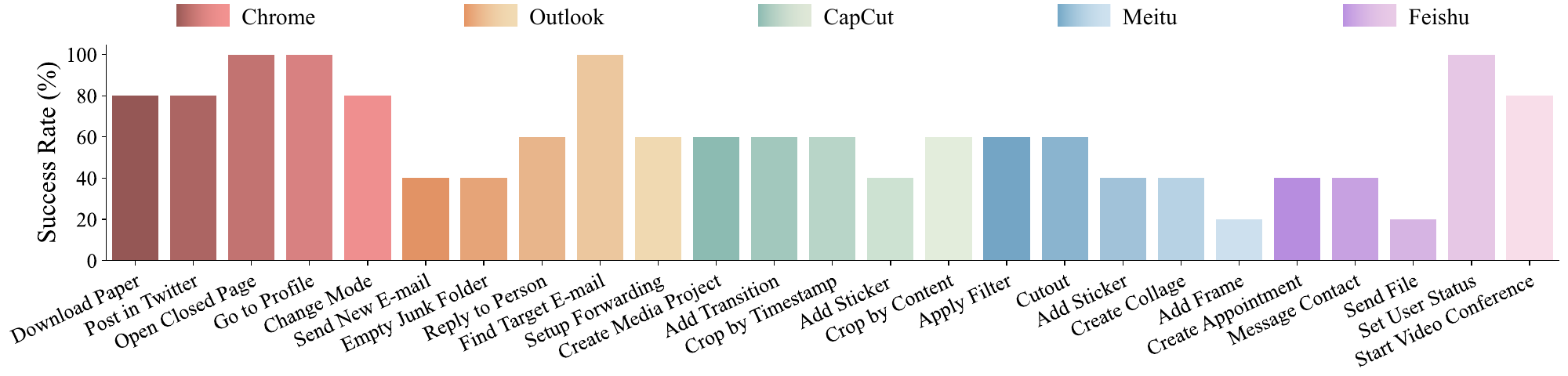}
  \vspace{-10pt}
  \caption{Success rates for tasks in software applications. Each task is run with a maximum of 30 steps.}
  \label{fig:software_barchart}
  \vspace{-15pt}
\end{figure}

\textbf{Software Applications.} 
Figure~\ref{fig:software_barchart} shows \projname's success rates across tasks for all five applications. 
Multiple tasks remain challenging. Even with a well-known GUI, like Chrome and Outlook, GPT-4o still cannot recognize specific UI items to interact with and also struggles with visual context. For example, forgetting to press the Save button in an open dialog, or not distinguishing between a nearby enabled button vs. a distant and disabled one (\eg when posting on Twitter). The phenomenon is more severe in the UI with non-standard layouts, like CapCut, Meitu, and Feishu. Lacking prior knowledge by GPT-4o leads to the failure of task inference and selecting the correct skills.

\begin{wraptable}{r}{0.6\textwidth}
\centering
\footnotesize
\setlength{\tabcolsep}{2pt}
\caption{Success rates (\%) of different methods on OSWorld tasks.}
\label{tab:osworld_main_results}
    \begin{tabular}{c|ccccc|c}
    \toprule
    Method 
    & \begin{tabular}[c]{@{}c@{}}Office \\ (117)\end{tabular} 
    & \begin{tabular}[c]{@{}c@{}}OS \\ (24) \end{tabular} 
    & \begin{tabular}[c]{@{}c@{}}Daily \\ (78) \end{tabular} 
    & \begin{tabular}[c]{@{}c@{}}Workfl-\\ow(101)\end{tabular} 
    & \begin{tabular}[c]{@{}c@{}}Professi-\\onal (49)\end{tabular}
    & \begin{tabular}[c]{@{}c@{}}All \\ (369)\end{tabular}
    \\ 
    \midrule
    \begin{tabular}[c]{@{}c@{}} GPT-4o\end{tabular}    
    & \begin{tabular}[c]{@{}c@{}} \textbf{3.58} \end{tabular} 
    & \begin{tabular}[c]{@{}c@{}}$8.33$ \end{tabular} 
    & \begin{tabular}[c]{@{}c@{}}$6.07$ \end{tabular} 
    & \begin{tabular}[c]{@{}c@{}} \textbf{5.58} \end{tabular} 
    & \begin{tabular}[c]{@{}c@{}} $4.08$\end{tabular} 
    & \begin{tabular}[c]{@{}c@{}} $5.03$\end{tabular} 
     \\
    \begin{tabular}[c]{@{}c@{}} GPT-4o+SoM\end{tabular}    
    & \begin{tabular}[c]{@{}c@{}} \textbf{3.58}\end{tabular} 
    & \begin{tabular}[c]{@{}c@{}} \textbf{20.83}\end{tabular} 
    & \begin{tabular}[c]{@{}c@{}}$3.99$ \end{tabular} 
    & \begin{tabular}[c]{@{}c@{}}$3.60$\end{tabular} 
    & \begin{tabular}[c]{@{}c@{}}$2.04$\end{tabular} 
    & \begin{tabular}[c]{@{}c@{}}$4.59$\end{tabular} 
     \\
    \begin{tabular}[c]{@{}c@{}}\projname\end{tabular}  
    & \begin{tabular}[c]{@{}c@{}} \textbf{3.58} \end{tabular} 
    & \begin{tabular}[c]{@{}c@{}} $16.67$\end{tabular} 
    & \begin{tabular}[c]{@{}c@{}} \textbf{6.55}\end{tabular} 
    & \begin{tabular}[c]{@{}c@{}} $5.48$ \end{tabular} 
    & \begin{tabular}[c]{@{}c@{}} \textbf{20.41}\end{tabular}
    & \begin{tabular}[c]{@{}c@{}}\textbf{7.81}\end{tabular} 
     \\
    \bottomrule
    \end{tabular}
\vspace{-12pt}
\end{wraptable}
\textbf{OSWorld.} Table~\ref{tab:osworld_main_results} shows that \projname achieves the overall highest success rate in OSWorld, compared to the baselines without relying on any internal APIs to provide extra grounding labels, \eg Set-of-Mark (SoM)~\citep{yang2023setofmark}. 
The information gathering module improves grounding for more precise action execution, increasing the performance. The self-reflection module greatly helps it to correctly predict infeasible tasks and subsequently fix mistakes, as exemplified in the professional domain results, where it achieves a $20.41\%$ success rate, significantly surpassing the baselines.

\vspace{-10pt}
\subsection{Ablation Study \& Baselines Comparison}
\vspace{-10pt}
Since no existing methods are fully applicable to the GCC setting, we select several representative methods with necessary adaptions to make them applicable to GCC, labeling them as "like" in Table ~\ref{tab:ablation_studies_results}.
Compared to \projname, React~\citep{yao2023react}-like method only has gather information, skill curation and action planning module, while Reflextion~\citep{shinn2023reflexion}-like method adds a self-reflection and episodic memory, compared to React-like. To show the necessity of multimodal input without access to APIs, we let GPT-4o describe the image and then feed the textual description to Voyager~\citep{wang2023voyager}-like as input. Additionally, experiments with GPT-4o and Claude 3 Opus~\citep{anthropicclaude3} as backbone are conducted. 
Due to the limitation of requests per minute, 
other prompting methods like self-consistency~\citep{wang2022self} and TOT~\citep{yao2024tree} are not considered.

As seen in Table \ref{tab:ablation_studies_results},
all the baseline methods can only complete simple and straightforward tasks without complex targets and time delays. Compared to React-like method, Reflextion-like method has better performance in the task of \emph{Follow Micah} and still fails to complete more complex tasks, emphasizing the importance of task inference and procedural memory. Voyager-like method that loses vision suffers to accomplish tasks and are the worst of all comparison methods. \projname with GPT-4o always has the best performance across all tasks. \projname with GPT-4o has the best performance, while Claude 3 Opus fails frequently due to unreliable OCR ability of the guidance, leading to incorrect skill generation and failures of complex tasks.

\begin{wraptable}{r}{0.4\textwidth}
\vspace{-15pt}
\centering
\includegraphics[width=\linewidth]{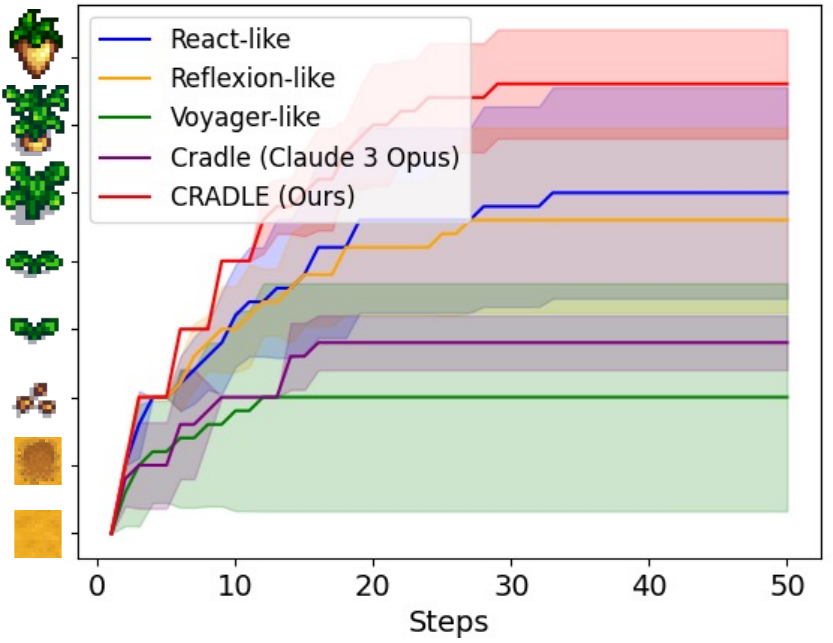}
\vspace{-15pt}
\caption{Performance of each baseline method in task \emph{Cultivation}. Only if the parsnip (shown on the top of the y-axis) is obtained will this run be counted as a success. }
\label{fig:stardew_ablation}
\end{wraptable}

Figure~\ref{fig:stardew_ablation} provides the detailed performance of each baseline method in the \emph{Cultivation} task in Stardew Valley.  Without task inference and episodic memory for summarization, even React-like and Reflexion-like methods sometimes managed to get the parsnip to sprout from the ground, they failed to successfully harvest it because GPT-4o failed to recognize the mature parsnip. Episodic summary can help \projname record the days of watering and know when the crop can be harvested. Voyager-like method struggles with getting out of the house and returning home due to the lack of visual input. Claude 3 Opus also has difficulties in localizing the position of the character and the crop. Moreover, it prefers moving characters much more frequently than GPT-4, resulting in the failure to position the character in front of the crop.

\begin{table}[h]
\centering
\caption{Ablation results for RDR2 and Stardew Valley Cultivation tasks. Numbers before the brackets are interaction steps averaged over five trials. N/A indicates failure in all five runs. For RDR2, each task is run at most 500 steps and for Stardew Valley, each task is run for at most 100 steps.}
\label{tab:ablation_studies_results}
\setlength{\tabcolsep}{2pt}
\begin{footnotesize}
    \begin{tabular}{c|ccccc|c}
    \toprule
    Method 
    & \begin{tabular}[c]{@{}c@{}}Follow \\ Dutch\end{tabular} 
    & \begin{tabular}[c]{@{}c@{}}Follow \\ Micah\end{tabular} 
    & \begin{tabular}[c]{@{}c@{}}Hitch \\ Horse\end{tabular} 
    & \begin{tabular}[c]{@{}c@{}}Protect \\ Dutch\end{tabular} 
    & \begin{tabular}[c]{@{}c@{}}Search \\ for Supplies\end{tabular} 
    & Cultivation \\ 
    \midrule
    \begin{tabular}[c]{@{}c@{}}React~\citep{yao2023react}-like (GPT-4o)\end{tabular}       
    & \begin{tabular}[c]{@{}c@{}}\textcolor{codegreen}{$15 \pm 2$} \textcolor{codegreen}{(\(\nicefrac{5}{5}\))}\end{tabular} 
    & \begin{tabular}[c]{@{}c@{}}\textcolor{codegreen}{$74 \pm 0$} \textcolor{codegreen}{(\(\nicefrac{1}{5}\))}\end{tabular}
    & \begin{tabular}[c]{@{}c@{}}N/A\end{tabular}
    & \begin{tabular}[c]{@{}c@{}}N/A\end{tabular}
    & \begin{tabular}[c]{@{}c@{}}N/A\end{tabular}
    & \begin{tabular}[c]{@{}c@{}}N/A\end{tabular} \\
    \begin{tabular}[c]{@{}c@{}}Reflextion~\citep{shinn2023reflexion}-like (GPT-4o)\end{tabular}       
    & \begin{tabular}[c]{@{}c@{}}\textcolor{codegreen}{$19 \pm 4$} \textcolor{codegreen}{(\(\nicefrac{5}{5}\))}\end{tabular} 
    & \begin{tabular}[c]{@{}c@{}}\textcolor{codegreen}{$58 \pm 14$} \textcolor{codegreen}{(\(\nicefrac{2}{5}\))}\end{tabular}
    & \begin{tabular}[c]{@{}c@{}}N/A\end{tabular}
    & \begin{tabular}[c]{@{}c@{}}N/A\end{tabular}
    & \begin{tabular}[c]{@{}c@{}}N/A\end{tabular}
    & \begin{tabular}[c]{@{}c@{}}N/A\end{tabular} \\
    \begin{tabular}[c]{@{}c@{}}Voyager~\citep{wang2023voyager}-like (GPT-4o)\end{tabular}       
    & \begin{tabular}[c]{@{}c@{}}\textcolor{codegreen}{$32 \pm 12$} \textcolor{codegreen}{(\(\nicefrac{3}{5}\))}\end{tabular} 
    & \begin{tabular}[c]{@{}c@{}}N/A\end{tabular}
    & \begin{tabular}[c]{@{}c@{}}N/A\end{tabular}
    & \begin{tabular}[c]{@{}c@{}}N/A\end{tabular}
    & \begin{tabular}[c]{@{}c@{}}N/A\end{tabular}
    & \begin{tabular}[c]{@{}c@{}}N/A\end{tabular} \\
    \begin{tabular}[c]{@{}c@{}}\projname (Claude 3 Opus)\end{tabular}       
    & \begin{tabular}[c]{@{}c@{}}\textcolor{codegreen}{$30 \pm 7$} \textcolor{codegreen}{(\(\nicefrac{5}{5}\))}\end{tabular} 
    & \begin{tabular}[c]{@{}c@{}}\textcolor{codegreen}{$52 \pm 17$} \textcolor{codegreen}{(\(\nicefrac{4}{5}\))}\end{tabular} 
    & \begin{tabular}[c]{@{}c@{}}N/A\end{tabular}
    & \begin{tabular}[c]{@{}c@{}}N/A\end{tabular}
    & \begin{tabular}[c]{@{}c@{}}N/A\end{tabular}
    & \begin{tabular}[c]{@{}c@{}}N/A\end{tabular} \\
    \midrule
    \begin{tabular}[c]{@{}c@{}}\projname (GPT-4o) \\ (Ours)\end{tabular}       
    & \begin{tabular}[c]{@{}c@{}}\textcolor{codegreen}{$\mathbf{13} \pm \mathbf{3}$} \\\textcolor{codegreen}{\textbf{(\({\nicefrac{5}{5}}\))}}\end{tabular} 
    & \begin{tabular}[c]{@{}c@{}}\textcolor{codegreen}{$\mathbf{33} \pm \mathbf{3}$} \\\textcolor{codegreen}{\textbf{(\({\nicefrac{5}{5}}\))}}\end{tabular} 
    & \begin{tabular}[c]{@{}c@{}}\textcolor{codegreen}{$\mathbf{26} \pm \mathbf{5}$} \\\textcolor{codegreen}{\textbf{(\({\nicefrac{4}{5}}\))}}\end{tabular} 
    & \begin{tabular}[c]{@{}c@{}}\textcolor{codegreen}{$\mathbf{461} \pm \mathbf{0}$} \\\textcolor{codegreen}{\textbf{(\({\nicefrac{1}{5}}\))}}\end{tabular} 
    & \begin{tabular}[c]{@{}c@{}}\textcolor{codegreen}{$\mathbf{134} \pm \mathbf{0}$} \\\textcolor{codegreen}{\textbf{(\({\nicefrac{1}{5}}\))}}\end{tabular} 
    & \begin{tabular}[c]{@{}c@{}}\textcolor{codegreen}{$\mathbf{24} \pm \mathbf{4}$} \\\textcolor{codegreen}{\textbf{(\({\nicefrac{4}{5}}\))}}\end{tabular} \\
    \bottomrule
\end{tabular}
\end{footnotesize}
\end{table}

\section{Limitations and Future Work.}
\label{sec:limitations}
Despite \projname's encouraging performance across games and software, several limitations remain. 
i) Due to the weaknesses of current LMM models, \projname struggles in recognizing out-of-distribution (OOD) icons and completing OOD tasks, such as games with non-realistic styles, \ie Stardew Valley. As LMMs evolve, they can further improve \projname's performance. 
ii) Audio, as an important modality, often plays an important role in games and software; however, it has not been considered in this work. The future work will be enabling \projname to process the audio and graphical input simultaneously.
iii) Most \projname's modules need to call LMM explicitly to process the input for best performance, resulting in frequent interactions with LMM and potentially high costs and long delays. The six modules represent a problem-solving mindset; as LMM capabilities improve, some or even all of these modules may be combined into a single request.
iv) In this work, we mainly focus on enabling foundation agents to interact with various software in a unified manner without taking training into consideration. As SIMA~\citep{raad2024scaling} has already shown promising results in a similar setting with the trained agents, we will let \projname autonomously explore and improve over environments through RL~\citep{tan2023true} or collect expert demonstrations for supervised learning~\citep{raad2024scaling}. 
v) Though \projname is generally applicable to any computer task, only a few with limited tasks are investigated in this work. We will extend it to a wider range of targets, go deeper into the complex games, and make it easier for users to adapt on their own. 

\projname holds great potential to improve effective general computer task completion
and boost research and deployment of foundation agents. However, there is also a risk of unintended or unsuitable usage, including developing game cheats, incorrect operations of software with harmful failures, or other negative agent behavior. Therefore, additional regulations or safeguards are required for secure and responsible deployments across digital and physical environments.

\section{Conclusion}
\label{sec:conclusion}
In this work, we introduce GCC, a general and challenging setting with a unified and standard interface for control of diverse video games and other software (via screenshots, and keyboard and mouse operations), paving the way towards general foundation agents across all digital world tasks. To properly address the challenges GCC presents, we propose a novel open-source framework, {\projname}, which exhibits strong performance in reasoning and performing actions to accomplish real missions or tasks in a set of complex video games and common software applications. To the best of our knowledge, \projname is the first framework that enables foundation agents to succeed in such a diverse set of environments without relying on any built-in APIs. The success of \projname greatly extends the reach of foundation agents and demonstrates the feasibility of converting any software, especially complex games, into benchmarks to evaluate agents' general intelligence and facilitate further data collection for self-improvement.  Although \projname still faces difficulties in certain tasks, it serves as a pioneering work to develop more powerful LMM-based agents across computer control tasks, combining both further framework enhancements and new advances in LMMs. 

\section{Acknowledgments and Disclosure of Funding}
We thank Ye Wang and Jiangxing Wang for their time and effort in helping us test the framework. Xinrun Xu is advised by Dr. Zhiming Ding from the Institute of Software, Chinese Academy of Sciences, and is supported by the National Key R\&D Program of China (No. 2022YFF0503900).

\section{Team Members and Contributions}

\subsection{Roles}
\textbf{Program Leads:} Zongqing Lu, Shuicheng Yan, and Bo An\\
\textbf{Team Lead:} Weihao Tan \\
\textbf{Framework  Co-Leads:} Börje F. Karlsson and Weihao Tan \\
\textbf{General Advisors}: Xinrun Wang and Börje F. Karlsson \\
\textbf{Core Contributors}: Weihao Tan, Wentao Zhang, Xinrun Xu, Haochong Xia, Ziluo Ding, Boyu Li, Bohan Zhou, Junpeng Yue, Jiechuan Jiang, Yewen Li, Ruyi An, Molei Qin, Chuqiao Zong, Longtao Zheng, Yujie Wu, Xiaoqiang Chai, Xinrun Wang, and Börje F. Karlsson

\subsection{Detailed Contributions}
\textbf{Framework Design.} Weihao Tan,  Börje F. Karlsson, Wentao Zhang, Ziluo Ding, and Xinrun Wang. 

\textbf{RDR2.} Implementation, experiments, and analysis by Weihao Tan, Börje F. Karlsson, Wentao Zhang, Ziluo Ding, Boyu Li, Bohan Zhou, Junpeng Yue, Haochong Xia, Jiechuan Jiang, Molei Qin, Longtao Zheng, Xinrun Xu, Yifei Bi, Pengjie Gu, and Yewen Li. Xinrun Wang provided insightful suggestions.

\textbf{Stardew Valley.} Implementation, experiments, and analysis by Weihao Tan, Wentao Zhang, Chuqiao Zong, Yujie Wu, Xiaoqiang Chai, Haochong Xia, Yewen Li, Ruyi An, and Molei Qin. Xinrun Wang, Long Tian, Chaojie Wang, and Börje F. Karlsson provided insightful suggestions.

\textbf{Cities: Skylines.} Implementation, experiments, and analysis by Weihao Tan, Wentao Zhang, Haochong Xia, and Ceyao Zhang. Xinrun Wang provided insightful suggestions.

\textbf{Dealer's Life 2.} Implementation, experiments, and analysis by Yewen Li, Ruyi An. Weihao Tan, Wentao Zhang, and Xinrun Wang provided insightful suggestions.

\textbf{Software Applications} (including \textbf{Chrome}, \textbf{Outlook}, \textbf{CapCut}, \textbf{Meitu}, and \textbf{Feishu}). Implementation, experiments, and analysis by Xinrun Xu, Börje F. Karlsson, and Xiyun Li. Weihao Tan and Xinrun Wang provided insightful suggestions.

\textbf{OSWorld.} Implementation, experiments, and analysis by Haochong Xia, Tianbao Xie, Pengjie Gu. Weihao Tan, Xinrun Xu, Xinrun Wang, and Börje F. Karlsson provided insightful suggestions.

\textbf{Paper Writing.} Weihao Tan, Xinrun Wang, Börje F. Karlsson, Wentao Zhang, Xinrun Xu, Yewen Li, Ruyi An, and Chuqiao Zong. 
Haochong Xia, Molei Qin, and Ceyao Zhang further contributed with writers in appendix organization.

\textbf{Organization.} Zongqing Lu, Shuicheng Yan, and Bo An provided directional research advice and organizational support.

\bibliography{neurips_data_2024}
\bibliographystyle{plain}

\clearpage

\renewcommand{\thepart}{}
\renewcommand{\partname}{}
\newpage

\appendix
\part{Appendix}
\parttoc
\clearpage

\section{General Implementation}
\label{app:general_implementation}
\input{appendix/general_implementation}

\section{Red Dead Redemption II}
\label{appx:rdr2}
\input{appendix/rdr2}

\section{Stardew Valley}
\label{appx:stardew}
\input{appendix/stardew}

\section{Dealer's Life 2}
\label{appx:dealer}
\input{appendix/dealer}

\section{Cities: Skylines}
\label{appx:skylines}
\input{appendix/skylines}

\section{Software Applications}
\label{appx:software}
\input{appendix/software}

\section{OSWorld}
\label{appx:osworld}
\input{appendix/osworld}

\section{Cradle Prompts}
\label{appx:prompts}

Here we exemplify the utilized prompts, for each module in the framework. All prompts and customizations are included in the relevant branch in \projname's open-source repository in GitHub~\footnote{https://github.com/BAAI-Agents/Cradle}.

\renewcommand\lstlistingname{Prompt}

\subsection{Prompts for RDR2}
\begin{lstlisting}[breaklines=true,caption={RDR2: Information Gathering prompt.}]
Assume you are a helpful AI assistant integrated with 'Red Dead Redemption 2' on the PC, equipped to handle a wide range of tasks in the game. Your advanced capabilities enable you to process and interpret gameplay screenshots and other relevant information.

<$few_shots$>

<$image_introduction$>

Current task:
<$task_description$>

Target_object_name: Assume you can use an object detection model to detect the most relevant object for completing the current task if needed. What object should be detected to complete the task based on the current screenshot and the current task? You should obey the following rules: 
1. The object should be relevant to the current target or the intermediate target of the current task. Just give one name without any modifiers. 
2. If no explicit weapon is specified on the weapon radial menu, prioritize choosing 'gun' as the weapon. 
3. If no explicit shoot target is specified, prioritize choosing 'person' as the target. 
4. If no explicit item is specified, only output 'null'.
5. If the object name belongs to the person type, replace it with 'person'. 
6. If there is no need to detect an object, only output "null". 
7. If you are on the trade, map, inventory, or satchel interfaces, only output 'null'.

Reasoning_of_object: Why was this object chosen, or why is there no need to detect an object? 

Description: Please describe the screenshot image in detail. Pay attention to any maps in the image, if any, especially critical icons, red paths to follow, or created waypoints. If there are multiple images, please focus on the last one. 

Screen_classification: Please select the class that best describes the screenshot among "Inventory", "Radial menu", "Satchel", "Map", "Trade", "Pause", and "General game interface without any menu". Output the class of the screenshot in the output of Screen_classification.

Reasoning_of_screen: Why was this class chosen for the current screenshot?

Movement: Does the current task require the character to go somewhere?

Noun_and_Verb: The number of nouns and verbs in the current task.

Task_horizon: Please judge the horizon of the current task, i.e., whether this task needs multiple or only one interaction.
There are two horizon types: long-horizon and short-horizon. For long-horizon tasks, the output should be 1. For short-horizon tasks, the output should be 0. You should obey the following rules: 
1. If the task contains only nouns without verbs, it is short-horizon.
2. If the task contains more than one verb, it is long-horizon.
3. If the task requires the character to go somewhere, it is long-horizon.
Short-horizon tasks are sub-goals during a long-horizon task, which only need one interaction. There are some examples of short-horizon tasks:
1. Pick up something: To complete this task, the character needs to execute the action "pick up" only once, so it is short-horizon.
2. Use or press [B] key: The character needs to press the key [B] only once to talk, so it is short-horizon.
3. Talk to somebody: The character needs to press a certain button once to complete this task, so it is short-horizon.
Long-horizon tasks are long-term goals, which usually need many interactions. There are some examples of long-horizon tasks.
1. Go outside: The character should go outside step by step, so it is long-horizon.
2. Approach something: The character should move closer to the target step by step, so it is long-horizon.
3. Keep away from something, shoot, take down, or battle with something: The character must engage in a series of interactions, so it is long-horizon.

Reasoning_of_task: Why do you make such a judgment of task_horizon?

You should only respond in the format described below and not output comments or other information.
Target_object_name:
Name
Reasoning_of_object:
1. ...
2. ...
...
Description:
The image shows...
Screen_classification:
Class of the screenshot
Reasoning_of_screen:
1. ...
2. ...
...
Movement:
Yes or No
Noun_and_Verb:
1 noun 1 verb
Task_horizon:
1
Reasoning_of_task:
1. ...
2. ...
...
\end{lstlisting}

\begin{lstlisting}[breaklines=true,caption={RDR2: Gather Text Information prompt.}]
Assume you are a helpful AI assistant integrated with 'Red Dead Redemption 2' on the PC, equipped to handle a wide range of tasks in the game. Your advanced capabilities enable you to process and interpret gameplay screenshots and other relevant information.

<$image_introduction$>

Information: List all text prompts on the screenshot from the top to the bottom, even the text prompt is one word.

All information should be categorized into one or more kinds of <$information_type$>. If you think a piece of information is both "A" and "B" categories, you should write information in both "A" and "B" categories. For example, "use E to drink water" could both be "Action Guidance" and "Task Guidance" categories.

Item_status: The helpful information to the current context in the game, such as the cash, amount of ammo, current using item, if the player is wanted, etc. This content should be pairs of status names and their values. For example, "cash: 100$". If there is no on-screen text and no item status, only output "null".

Environment_information: The information about the location, time, weather, etc. This content should be pairs of status names and their values. For example, "location: VALENTINE". If there is no on-screen text and environment information, only output "null".

Notification: The game will give notifications showing the events in the world, such as obtaining items or rewards, completing objectives, and becoming wanted. Besides, it also contains valuable notifications of the game's mechanisms, such as "Health is displayed in the lower left corner". The content must be the on-screen text. If there is no on-screen text or notification, only output "null".

Task_guidance: The content should obey the following rules:
1. The content of task guidance must be an on-screen text prompt, including the menu and the general game interface.
2. The game will give guidance on what should be done to proceed with the game, for example, "follow Tom". This is task guidance.
3. The game will give guidance on how to perform a task using keyboard keys or mouse buttons, for example, "use E to drink water". This is task guidance.
4. If no on-screen text prompt or task guidance exists, only output "null". Never derive the task guidance from the dialogue or notifications.

Action_guidance: The game will give guidance on how to perform a task using keyboard keys or mouse buttons; you must generate the code based on the on-screen text. The content of the code should obey the following code rules:
1. You should first identify the exact keyboard or mouse key represented by the icon on the screenshot. 'Ent' refers to 'enter'. 'RM' refers to 'right mouse button'. 'LM' refers to 'left mouse button'. You should output the full name of the key in the code.
2. You should refer to different examples strictly based on the word used to control the key, such as 'use', 'hold', 'release', 'press', and 'click'.
3. If 'use' or 'press' is in the prompt to control the keyboard key or mouse button, io_env.key_press('key', 2) or io_env.mouse_click('button', 2) must be used to act on it. Refer to Examples 1, 2, and 3.
4. If there are multiple keys, io_env.key_press('key1,key2', 2) must be used to act on it. Refer to Example 4.
5. If 'hold' is in the prompt to control the keyboard key or mouse button, it means keeping the key held with io_env.key_hold or the button held with io_env.mouse_hold (usually indefinitely, with no duration). If you need to hold it briefly, specify a duration argument. Refer to Examples 5 and 6.
6. All durations are set to a minimum of 2 seconds by default. You can choose a longer or shorter duration. If it should be indefinite, do not specify a duration argument.
7. The name of the created function should only use phrasal verbs, verbs, nouns, or adverbs shown in the prompt and should be in the verb+noun or verb+adverb format, such as drink_water, slow_down_car, and ride_faster. Note that words that do not show in the prompt are prohibited.

This is Example 1. If "press" is in the prompt and the text prompt on the screenshot is "press X to play the card", your output should be:
```python
def play_card():
    """
    press "x" to play the card
    """
    io_env.key_press('x', 2)
```
This is Example 2. If the instructions involve the mouse and the text prompt on the screenshot is "use the left mouse button to confirm", your output should be:
```python
def confirm():
    """
    use "left mouse button" to confirm
    """
    io_env.mouse_click("left mouse button")
```
This is Example 3. If "use" is in the prompt and the text prompt on the screenshot is "use ENTER to drink water", your output should be:
```python
def drink_water():
    """
    use "enter" to drink water
    """
    io_env.key_press('enter', 2)
```
This is Example 4. If "use" is in the prompt and the text prompt on the screenshot is "use W and J to jump the barrier", your output should be:
```python
def jump_barrier():
    """
    use "w" and "j" to jump the barrier
    """
    io_env.key_press('w,j', 3)
```
This is Example 5. If "hold" is in the prompt and the text prompt on the screenshot is "hold H to run", your output should be:
```python
def run():
    """
    hold "h" to run
    """
    io_env.key_hold('h')
```
This is Example 6. If the instructions involve the mouse and the text prompt on the screenshot is "hold the right mouse button to focus on the target", your output should be:
```python
def focus_on_target():
    """
    hold "right mouse button" to focus
    """
    io_env.mouse_hold("right mouse button")
```
This is Example 7. If "release" is in the prompt and the text prompt on the screenshot is "release Q to drop the items", your output should be:
```python
def drop_items():
    """
    release "q" to drop the items
    """
    io_env.key_release('q')
```

Dialogue: Conversations between characters in the game. This content should be in the format of "character name: dialogue". For example, "Arthur: I'm fine". If there is no on-screen text or dialogue, only output "null".

Other: Other information that does not belong to the above categories. If there is no on-screen text, only output "null".

Reasoning: The reasons for classification for each piece of information.
If the on-screen text prompt is an instruction on how to perform a task using keyboard keys or mouse buttons, it should also classified as action guidance and task guidance.
For action guidance, which code rules should you follow based on the word used to control the key or button, such as press, hold, release, and click?

The information should be in the following categories, and you should output the following content without adding any other explanation:
Information:
1. ...
2. ...
...
Reasoning:
1. ...
2. ...
...
Item_status:
Item_status is ...
Environment_information:
Environment information is ...
Notification:
Notification is ...
Task_guidance:
Task is ...
Action_guidance:
```python
Python code to execute
```
```python
Python code to execute
```
...
Dialogue:
Dialogue is ...
Other:
Other information is ...
\end{lstlisting}

\begin{lstlisting}[breaklines=true,caption={RDR2: Self-Reflection prompt.}]
Assume you are a helpful AI assistant integrated with 'Red Dead Redemption 2' on the PC, equipped to handle a wide range of tasks in the game. Your advanced capabilities enable you to process and interpret gameplay screenshots and other relevant information. Your task is to examine these inputs, interpret the in-game context, and determine whether the executed action takes effect. 

Current task:
<$task_description$>

Last executed action:
<$previous_action$>

Implementation of the last executed action:
<$action_code$>

Error report for the last executed action:
<$executing_action_error$>

Reasoning for the last action:
<$previous_reasoning$>

Valid action set in Python format to select the next action:
<$skill_library$>

<$image_introduction$>

Reasoning: You need to answer the following questions step by step to get some reasoning based on the last action and sequential frames of the character during the execution of the last action.
1. What is the last executed action not based on the sequential frames?
2. Was the last executed action successful? Give reasons. You should refer to the following rules:
- If the action involves moving forward, it is considered unsuccessful only when the character's position remains unchanged across sequential frames, regardless of background elements and other people.
3. If the last action is not executed successfully, what is the most probable cause? You should give only one cause and refer to the following rules:
- The reasoning for the last action could be wrong.
- Not holding enough time should not be considered in this part.
- If it is an interaction action, the most probable cause was that the action was unavailable or not activated at the current place.
- If it is a movement action, the most probable cause was that you were blocked by seen or unseen obstacles.
- If there is an error report, analyze the cause based on the report.

You should only respond in the format as described below:
Reasoning:
1. ...
2. ...
3. ...
...
\end{lstlisting}

\begin{lstlisting}[breaklines=true,caption={RDR2: Task Inference prompt.}]
Assume you are a helpful AI assistant integrated with 'Red Dead Redemption 2' on the PC, equipped to handle a wide range of tasks in the game. You will be sequentially given <$event_count$> screenshots and corresponding descriptions of recent events. You will also be given a summary of the history that happened before the last screenshot. You should assist in summarizing the events for future decision-making.

The following are <$event_count$> successive screenshots and corresponding descriptions:

<$image_introduction$>

The following is the summary of history that happened before the last screenshot:
<$previous_summarization$>

Current task:
<$task_description$>

Info_summary: Based on the above input, please make a summary from the screenshots with descriptions and the history in no less than 10 sentences, following the rules below.
1. Summarize the tasks from the history and the current task, with a special note on the method of crucial press operations.
2. Summarize the entities and behaviors mentioned in the successive descriptions.
3. If entities and behaviors in the history and screenshots are missed in the descriptions, please add them to the summarization.  
4. Organize the summarization as a story in order of time, including the past entities and behaviors.
5. Only give descriptions; do not provide suggestions.

Entities_and_behaviors: Entities and behaviors which are summarized, e.g., The entities include the player's character, the target character, and horses for both the player and the target. The behaviors consist of the player character riding horseback, following the target on horseback, and moving forward to maintain a distance behind the target.

The output should be in the following format:
Info_summary:
The summary is...
Entities_and_behaviors:
The summary is...
\end{lstlisting}

\begin{lstlisting}[breaklines=true,caption={RDR2: Action Planning prompt.}]
You are a helpful AI assistant integrated with 'Red Dead Redemption 2' on the PC, equipped to handle various tasks in the game. Your advanced capabilities enable you to process and interpret gameplay screenshots and other relevant information. By analyzing these inputs, you gain a comprehensive understanding of the current context and situation within the game. Utilizing this insight, you are tasked with identifying the most suitable in-game action to take next, given the current task. You control the game character and can execute actions from the available action set. Upon evaluating the provided information, your role is to articulate the precise action you would deploy, considering the game's present circumstances, and specify any necessary parameters for implementing that action.

Here is some helpful information to help you make the decision.

Current task:
<$task_description$>

Memory examples:
<$memory_introduction$>

<$few_shots$>

<$image_introduction$>

Last executed action:
<$previous_action$>

Reasoning for the last action:
<$previous_reasoning$>

Self-reflection for the last executed action:
<$previous_self_reflection_reasoning$>

Summarization of recent history:
<$info_summary$>

Valid action set in Python format to select the next action:
<$skill_library$>

Minimap information:
<$minimap_information$>

Based on the above information, you should first analyze the current situation and provide the reasoning for what you should do for the next step to complete the task. Then, you should output the exact action you want to execute in the game. You should respond to me with:

Reasoning: You should think step by step and provide detailed reasoning to determine the next action executed on the current state of the task. You need to answer the following questions step by step. You cannot miss the question number 13:
    1. Only answer this question when the radial menu, trade, map, satchel or inventory interfaces are open. You should first describe each item in the screen line by line, from the top left and moving right. Is the target item in the current screen?
    2. Only answer this question when the radial menu, trade, map, satchel or inventory interfaces are open. Which item is selected currently?
    3. Only answer this question when the character is visible in the screenshot of the current step. Where is the character in the screenshot of the current step?
    4. Where is the target in the screenshot of the current step based on the task description, on the left side or on the right side? Does it appear in the previous screenshots?
    5. Are there any bounding boxes with coordinates values and object labels, such as "door x = 0.5, y = 0.5", shown in the screenshot? The answer must be based only on the screenshot of the current step, not on any previous steps. If the answer is no, ignore the questions 6 to 8.
    6. You should first describe each bounding box, from left to right. Which bounding box is more relevant to the target? 
    7. What is the value x of the most relevant bounding box only in the current screenshot? The value is the central coordination (x,y) of the central point of the box.
    8. Based on the few shots and the value x, where is the relevant bounding box in the current screenshot? Clearly on the left side, slightly on the left side, in the center, slightly on the right side, or clearly on the right side?
    9. Only answer this question when the radial menu, trade, map, satchel or inventory interfaces are not open. Summarize the contents of recent history, mainly focusing on the historical tasks and behaviors.
    10. Only answer this question when the radial menu, trade, map, satchel or inventory interfaces are not open. Summarize the content of self-reflection for the last executed action, and do not be distracted by other information.
    11. What was the previous action? If the previous action was a turn, was it a left or a right turn? If the previous action was a movement, were you blocked? 
    12. List conditions in action rule 12 and which condition is satisfied. Only when you do not satisfy any conditions, summarize the content of the minimap information. 
    13. This is the most critical question. Based on the action rules and self-reflection, what should be the most suitable action in the valid action set for the next step? You should analyze the effects of the action step by step.

Actions: The best action, or short sequence of actions without gaps, to execute next to progress in achieving the goal. Pay attention to the names of the available skills and to the previous skills already executed, if any. You should also pay more attention to the following action rules:
    1. You should output actions in Python code format and specify any necessary parameters to execute that action. If the function has parameters, you should also include their names and decide their values, like "move(duration=1)". If it does not have a parameter, just output the action, like "mount_horse()".
    2. Given the current situation and task, you should only choose the most suitable action from the valid action set. You cannot use actions that are not in the valid action set to control the character.
    3. If the target is not on the radial menu, trade, satchel or inventory interfaces, you MUST choose the skill 'view_next_page'. For the map, ignore the skill 'view_next_page'.
    4. If the minimap information exists, it may include angle information for red points, yellow points, or yellow regions. Angle information specifies the direction of the corresponding point or area. A negative angle indicates the left side, while a positive value signifies the right side. If the angle is 30, the corresponding point or area is 30 degrees to the character's right. If the angle is -50, the corresponding point or area is 50 degrees to the character's left. Do not doubt the correctness of these angles; you can refer to them when you approach these points or regions.
    5. When you decide to control the character to move, if the relevant bounding box is clearly on the left side in the current screenshot, you MUST turn left with a big degree. If the relevant bounding box is slightly on the left side in the current screenshot, you MUST turn left with a small degree. If the relevant bounding box is clearly on the right side in the current screenshot, you MUST turn right with a big degree. If the relevant bounding box is slightly on the right side in the current screenshot, you MUST turn right with a small degree. If the relevant bounding box is on the central side of the current screenshot, you can choose to move forward.
    6. When you decide to control the character to move, if yellow regions or yellow points exist in minimap information, they are related to the current task or instruction. This implies that you should approach within the yellow region or approach the yellow points. You can refer to the corresponding angle information when deciding to approach these regions or points. If red points exist in the minimap information, they are also related to the current task or instruction. This implies that you should turn towards them, and you can also refer to the corresponding angle information. 
    7. When you decide to control the character to move, if minimap information does not exist, the 'theta' you use to turn MUST be more than 10 degrees and less than 60 degrees.
    8. When you decide to control the character to move, if you are in a normal road condition, the 'duration' you use to move forward should be 1 second. If you have bad road conditions, such as snow, and grass, that can slow you down, the 'duration' you use to move forward should be 2 seconds.
    9. When you are exploring or searching a place, if you are leaving the place, you MUST make a sharp turn to face the inside of the place. Any values for degrees are allowed.
    10. If upon self-reflection you think the last action was unavailable at the current place, you MUST move to another place.
    11. If upon self-reflection you think you were blocked, you MUST make a moderate turn in the same direction as the previous turn action and move forward, so that you can pass obstacles. 
    12. The conditions to ignore the minimap information for decision-making are: 1. When self-reflection implies you were blocked. 2. When you were inside the highlighted area in the minimap. If any of the conditions satisfied, you must ignore the minimap information for decision-making even if it is relevant to the current task.
    13. When you are indoors, or the current task does not imply following, you MUST not use the follow action.
    14. When you are outdoors, and the current task implies following, you MUST use the follow action.
    15. If you were dead or the game failed, you MUST retry from the checkpoint, and MUST NOT restart the mission.

You should only respond in the format described below, and you should not output comments or other information:
Reasoning:
1. ...
2. ...
3. ...
Actions:
```python
    action(args1=x,args2=y)
```
\end{lstlisting}

\subsection{Prompts for Cities: Skylines}

\begin{lstlisting}[breaklines=true,caption={Skylines: Information Gathering prompt.}]
Assume you are a helpful AI assistant integrated with 'Cities: Skylines' on the PC, equipped to handle a wide range of tasks in the game. Your advanced capabilities enable you to process and interpret gameplay screenshots and other relevant information.

<$image_introduction$>

Current task:
<$task_description$>

Description: Please analyze and describe the screenshot image in detail and then provide an overall image description. Pay attention to anything related to the task. If there are specific features such as characters or text, mention these as well.

Budget: Bank Balance is shown at the bottom of the screenshot.

Population: The population of the city is shown at the bottom of the screenshot, next to the budget.

Error_message: If there are some in-game error messages, which are usually in red color, such as "Space already occupied!", extract the text, otherwise, only output "null". 

Construction_information: If there is some in-game construction information, which is usually in blue colors, such as "Construction cost: 2500 Estimated production:0 m^3/week" and "Construction cost: 2500 Shoreline recommended", extract the text, otherwise, only output "null". 

Other: Other information that does not belong to the above categories. If none of them applies, only output "null".

You should only respond in the format described below and not output comments or other information.
Description:
The image shows...
Budget:
The amount of budget 
Population:
The amount of population
Error_message:
The text of the error message
Construction_information:
The text of the construction information
Other:
Other information is
\end{lstlisting}

\begin{lstlisting}[breaklines=true,caption={Skylines: Self-Reflection prompt.}]
Assume you are a helpful AI assistant integrated with 'Cities: Skylines' on the PC, equipped to handle a wide range of tasks in the game. Your advanced capabilities enable you to process and interpret gameplay screenshots and other relevant information. Your task is to examine these inputs, interpret the in-game context, and determine whether the executed action takes effect.

Target task:
<$task_description$>

Current subtask for completing the target task:
<$subtask_description$>

Current coordinates:
<$coordinates$>

Last executed action for completing the subtask:
<$actions$>

Error message for the last executed action:
<$error_message$>

Construction information:
<$construction_information$>

Summarization of recent history:
<$history_summary$>

<$image_introduction$>

Reasoning: You MUST answer the following questions step by step to get some reasoning based on the last action and sequential frames during the execution of the last action. 
1. What is the executed action? Please answer this question not based on the sequential frames.
2. Is the construction information provided in the information shown above? If yes, what is it?
3. Was the last executed action successful? Give reasons. You should refer to the following rules:
- Buildings and roads cannot be built on the river.
- Water pumping station and water drain pipe need to be built as close as possible to the river.
- If you are try_place a water pumping station and the construction information provided above shows that the estimated production is 0 m^3/week, then it means that it is not close enough to the river. So you need to try_place to place the building to another place. If the estimated production is not 0 m^3/week, or the construction information is not provided, regard this action as a success. You should only refer to the textual construction information instead of extracting it from the sequential frames.
- If you are try_place a water drain pipe and the construction information shows that shoreline is recommended. Then it means that it is not close enough to the river. So you need to try_place to place the building in another place. 
- Roads are prohibited from crossing together and do not build roads on water.
4. If the last action is not executed successfully, what is the most probable cause? How to improve this action? You should give only one cause and refer to the following rules:
- The reasoning for the last action could be wrong.
- If there is an error message for the last executed action provided in the above information, analyze the cause based on the report, otherwise, you should regard that there are no error messages. You are not allowed to guess the error message by yourself.
5. Is the subtask completed? Give your reasons. You MUST remember that action starts with "try_place" can NEVER complete the subtask. Only "confirm_placement()" can make the building happen and complete the task. If you want to make any confirmation, regard it as a success.
6. Do you think the subtask is reasonable? Give your reasons.

Success: You need to output whether the last action was executed successfully or not.
- If the last action is successful, you should only output 'True'. Otherwise, you should only output 'False'.

You should only respond in the format described below.
Reasoning:
1. ...
2. ...
3. ...
4. ...
5. ...
6. ...
...
Success:
True
...
\end{lstlisting}

\begin{lstlisting}[breaklines=true,caption={Skylines: Task Inference prompt.}]
Assume you are a helpful AI assistant integrated with 'Cities: Skylines' on the PC, equipped to handle a wide range of tasks in the game. You will also be given a summary of the history that happened before the last screenshot. You should assist in summarizing the events for future decision-making and also propose a new subtask, which is the most suitable subtask for the current situation, given the target task.

Here is some helpful information to help you do the summarization and propose the subtask.

Current task:
<$task_description$>

Previous proposed subtask for the task:
<$subtask_description$>

Previous reasoning for proposing the subtask:
<$subtask_reasoning$>

<$image_introduction$>

Current budget:
<$budget$>

Current population:
<$population$>

Last executed action:
<$actions$>

Self-reflection for the last executed action:
<$self_reflection_reasoning$>

Error message for the last action:
<$error_message$>

The following is the summary of history that happened before the last screenshot:
<$previous_summarization$>

The task can be decomposed into the following subtasks:
1. Start from the Highway entry: Build a road from the highway entry in grid (4, 2) vertically northwards towards grid (3,1).
2. Extend Horizontally to the Left (1,1): From the endpoint in grid (1,1), construct a road horizontally to the left, spanning across grids (3,1) and (2,1), and ending at the center of grid (1,1).
3. Build a Road Down to the bottom of Grid (2,2): Start from grid (1,1) and construct the road to the top of grid (2, 3).
4. Extend Eastward to Grid (3,3): From the bottom of grid (2,2), build a road eastward to reach the center of grid (3,3).
5. Connect the road to the Highway Exit: Extend the end of the road from grid (3,3) to the exit of the highway, completing the road loop.
6. Install a Water Pumping Station near the River at the top-left corner of grid (2,3): Place the water pumping station near the river in grid (2,3) to ensure an adequate water supply.
7. Position a Water Drain Pipe near the River at the top-left corner of grid (2,3): Install a water drain pipe slightly downstream from the pumping station but within the same grid to prevent water contamination.
8. Lay Water Pipes: Connect the water pumping station to the water drain pipe using water pipes. Additionally, ensure all roads built are covered with water pipes to provide water access across the entire area.
9. Erect Wind Turbines for Power: Construct several wind turbines near the water pumping station and along the roads to provide sustainable electricity to the area.
10. Designate Residential Zones: Allocate spaces adjacent to the roads for residential zones to foster community living.
11. Establish Industrial Zones: Set aside areas near the roads for industrial purposes, ideally in parts of the grid further from residential zones to manage noise and pollution.
12. Create Commercial Zones: Develop commercial zones near the roads to provide services and retail options for the residents and workers in the area.
13. Make sure all the zones near roads are built with Residential Zones, Industrial Zones or Industrial Zones.
14. Build more roads and zones and ensure water and electricity supply.

History_summary: Summarize what happened in the past experience, especially the last step according to the decision-making reasoning and self-reflection reasoning for the last executed action. The summarization needs to be precise, concrete and highly related to the task and follow the rules below.
1. Summarize the tasks from the history and the current task. What is the current progress of the task?
2. Which subtask has been completed? Which subtasks are not?

Subtask_reasoning: According to the task decomposition, analyze the current progress step by step and then decide whether the previous subtask is finished and whether it is necessary to propose a new subtask. The subtask should be straightforward, contribute to the target task and be most suitable for the current situation, which should be completed within a few actions. You should respond to me with:
1. What is the previous subtask? Which step it is for in the task decomposition?
2. According to the reasoning of self-reflection, is the previous subtask completed? Note that the success of the action does not mean the success of the subtask. You should strictly follow the reasoning of whether the subtask is completed in the self-reflection. If yes, you should move to the next step and propose it as the new subtask. If not, you should continue the previous subtask without changing anything.  Please do not make any assumptions if they are not mentioned in the above information. You should assume that you are doing the task from scratch. Please strictly follow the description and requirements in the current task. 
3. The proposed subtask needs to be precise and concrete within one sentence. It should not be related to any skills. 
4. To enable water supply, you should first build a water pumping station and then build a water drain pipe near the river, and finally use water pipes to connect them with the roads. And ensure the water pipes cover all the roads.
5. The water pumping station and water drain pipe also need electricity to work. So you also need to provide electricity for them.
6. If you want to build roads for the village at the beginning, make sure to mention that the road needs to be as long as possible and use several roads to form a large square for the village.

Subtask: According to the subtask reasoning, determine and output the most suitable subtask for the current situation. You MUST output the subtask in the output.

You should only respond in the format described below, and you should not output comments or other information.
History_summary:
The summary is ...
Subtask_reasoning:
1. ...
2. ...
3. ...
Subtask:
The current subtask is ...
\end{lstlisting}

\begin{lstlisting}[breaklines=true,caption={Skylines: Action Planning prompt.}]
You are a helpful AI assistant integrated with 'Cities: Skylines' on the PC, equipped to handle various tasks in the game. Your advanced capabilities enable you to process and interpret gameplay screenshots and other relevant information. By analyzing these inputs, you gain a comprehensive understanding of the current context and situation within the game. Utilizing this insight, you are tasked with identifying the most suitable in-game action to take next, given the current task. You control the game character and can execute actions from the available action set. Upon evaluating the provided information, your role is to articulate the precise action you would deploy, considering the game's present circumstances, and specify any necessary parameters for implementing that action.

Here is some helpful information to help you make the decision.

Current task:
<$subtask_description$>

Coordinates of constructed buildings:
<$coordinates$>

The latest successful action that builds the building. If you want to try_place a road, and the endpoint (x2, y2), of the latest successful action is also try_place a road. Then you MUST use the end point of the constructed road as the start point of your new road. 
<$last_success_try_place_action$>

Current budget:
<$budget$>

Current population:
<$population$>

Last executed action:
<$actions$>

Self-reflection reasoning for the last executed action:
<$self_reflection_reasoning$>

Error message for the last action:
<$error_message$>

Construction information for the last action:
<$consruction_information$>

Summarization of recent history:
<$history_summary$>

Valid action set in Python format to select the next action:
<$skill_library$>

<$image_introduction$>

Based on the above information, analyze the current situation and provide the reasoning for what you should do for the next step to complete the task. Then, you should output the exact action you want to execute in the game. You should respond to me with:

Reasoning: You should think step by step and provide detailed reasoning to determine the next action executed on the current state of the task. You need to answer the following questions step by step. You cannot miss the last question:
    1. What is the current task? What are the requirements to achieve the goal?
    2. According to the self-reflection reasoning, is the last action executed successfully?
    3. If you want to place anything, do you already open the corresponding menu? Otherwise, you need to open the right menu first in this step rather than doing anything else. If you have not already opened the corresponding menu, skip answering questions 4, 5, 6, 7, 8 and 9.
    4. Does the previous action "try_place" something? If there is an error message showing that the space is already occupied or the last action failed according to the self-reflection reasoning, you should use the same action with different parameters as the position of it to try again. The difference needs to be significant enough with at least 100 pixels of change for the position of the input points. If there is no error message, you should only output confirm_placement() or cancel_placement() to approve or cancel the placement. You should not call anything else.
    5. Does the previous action open any menu? Then you should "try_place" something according to the task description instead of using "confirm_placement". 
    6. If you want to place a building, which grid do you plan to place the building in? What is the exact pixel position of it?
    7. If you want to place a road, which grids do you plan to make it cross? Which grids are the start point and end point in, respectively? What are the exact pixel positions of them? You MUST use one of the endpoints of the constructed road shown in the coordinates information as the start point of the new road. If you want to try_place a road, and the endpoint (x2, y2), of the latest successful action is also try_place a road. Then you MUST use the end point of the constructed road as the start point of your new road. 
    8. If you want to place a zone, which grids do you plan to make it cover? You should only use the vertices coordinates of the corresponding grids as the parameter for the action. Zones cannot cover each other. 
    9. If you want to place a Water Pipe, the start point should be the position of Water Pumping Station, Water Drain Pipe, the start point of a built Water Pipe or the end point of a built Water Pipe.
    10. This is the most critical question. Based on the action rules and self-reflection, what should be the most suitable action in the valid action set for the next step? You should analyze the effects of the action step by step. You should not repeat the previous action again. Do not try to verify whether the previous action succeeded.
    11. Do all the selected actions exist in the valid action set? If no, regenerate the action and give the reasons.
    12. If you are placing a road, is the road more than 300 pixels long? Otherwise, regenerate the action and give reasons.

Actions: The requirements that the generated action needs to follow. The best action, or short sequence of actions without gaps, to execute next to progress in achieving the goal. Pay attention to the names of the available skills and to the previous skills already executed, if any. You should also pay more attention to the following action rules:
    1. You should output actions in Python code format and specify any necessary parameters to execute that action. If the function has parameters, you should also include their names and decide their values, like "move_right(duration=1)". If it does not have a parameter, just output the action, like "open_map()".
    2. Given the current situation and task, you should only choose the most suitable action from the valid action set. You cannot use actions that are not in the valid action set to control the character.
    3. You MUST NOT output more than one skill in the actions.
    4. If you want to build a village, you should follow these rules:
     4.1 Build roads correctly.
      - If you have not opened the road tool, you should open the menu. If you have already opened the menu, you should not open it again.
      - Newly built roads must be connected to the existing roads.
      - Determine in which grid the starting point of the newly built road is located, and identify the pixel position of the starting point.
      - Build the road in the correct direction.
    5. You MUST NOT repeat the previous action with the same parameters again if you think the previous action fails.
    6. Your action should strictly follow the analysis in the reasoning. Do not output any additional action not mentioned in the reasoning.
    7. Please do not directly connect the entrance of the highway with the exit of the highway at the beginning. To make the village as large as possible. You should build roads in the wild and connect them with each other.
    8. If you are placing a road, the road needs to be at least 300 pixels long.

You should only respond in the format described below, and you should not output comments or other information.
Reasoning:
1. ...
2. ...
3. ...
...
Actions:
```python
    action(args1=x,args2=y)
```
\end{lstlisting}

\subsection{Prompts for Stardew Valley}

\begin{lstlisting}[breaklines=true,caption={Stardew: Information Gathering Cultivation prompt.}]
Assume you are a helpful AI assistant integrated with 'Stardew Valley' on the PC, equipped to handle a wide range of tasks in the game. Your advanced capabilities enable you to process and interpret gameplay screenshots and other relevant information.

<$image_introduction$>

Current task:
<$task_description$>

Description: Please analyze and describe the screenshot image in a grid-by-grid format and then provide an overall image description. Pay attention to anything related to the task. The image is divided into a 3x5 grid, each cell having its own coordinates. For each grid cell, describe the contents in detail, focusing on any critical icons, or objects present in that particular segment. If there are specific features such as characters or text, mention these as well. After completing the description for one cell, proceed to the next, for example, 'In grid (1,1), [description]. In grid (1,2), [description].' and so on until the entire image is covered.

Date_time: The date and time information in the game are shown on the upper-right of the screenshot, in grid (1, 5). An example of the date and time information is "Wed 10, 5:10 pm".

Energy: The current energy remains for the character doing actions. The energy bar is shown on the bottom-right of the screenshot, in grid (3, 5). The full energy is 270. An example of the energy information is "150/270".

Weather: The current weather information in the game, the weather is one from "Sunny", "Rainy", "Windy", "Snowy", "Stormy", "Festival", "Wedding", and "null". If none of them applies, only output "null".

Dialog: If there are some dialogs shown in the screenshot, extract the text of the conversation, like "Shopkeeper: What do you want to buy?", otherwise, only output "null".

Other: Other information that does not belong to the above categories. If none of them applies, only output "null".

You should only respond in the format described below and not output comments or other information.
Description:
In grid (1,1), ...
In grid (1,2), ...
...
In grid (3,5), ... 
Overall, the image shows...
Date_time:
Date and time information
Energy:
The number of energy remains showing in the energy bar
Weather:
Weather information
Dialog:
Dialog text
Other:
Other information is ...
\end{lstlisting}

\begin{lstlisting}[breaklines=true,caption={Stardew: Self-Reflection Cultivation prompt.}]
Assume you are a helpful AI assistant integrated with 'Stardew Valley' on the PC, equipped to handle a wide range of tasks in the game. Your advanced capabilities enable you to process and interpret gameplay screenshots and other relevant information. Your task is to examine these inputs, interpret the in-game context, and determine whether the executed action takes effect.

Target task:
<$task_description$>

Current subtask for completing the target task:
<$subtask_description$>

The reasoning for proposing the current subtask:
<$subtask_reasoning$>

Last executed action for completing the subtask:
<$previous_action$>

Reasoning for the last action:
<$previous_reasoning$>

Current date and time:
<$date_time$>

Previous toolbar information:
<$previous_toolbar_information$>

Current toolbar information:
<$toolbar_information$>

Summarization of recent history:
<$history_summary$>

<$image_introduction$>

Reasoning: You need to answer the following questions step by step to get some reasoning based on the last action and sequential frames of the character during the execution of the last action.
1. What is the executed action? Please answer this question not based on the sequential frames.
2. Was the executed action successful? Give reasons. You should refer to the following rules:
- If the action involves moving forward, it is considered unsuccessful only when the character's position remains unchanged across sequential frames, regardless of background elements and other people.
- If you are not 100% sure that the action fails, regard it as success. 
3. If the last action is not executed successfully, what is the most probable cause? You should give only one cause and refer to the following rules:
- The reasoning for the last action could be wrong.
- If it is an interaction action, the most probable cause was that the action was unavailable at the current place, then you should move to a new place.
- If it is a movement action, the most probable cause was that you were blocked by seen or unseen obstacles.
- If there is an error report, analyze the cause based on the report.
4. Is the subtask completed? Give your reasons. If you want to make any confirmation, regard it as a success.
5. Is the target task completed? Give your reasons.
6. Do you think the subtask is reasonable? Give your reasons.

You should only respond in the format described below.
Reasoning:
1. ...
2. ...
3. ...
...
\end{lstlisting}

\lstinputlisting[breaklines=true,caption={Stardew: Task Inference Cultivation prompt.}]{prompt/stardew_task_inference_cultivation_prompt.txt}

\lstinputlisting[breaklines=true,caption={Stardew: Action Planning Cultivation prompt.}]{prompt/stardew_action_planning_cultivation_prompt.txt}

\begin{lstlisting}[breaklines=true,caption={Stardew: Information Gathering Farm Clearup prompt.}]
Assume you are a helpful AI assistant integrated with 'Stardew Valley' on the PC, equipped to handle a wide range of tasks in the game. Your advanced capabilities enable you to process and interpret gameplay screenshots and other relevant information.

<$image_introduction$>

Current task:
<$task_description$>

Description: Please analyze and describe the screenshot image in a grid-by-grid format and then provide an overall image description. Pay attention to anything related to the task. The image is divided into a 3x5 grid, each cell having its own coordinates. For each grid cell, describe the contents in detail, focusing on any critical icons, or objects present in that particular segment. If there are specific features such as characters or text, mention these as well. After completing the description for one cell, proceed to the next, for example, 'In grid (1,1), [description]. In grid (1,2), [description].' and so on until the entire image is covered.

Date_time: The date and time information in the game are shown on the upper-right of the screenshot, in grid (1, 5). An example of the date and time information is "Wed 10, 5:10 pm".

Energy: The current energy remains for the character doing actions. The energy bar is shown on the bottom-right of the screenshot, in grid (3, 5). The full energy is 270. An example of the energy information is "150/270".

Weather: The current weather information in the game, the weather is one from "Sunny", "Rainy", "Windy", "Snowy", "Stormy", "Festival", "Wedding", and "null". If none of them applies, only output "null".

Dialog: If there are some dialogs shown in the screenshot, extract the text of the conversation, like "Shopkeeper: What do you want to buy?", otherwise, only output "null".

Other: Other information that does not belong to the above categories. If none of them applies, only output "null".

You should only respond in the format described below and not output comments or other information.
Description:
In grid (1,1), ...
In grid (1,2), ...
...
In grid (3,5), ... 
Overall, the image shows...
Date_time:
Date and time information
Energy:
The number of energy remains showing in the energy bar
Weather:
Weather information
Dialog:
Dialog text
Other:
Other information is ...
\end{lstlisting}

\lstinputlisting[breaklines=true,caption={Stardew: Self-Reflection Farm Clearup prompt.}]{prompt/stardew_self_reflection_farm_clearup_prompt.txt}

\lstinputlisting[breaklines=true,caption={Stardew: Task Inference Farm Clearup prompt.}]{prompt/stardew_task_inference_farm_clearup_prompt.txt}

\lstinputlisting[breaklines=true,caption={Stardew: Action Planning Farm Clearup prompt.}]{prompt/stardew_action_planning_farm_clearup_prompt.txt}

\lstinputlisting[breaklines=true,caption={Stardew: Information Gathering Shopping prompt.}]{prompt/stardew_information_gathering_shopping_prompt.txt}

\lstinputlisting[breaklines=true,caption={Stardew: Self-Reflection Shopping prompt.}]{prompt/stardew_self_reflection_shopping_prompt.txt}

\lstinputlisting[breaklines=true,caption={Stardew: Task Inference Shopping prompt.}]{prompt/stardew_task_inference_shopping_prompt.txt}

\lstinputlisting[breaklines=true,caption={Stardew: Action Planning Shopping prompt.}]{prompt/stardew_action_planning_shopping_prompt.txt}

\subsection{Prompts for Dealer's Life 2}

\lstinputlisting[breaklines=true,caption={Dealer's Life 2: Information Gathering prompt.}]{prompt/dealer_information_gathering_prompt.txt}

\lstinputlisting[breaklines=true,caption={Dealer's Life 2: Self Reflection prompt.}]{prompt/dealer_self_reflection_prompt.txt}

\lstinputlisting[breaklines=true,caption={Dealer's Life 2: Task Inference prompt.}]{prompt/dealer_task_inference_prompt.txt}

\lstinputlisting[breaklines=true,caption={Dealer's Life 2: Action Planning prompt.}]{prompt/dealer_action_planning_prompt.txt}

\subsection{Prompts for Software Applications}
\lstinputlisting[breaklines=true,caption={Chrome: Information Gathering prompt.}]{prompt/01_Chrome_gather_information_prompt.txt}
\lstinputlisting[breaklines=true,caption={Chrome: Self-Reflection prompt.}]{prompt/01_Chrome_self_reflection_prompt.txt}

\lstinputlisting[breaklines=true,caption={Chrome: Task Inference prompt.}]{prompt/01_Chrome_information_summary_prompt.txt}

\lstinputlisting[breaklines=true,caption={Chrome: Action Planning prompt.}]{prompt/01_Chrome_decision_making_prompt.txt}

\lstinputlisting[breaklines=true,caption={Outlook: Information Gathering prompt.}]{prompt/02_Outlook_gather_information_prompt.txt}
\lstinputlisting[breaklines=true,caption={Outlook: Self-Reflection prompt.}]{prompt/02_Outlook_self_reflection_prompt.txt}
\lstinputlisting[breaklines=true,caption={Outlook: Task Inference prompt.}]{prompt/02_Outlook_information_summary_prompt.txt}
\lstinputlisting[breaklines=true,caption={Outlook: Action Planning prompt.}]{prompt/02_Outlook_decision_making_prompt.txt}

\lstinputlisting[breaklines=true,caption={Capcut: Information Gathering prompt.}]{prompt/03_Capcut_gather_information_prompt.txt}
\lstinputlisting[breaklines=true,caption={Capcut: Self-Reflection prompt.}]{prompt/03_Capcut_self_reflection_prompt.txt}
\lstinputlisting[breaklines=true,caption={Capcut: Task Inference prompt.}]{prompt/03_Capcut_information_summary_prompt.txt}
\begin{lstlisting}[breaklines=true,caption={Capcut: Screen Classification prompt.}]
You are an assistant who assesses my progress in playing Red Dead Redemption 2 on the PC and provides expert guidance. Imagine you are playing Red Dead Redemption 2 with the keyboard and mouse, the image is the screenshot of your computer.

Given the classes, please select the class that best describes the screenshot.
<classes>

You must follow the following criteria:
(1) The output should only be a JSON file. You should not add any other explanation text along with the JSON.
(2) You should choose one class for the value of "class".
(3) Do not change the "type": "screen_classification" in your output.

The output format should be as follows:
Classes:
map
\end{lstlisting}
\lstinputlisting[breaklines=true,caption={Capcut: Action Planning prompt.}]{prompt/03_Capcut_decision_making_prompt.txt}

\lstinputlisting[breaklines=true,caption={Meitu: Information Gathering prompt.}]{prompt/04_Xiuxiu_gather_information_prompt.txt}
\lstinputlisting[breaklines=true,caption={Meitu: Self Reflection prompt.}]{prompt/04_Xiuxiu_self_reflection_prompt.txt}
\lstinputlisting[breaklines=true,caption={Meitu: Task Inference prompt.}]{prompt/04_Xiuxiu_information_summary_prompt.txt}
\lstinputlisting[breaklines=true,caption={Meitu: Action Planning prompt.}]{prompt/04_Xiuxiu_decision_making_prompt.txt}

\lstinputlisting[breaklines=true,caption={Feishu: Information Gathering prompt.}]{prompt/05_Feishu_gather_information_prompt.txt}
\lstinputlisting[breaklines=true,caption={Feishu: Self Reflection prompt.}]{prompt/05_Feishu_self_reflection_prompt.txt}
\lstinputlisting[breaklines=true,caption={Feishu: Task Inference prompt.}]{prompt/05_Feishu_information_summary_prompt.txt}
\lstinputlisting[breaklines=true,caption={Feishu: Action Planning prompt.}]{prompt/05_Feishu_decision_making_prompt.txt}

\end{document}

%% file: appendix/general_implementation.tex
\label{appx:general_implementation}
Here we introduce the general implementation details of \projname. For specialized implementations addressing issues unique to their own environment, please refer to the corresponding section.

\textbf{Backbone Model.} We employ GPT-4o~\citep{openaigpt4o}, currently one of the most capable LMM models, as the framework's backbone model. If not mentioned explicitly, all the experiments are done with \emph{gpt-4o-2024-05-13}. Temperature is set to 0 to lower the variance of the text generation.

\textbf{Observation Space.} \projname only takes a video clip, which records the progress of execution of the last action, as input. To lower the frequency of interaction with backbone models and reduce the strain on the computer, video is recorded at 2 fps (a screenshot every 0.5 seconds), which proves to be sufficient in most cases for information gathering without missing any important information.

\textbf{Action Space.}  For the action space, it includes all possible keyboard and mouse operations, including \texttt{key\_press}, \texttt{key\_hold}, \texttt{key\_release}, \texttt{mouse\_move}, \texttt{mouse\_click}, \texttt{mouse\_hold}, \texttt{mouse\_release}, and \texttt{wheel\_scroll}, which can be combined in different ways to form combos and shortcuts, use keys in fast sequence, or coordinate timings. We choose to use Python code to simulate these operations and encapsulate them into an \texttt{io\_env} class. Skill code needs to be generated by the agent in order to utilize such functions and affordances so executed actions take effect. Table~\ref{tab:action_space} illustrates \projname's action space.

\begin{table}[ht]
\centering
\caption{Action space in the \projname framework, including action attributes. Coordinate system is either \emph{absolute} or \emph{relative}. Actions with durations can be either \emph{synchronous} or \emph{asynchronous}.}\label{tab:action_space}
\begin{tabular}{c|cl}
\toprule
\textbf{Type} & \textbf{Action} & \textbf{Attributes} \\ \midrule
\multirow{14}{*}{\textbf{Keyboard}} & \multirow{2}{*}{Key Press} & Key name (string), \\
                  && Key press duration (seconds:float)\\ \cline{2-3} 
                  & Key Hold & Key name (string) \\ \cline{2-3} 
                  & Key Release & Key name (string) \\ \cline{2-3} 
                  & \multirow{3}{*}{Key Combo} & Key names (strings), \\ 
                  && Key combo duration (seconds:float), \\
                  && Wait behaviour (sync/async) \\\cline{2-3} 
                  & \multirow{3}{*}{Hotkey} & Key names (strings), \\
                  && Hotkey sequence duration (seconds:float), \\
                  && Wait behaviour (sync/async) \\ \cline{2-3} 
                  & \multirow{2}{*}{Text Type} & String to type (string), \\
                  && Typing duration (seconds:float) \\ \midrule
\multirow{19}{*}{\textbf{Mouse}} & \multirow{2}{*}{Button Click} & Mouse button (left/middle/right), \\
                  && Button click duration (seconds:float) \\ \cline{2-3} 
                  & Button Hold & Mouse button (left/middle/right) \\ \cline{2-3} 
                  & Button Release & Mouse button (left/middle/right) \\ \cline{2-3} 
                  & \multirow{4}{*}{Move} & Mouse position (width:int, height:int), \\ 
                  && Mouse speed (seconds:float), \\ 
                  && Coordinate system (relative/absolute), \\
                  && Tween mode (enum)~\footnote{Tweening is not currently utilized in the prototype. But this is a OS-supported functionality that can greatly enhance user experience.} \\ \cline{2-3} 
                  & \multirow{3}{*}{Scroll} & Orientation (vertical), \\
                  && Distance (pixels:int), \\
                  && Duration (seconds:float) \\ \midrule 
\textbf{Wait} & Noop & - \\
\bottomrule
\end{tabular}
\end{table}

It is important to note that, while some works (\eg AssistantGUI~\citep{gao2023assistgui},  OmniACT~\citep{kapoor2024omniact} and OSWorld~\citep{xie2024osworld}) use \emph{PyAutoGUI}~\footnote{Python library that provides a cross-platform GUI automation module - \url{https://github.com/asweigart/pyautogui}} for keyboard and mouse control, this approach does not work in all applications, particularly in modern video games using DirectX~\footnote{Microsoft DirectX graphics provides a set of APIs for high-performance multimedia apps - \url{https://learn.microsoft.com/en-us/windows/win32/directx}}. Moreover, such work chooses to expose a subset of the library functionality in its action space, ignoring dimensions like press duration and movement speed, which are critical in many scenarios (\eg RDR2, for opening the weapon wheel and changing view).

To ensure wide game and software compatibility and accommodate different operating systems, in our current implementation we use the similar \emph{PyDirectInput} library~\footnote{Python library encapsulating Microsoft's \emph{DirectInput} calls for convenience manipulating keyboard keys - \url{https://github.com/learncodebygaming/pydirectinput}} and \emph{PyAutoGUI} for keyboard control, utilize \emph{AHK}~\footnote{A fully typed Python wrapper around AutoHotkey to keyboard and mouse control - \url{https://github.com/spyoungtech/ahk}} and write our own abstraction (using the \emph{ctypes} library~\footnote{Python library that provides C compatible data types, and allows calling functions in DLL/.so binaries - \url{https://docs.python.org/3/library/ctypes.html}}) to send low-level mouse commands to the operating system for mouse control. For increased portability and ease of maintenance, all keyboard and mouse control is encapsulated in a class, called \emph{IO\_env}.

Notably, our low-level control wrapper is adapted for both MacOS and Windows systems, making the OS transparent to us. At the software window level, we implemented automatic switching between the target software window and the window running the agent (using Python \emph{ctypes} for Windows and \emph{AppleScript} for MacOS~\footnote{AppleScript is a scripting language created by Apple, which allows users to directly control scriptable applications, as well as parts of MacOS - \url{https://developer.apple.com/library/archive/documentation/AppleScript/Conceptual/AppleScriptLangGuide/introduction/ASLR_intro.html}}).

\textbf{Procedure Memory.} This memory stores pre-defined basic skills and the generated skills captured from the \emph{Skill Curation}. However, as we continuously obtain new skills during game playing, the number of skills in procedural memory keeps increasing, and it is hard for GPT-4o to precisely select the most suitable skill from the large memory. Thus, similar to Voyager~\citep{wang2023voyager}, we use OpenAI's \emph{text-embedding-ada-002 model}~\citep{openaiada} to generate embeddings for each skill and 
store pre-defined basic skills and any generated skills captured from \emph{Skill Curation}, along with their embeddings in a procedural memory. We retrieve a subset of skills, that are relevant to the given task, and then let GPT-4o select the most suitable one from the subset. In the skill retrieval, we pre-compute the embeddings of the documentations (code, comments and descriptions) of skill functions, which describe the skill functionality, and compute the embedding of the given task. Then we compute the cosine similarities between the skill documentation embeddings and the task embedding. The higher similarity means that the skill's functionality is more relevant to the given task. We select the top K skills with the highest similarities as the subset. Using similarity matching to select a small candidate set simplifies the process of choosing skills.

\textbf{Episodic Memory.} This memory stores all the useful information provided by the environment and LMM, which consists of short-term memory and long-term summary. 

The short-term memory stores the screenshots within the recent k interactions in game playing and the corresponding information from other modules, \eg screenshot descriptions, task guidance, actions, and reasoning. We set k to five, and it can be regarded as the memory length. Information stored over k interactions ago will be forgotten from direct short-term memory. Empirically, we found that recent information is crucial for decision-making, while a too-long memory length would cause hallucinations. In addition, other modules continuously retrieve recent information from short-term memory and update the short-term memory by storing the newest information. 

For some long-horizon tasks, short-term memory is not enough. This is because the completion of a long-horizon task might require historical information from a long steps ago. For example, the agent might do a series of short-horizon tasks during a long-horizon task, which makes the original long-horizon task forgotten in short-term memory. To maintain the long-term valuable information while avoiding the long-token burden of GPT-4o, we propose a recurrent information summary as long-term memory, which is the text summarization of experiences in game playing, including the ongoing task, the past entities that the player met, and the past behaviors of the player and NPCs. 

In more detail, we provide GPT-4o with the summarization before the current screenshot and the recent screenshots with corresponding descriptions, and GPT-4o will make a new summarization by organizing the tasks, entities, and behaviors in the time order with sentence number restriction. Then we update the summarization to be the newly generated one, which includes the information in the current screenshot. The recurrent summarization update, inspired by RNN, achieves linear-time inference by preserving a hidden state that encapsulates historical input. This method ensures the compactness of summarization token lengths and recent input data. Furthermore, the incorporation of long-term memory enables the agent to effectively retain crucial information over extended periods, thereby enhancing decision-making capabilities.

\textbf{Information Gathering.} Given the video clip as input, we mainly depend on GPT-4o's OCR capabilities to extract textual information in the keyframes, which usually contain critical guidance and notifications for the current situation. We also rely on GPT-4o's visual understanding to analyze the visual information in the frames. Besides, we augment LMMs' visual understanding via some tools, like template matching~\citep{brunelli2009template}, Grounding DINO~\citep{liu2023grounding}, and SAM~\citep{kirillov2023segment}, to provide additional grounding for object detection and segmentation. Some visual prompting tricks, like drawing axes and colorful directional bands, are also applied to enhance the GPT-4o's visual ability.

\textbf{Task Inference.} After reflecting on the outcome of the last executed action, We let GPT-4o analyze the current situation to infer the most suitable task for the current moment and estimate the highest priority task to perform and when to stop an ongoing task and start a new one.

\textbf{Skill Curation.} 
GPT-4o is required to strictly follow the provided interfaces and examples to generate the corresponding code for new skills. Moreover, GPT-4o is required to include documentation/comments within the generated code, delineating the functionality of each skill. \emph{Procedural Memory} where skills are stored will then check whether the code is valid, whether the format of documentation is right, and whether any skill with the same name already exists. If all conditions are passed, the newly generated skill is persisted for future utilization.

\textbf{Action Planning.}
GPT-4o needs to select the appropriate skills from the curated skill set and instantiate these skills into a sequence of executable actions by specifying any necessary parametric aspects (\eg duration, position, and target) according to the current task and history information. The generated action is then fed to the \textit{\textbf{Executor}} for interaction with the environment. 

%% file: appendix/rdr2.tex
\subsection{Introduction to RDR2}
\label{sec:appx_rdr2_game_intro}

Red Dead Redemption II (RDR2) is an epic AAA Western-themed action-adventure game by Rockstar Games. As one of the most famous and highest-selling games in the world, it is widely acknowledged for its movie-like realistic scenes, rich storylines, and immersive open-ended world. The game applies a typical role-playing game (RPG) control system, played from a first- or third-person perspective, which uses WASD for movement, mouse control for view changing, first- or third-person shooting for combat, and inventory and manipulation. 

For most of the game, players need to control the main character, Arthur Morgan, upon choosing to complete mission scenarios following the main storyline. Otherwise, they can freely explore the interactive world, such as going hunting, fishing, chatting with non-player characters (NPCs), training horses, witnessing or partaking in random events, and participating in side quests. As the main storyline progresses, different skills are gradually unlocked. As a close-source commercial game, no APIs are available for obtaining additional game-internal information nor pre-defined automation actions. Following its characteristics, this game serves as a fitting and challenging environment for the GCC setting and a comprehensive benchmark for embodiment.

\subsection{Objectives}
\label{sec:rdr2_tasks}
\begin{figure}[ht]
\centering
\centering
\includegraphics[width=0.9\textwidth]{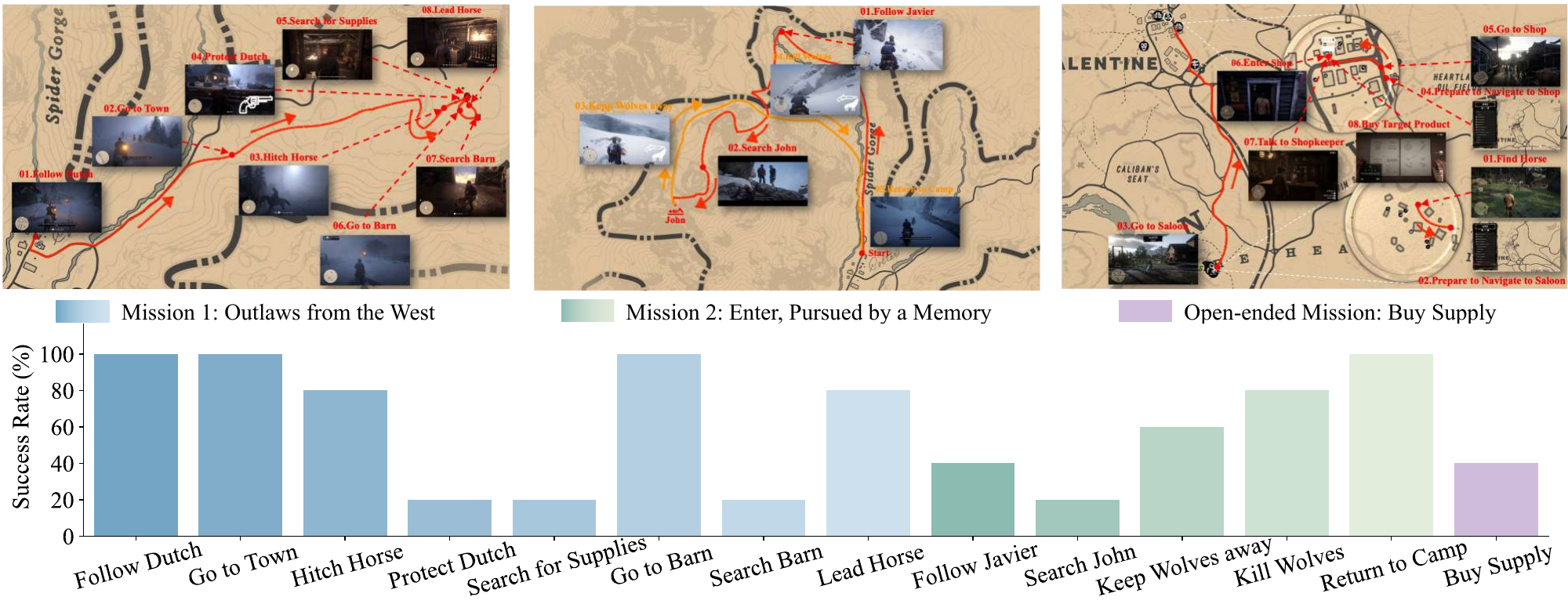}
\caption{Trajectory and success rates of 13 main storyline tasks and 1 open-world task in RDR2.}
\label{fig:rdr2_trajectories}
\vspace{-10pt}
\end{figure}

In Chapter 1 of RDR2, the first two missions of the main storyline are \emph{Outlaws from the West} and \emph{Enter, Pursued by a Memory}. These missions serve as the tutorial content for RDR2, guiding players step-by-step into the role of Arthur. They immerse the player in the story's development while teaching the game's controls and mechanics.

We divided Mission 1 and Mission 2 into 8 and 5 tasks respectively based on the checkpoints within each mission. Each checkpoint may present failure scenarios. For example, in Mission 1, there are six failure scenarios: i) Assaults, kills, or abandons Dutch or Micah; ii) Allows Dutch or Micah to be killed;
iii) Abandons the homestead; iv) Assaults, kills, or abandons their horse; v) Assaults, kills, or abandons the horse in the barn; vi) Dies. We categorized each sub-task as either "Easy" or "Hard" based on the likelihood of failure at each checkpoint and the need to retry the checkpoint.

To evaluate \projname's capabilities in an open-world environment, Mission 3 is designed as a hard open-ended task. Unlike the first two tutorial missions, it does not include any checkpoints. Consequently, the entire Mission 3 is treated as a single, comprehensive task. Although we do not subdivide Mission 3 into finer tasks, we aim to identify key points to facilitate a clearer understanding of Mission 3 for the reader.

Tables~\ref{tab:subtask_list} and ~\ref{tab:open_ended_subtask_list} provide a brief introduction of each task in the first two missions of the main storyline and an open-ended mission, along with approximate estimates of their difficulty. Due to GPT-4o's poor performance in spatial understanding and fine-manipulation skills, it can be challenging for our agent to perform certain actions, like entering or leaving a building, or going to precise indoor locations to retrieve specific items. Additionally, the high latency of GPT-4o's responses also makes it harder for an agent to deal with time-sensitive events, \eg during combat.


\begin{table}[ht]
\centering
\caption{Tasks in the first two missions of RDR2. In the tutorial guide, the prompt text \emph{Start Dialogue} signifies the end of the previous checkpoint and the beginning of the current checkpoint. \emph{Difficulty} refers to how hard to accomplish the corresponding tasks. Figures~\ref{fig:app_mission1_tasks} and ~\ref{fig:app_mission2_tasks} showcase snapshots of each task (specific sub-figures marked in parenthesis in the table). The maximal number of steps (agent takes one action) for each task is 500.}
\label{tab:subtask_list}
\resizebox{\textwidth}{!}{
\begin{tabular}{p{4cm}p{8cm}p{4cm}p{1cm}}
\toprule
Mission 1: Outlaws from the West & Description & Start Dialogue & Difficulty \\ 
\midrule

Follow Dutch (Fig. \ref{subfig:M01_C01_Follow_Dutch}) & Arthur follows Dutch on horseback into the snow to find their scouting gang members. & Use [W] to Follow Dutch & Easy \\

Go to Town (Fig. \ref{subfig:M01_C02_Follow_Micah}) & Arthur rides his horse, following Micah to the vicinity of a little homestead Micah discovered. & Hold [W] to match speed with Dutch and Micah & Easy \\

Hitch Horse (Fig. \ref{subfig:M01_C03_Hitch_Horse}) & Arthur hitches the horse to the hitching post, then goes to the old shed and takes cover. & Hold [E] to hitch your horse & Easy \\

Protect Dutch (Fig. \ref{subfig:M01_C04_Protect_Dutch}) & Arthur uses his gun to shoot all of the O'Driscolls inhabiting the house and protect Dutch. & Use [W] to peak out of cover & Hard \\

Search for Supplies (Fig. \ref{subfig:M01_C05_Search_for_Supplies}) & Arthur follows Dutch to the house to search for supplies. & Hold [R] near items to pick the up while searching house. & Hard \\

Go to Barn (Fig. \ref{subfig:M01_C06_Go_to_the_Barn}) & Arthur follows Dutch's directions and goes to the barn to see if there's anything inside. & Dutch: Micah, Arthur, keep looking for stuff & Easy \\

Search Barn (Fig. \ref{subfig:M01_C07_Search_the_Barn}) & Arthur searches the barn and defeats the O'Driscoll hiding inside. & [F] Attack the O’Driscoll & Hard \\

Lead Horse (Fig. \ref{subfig:M01_C08_Lead_the_Horse}) & Arthur calms the horse and takes it out of the barn. & Hold [Right Mouse Button] to focus on the horse & Easy \\
\bottomrule

\toprule
Mission 2: Enter, Pursued by a Memory  & Description & Start Dialogue & Difficulty \\ 
\midrule
Follow Javier (Fig. \ref{subfig:M02_C01_Follow_Javier}) & Arthur rides his horse following Javier up the mountain through the blizzard searching for John's trail. & Follow Javier & Hard \\

Search John (Fig. \ref{subfig:M02_C02_Find_John}) & After dismounting, Arthur followed Javier over slopes and ledges to find John and carry him away. & Javier: Down this way & Hard \\

Keep Wolves away (Fig. \ref{subfig:M02_C03_Shoot_Wolves}) & Arthur manages to shoot all of the wolves before they can attack Javier and John. & Keep the wolves away from Javier and John & Hard \\

Kill Wolves (Fig. \ref{subfig:M02_C04_Eliminate_Wolves}) & Three people ride horses down the mountain. Arthur eliminate the wolves, protecting Javier and John ahead. & Javier: Come on, let’s get back to the others & Hard \\

Return to Camp (Fig. \ref{subfig:M02_C05_Return_to_Camp}) & Arthur followed Javier on horseback back to camp. & Yea…c’mon. Let’s push hard and get back & Easy \\
\bottomrule

\end{tabular}
}
\end{table}

\begin{table}[ht]
\centering
\caption{Key points in the open-ended mission, \emph{Buy Supply} in RDR2. Figure~\ref{fig:app_mission3_tasks} showcases snapshots of key points (specific sub-figures marked in parenthesis in the table).}
\label{tab:open_ended_subtask_list}
\resizebox{0.9\textwidth}{!}{
\begin{tabular}{ll}
\toprule
Mission 3: Buy Supply & Description \\ 
\midrule
Find Horse (Fig. \ref{subfig:M03_C01_Find_Horse}) & Find and mount the horse in the camp. \\

Prepare to Navigate to Saloon (Fig. \ref{subfig:M03_C02_Generate_Navigation_to_Saloon}) & Open map, find the saloon and create waypoint. \\

Go to Saloon (Fig. \ref{subfig:M03_C03_Go_to_Saloon}) & Ride horse to the saloon. \\

Prepare to Navigate to Shop (Fig. \ref{subfig:M03_C04_Generate_Navigation_to_Shop}) & Open map, find the general store and create waypoint. \\

Go to Shop (Fig. \ref{subfig:M03_C05_Go_to_Shop}) & Ride horse to the shop. \\

Enter Shop  (Fig. \ref{subfig:M03_C06_Enter_Shop}) & Dismount the horse and enter the shop. \\
Talk to Shopkeeper(Fig. \ref{subfig:M03_C07_Talk_to_Shopkeeper}) & Approach the shopkeeper and talk. \\
Buy Target Product (Fig. \ref{subfig:M03_C08_Buy_Target_Product}) & Open the menu, find and buy the target product. \\
\bottomrule
\end{tabular}
}
\end{table}

\input{rdr2_figure/mission1}

\input{rdr2_figure/mission2}

\input{rdr2_figure/mission3}

\begin{figure}[!t]
    \centering
    \includegraphics[width=1.0\linewidth]{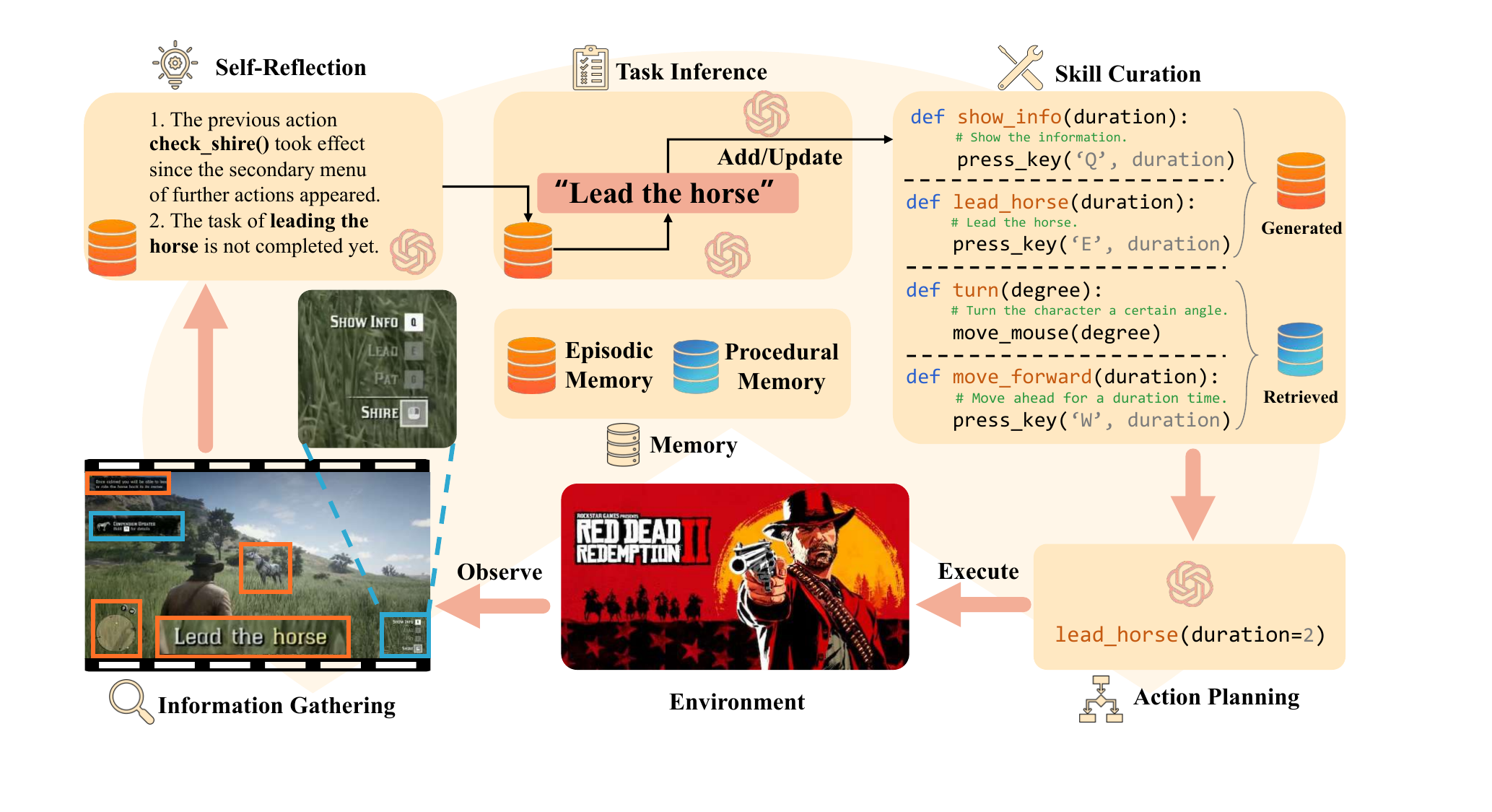}
    \caption{The detailed illustration of how \projname is instantiated as a game agent to play RDR2.}
    \label{fig:dataflow}
\end{figure}

\subsection{Implementation Details} 
\label{sec:implementation_details_RDR2}
Our experiments are based on the latest version of RDR2, `Build 1491.50'. As shown in Figure \ref{fig:dataflow}, strictly following the GCC setting, our agent takes the video of the screen as input and outputs keyboard and mouse operations to interact with the computer and the game. An observation thread is responsible for the collection of video frames from the screen and each video clip records the whole in-game process since executing the last action. 

\textbf{Information Gathering.} To extract keyframes from the video observation, we utilize the VideoSubFinder tool~\footnote{VideoSubFinder standalone tool - \url{https://sourceforge.net/projects/videosubfinder/}}, a professional subtitle discovery and extraction tool. These keyframes usually contain rich meaningful textual information in the game, which are highly relevant to the completion of tasks and missions (such as character status, location, dialogues, in-game prompts and tips, etc.) We use GPT-4o to extract and categorize all the meaningful contexts in these keyframes and perform OCR, and call this processing "gathering text information". Then, to save interactions with GPT-4o, we only let GPT-4o provide a detailed description of the last frame of the video. 

While GPT-4o exhibits impressive visual understanding abilities across various CV tasks, we find that it struggles with spatial reasoning and recognizing some game-specific icons. To address these limitations, we add a visual augmentation sub-module within our \emph{Information Gathering} module. This augmentation step serves two main purposes: i) utilize Grounding DINO~\citep{liu2023grounding}, an open-set object detector, to output precise bounding boxes of possible targets in an image and serve as spatial clues for GPT-4o; and ii) perform template matching~\citep{brunelli2009template} to provide icon recognition grounding truth for GPT-4o when interpreting instructions or menus shown on screen. As LMM capabilities mature, it should be possible to disable such augmentation.

\textbf{Self-Reflection.} The reflection module mainly serves to evaluate whether the previously executed action was successfully carried out and whether the current executing task is finished. To achieve this, we uniformly sample at most 8 sequential frames from the video observation since the execution of the last action and use GPT-4o to estimate the success of its execution. Additionally, we expect GPT-4o can also provide analysis for any failure of the last action (\eg the move-forward action failed and the cause could be the agent was blocked by an obstacle). With such valuable information as input for \emph{Action Planning}, including the failure/success of the last action and the corresponding analysis, the agent is capable of attempting to remedy an inappropriate decision or action execution. 

Moreover, some actions require prolonged durations, such as holding down specific keys, which can coexist or interfere with other actions decided by subsequent decisions. Consequently, the reflection module must also decide whether an ongoing action should continue to be executed. Furthermore, self-reflection can be leveraged to dissect why the last action failed to bring the agent close to the target task completion, better understand the factors that led to the successful completion of the preceding task, and so on. 

Besides, we observe that instead of providing GPT-4o with sequential high-resolution images for self-reflection, low-resolution images make it easier for GPT-4o to understand the relation among the sequential screenshots and capture dynamic changes, resulting in a significantly higher success rate of detecting whether the action is executed successfully and take any effect. We hypothesize that since a high-resolution image can cost as many as 2000 tokens, too many high-resolution images make GPT-4o fail to capture the overall changes across screenshots and be caught up in the local details.

\textbf{Task Inference.} During gameplay, we let GPT-4o propose the current task to perform whenever it believes it is time to start a new task. GPT-4o also outputs whether the task is a long- or short-horizon task when proposing a new task. Long-horizon tasks, such as traveling to a location, typically require multiple iterations, whereas short-horizon tasks, like picking up an item or conversing with someone, involve fewer iterations. The agent will follow the newly generated task for the next 3 interactions. After 3 interactions, the agent returns to the last long-horizon task in the stack. Deciding on a binary task horizon is much easier and more robust for GPT-4o, than re-planning at every iteration. Since a long-horizon task frequently includes multiple short-horizon sub-tasks, this implementation also helps avoid forgetting the long-horizon tasks under execution.

\begin{figure*}
\centering
\includegraphics[width=1\linewidth]{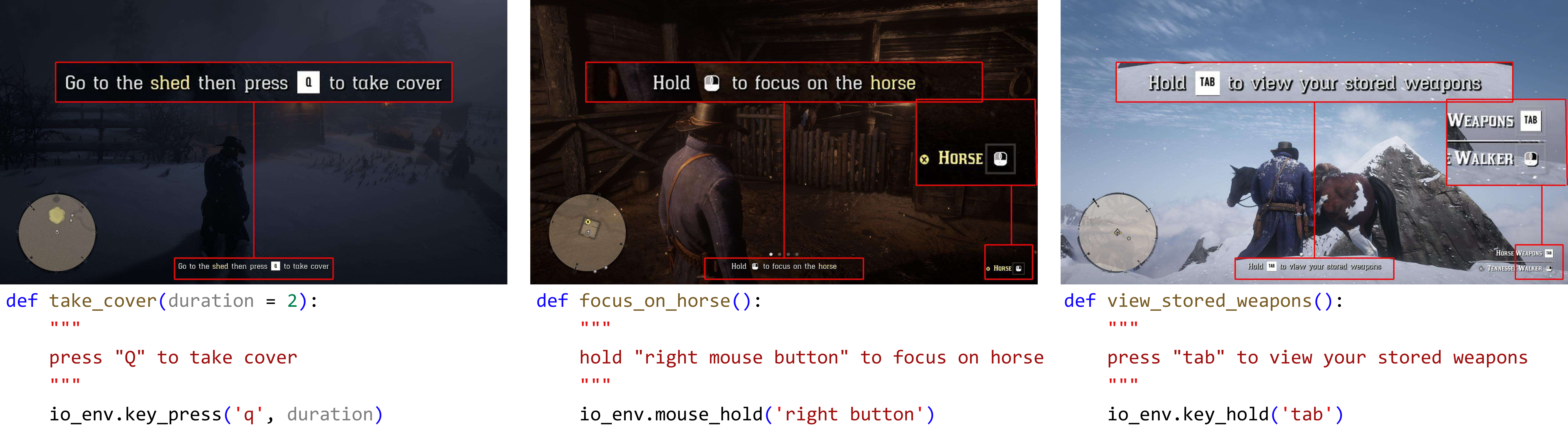}
\caption{Skill code generation based on in-game instructions. As the storyline progresses, the game will continually provide prompts on how to use a new skill via keystrokes or utilizing the mouse.} 
\label{fig:skills}
\end{figure*}

\textbf{Skill Curation.} 
As shown in Figure~\ref{fig:skills}, during gameplay, instructions often appear on the screen, such as "press [Q] to take over" and "hold [TAB] to view your stored weapons", which serve as essential directives for completing current and future tasks proficiently. To save interactions with GPT-4o, we implement a simple version of this module inside \emph{Information Gathering} to reduce interactions with GPT-4o. When GPT-4o detects and classifies some instructional text in the recent observation, which usually contains key and button hints, it will directly generate the corresponding code and description.

\textbf{Action Planning.} Upon execution of this module, we first retrieve the top k relevant skills for the task from procedural memory, alongside the newly generated skills. We then provide GPT-4o with the current task, the set of retrieved skills, and other information collected in \emph{Information Gathering} that may be helpful for decision-making (\textit{e.g.}, recent screenshots with corresponding descriptions, previous decisions, and examples) and let it suggest which skills should be executed. We also request that GPT-4o provide the reasons for choosing these skills, which increases the accuracy, stability, and explainability of skill selection and thus greatly improves framework performance. While GPT-4o sometimes may generate a sequence of actions, we currently only execute the first one, and perform \emph{Self-Reflection}, since we observe a tendency for the second action to usually suffer from severe hallucinations.

\textbf{Action Execution}. Unlike the conventional mouse operation in standard software, where the cursor is restricted to a 2D grid and remains visible on the screen to navigate and interact with elements, the utilization of the mouse in 3D games like RDR2 introduces a varied control scheme. In menu screens, the mouse behaves traditionally, offering familiar point-and-click functionality. However, during gameplay, the mouse cursor disappears, requiring players to move the mouse according to specific action semantics. For example, to alter the character's viewpoint, the player needs to map the actual mouse movement to in-game direction angle changes, which differ in magnitude in the X and Y axes. Another special transition applies to shooting mode, where the front sight is fixed at the center of the screen, and players must maneuver the mouse to align the sight with target enemies. This nuanced approach to mouse control in different contexts adds an extra layer of challenge to general computer handling, showcasing the adaptability required in game environments, compared to regular software applications.

\textbf{Procedural Memory}. In our target setting, We intend to let the agent learn all skills from scratch, to the extent possible for the main storyline missions. The procedural memory is initialized with only preliminary skills for basic movement, which are not clearly provided by the in-game tutorial and guidance.
\begin{itemize}
    \item \emph{turn(degree)}, \emph{move\_forward(duration)}: Since the game does not precisely introduce how to move in the world through in-game instructions, we provide these two basic actions in advance, so GPT-4o can perform basic mobility, while greatly reducing the number of calls to the model.
    \item \emph{shoot(x, y)}: RDR2 also does not provide detailed instructions on how to aim and shoot. Moreover, due to limitations with GPT-4o spatial reasoning and the need to sometimes augment images with object bounding boxes, we provide such basic skill for the agent to complete relevant tasks.
    \item \emph{select\_item\_at(x, y)}: Similarly to \emph{shoot()}, due to the lack of instructions, we provide such skill for the agent to move the mouse to a certain place to select a given item.
\end{itemize}

Beyond these basic atomic low-level actions, we introduce a few composite skills to facilitate the game playing progress. The agent should be able to complete tasks using only the basic skills above and the skills it learns, but these composite skills streamline the process by greatly reducing calls to the backend model.

\begin{itemize}
    \item \emph{turn\_and\_move\_forward(degree, duration)}: This skill is just a simple composition of \emph{turn()} and \emph{move\_forward()} to save frequent calls to GPT-4o in a common sequence.
    \item \emph{follow(duration)} and \emph{navigate\_path(duration)}: In RDR2, tasks often guide players to follow NPCs or generated paths (red lines) in the minimap to certain locations. This can be reliably accomplished via the basic movement skills, but requires numerous interactions with GPT-4o. To control both cost and time budgets involving GPT-4o's responses, we leverage the information shown in the minimap to implement a composite skill to follow target NPCs or red lines for a short set of game iterations. The default duration is 20 iterations. Increasing the duration can dramatically improve the performance in task \emph{Follow Dutch}, \emph{Follow Javier} and \emph{Killing Wolves} but significantly decrease the success rate of \emph{Search John} since this task requires frequent exchange of the skills between climbing and following. 
    \item \emph{fight()}: As output of an interaction with GPT-4o, the agent will only take one action per step. However, though the action is generated correctly, specifically in fight scenarios, the action frequency may not be high enough to defeat an opponent. In order to allow sub-second punches, we provide a pre-defined action that wraps this multi-action punching, which can be selected by GPT-4o to effectively win fights.
\end{itemize}

For the open-ended mission, since the agent skips all the tutorials in Chapter I, we provide all the necessary skills in the procedural memory at the beginning of the mission.

\textbf{Episodic Memory.} This module stores all the useful information, \eg input and output of GPT-4o. 
In each iteration, after the self-reflection, we will request GPT-4o to summary the event that happened in the last action and the past experiences.

\textbf{Game Pause.} To prevent in-game time from passing in real-time games like RDR2, we have to pause the game while waiting for LMMs' response. The time interval between two consecutive actions can be as long as one minute. In RDR2, after the agent finishes executing outputted actions, \emph{esc} will be automatically pressed to pause the game and when the agent determines the next action, \emph{esc} will be automatically pressed again to unpause the game. Note that there will be an animation lasting up to 0.5 seconds for both pausing and unpausing. During this animation, we can not control the character, but the dynamics of the game world keep changing, \eg the wolves are still moving. It introduces additional challenges for the tasks that require precise timing, like combat. 

\begin{figure}[htp]
    \centering \includegraphics[width=0.9\linewidth]{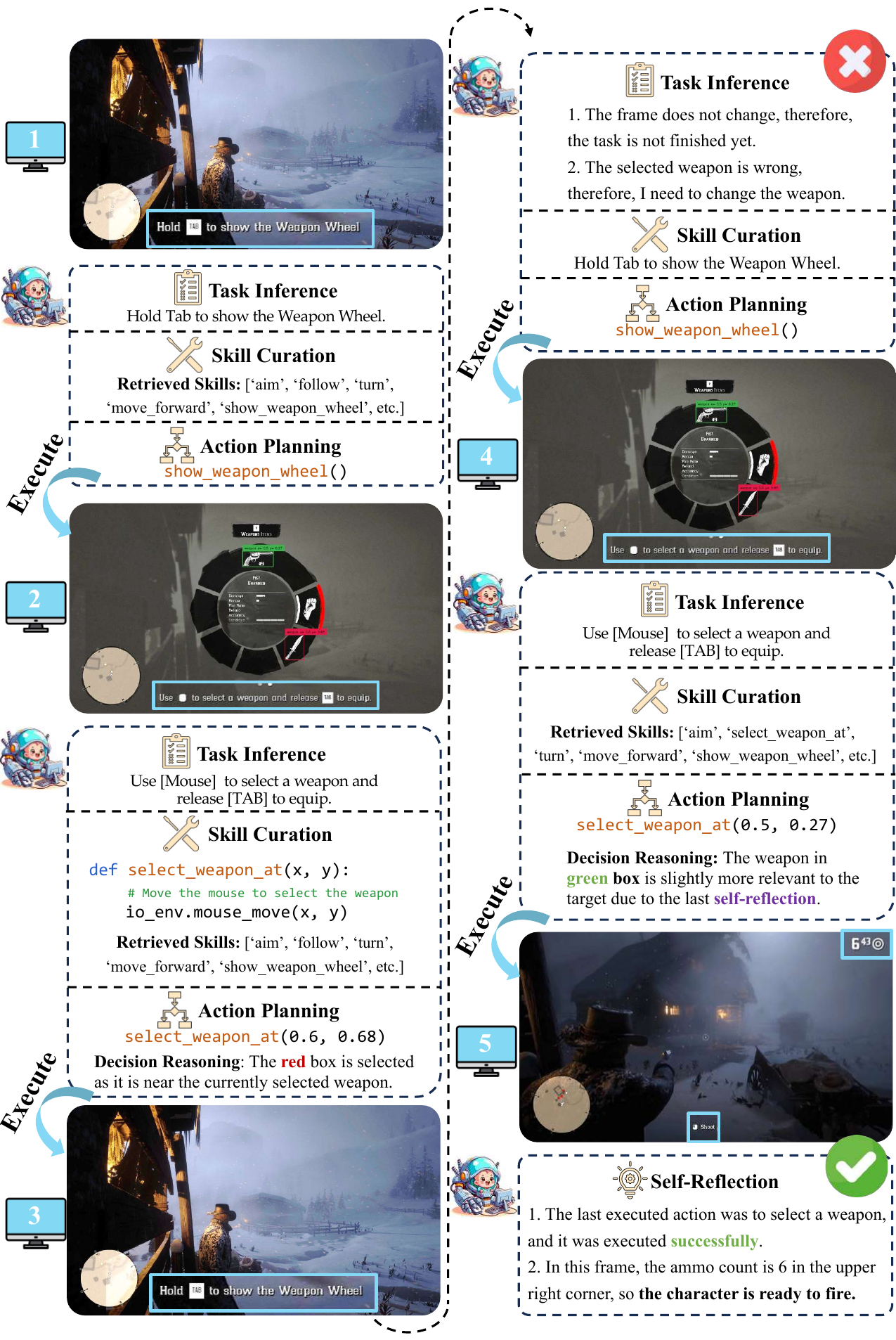}
    \caption{Case study of self-reflection on re-trying a failed task. Task instruction and context require the agent to equip the gun. A wrong weapon (knife) is first selected, but the agent equips the gun after self-reflection.
    Only relevant modules are shown for better readability, though all modules (Figure~\ref{fig:system}) are executed per iteration.}
    \label{fig:example_self_reflection_RDR2}
\end{figure}

\subsection{Case Studies}

Here we present a few game-specific case studies for more in-depth discussion of the framework capabilities and the challenges of the GCC setting.

\subsubsection{Self-Reflection}
Self-reflection is an essential component in \projname as it allows our framework reasoning to correct previous mistakes or address ineffective actions taken in-game. Figure~\ref{fig:example_self_reflection_RDR2} provides an example of the self-reflection module. The task requires the agent to select a weapon to equip, in the context of the ``Protect Dutch'' task. Initially, the agent selects a knife as its weapon by chance, but since the game requires a gun to be chosen, this is incorrect and the game still prompts the player to re-open the weapon wheel. The self-reflection module is able to determine that the previous action was incorrect and on a subsequent iteration the agent successfully opts for the gun, correctly fulfilling the task requirement and advancing to the next stage in the story. 

\subsubsection{Skill Curation}
For skill curation, we first provide GPT-4o with examples of general mouse and keyboard control APIs, \textit{e.g.}, io\_env.key\_press and io\_env.mouse\_click. Figure~\ref{fig:skills} shows that GPT-4o can capture and understand the prompts appearing on screenshots, \ie icons and text, and strictly follow the provided skill examples using our IO interface to generate correct skill code. Moreover, GPT-4o also generates comments in the code to demonstrate the functionality of this skill, which are essential for computing similarity and relevance with a given task during skill retrieval. The quality of the generated comment directly determines the results of skill retrieval, and further impacts reasoning to action planning. Curation can also re-generate code for a given skill, which is useful if GPT-4o wrongly recognized a key or mouse button in a previous iteration.

\subsubsection{Action Execution and Feedback} \label{sec:env_feedback} 
Proper reasoning about environment feedback is critical due to the generality of the GCC setting and the level of abstraction to interact with the complex game world. The semantic gaps between the execution of an action, its effects in the game world, and observing the relevant outcomes for further reasoning lead to several potential issues that \projname needs to deal with. Such issues can be categorized into four major cases:

\textbf{Lack of grounding feedback.}
In many situations, due to the lack of precise information from the environment, it can be difficult for the system to deduce the applicability or outcome of a given action. For example, when picking an item from the floor, the action may fail due to the distance to the object not yet being close enough. Or, if within pick up range, the chosen action may not exactly apply due to other factors (\eg character's package is full).

Even if the right action is selected and executed successfully, the agent still needs to figure out its results from the partial visual observation of the game world. If the agent needs to pick or manipulate an object that is occluded from view, the action may execute correctly, but no outcome can be seen. 

A representative example in RDR2 happens when the agent tries to pick up its gun from the floor after a fight. Getting to the right distance, without completely occluding the object, can lead to multiple re-trials. Figure~\ref{fig:grounding_a} showcases a situation where, though the character is already standing near the gun (as seen in the minimap), it's still not possible to pick it up.

Previous efforts~\citep{wang2023describe, wang2023voyager} that utilize in-game state APIs unreasonably bypass such issues by leveraging internal structured information from the game and the full semantics of responses (data) or failures (error messages).

\textbf{Imprecise timing in IO-level calls.}
This issue is caused by the ambiguity in the game instructions or differences in specific in-game action behaviors, where even the execution of a correct action may fail due to minor timing mismatches. For example, when executing an action like `open cabinet', which requires pressing the [R] key on the keyboard, if the press is too fast, no effect happens in the game world. However, as there is no visual change in the game nor other forms of feedback, it can be difficult for GPT-4o to figure out if an inappropriate action was chosen at this game state or if the minor timing factor was the problem. Pressing the key for longer triggers an animation around the button (only if the helper menu is on screen), but this is easily missed and any key release before the circle completes also results in no effect. Figure~\ref{fig:grounding_b} illustrates the situation. 

The same problem also manifests in other situations in the game, where pressing the same key for longer triggers a completely different action (\eg lightly pressing the [Left Alt] key vs. holding it for longer).

\textbf{Change in the semantics of key and button.}
A somewhat similar situation occurs when the same keyboard key or mouse button gets attributed different semantics in different situations (or even in a multi-step action). GPT-4o may decide to execute a given skill, but the original semantics no longer hold. The lack of in-game effect parallels the previous situations. Worse yet, an undesired effect will confuse the system regarding the correct action being selected or not.

For example, when approaching a farm in the beginning of the game, the agent needs to hitch the horse to a pole to continue. The operation to perform the action consists of pressing the [E] key near a hitching post (as shown in Figure~\ref{fig:grounding_c}). However, the same [E] key press is the only constituting step in other actions with different semantics, like \emph{dismount the horse} or \emph{open the door}. Wrongly triggering a horse dismount at the situation shown in the figure can lead to undesired side effects, \ie it may mislead the system about the actual effects of the action or affect the planning of which next actions to perform. 

\textbf{Interference issues.}
Lastly, completion of some actions requires the correct execution of multiple steps sequentially, which could be interrupted in many ways not related to the agent's own actions. Without the use of APIs that expose internal states or other forms of feedback, it is much harder for the agent to decide when to repeat sub-actions or try different strategies. For example, if the agents gets shot and loses aim while in combat, or an unrelated in-game animation is triggered mid-action, canceling it.

Since there is no direct environment feedback, the agent needs to carefully analyze the situation and try to infer if any action step needs re-execution.

\newcommand{\subfigwidth}{0.32}
\begin{figure}
  \centering
  \begin{subfigure}{\subfigwidth\textwidth}
    \centering
    \includegraphics[width=\linewidth]{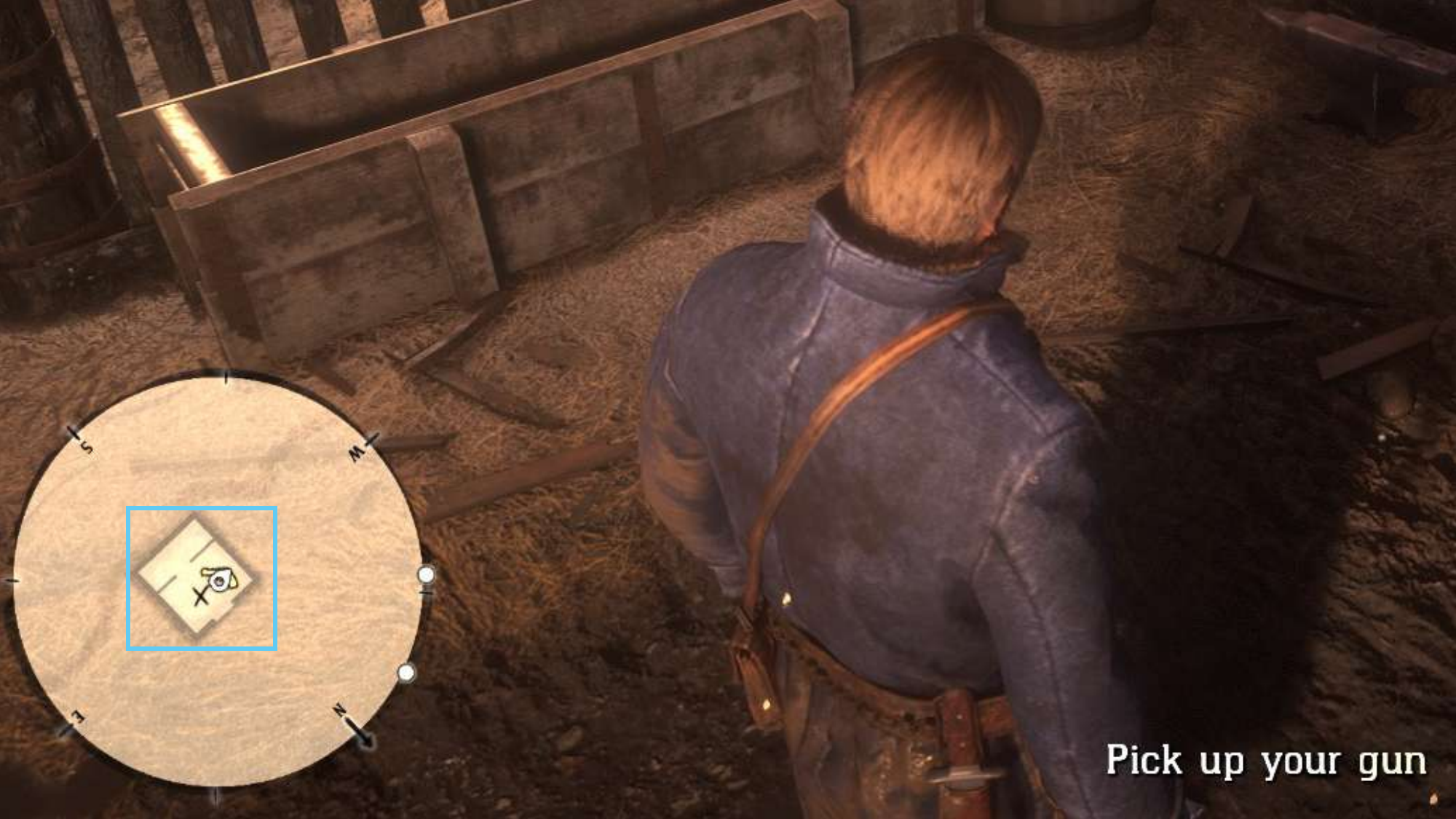}
    \caption{`Pick gun' unavailable} \label{fig:grounding_a}
  \end{subfigure}
  \begin{subfigure}{\subfigwidth\textwidth}
    \centering
    \includegraphics[width=\linewidth]{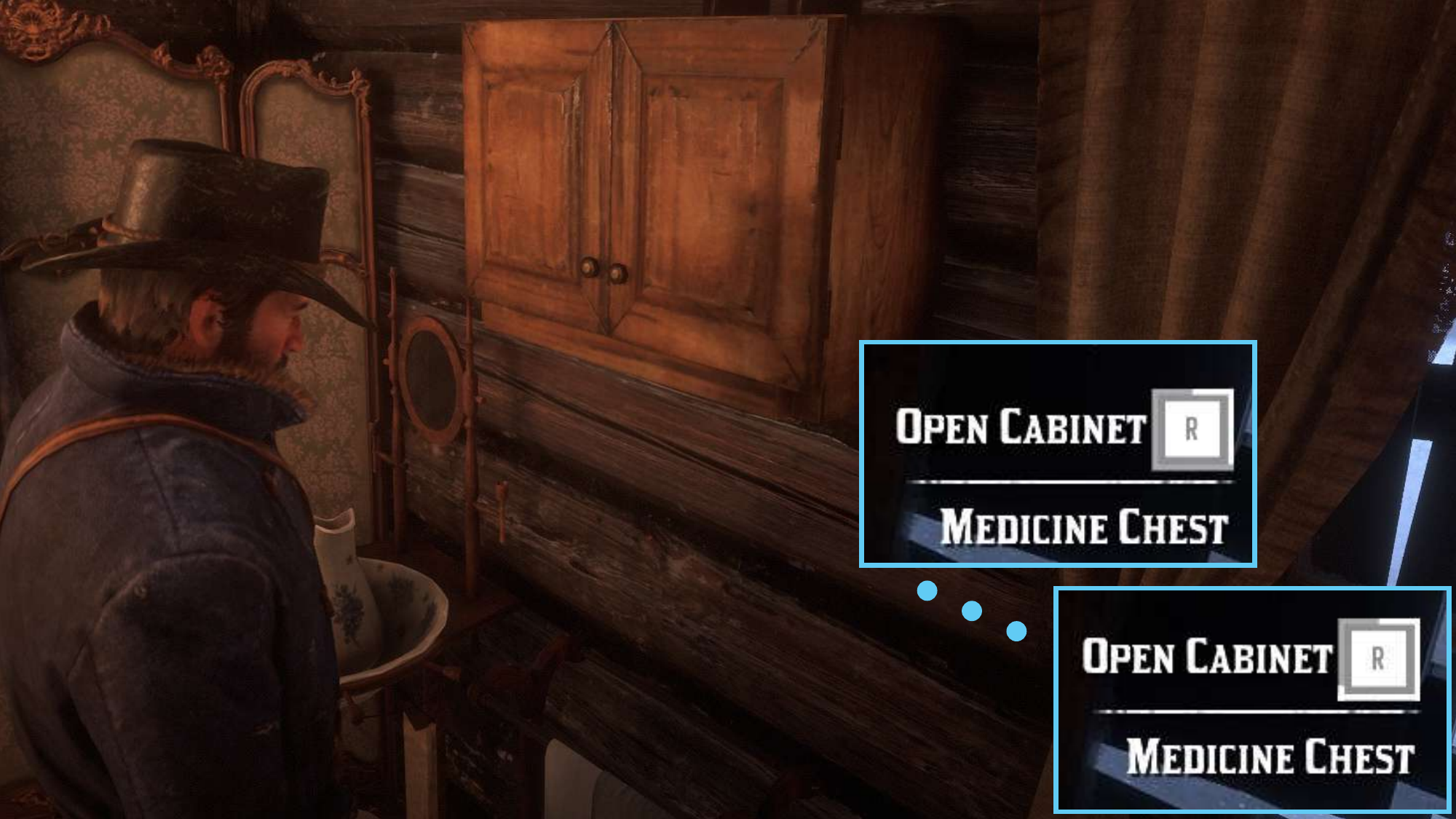}
    \caption{`Open cabinet' press timing} \label{fig:grounding_b}
  \end{subfigure}
  \begin{subfigure}{\subfigwidth\textwidth}
    \centering
    \includegraphics[width=\linewidth]{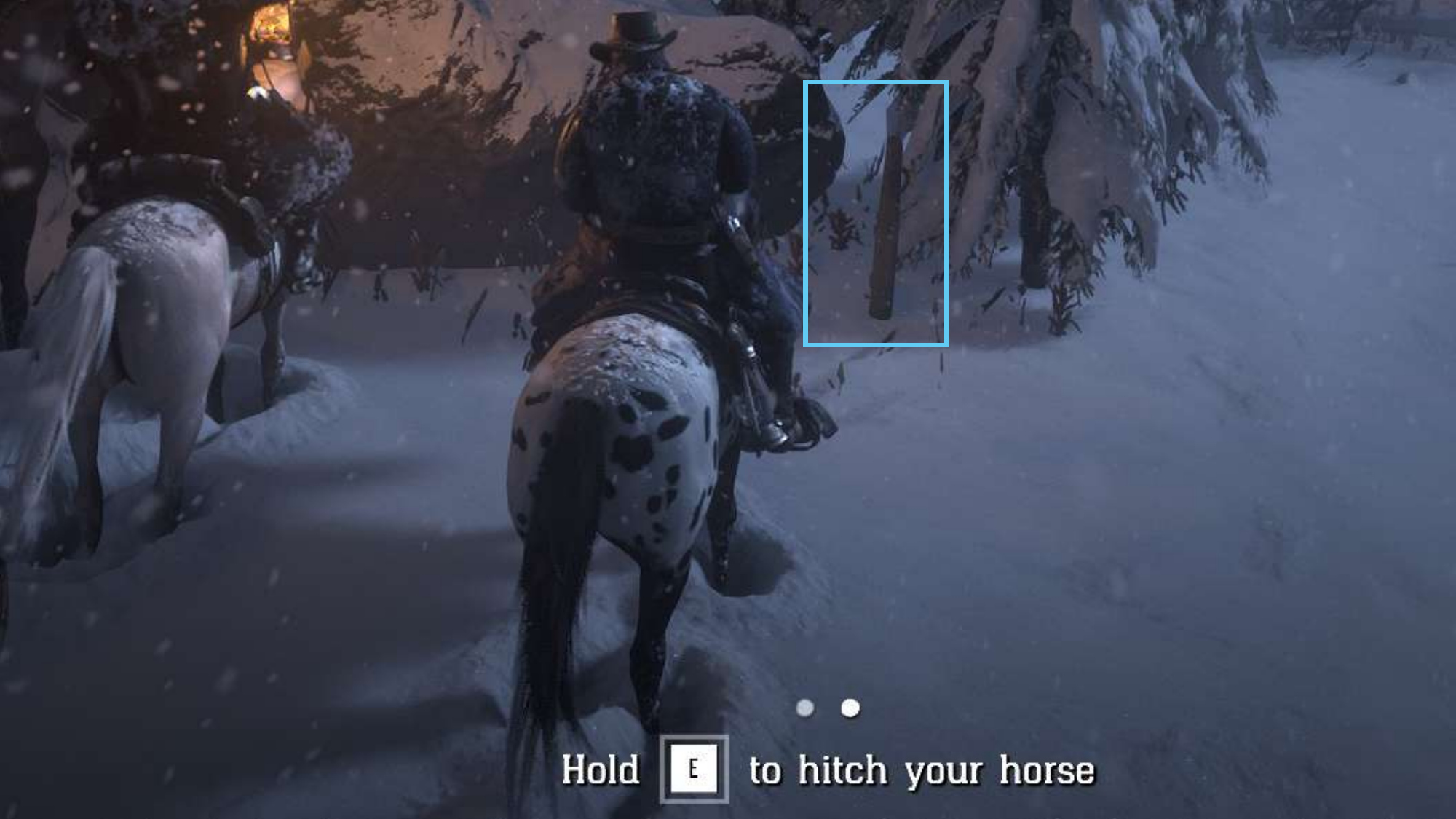}
    \caption{`Hitch horse' re-use of [E] key} \label{fig:grounding_c}
  \end{subfigure}
  \caption{Examples of action execution uncertainty. Lack of environmental feedback to actions and semantic gaps between action intent and game command can lead to challenging situations for agent reasoning.}
  \label{fig:env_feedback_issues}
\end{figure}

\newcommand{\widthofGPT}{0.95}
\begin{figure}[p]
  \centering
  \begin{subfigure}{\widthofGPT\textwidth}
    \centering
    \includegraphics[width=\textwidth]{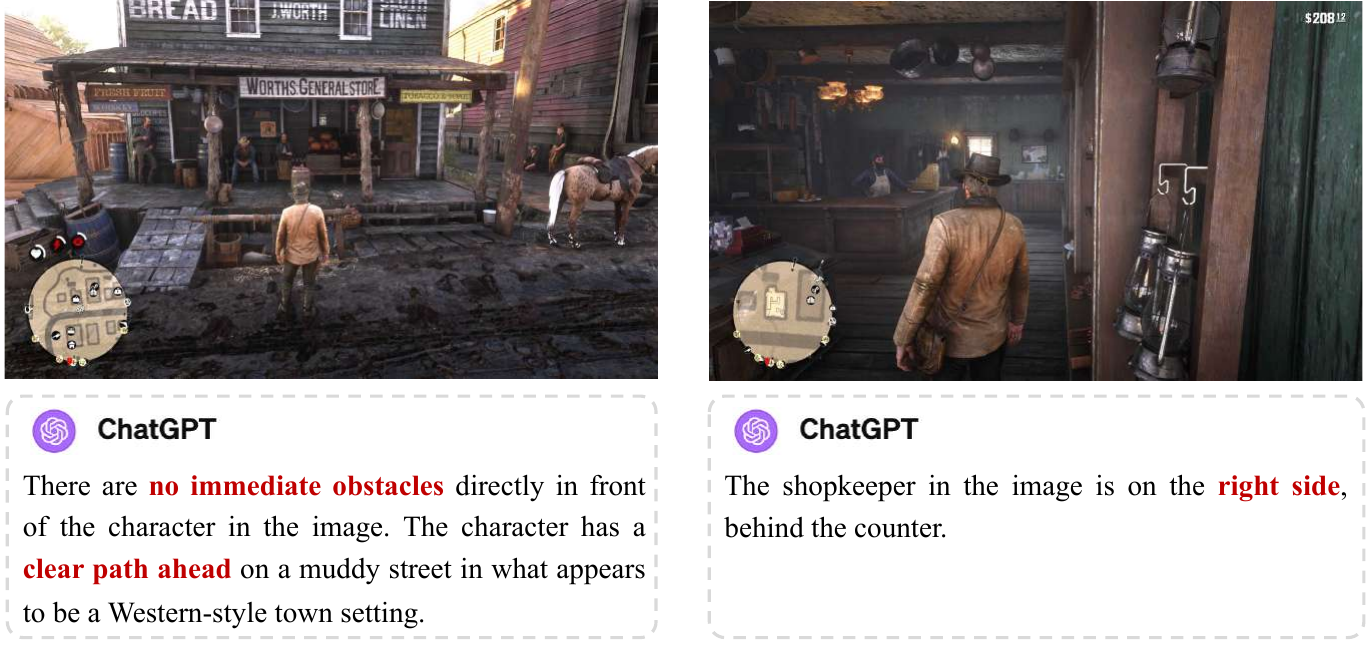}
    \caption{GPT-4V fails to recognize obstacles in the environment and the position of the shopkeeper.}
    \label{fig:GPT-4V_case_study_1}
    \vspace{0.5cm}
  \end{subfigure}
  \begin{subfigure}{\widthofGPT\textwidth}
    \centering
    \includegraphics[width=\textwidth]{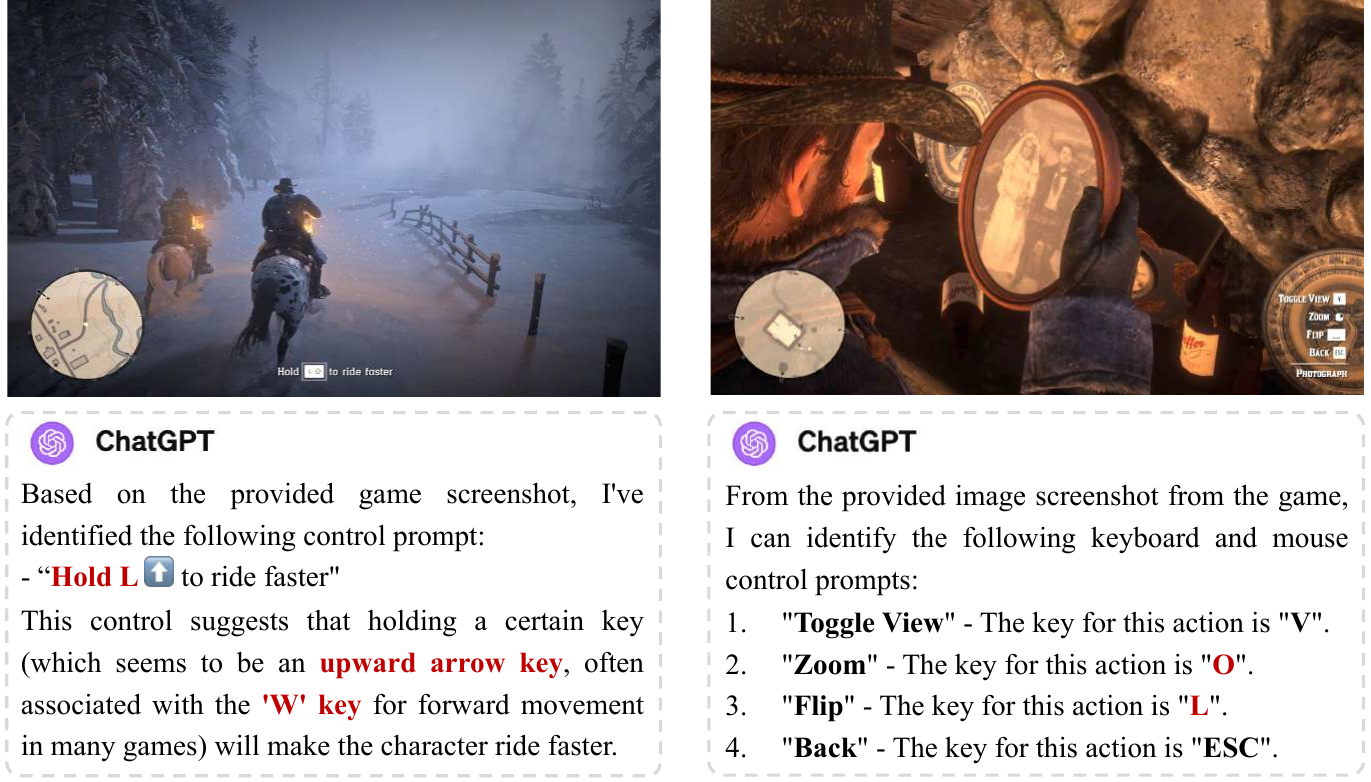}
    \caption{GPT-4V struggles to recognize the icons for keys on keyboard and mouse buttons. }
    \label{fig:GPT-4V_case_study_2}
    \vspace{0.5cm}
  \end{subfigure}
  \begin{subfigure}{\widthofGPT\textwidth}
    \centering
    \includegraphics[width=\textwidth]{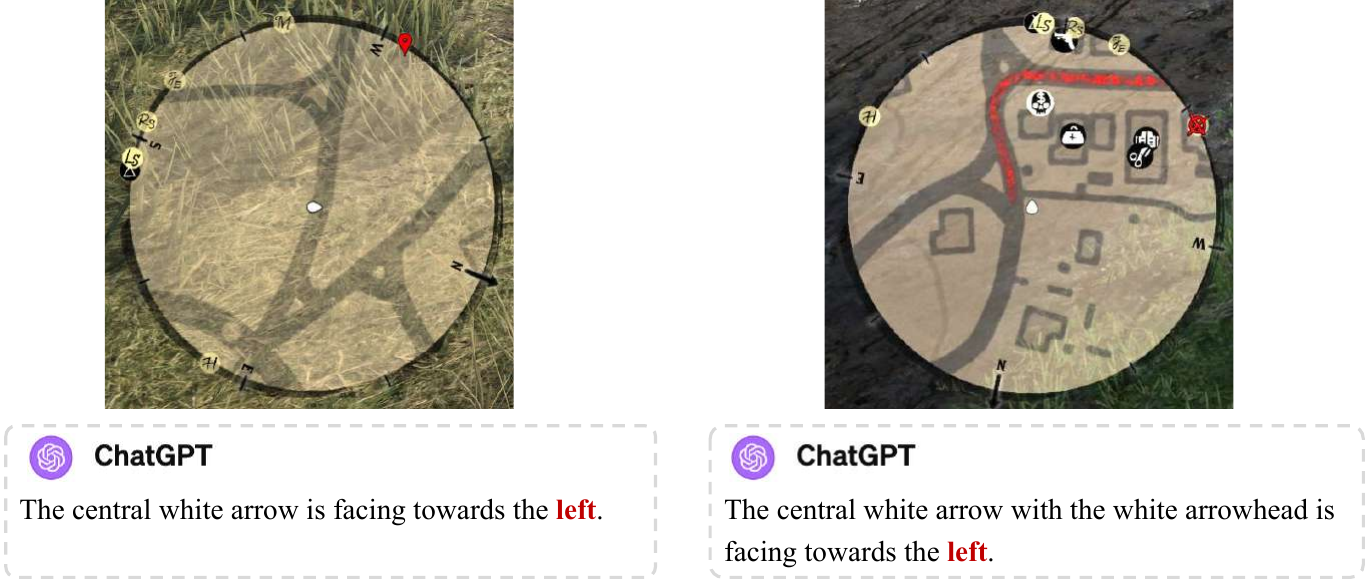}
    \caption{GPT-4V cannot understand the correct direction of arrow points, \ie character, towards in the mini-map.}
    \label{fig:GPT-4V_case_study_3}
  \end{subfigure}
  \caption{Example situations of GPT-4V's limitations in understanding visual information from the game.}
\end{figure}

\begin{figure}[p]
  \centering
  \begin{subfigure}{\widthofGPT\textwidth}
    \centering
    \includegraphics[width=\textwidth]{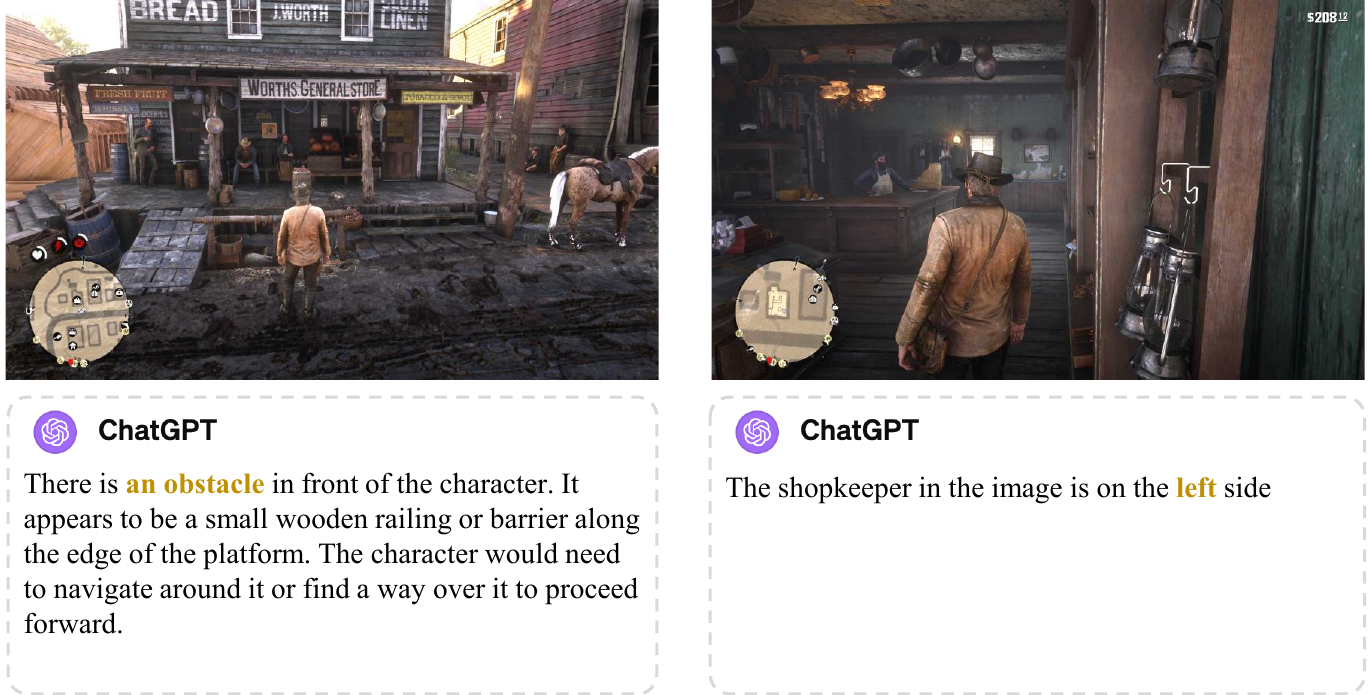}
    \caption{GPT-4o can recognize obstacles in the environment and the position of the shopkeeper.}
    \label{fig:GPT-4o_case_study_1}
  \end{subfigure}
  \begin{subfigure}{\widthofGPT\textwidth}
    \centering
    \includegraphics[width=\textwidth]{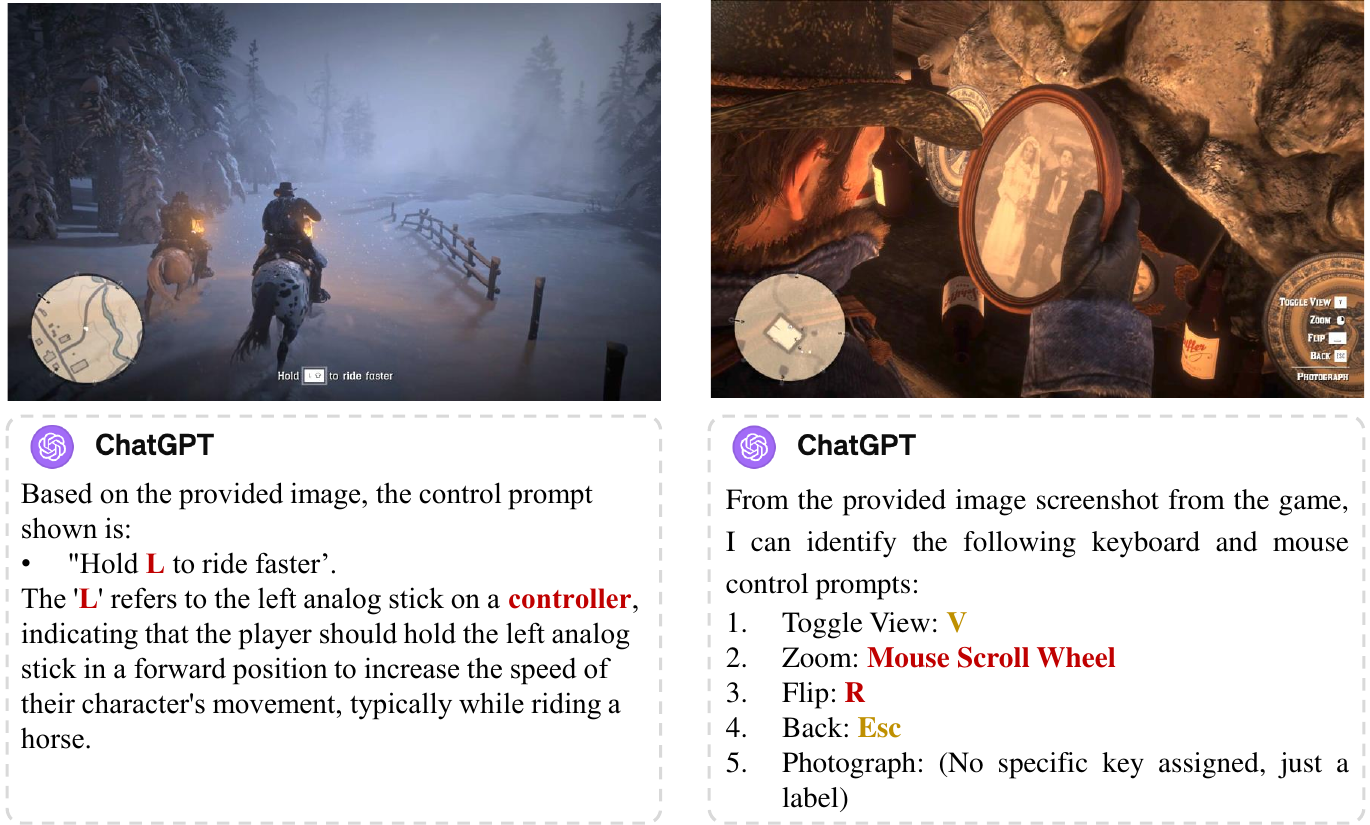}
    \caption{GPT-4o struggles to recognize the icons for keys on keyboard and mouse buttons. }
    \label{fig:GPT-4o_case_study_2}
  \end{subfigure}
  \begin{subfigure}{\widthofGPT\textwidth}
    \centering
    \includegraphics[width=\textwidth]{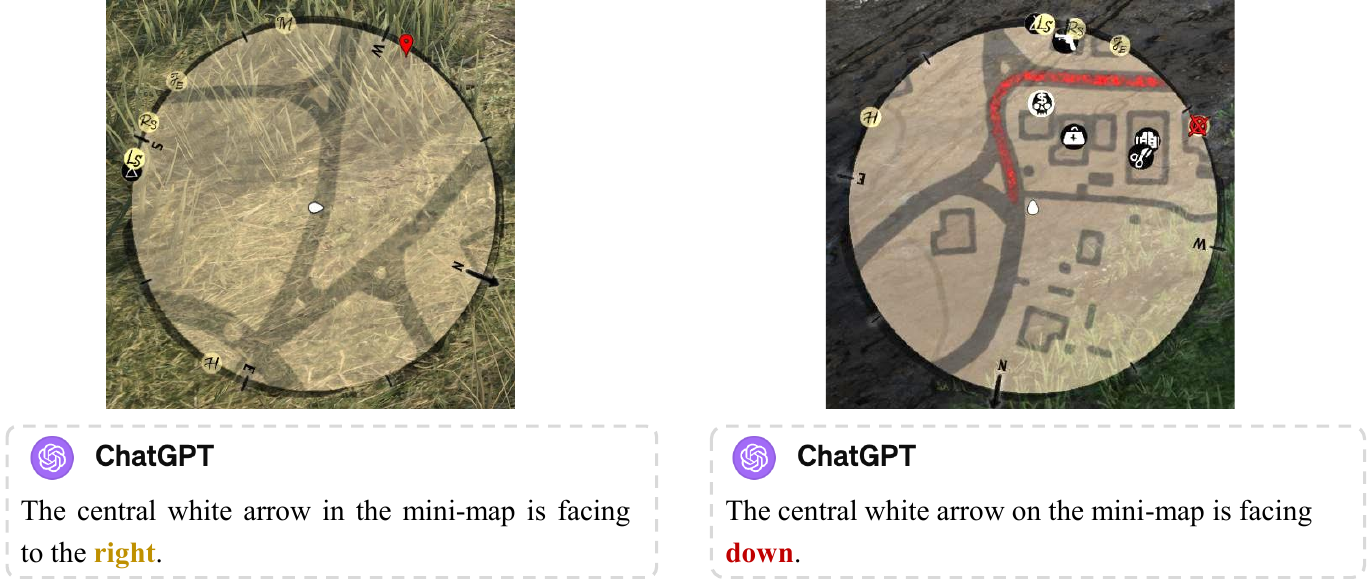}
    \caption{GPT-4o occasionally accurately determines the direction of arrow points, \ie character, towards in the mini-map.}
    \label{fig:GPT-4o_case_study_3}
  \end{subfigure}
  \caption{Example situations of GPT-4o's limitations in understanding visual information from the game.}
\end{figure}

\subsection{Limitations of GPT-4o and GPT-4V}
Deploying \projname in a complex game like RDR2 requires the backbone LMM model to handle multimodal input, which revealed several limitations of both GPT-4V and GPT-4o, necessitating external tools to enhance overall framework performance. Initial tests and exploration were performed using GPT-4V, as GPT-4o was not yet available. These tests highlighted significant weaknesses in spatial perception, icon understanding, history processing, and world understanding. Upon the release of GPT-4o, further testing demonstrated some notable improvements in spatial perception. However, enhancements in other areas remained marginal, while some regressions were also observed, all indicating the need for additional tools to aid decision-making.

\textbf{Spatial Perception.} As shown in Figure~\ref{fig:GPT-4V_case_study_1} and~\ref{fig:GPT-4o_case_study_1}, GPT-4V's spatial-visual recognition capability is insufficient for precise fine-grained control, particularly in detecting whether the character is being or going to be blocked and in estimating the accurate relative positions of target objects. In contrast, GPT-4o exhibits a significant enhancement in spatial perception, capable of recognizing obstacles ahead and estimating the approximate relative positions between objects. However, both models require supplementary information, such as bounding boxes of potential target objects, to make fine-grained decisions.
These led to the need to augment certain images to provide auxiliary visual clues for decision-making, \ie bounding boxes of possible target objects. 

\textbf{Icon Understanding.} Both GPT-4o and GPT-4V struggle with domain-specific concepts, such as unique icons within the game, which may represent specific targets or refer to certain mouse and key actions. As shown in Figure~\ref{fig:GPT-4V_case_study_2} and~\ref{fig:GPT-4o_case_study_2}, GPT-4V and GPT-4o fail to recognize the left shift, right mouse button, and space icons. Attempts to incorporate few-shot learning to improve image understanding cannot be generalized. Therefore, we match prepared pattern templates, \eg icon images, against each screenshot to continuously detect and highlight any appearing icons.

Figure~\ref{fig:GPT-4V_case_study_3}
 and Figure~\ref{fig:GPT-4o_case_study_3}  also demonstrate that although GPT-4o performs better than GPT-4V in understanding the mini-map, it still fails to consistently interpret this crucial information regarding the position and direction of the character. This failure in localization leads the agent to sometimes get lost in the town and miss the task target. While the aforementioned issues can be slightly alleviated by providing additional few-shot examples, a significant improvement is only achieved by cropping the image and providing GPT-4o with the exact region containing the icon to be recognized. This dependency on precise input makes the issue challenging and unreliable for decision-making. Although the above issues can be slightly alleviated by providing additional few-shot examples, it can only have an obvious effect if we crop the image and provide the GPT-4o with the region exactly containing the icon to be recognized, which makes the issue intractable.

\textbf{History Processing.} Moreover, both GPT-4o and GPT-4V can easily get distracted by irrelevant information in longer contexts, resulting in hallucinations. For example, when action planning utilizes too many historical screenshots, they may confuse past and present frames. Additionally, performance fluctuates and both model versions frequently generate output not adhering to the rules in the provided prompts. To mitigate the issue of hallucinations, we more strictly control input information by further summarizing long-term memory.

\textbf{World Understanding.} Lastly, the absence of an RDR2 world model limits GPT-4V and GPT-4o's understanding of the consequences of its actions in the game. This often results in inappropriate action selection, such as overestimating the necessary adjustments for aligning targets or misjudging the duration required for certain actions. To alleviate this problem, we introduced extra prompt rules regarding action parameters and more flexibility into the self-reflection module.

%% file: rdr2_figure/mission1.tex
\begin{figure}[ht]
  \centering
  \begin{subfigure}{0.32\textwidth}
    \centering
    \includegraphics[width=\linewidth]{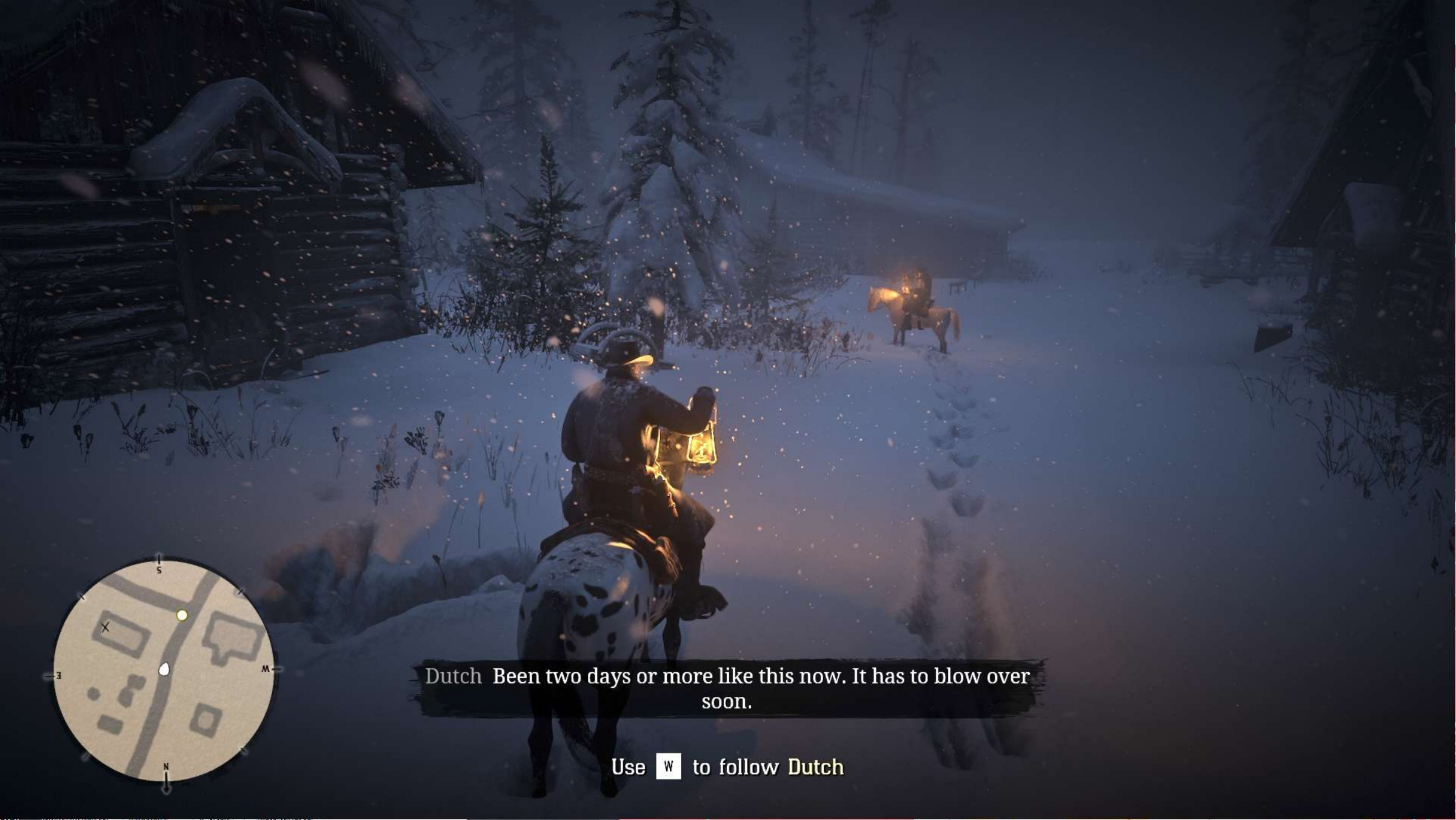}
    \caption{Follow Dutch} 
    \label{subfig:M01_C01_Follow_Dutch}
  \end{subfigure}
  \begin{subfigure}{0.32\textwidth}
    \centering
    \includegraphics[width=\linewidth]{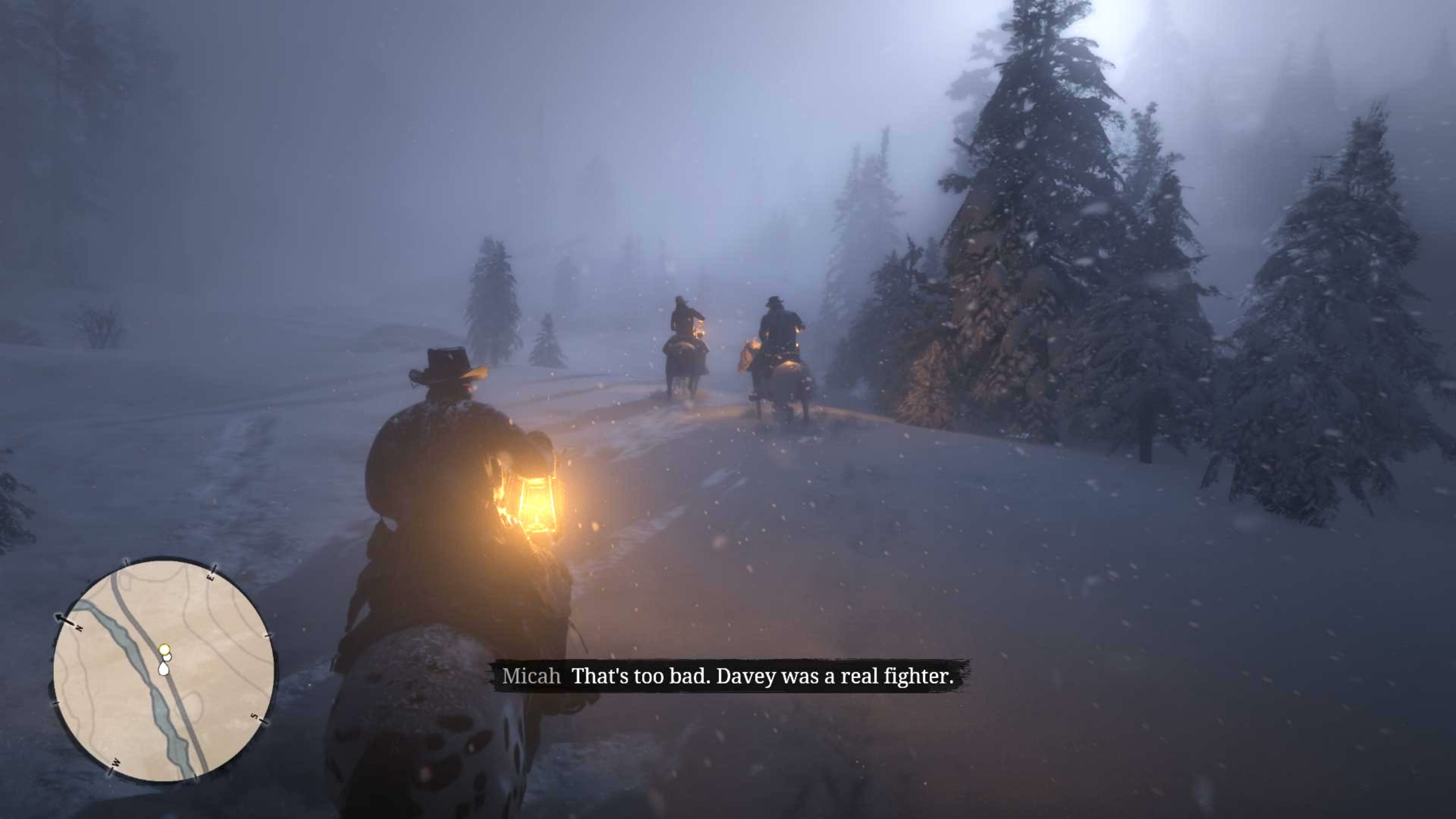} 
    \caption{Go to Town} 
    \label{subfig:M01_C02_Follow_Micah}
  \end{subfigure}
  \begin{subfigure}{0.32\textwidth}
    \centering
    \includegraphics[width=\linewidth]{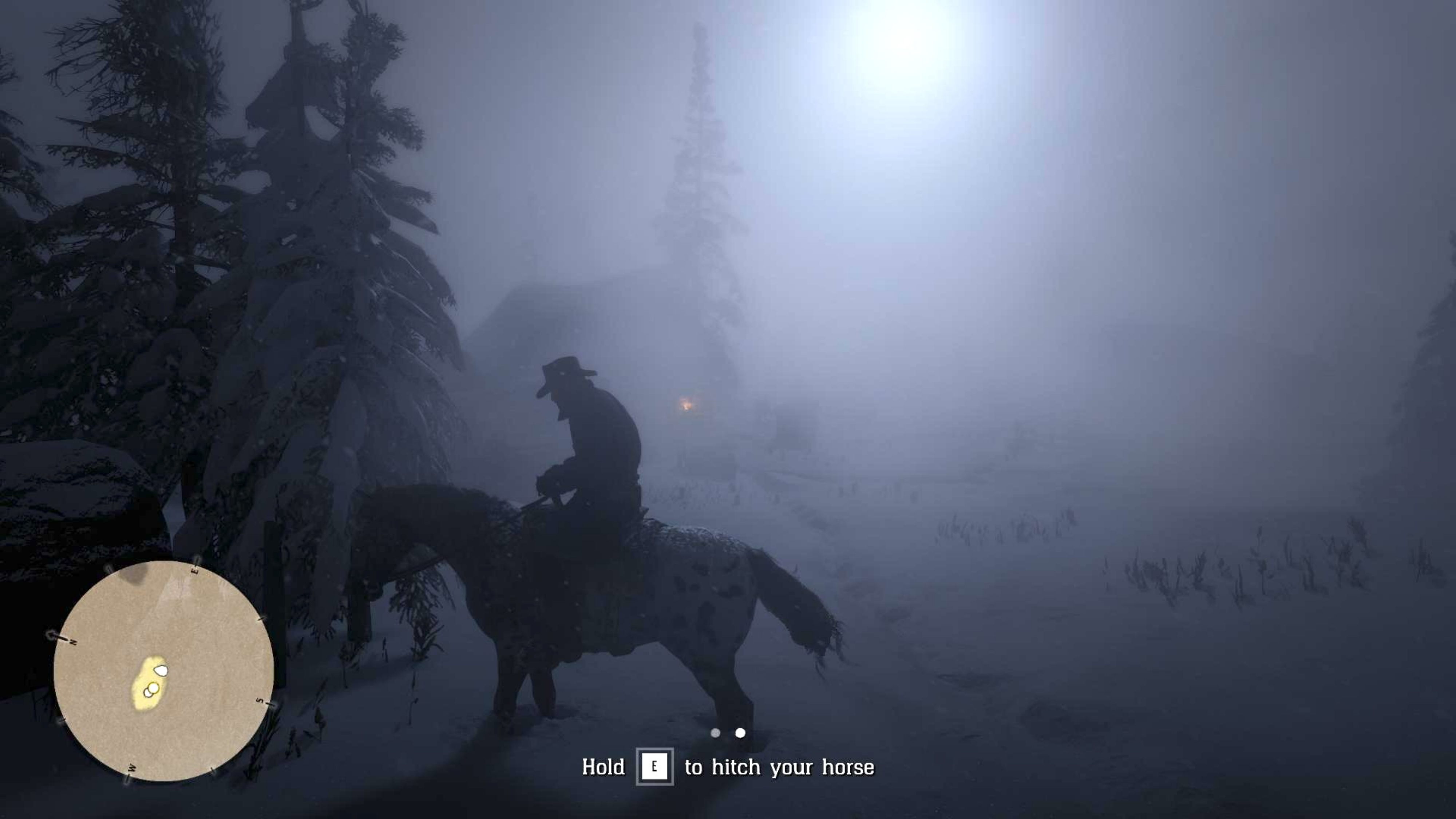}
    \caption{Hitch Horse} 
    \label{subfig:M01_C03_Hitch_Horse}
  \end{subfigure}
  \begin{subfigure}{0.32\textwidth}
    \centering
    \includegraphics[width=\linewidth]{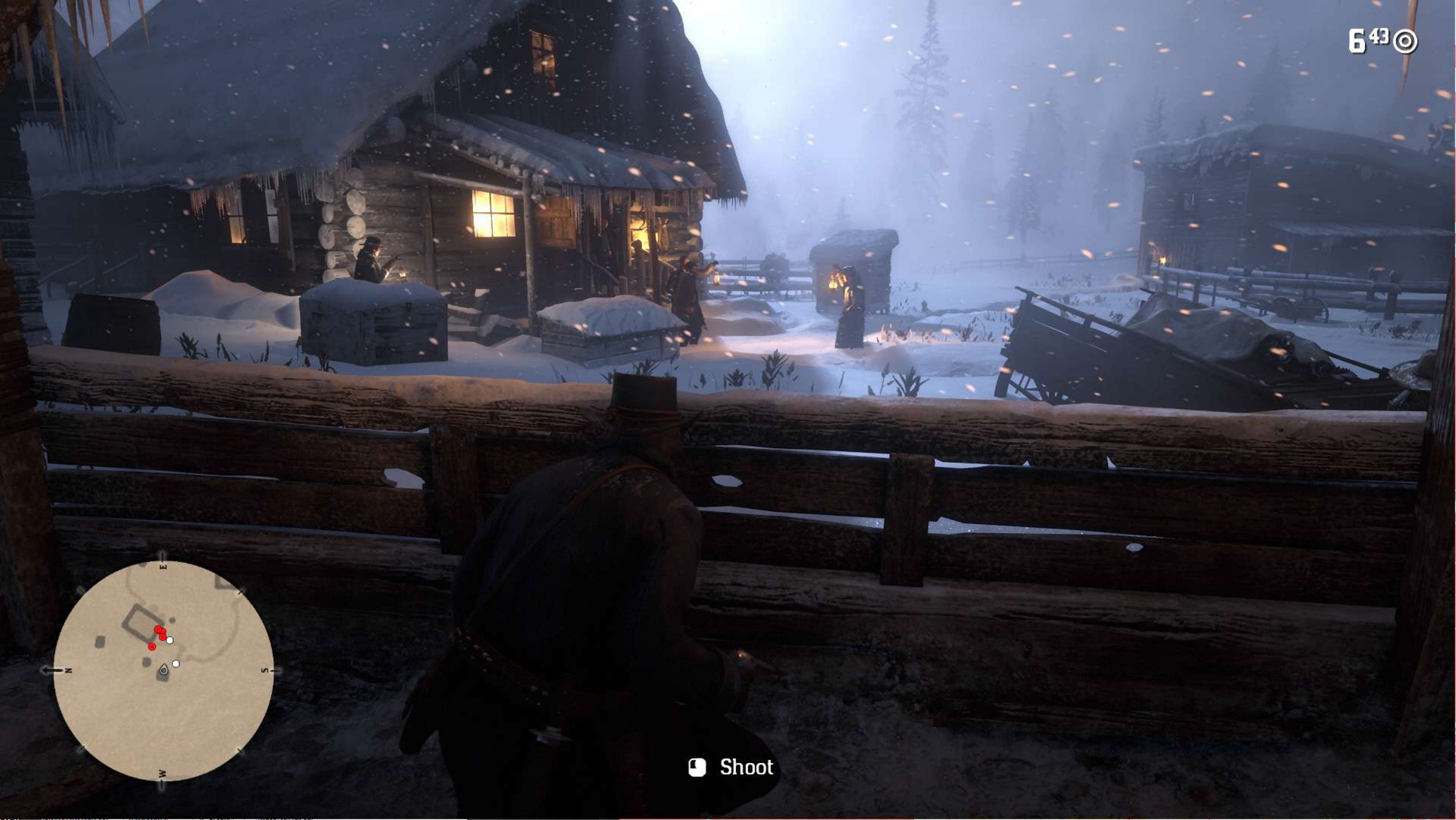}
    \caption{Protect Dutch} 
    \label{subfig:M01_C04_Protect_Dutch}
  \end{subfigure}
  \begin{subfigure}{0.32\textwidth}
    \centering
    \includegraphics[width=\linewidth]{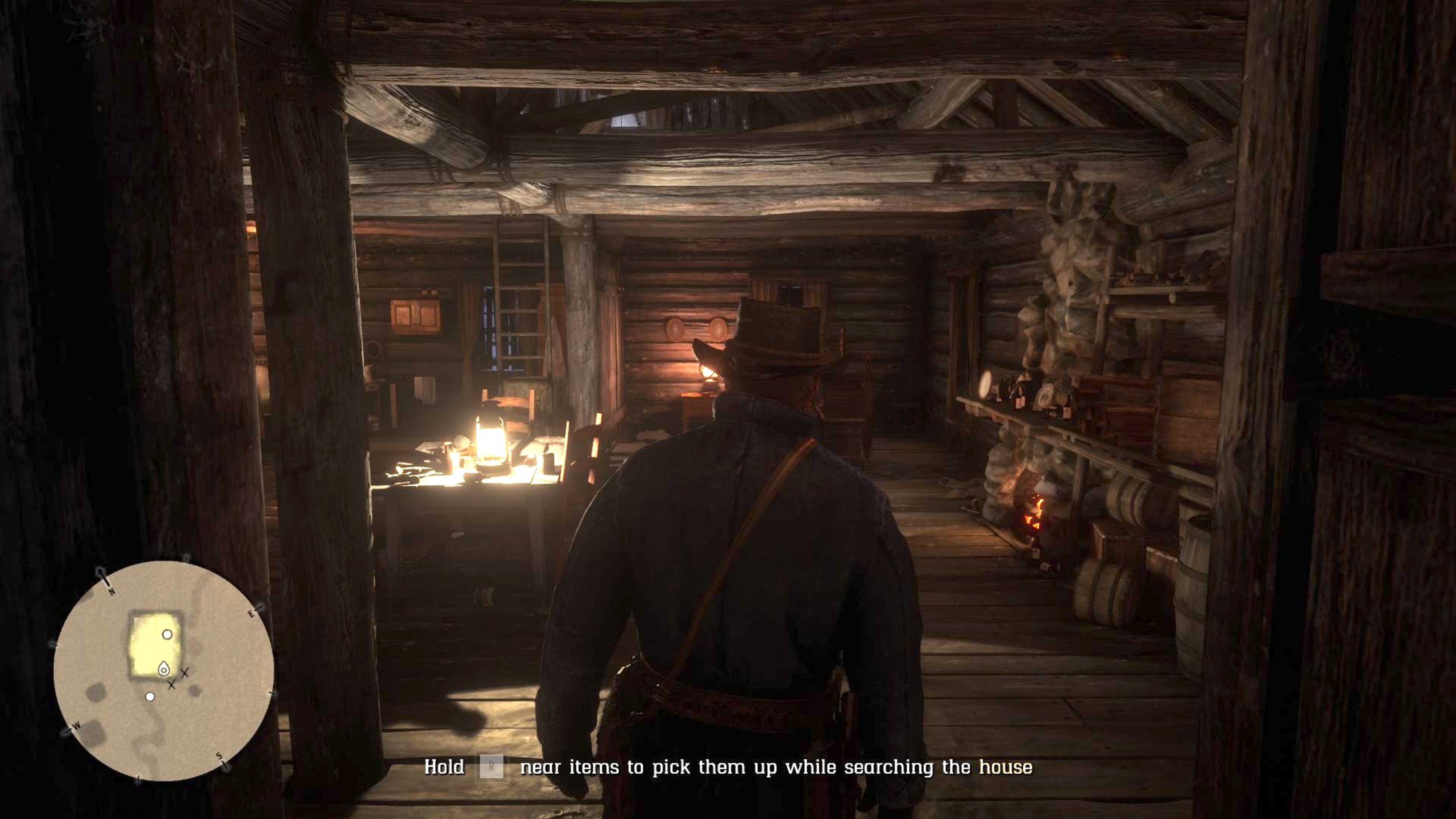}
    \caption{Search for Supplies} \label{subfig:M01_C05_Search_for_Supplies}
  \end{subfigure}
  \begin{subfigure}{0.32\textwidth}
    \centering
    \includegraphics[width=\linewidth]{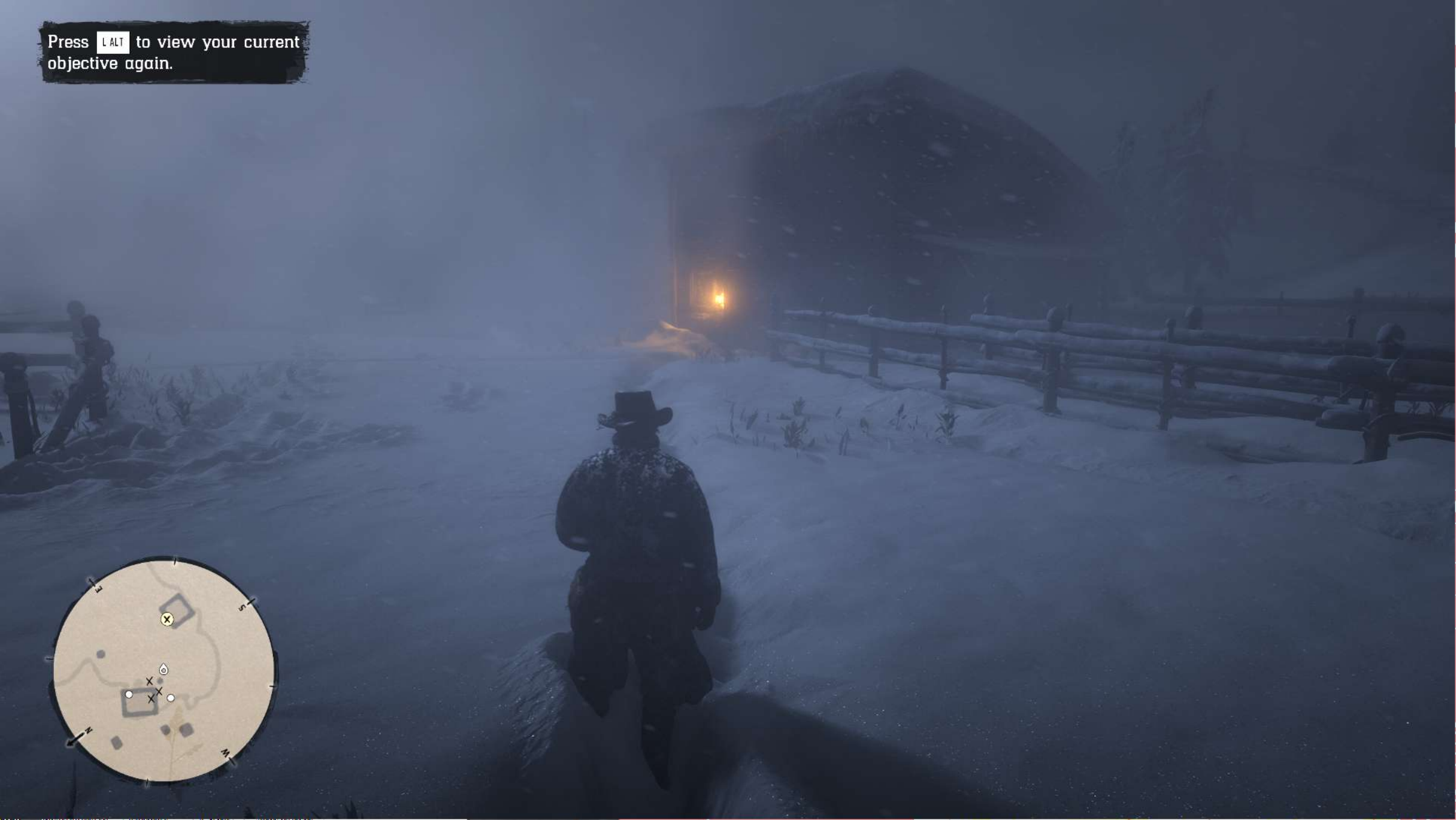}
    \caption{Go to Barn} \label{subfig:M01_C06_Go_to_the_Barn}
  \end{subfigure}
  \begin{subfigure}{0.32\textwidth}
    \centering
    \includegraphics[width=\linewidth]{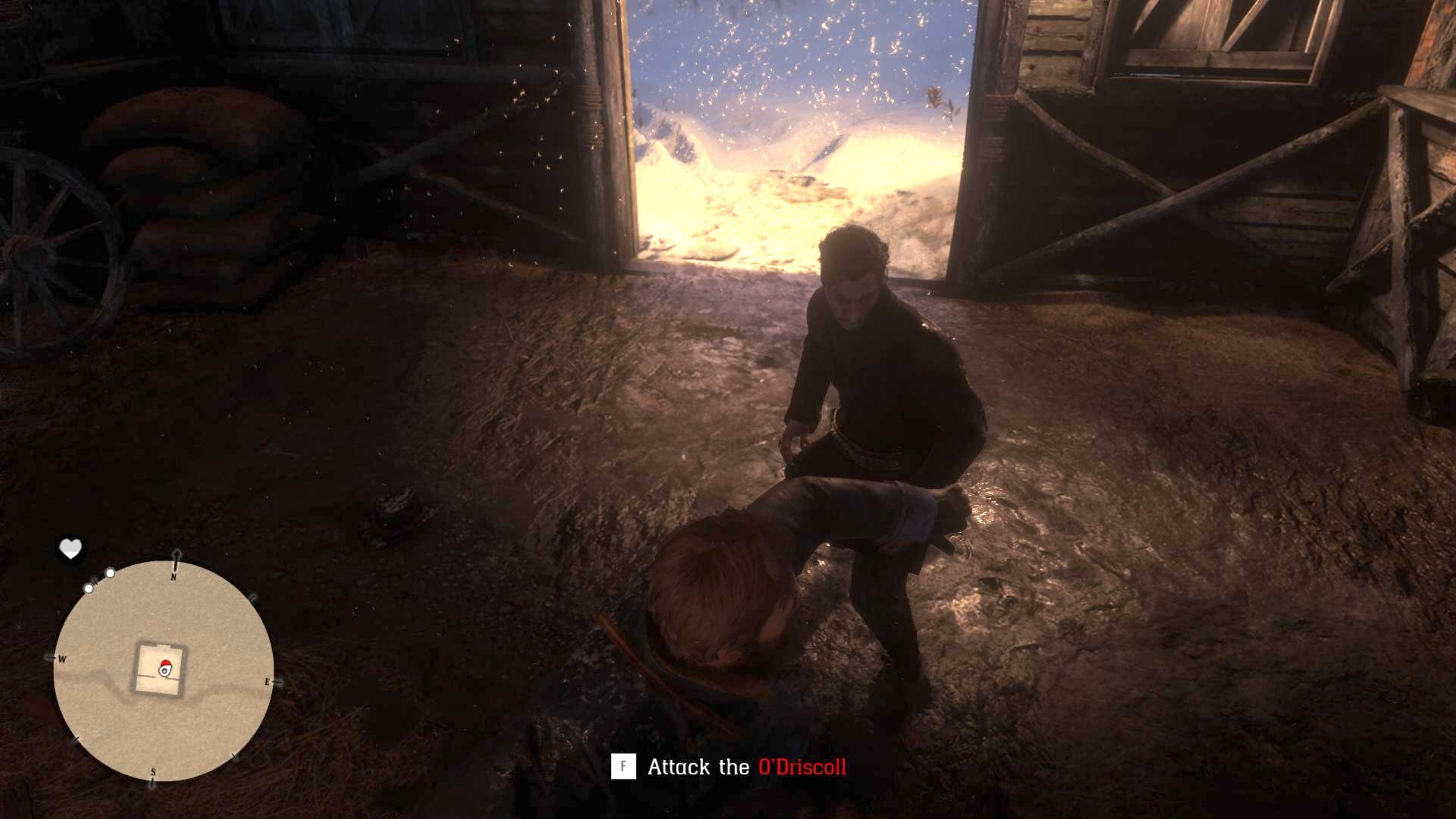}
    \caption{Search Barn} \label{subfig:M01_C07_Search_the_Barn}
  \end{subfigure}
  \begin{subfigure}{0.32\textwidth}
    \centering
    \includegraphics[width=\linewidth]{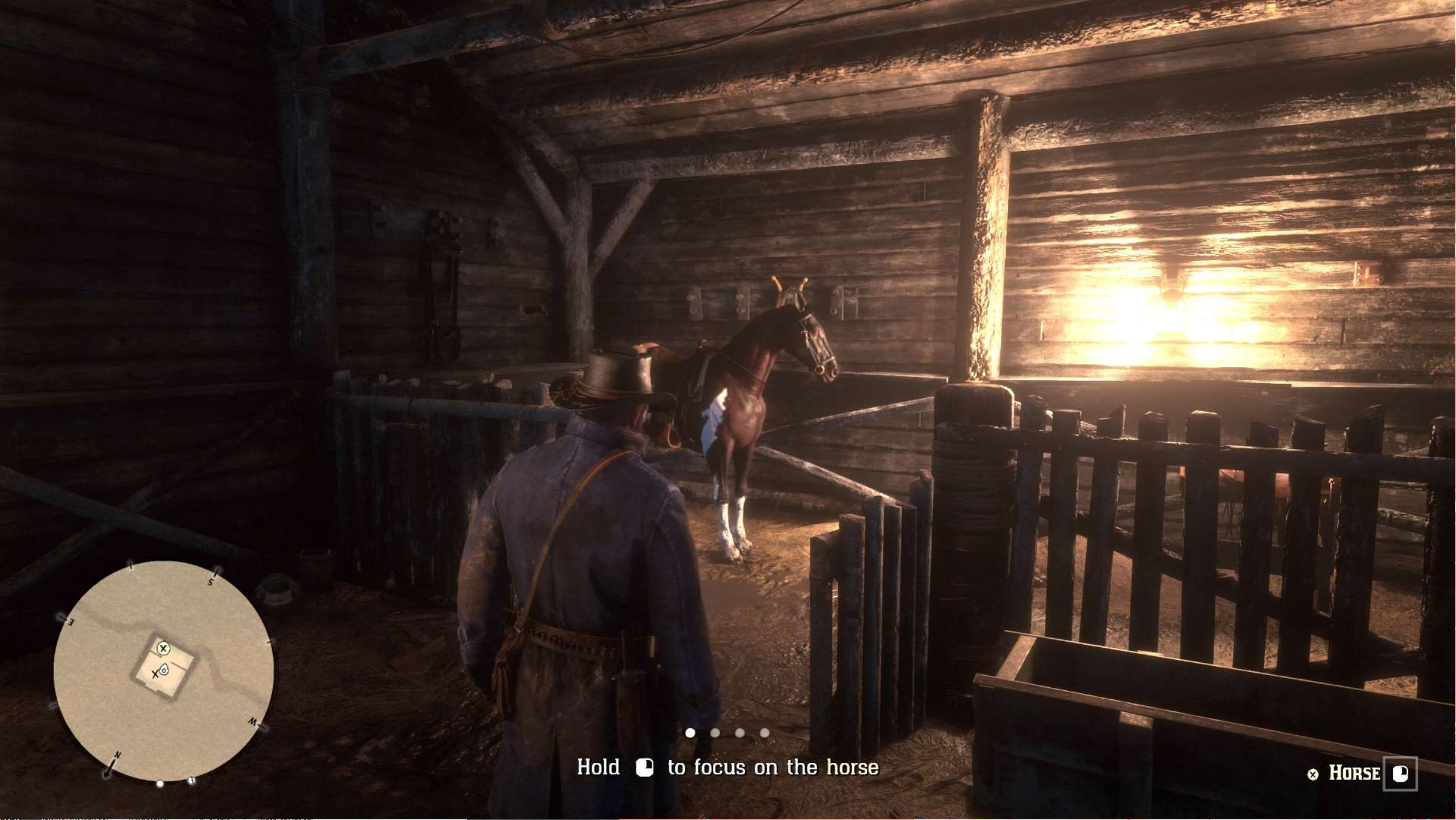}
    \caption{Lead Horse} \label{subfig:M01_C08_Lead_the_Horse}
  \end{subfigure}
  \caption{Image examples of tasks in the first mission of \emph{Outlaws from the West}. (The picture has been brightened for easier reading.)} 
  \label{fig:app_mission1_tasks}
\end{figure}

%% file: rdr2_figure/mission2.tex
\begin{figure}[htbp]
  \centering
    \begin{subfigure}{0.32\textwidth}
    \centering
    \includegraphics[width=\linewidth]{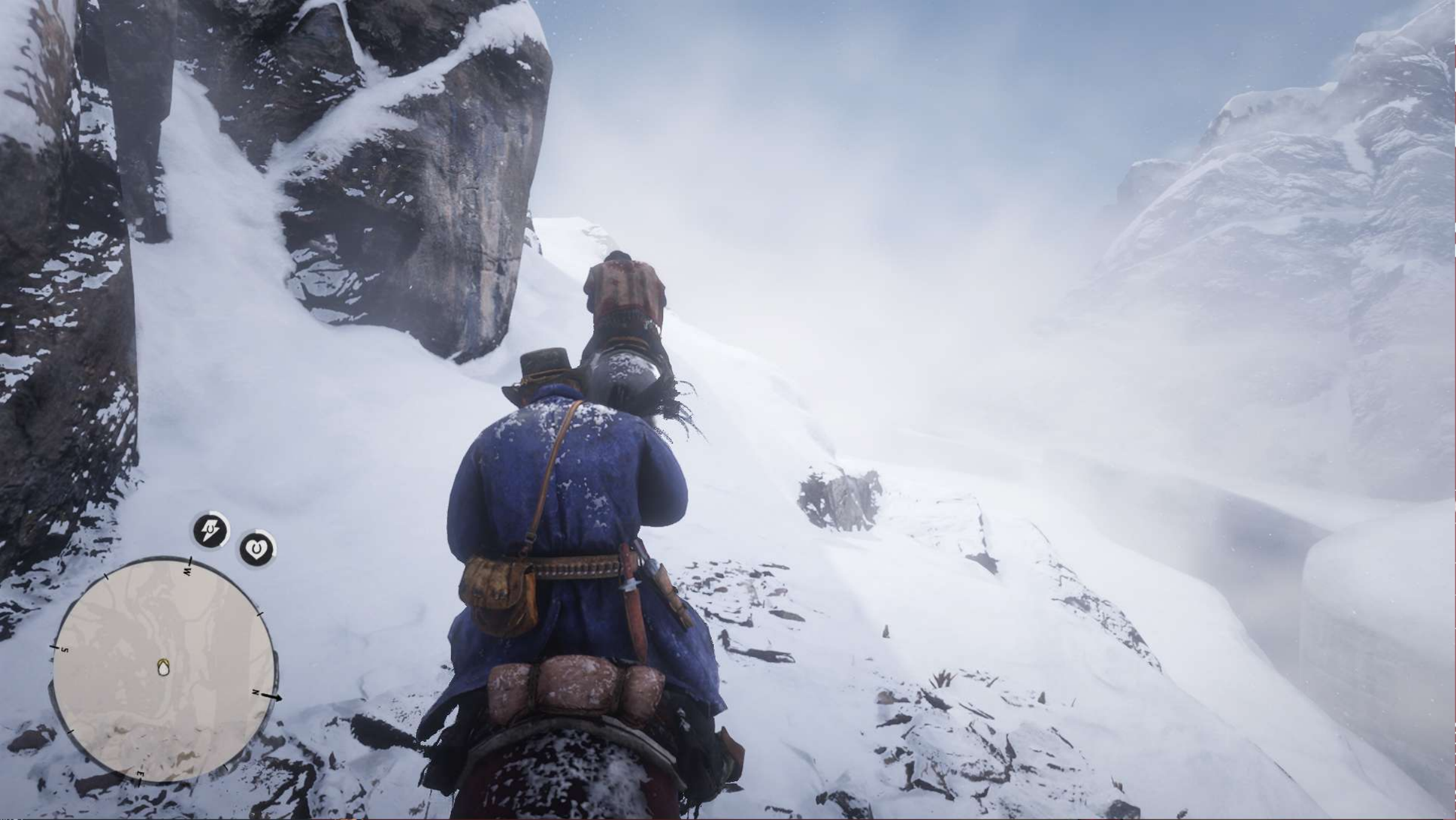}
    \caption{Follow Javier} \label{subfig:M02_C01_Follow_Javier}
  \end{subfigure}
  \begin{subfigure}{0.32\textwidth}
    \centering
    \includegraphics[width=\linewidth]{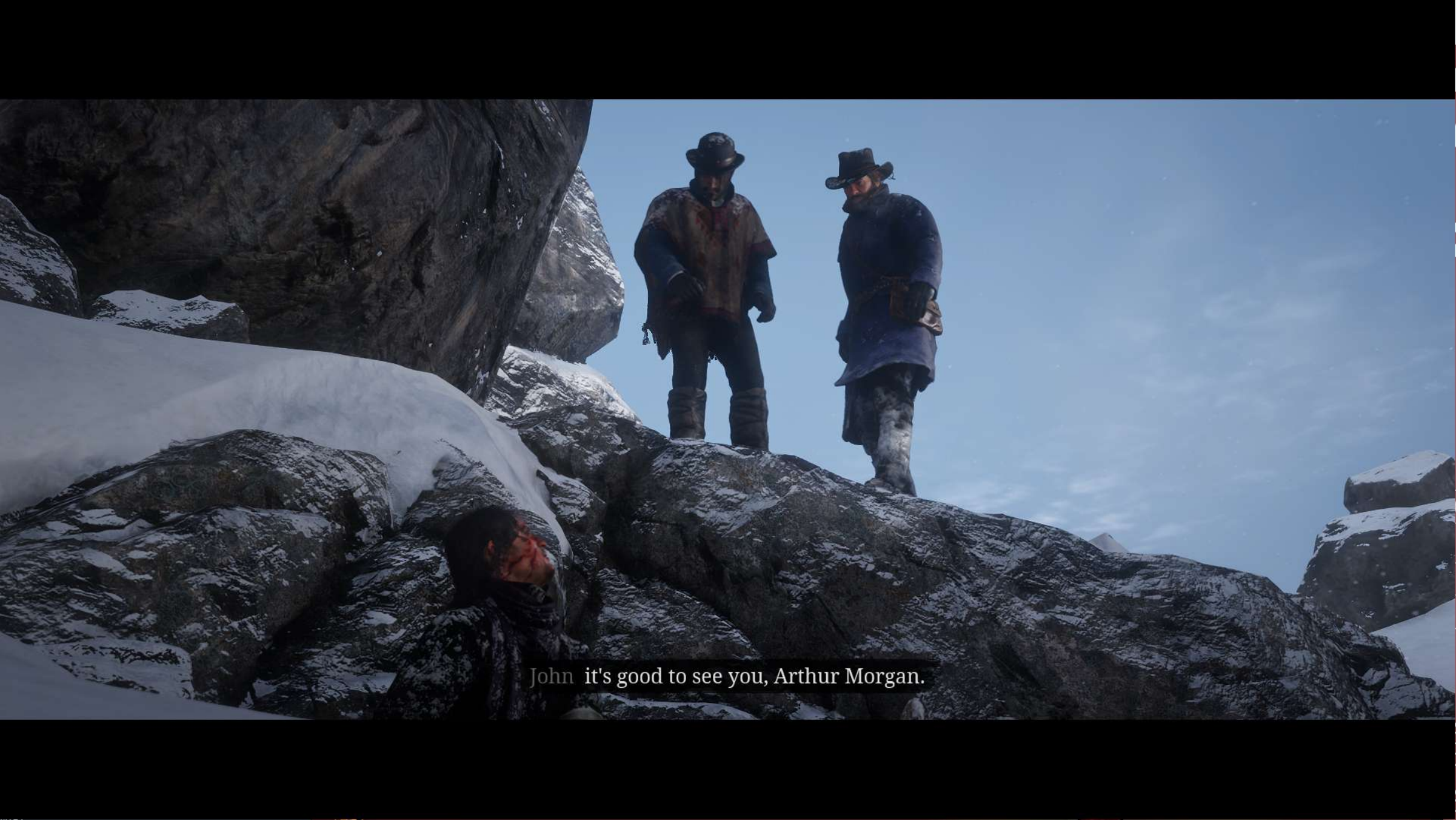}
    \caption{Search John} \label{subfig:M02_C02_Find_John}
  \end{subfigure}
  \begin{subfigure}{0.32\textwidth}
    \centering
    \includegraphics[width=\linewidth]{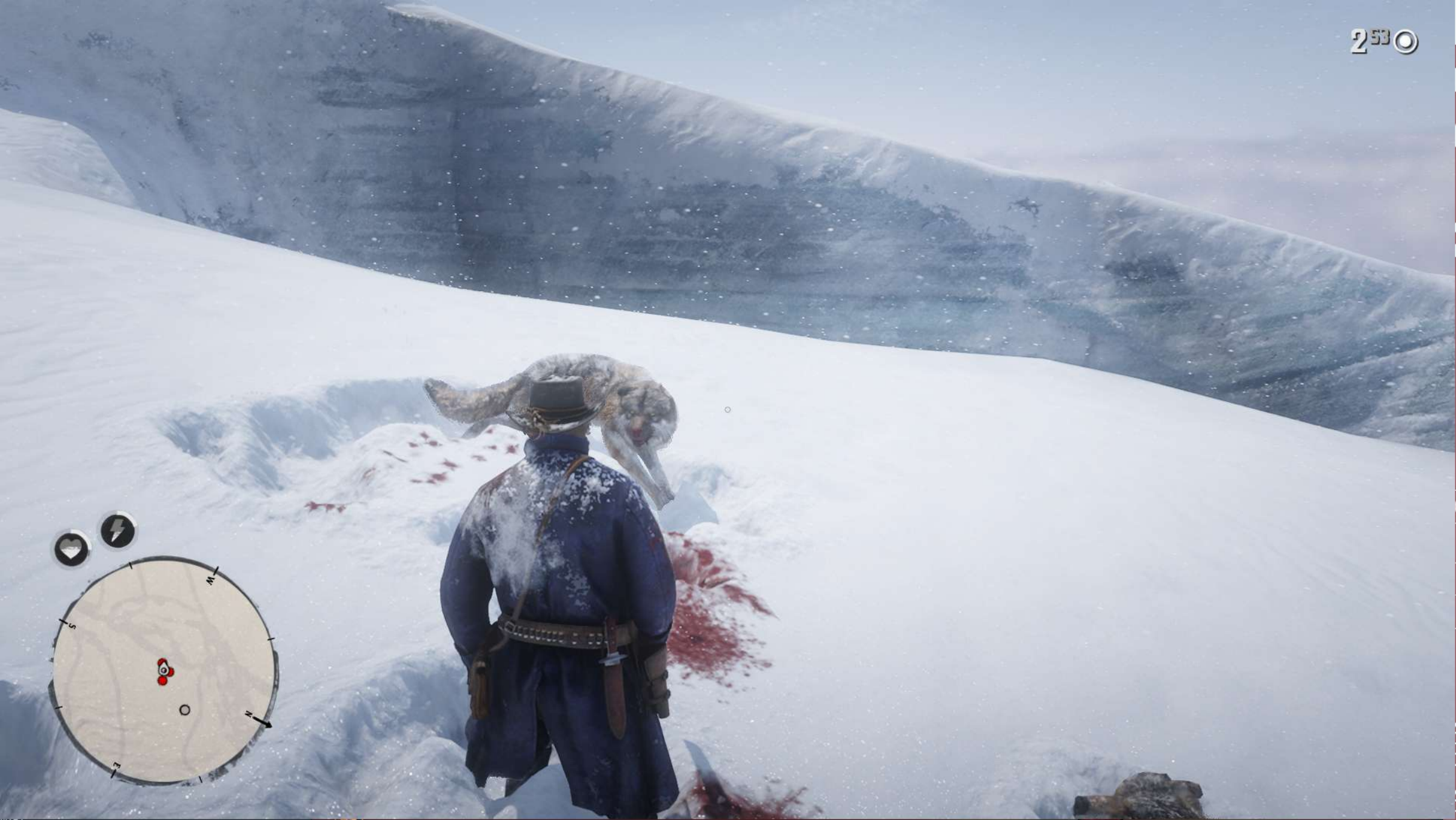}
    \caption{Keep Wolves Away} \label{subfig:M02_C03_Shoot_Wolves}
  \end{subfigure}
  \begin{subfigure}{0.32\textwidth}
    \centering
    \includegraphics[width=\linewidth]{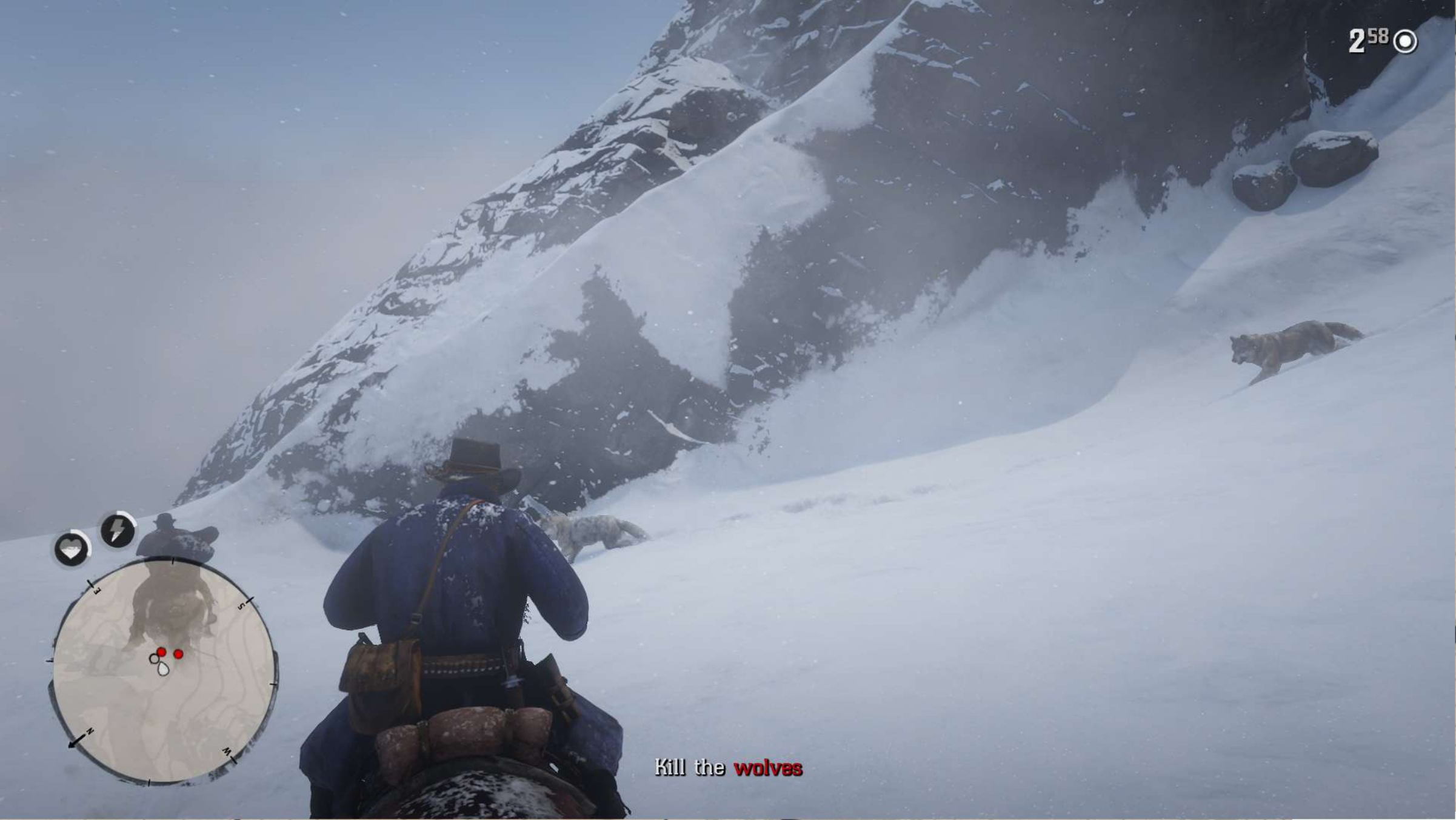}
    \caption{Kill Wolves} \label{subfig:M02_C04_Eliminate_Wolves}
  \end{subfigure}
  \begin{subfigure}{0.32\textwidth}
    \centering
    \includegraphics[width=\linewidth]{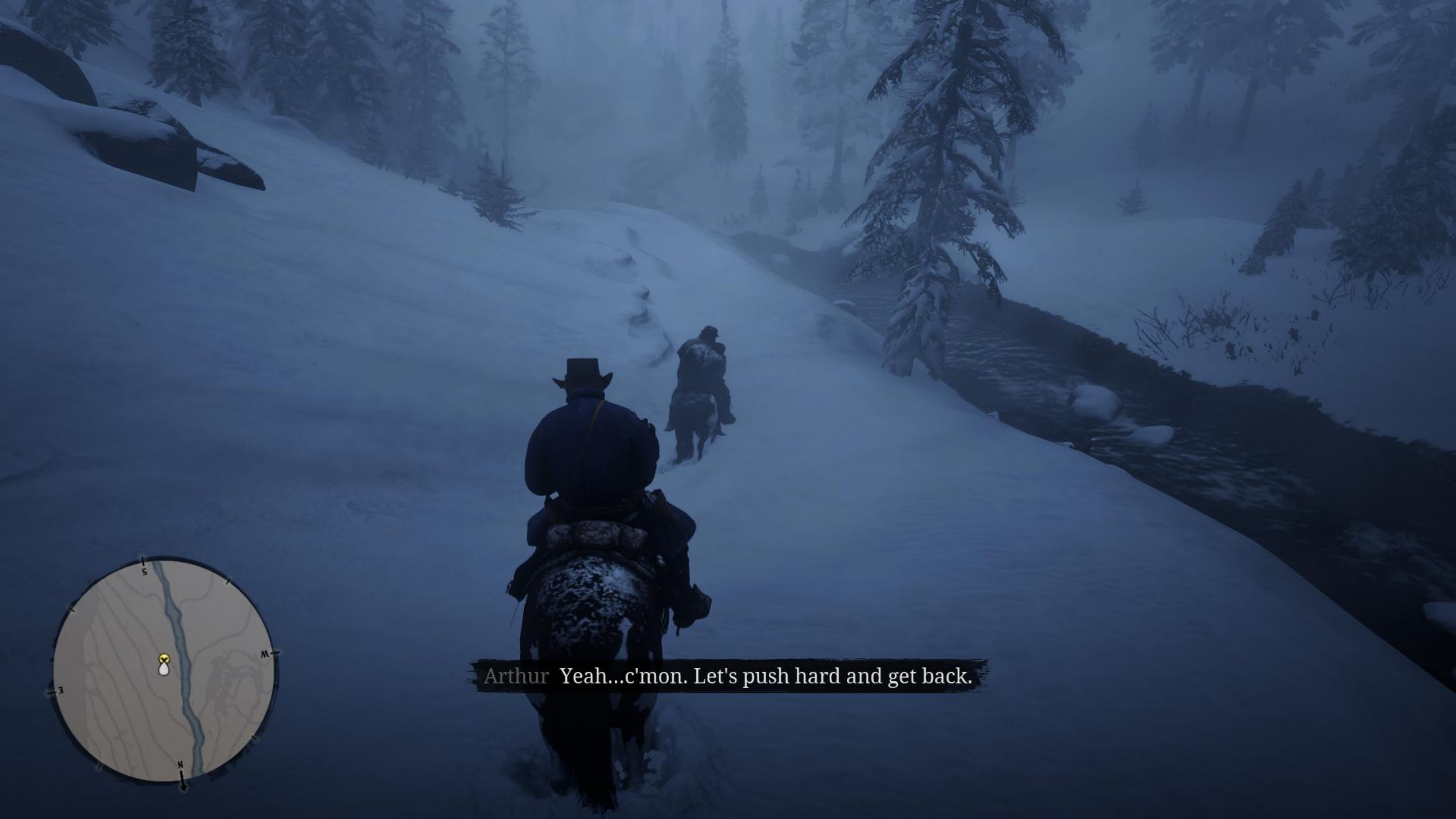}
    \caption{Return to Camp} \label{subfig:M02_C05_Return_to_Camp}
  \end{subfigure}
  \caption{Image examples of tasks in the second mission of \emph{Enter, Pursued by a Memory}.} \label{fig:app_mission2_tasks}
\end{figure}

%% file: rdr2_figure/mission3.tex
\begin{figure}[htbp]
  \centering
  \begin{subfigure}{0.32\textwidth}
    \centering
    \includegraphics[width=\linewidth]{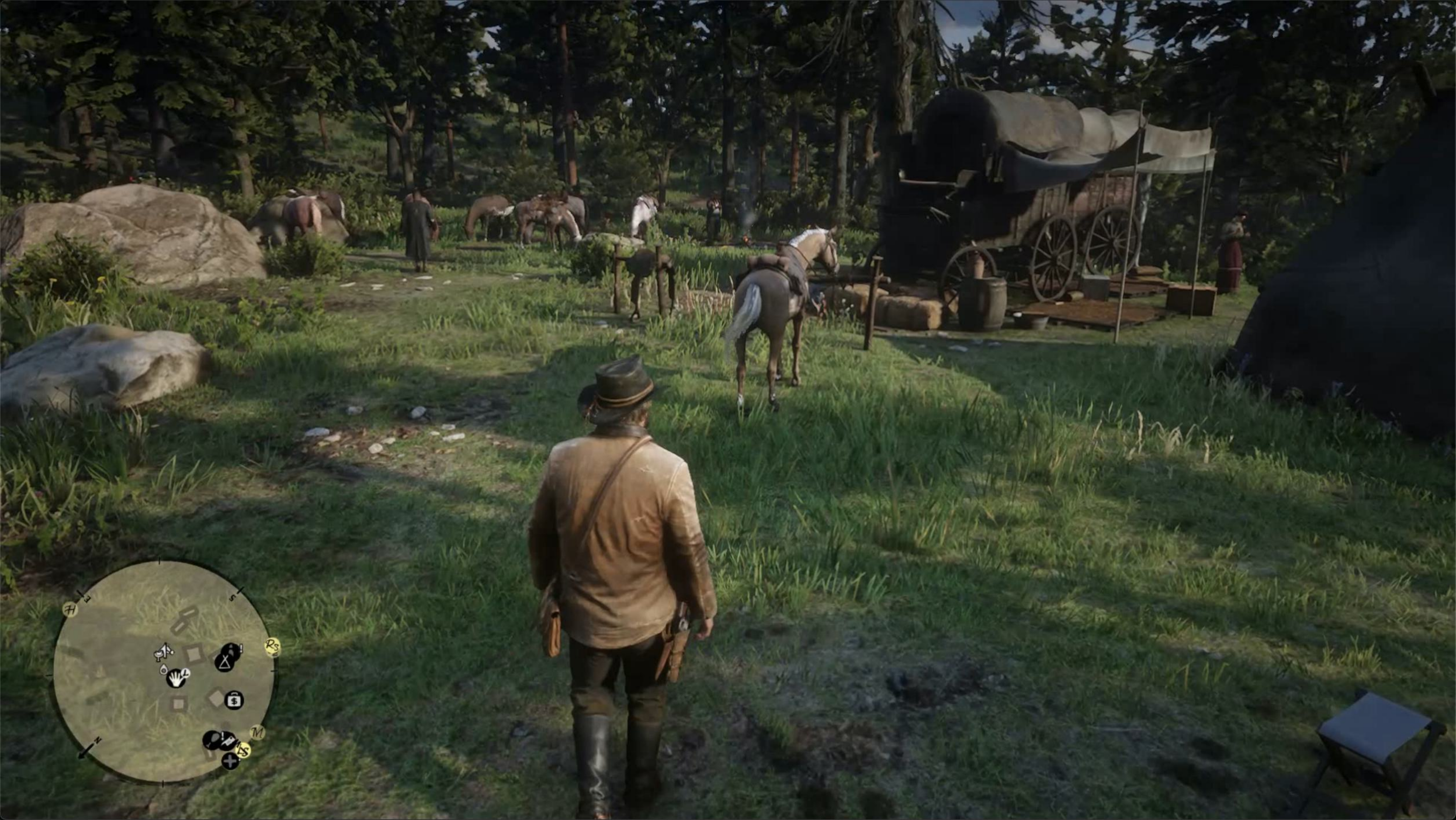}
    \caption{Find Horse} \label{subfig:M03_C01_Find_Horse}
  \end{subfigure}
  \begin{subfigure}{0.32\textwidth}
    \centering
    \includegraphics[width=\linewidth]{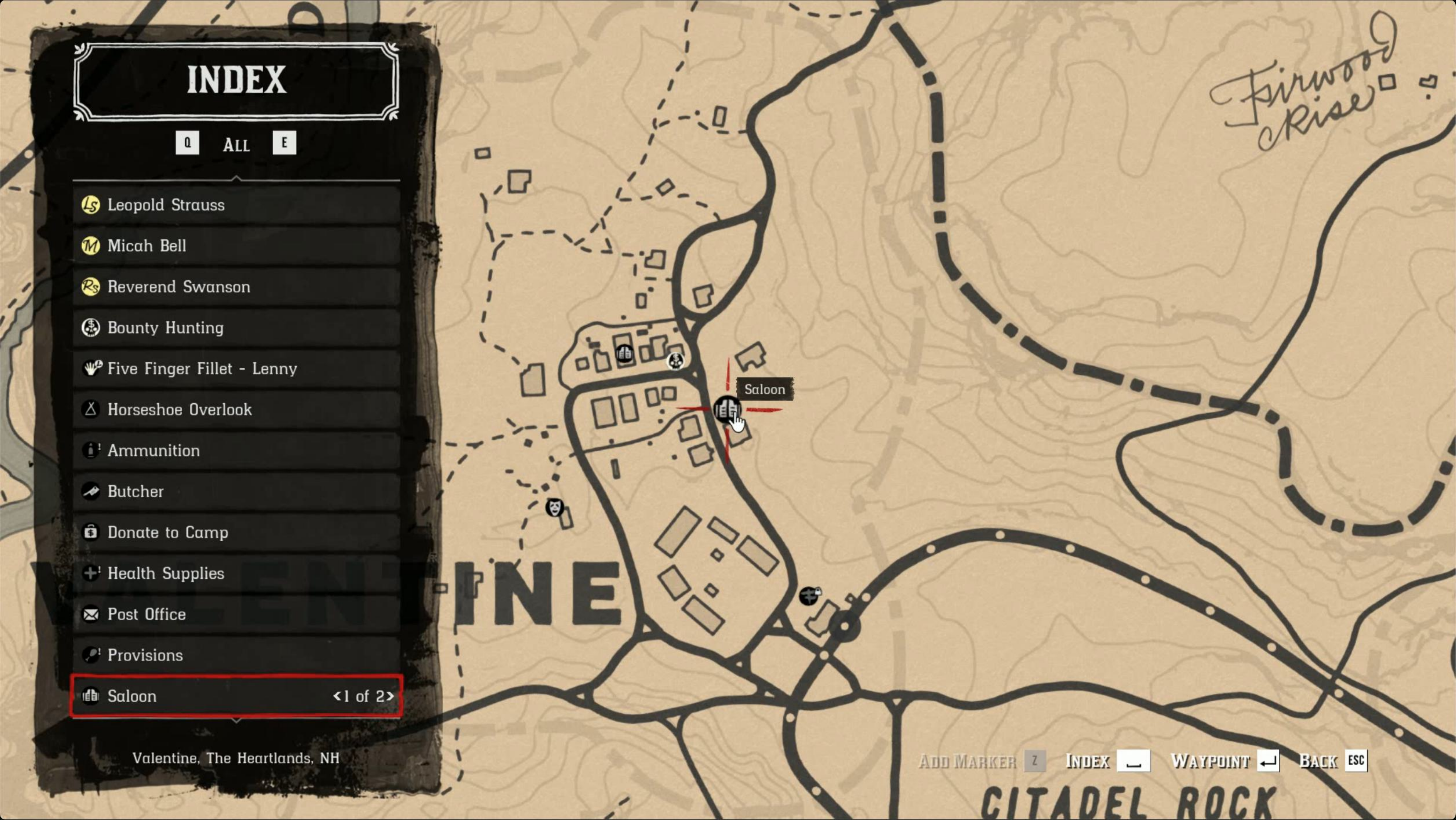}
    \caption{Prepare to Navigate to Saloon} \label{subfig:M03_C02_Generate_Navigation_to_Saloon}
  \end{subfigure}
  \begin{subfigure}{0.32\textwidth}
    \centering
    \includegraphics[width=\linewidth]{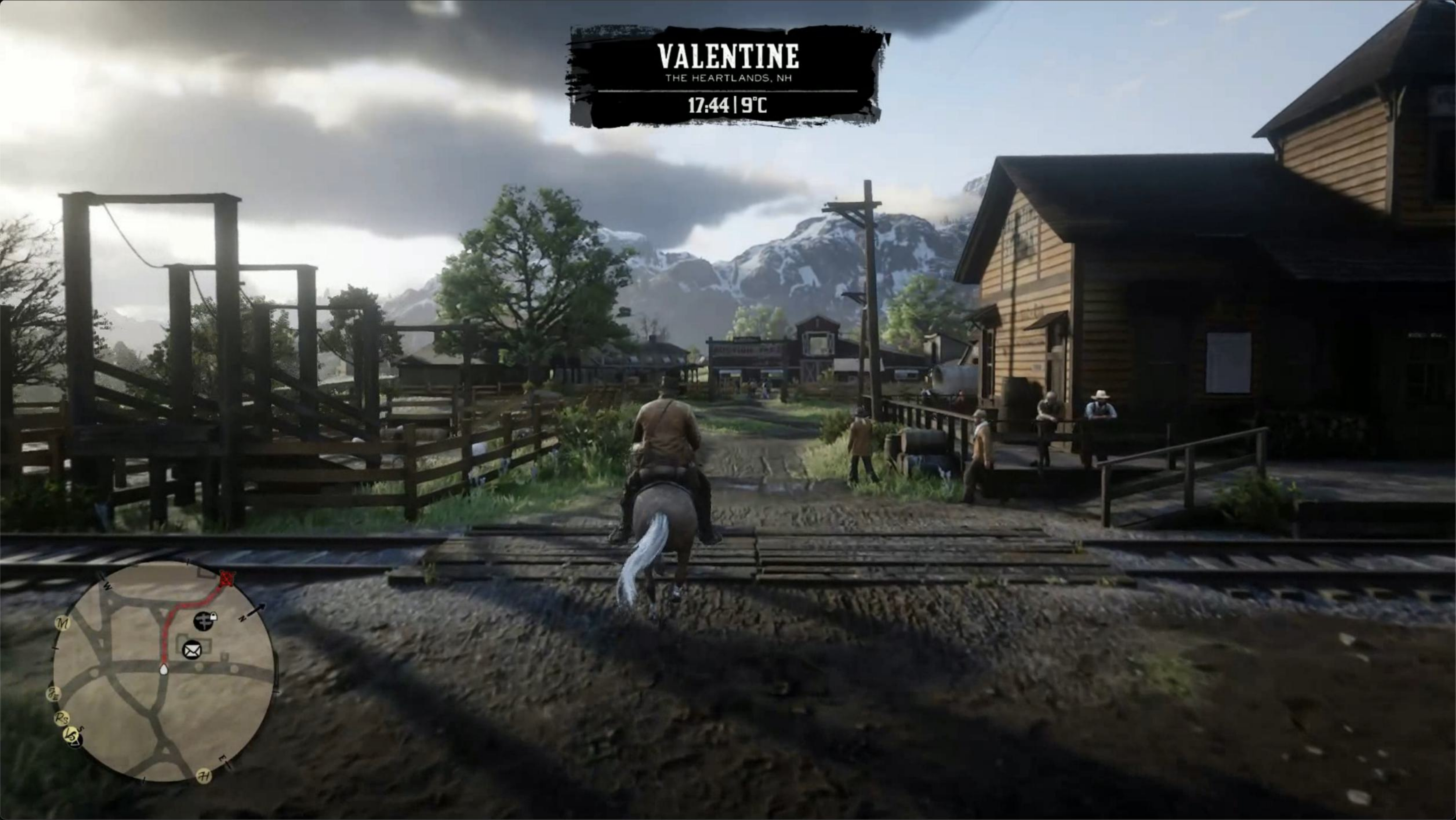}
    \caption{Go to Saloon} \label{subfig:M03_C03_Go_to_Saloon}
  \end{subfigure}
  \begin{subfigure}{0.32\textwidth}
    \centering
    \includegraphics[width=\linewidth]{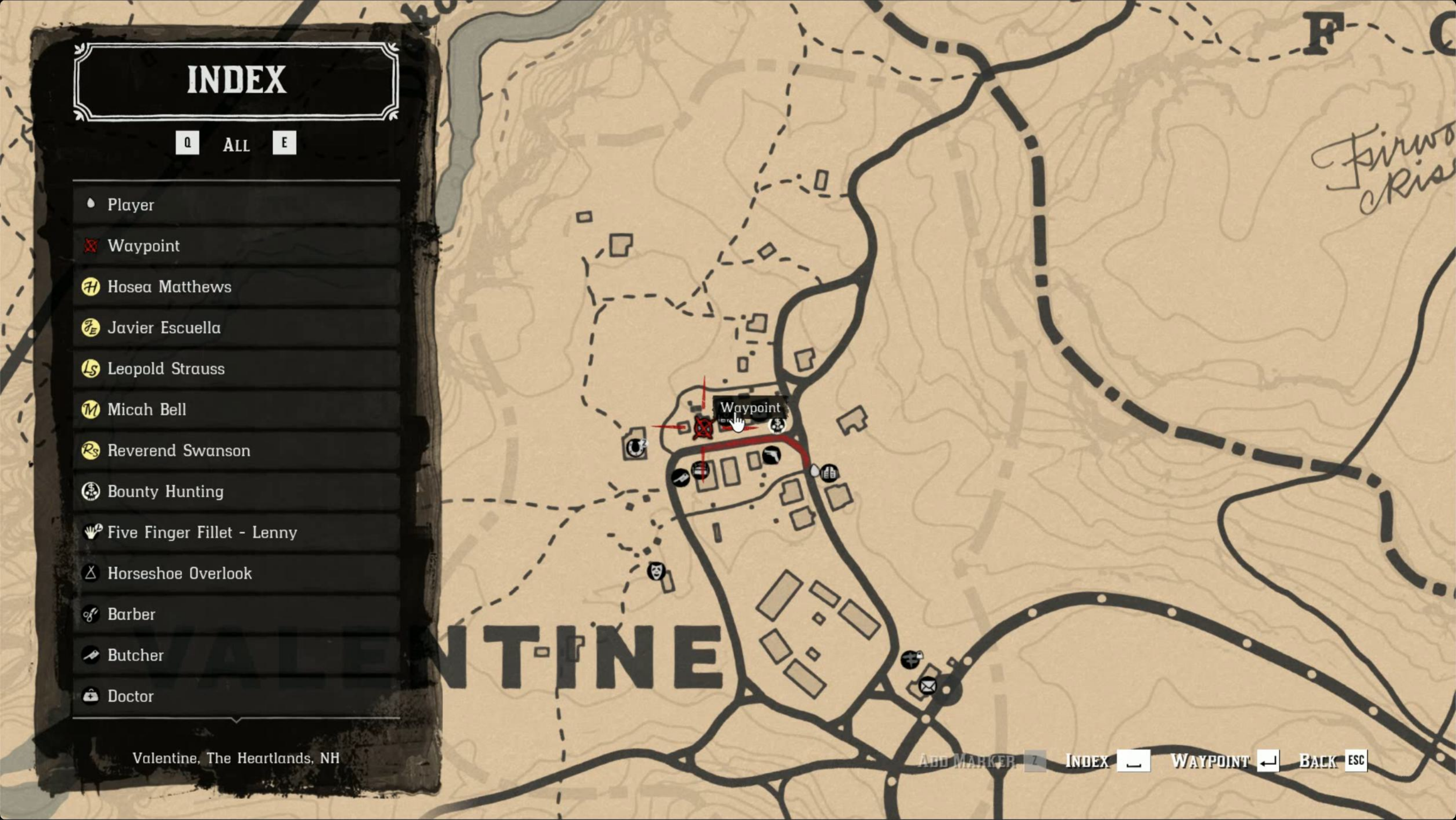}
    \caption{Prepare to Navigate to Shop} \label{subfig:M03_C04_Generate_Navigation_to_Shop}
  \end{subfigure}
  \begin{subfigure}{0.32\textwidth}
    \centering
    \includegraphics[width=\linewidth]{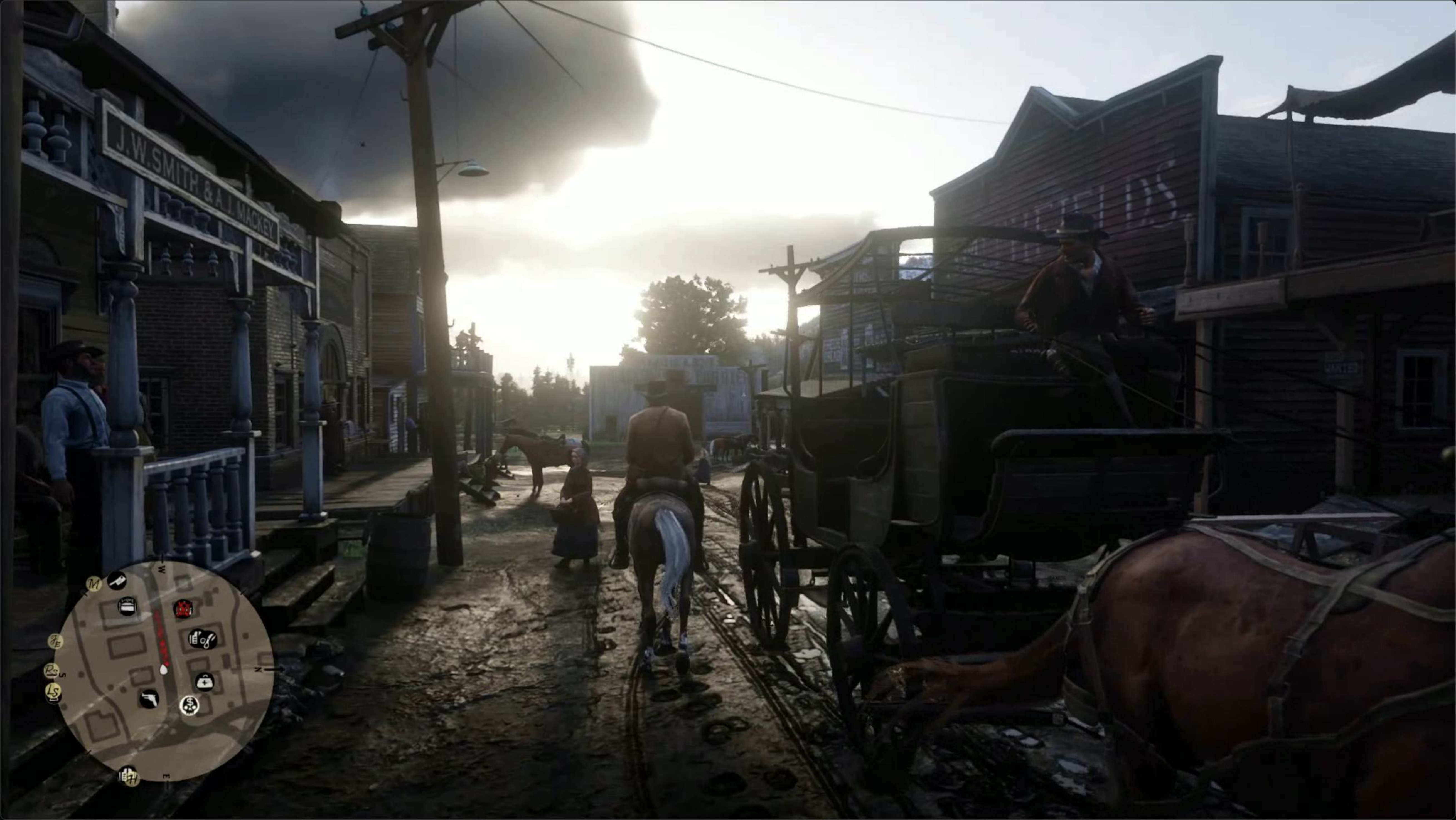}
    \caption{Go to Shop} \label{subfig:M03_C05_Go_to_Shop}
  \end{subfigure}
  \begin{subfigure}{0.32\textwidth}
    \centering
    \includegraphics[width=\linewidth]{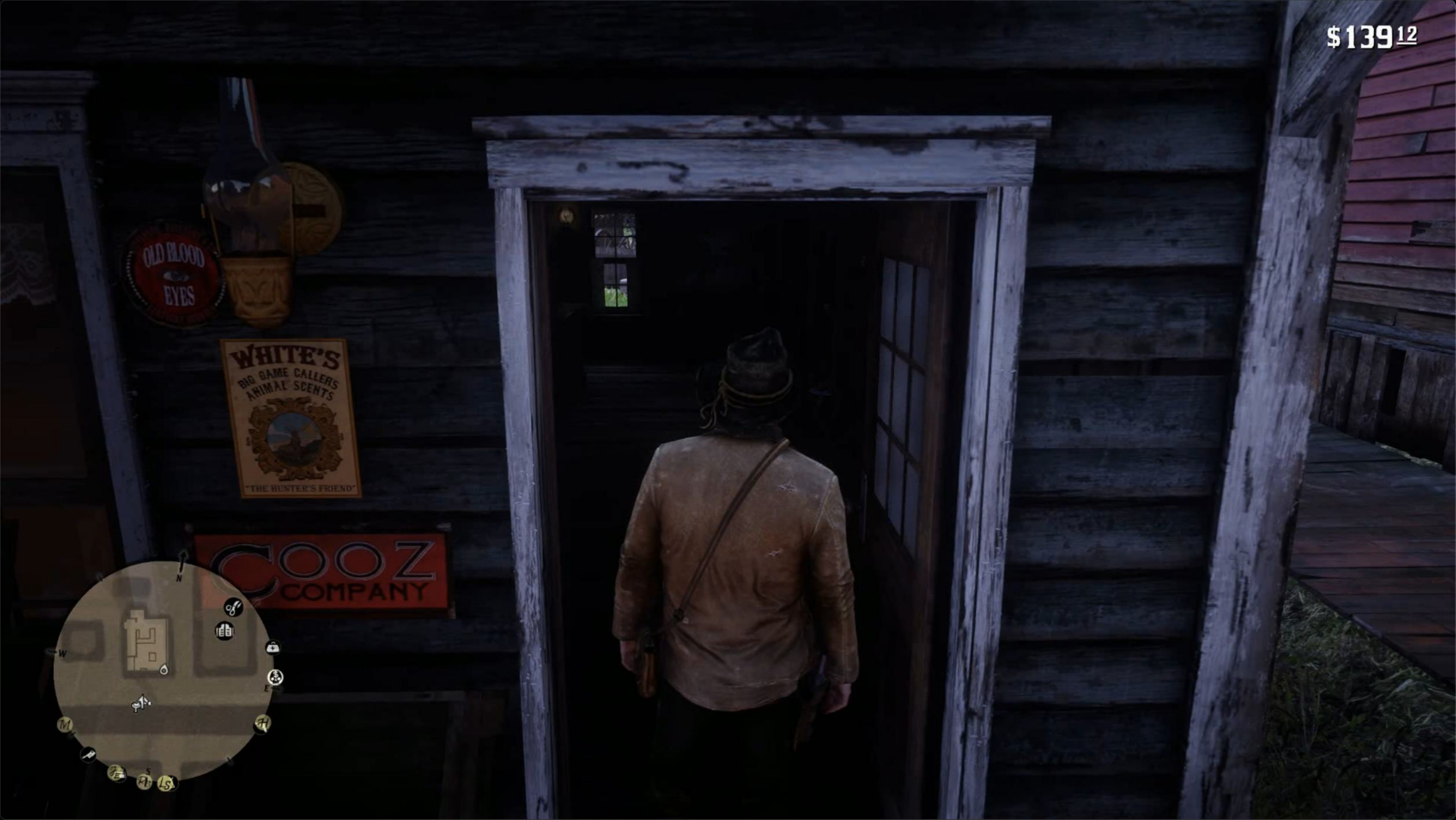}
    \caption{Enter Shop} \label{subfig:M03_C06_Enter_Shop}
  \end{subfigure}
  \begin{subfigure}{0.32\textwidth}
    \centering
    \includegraphics[width=\linewidth]{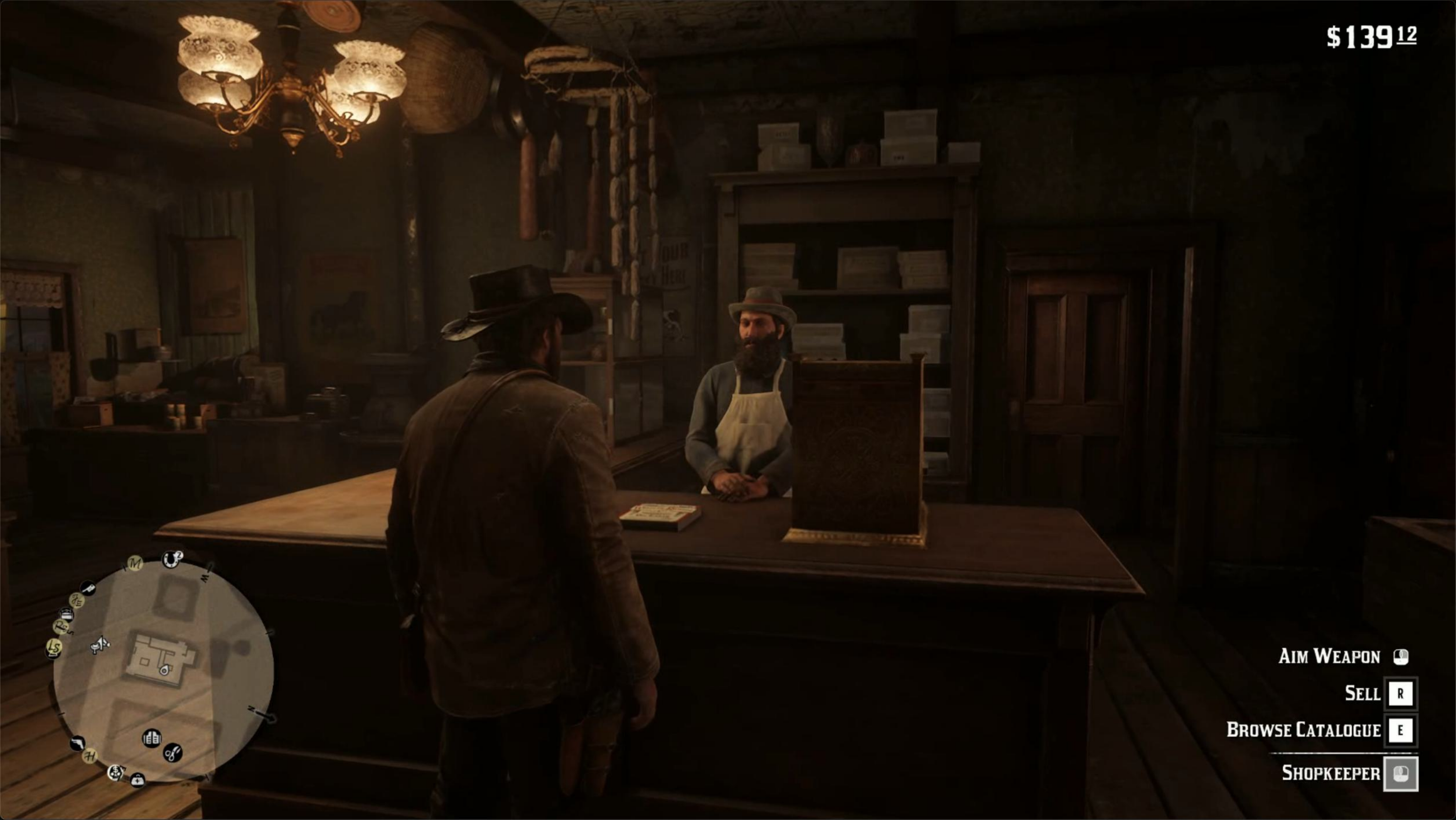}
    \caption{Talk to Shopkeeper} \label{subfig:M03_C07_Talk_to_Shopkeeper}
  \end{subfigure}
  \begin{subfigure}{0.32\textwidth}
    \centering
    \includegraphics[width=\linewidth]{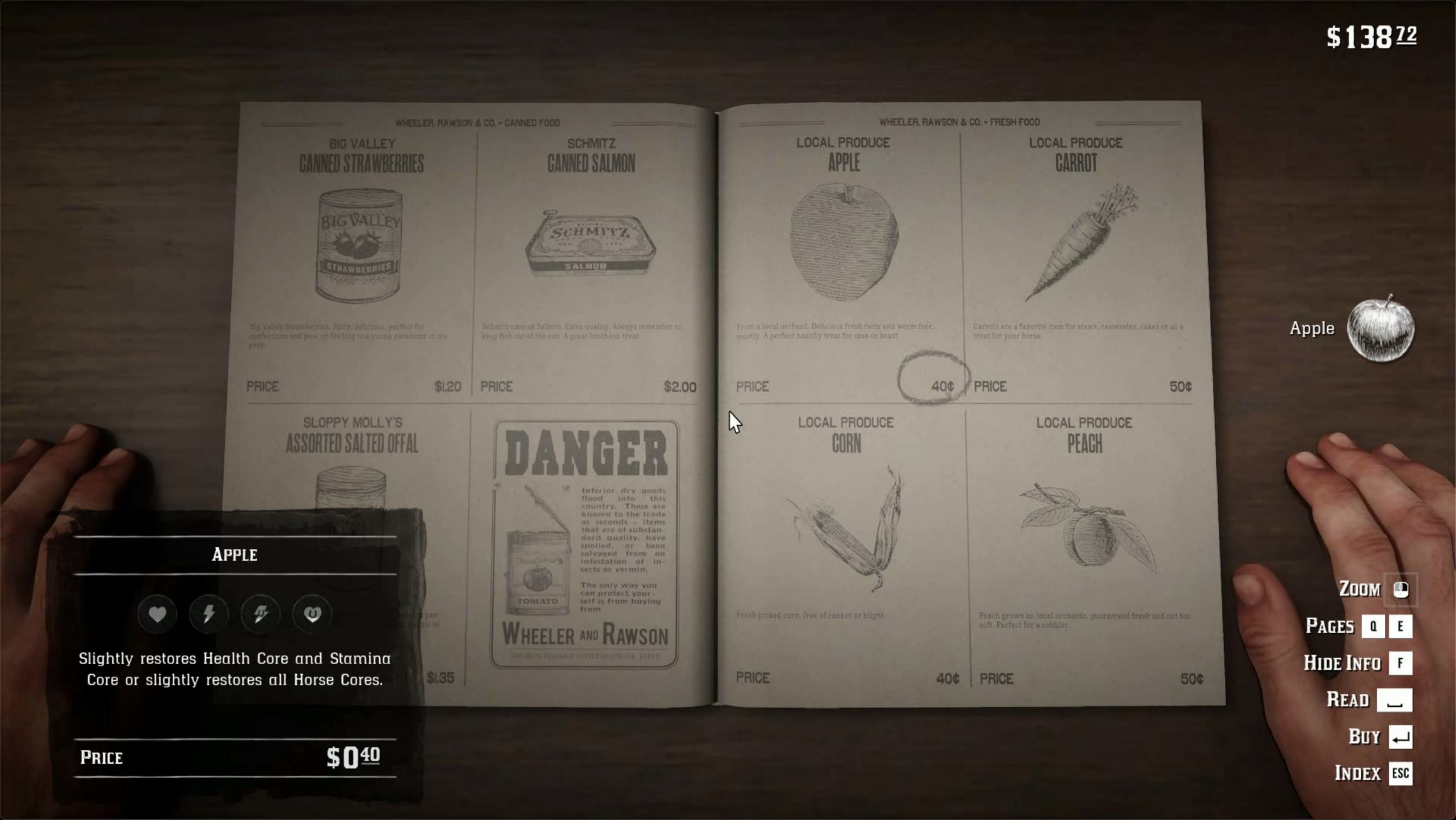}
    \caption{Buy Target Product} \label{subfig:M03_C08_Buy_Target_Product}
  \end{subfigure}
  \caption{Image examples of key points in the open-ended task of \emph{Buy Supply}.}
  \label{fig:app_mission3_tasks}
\end{figure}

%% file: appendix/stardew.tex
\subsection{Introduction to Stardew Valley}
\label{sec:stardew_game_intro}

Stardew Valley is an open-ended country-life RPG game developed by ConcernedApe, which has a 98\% positive rating on Steam and is rated as Overwhelmingly Positive. Players take on the role of a character disillusioned with city life who inherits a dilapidated farm from their late grandfather. Initially, the farmland is overrun with boulders, trees, stumps, and weeds, which players must clear to make way for crops, buildings, and placeable items. The main goal is to restore and expand the farm through activities such as planting crops, raising animals, mining, fishing, and crafting. Additionally, players can interact with NPCs in town, forming relationships that can lead to marriage and children. Players complete quests for money or to restore the town's Community Center by completing "bundles," which reward items like seeds and tools and unlock new areas and game mechanics. All activities are balanced against the character's health, energy, and the game's clock. Food provides buffs, health, and energy. The game features a simplified calendar with four 28-day months representing each season, affecting crop growth and activities. Compared to RDR2, this game is more lightweight and easy to control. This game features a wealth of production and social activities, presenting a comprehensive test of an agent's abilities, which is an ideal platform to observe and evaluate agents' comprehensive behaviors and abilities, like in the Generative Agents~\citep{park2023generative}. We use the latest version (1.6.8) of the game to conduct all the experiments. 

\subsection{Objectives}

\begin{figure}[ht]
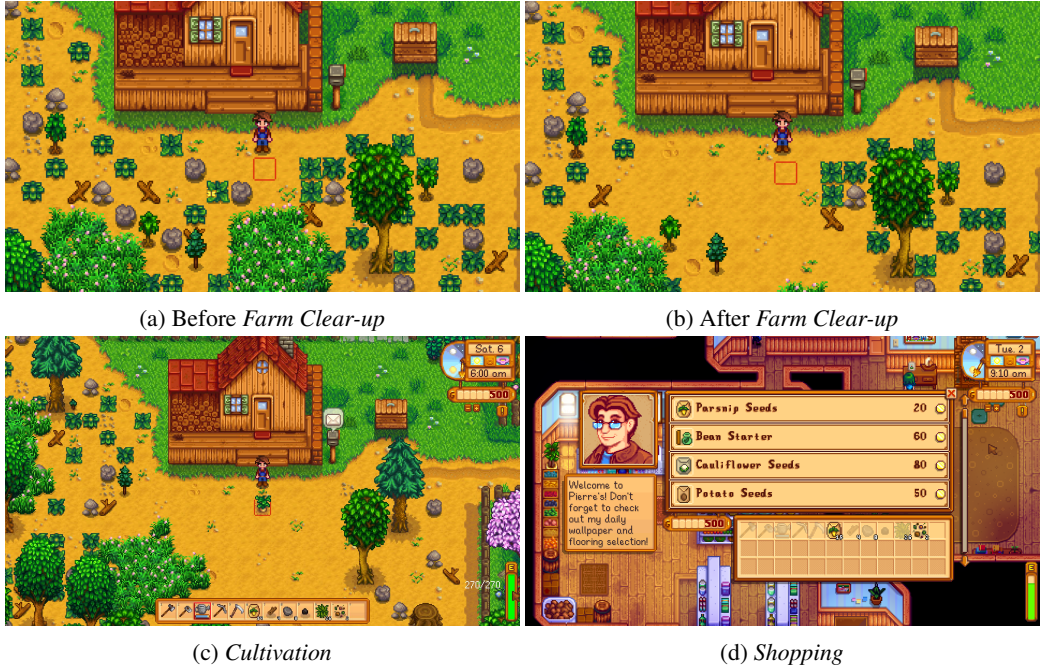

\centering
\begin{minipage}{0.98\textwidth}
\begin{subfigure}[b]{0.5\textwidth}
\centering
\includegraphics[width=\textwidth]{figures/stardew_start_zoomin.pdf}
\caption{Before \emph{Farm Clear-up}}
\label{appx_fig:stardew_clearup_start}
\end{subfigure}
\begin{subfigure}[b]{0.5\textwidth}
\centering
\includegraphics[width=\textwidth]{figures/stardew_result_zoomin.pdf}
\caption{After \emph{Farm Clear-up}}
\label{appx_fig:stardew_clearup_end}
\end{subfigure}
\begin{subfigure}[b]{0.5\textwidth}
\centering
\includegraphics[width=\textwidth]{figures/stardew_cultivation.pdf}
\caption{\emph{Cultivation}}
\label{appx_fig:stardew_cultivation}
\end{subfigure}
\begin{subfigure}[b]{0.5\textwidth}
\centering
\includegraphics[width=\textwidth]{figures/stardew_shopping.pdf}
\caption{\emph{Shopping}}
\label{appx_fig:stardew_shopping}
\end{subfigure}
\caption{Three tasks in Stardew Valley.}
\label{appx_fig:stardew_tasks}
\end{minipage}
\end{figure}

We find that GPT-4o surprisingly struggles with accurately recognizing and locating objects near the player in this 2D game. This leads to difficulties for the agent to interact with objects or people, as it requires the player to stand precisely in front of them in the grid (\eg when entering doors, using a pickaxe to break stones). Even some basic tasks are already challenging enough for current agents in this game.
Therefore, as shown in Figure~\ref{appx_fig:stardew_tasks}, we evaluate three essential tasks in the early stages of the game: 
\begin{itemize}
    \item \textbf{Farm Clearup.} Clear the obstacles on the farm, such as weeds, stones, and trees, as much as possible to prepare for farming. This task requires agents to move precisely to be in front of the obstacles, identify the type of obstacles correctly and select corresponding tools to deal with them.
    \item \textbf{Cultivation.} Use the hoe to till the soil, use a parsnip seed packet on the tilled soil to sow a crop, water the crop every day and harvest at least one parsnip. This task requires long-horizontal memory and reasoning. 
    \item \textbf{Shopping.} Go to the general store in the town, which is on the other map, to buy more seeds and return home. This task is used to evaluate agents' long-distance navigation ability. 
\end{itemize}
For each task, the maximal steps is 100.

\subsection{Implementation Details} 
\label{sec:implementation_details_stardew}

\begin{figure}[htp]
    \centering \includegraphics[width=\linewidth]{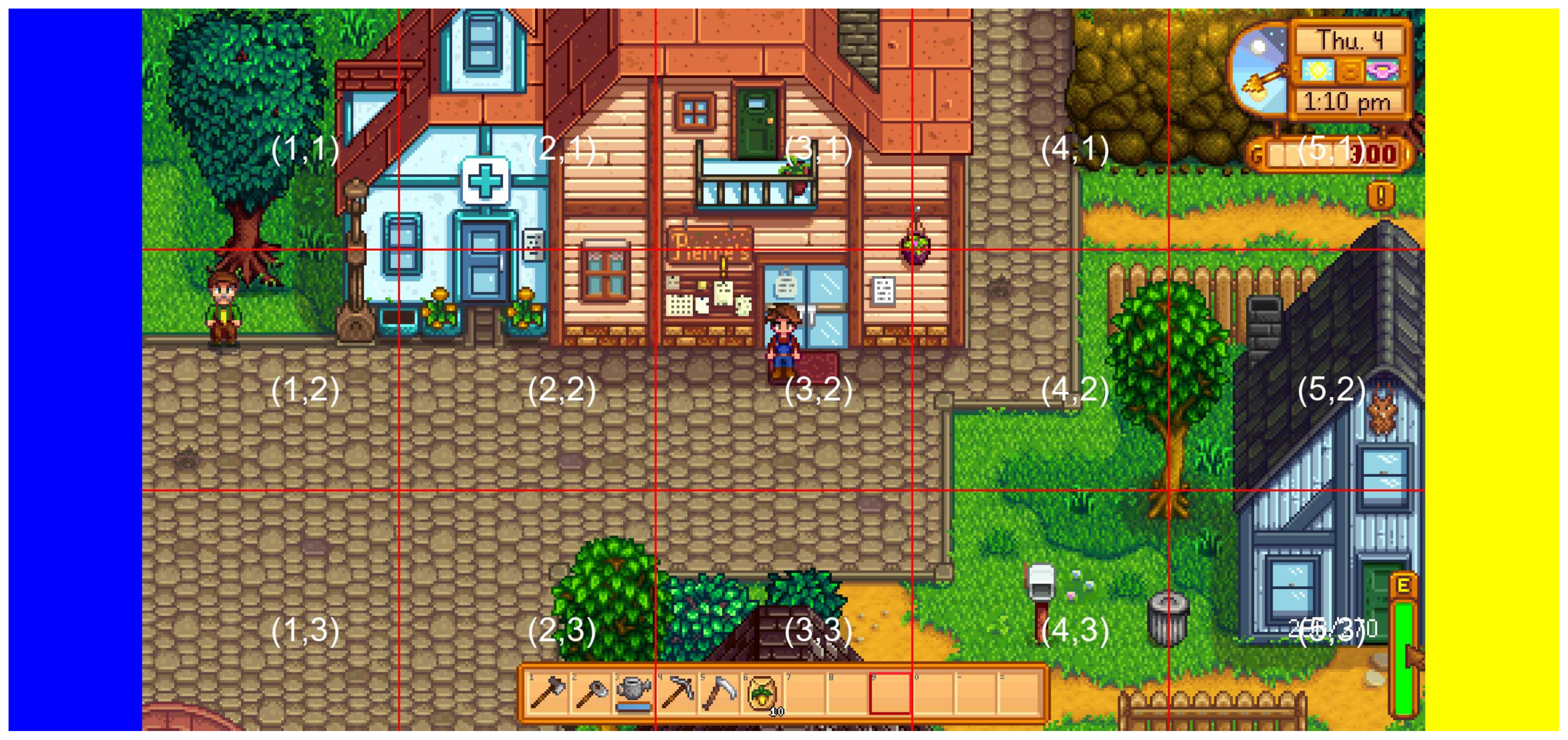}
    \caption{Augmented screenshot via visual prompting. The full screenshot is divided into \(3 \times 5\) grids and each grid has a unique white coordinate. Additionally, we augment all input images with color bands, with the prompt, "The blue band represents the left side and the yellow band represents the right side", which significantly improves GPT-4o's ability to accurately distinguish left from right. }
    \label{appx_fig:stardew_visual_prompting}
\end{figure}

\textbf{Visual Prompting.} As a cartoon-style pixel game, the game screen of Stardew is quite different from the real world. Although GPT-4o can observe coarse-grained information from screenshots, more fine-grained information is required to complete tasks. Therefore, as shown in Figure~\ref{appx_fig:stardew_visual_prompting}, we divide each screenshot into \(3 \times 5\) grids and require GPT-4o to describe the screenshot in a grid-by-grid format. We empirically find that it can result in a more precise and accurate description. And GPT-4o can also make better control based on the grids. In addition, we also augment the image with two blue and yellow bands on the left and right sides, respectfully, with the prompt, "The blue band represents the left side and the yellow band represents the right side". Our empirical results show that this method significantly improves GPT-4o's ability to accurately distinguish left from right.

\textbf{Information Gathering.} As mentioned in the introduction of visual prompting, we let GPT-4o describe the image grid by grid, which is helpful in locating the position of the character, surrounding objects and buildings and facilitates the understanding of the relative positions among them for GPT-4o.
Besides, while compared to GPT-4V, GPT-4o is able to recognize most of the icons and their quality in the toolbar shown at the bottom of the screenshot, GPT-4o cannot output the items in the inventory sequentially one by one as it always skips a few in between. We have to clip the box for each item out of the toolbar and feed them to GPT-4o independently, augmented with template matching, for recognition, which turns out to be more accurate. The success of recognition of the tools in the toolbar is critical to tasks like \textbf{Farm Clearup} and \textbf{Cultivation}.

\textbf{Self-Reflection.} The duration of actions in Stardew is usually much shorter than in RDR2, so we only use the first and last frame from the video observation to reduce the number of tokens used per request. Additionally, we provide some helpful prior information for GPT-4o. For example, a screenshot of the inside of the store is provided to check whether the store was successfully entered. This is useful because there are many other buildings near the store, and sometimes GPT-4o controls the character to enter the wrong one. However, this is not realized if the screenshot is not provided.

\textbf{Skill Curation.}
For skill curation, as mentioned in Figure~\ref{fig:skill_gen}, we mainly rely on the in-game manual to generate atomic skills, like \emph{move\_up()}, \emph{do\_action()} and \emph{use\_tool()}. In addition, to handle the challenges of locating objects, especially doors,  we have a special set of composite skills specifically for Stardew. \textit{e.g.}, go\_through\_door, buy\_item, get\_out\_of\_house and enter\_door\_and\_sleep. With the restrictions of GPT-4o in fine-grained control, we designed go\_through\_door composite skills for the agent to control the game character to accurately reach various doors and successfully enter, such as the house and the store door. and in order to buy certain items such as parsnip seeds, we designed the composite skills buy\_item to control the game character to interact with the salesman and buy parsnip seeds. similarly, we designed the get\_out\_of\_house and enter\_door\_and\_sleep composite skills to accurately exit the house from the bed and enter the house and walk to the bed. 

\textbf{Action Planning.} In this game, we let GPT-4o output at most two skills in a single action every time, which turns out to be efficient. The agent usually needs to select the correct tool first and then use the tool or do action.

\textbf{Procedure Memory.} Procedure Memory is used to store and retrieve skills in code form. In order for agents to quickly get started and complete some special tasks in Stardew, we have predefined skills in Procedure Memory. These skills are divided into atomic and composite skills. atomic skill consists of basic operations such as moving, selecting tools, etc. The description of all the atomic skills is listed as follows:
\begin{itemize}
    \item \emph{do\_action()}: The function to perform a context-specific action on objects or characters. 
    \item \emph{use\_tool()}: The function to execute an in-game action commonly assigned to using the character's current selected tool.
    \item \emph{move\_up(duration)}: The function to move the character upward (south) by pressing the 'w' key for the specified duration.
    \item \emph{move\_down(duration)}: The function to move the character downward (north) by pressing the 'w' key for the specified duration.
    \item \emph{move\_left(duration)}: The function to move the character left (west) by pressing the 'w' key for the specified duration.
    \item \emph{move\_right(duration)}: The function to move the character right (east) by pressing the 'w' key for the specified duration.
    \item \emph{select\_tool(key)}: The function to select a specific tool from the in-game toolbar based on the given tool number.
\end{itemize}

and the composite skills are designed for the agent to complete a variety of special tasks. The description of all the composite skills is listed as follows:
\begin{itemize}
    \item \emph{buy\_item()}: The function to interact with the salesman and buy the item. 
    \item \emph{enter\_door\_and\_sleep()}: The function to enter the house and walk to the bed. 
    \item \emph{get\_out\_of\_house()}: The function to accurately exit the house from the bed
    \item \emph{go\_through\_door()}: The function to reach and enter all kinds of doors.
\end{itemize}

\textbf{Game Pause.} The game will pause automatically when the game window is not focused. So when the character finishes executing actions, we will activate another window, \eg code window, to pause the game and stop the passage of the time in the game.

\subsection{Case Studies}
Here we present a few game-specific case studies to further discuss \projname's self-reflection and task-inference processes in the GCC setting.

\begin{figure}[htp]
    \centering \includegraphics[width=0.9\linewidth]{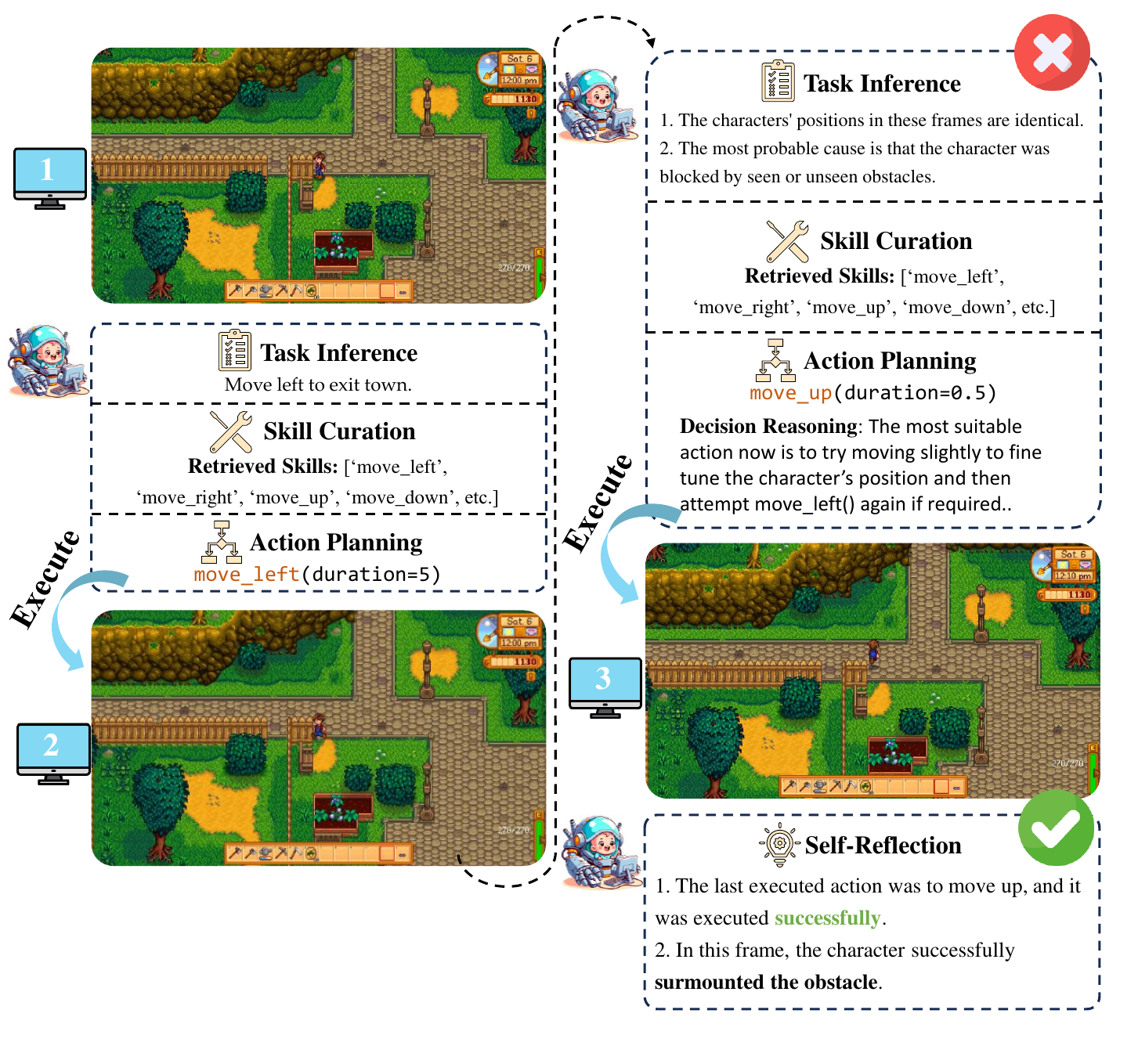}
    \caption{Case study of self-reflection on re-trying a failed task. Task instruction and context require the agent to exit town. A wrong direction is first selected, but the agent moves up after self-reflection.
    Only relevant modules are shown for better readability, though all modules (Figure~\ref{fig:system}) are executed per iteration.}
    \label{fig:example_exit_town}
\end{figure}

\subsubsection{Self-Reflection}
The Self-reflection module plays an important role in the completion of game missions in Stardew, giving our framework the ability to determine if the actions performed are complete and effective and to correct the errors of invalid actions. In the "Purchasing Seeds" task, the Agent is asked to return home from the store after purchasing items. At the "Home is on the left side of the store" prompt, the Agent controls the character to go left, but there are obstacles to keep going left, and the character must go up to circumnavigate the obstacles. As shown in the Fingure~\ref{fig:example_exit_town}, the role will initially be stuck at the obstacle and cannot continue to the left. Through Self-Reflection, the Agent can judge that it is currently in a state of obstruction, and moving to the left cannot be implemented smoothly. Therefore, the agent can adjust the direction upward to bypass the obstacle and enable the role to continue to the left until it returns home.

\subsubsection{Task-inference}
\begin{figure}[htp]
    \centering \includegraphics[width=\linewidth]{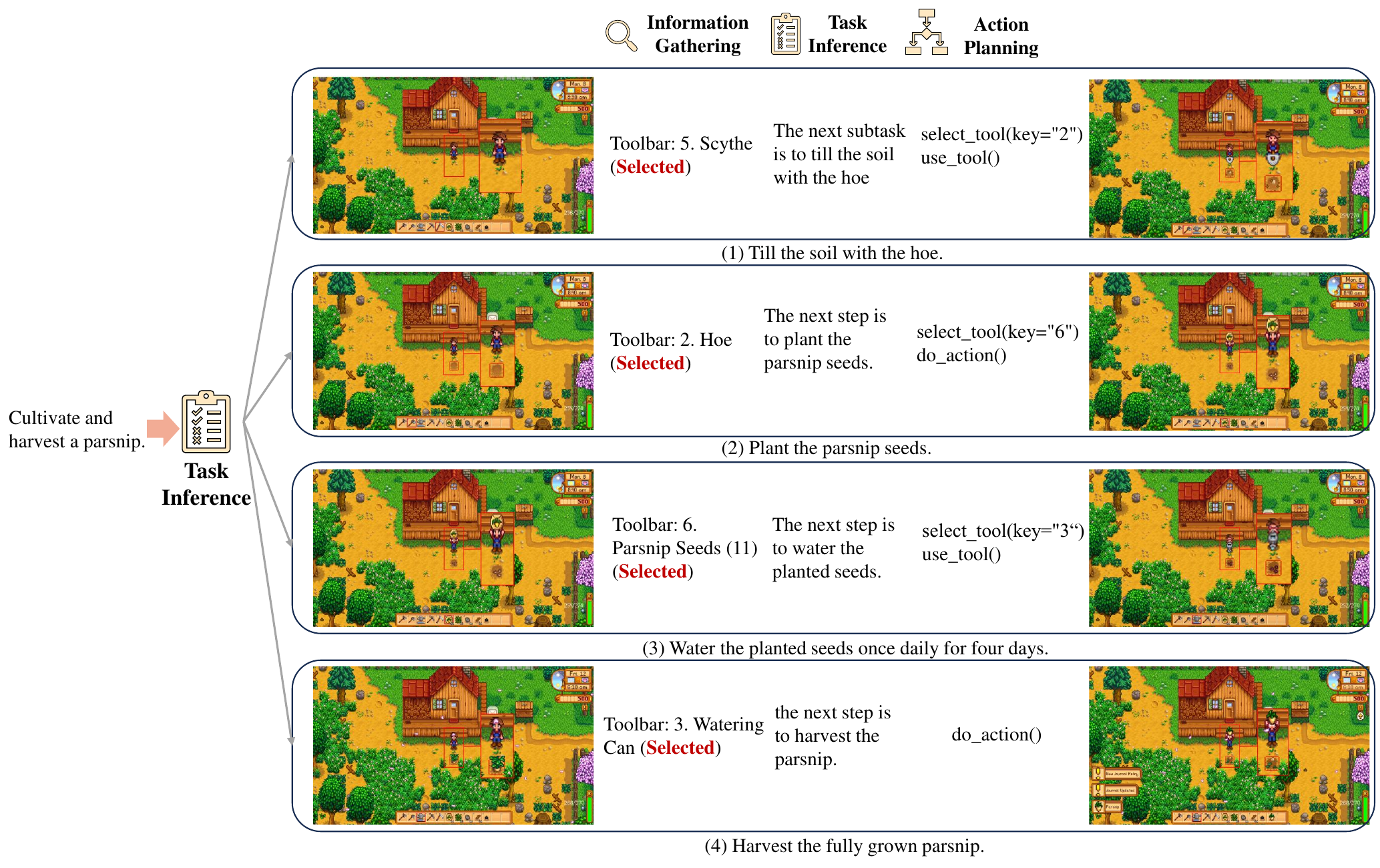}
    \caption{Case study of task inference on decomposing a task into specific sub-tasks. The complete task is to cultivate and harvest a parsnip. \projname decomposes the task into four sub-tasks by task inference. Only relevant modules are shown for better readability, though all modules (Figure~\ref{fig:system}) are executed per iteration.}
    \label{fig:example_cultivate}
\end{figure}

Task Inference is a very effective module for completing game quests in Stardew. Its function is to decompose a vague and grand task into a specific sub-task, which effectively guides the Agent to complete the overall task. For example, in the Farming task, as shown in Figure~\ref{fig:example_cultivate}, the task that the character needs to complete is "cultivate and harvest a parsnip." This is a complete but vague task. Through the Task Inference module, the Agent breaks down the task into (1) till the soil with the hoe, (2) plant the parsnip seeds, (3) water the planted seeds once daily for four days, (4) harvest the fully grown parsnip. This enables the Agent to know more clearly the steps needed to complete and finish the task successfully.

\subsection{Limitations of GPT-4o}
\textbf{Fine-grained Control.} Stardew Valley requires that players are positioned precisely to interact with objects, such as doors and NPCs. However, it is difficult for GPT-4o to take a pixel-level precise action. For example, GPT-4o can not take a precise movement even though the speed at which the figure moves is known. To alleviate this problem, we make some composite skills that use template-matching to complete some complex interaction tasks, such as purchasing items.

\textbf{Perception in a 2D virtual world .}
In Stardew Valley, it's common for a character to be blocked by rocks or trees, and GPT-4o fails to tell if a character is blocked by looking at the image once, and can't predict if the next move will be blocked, which is very easy for a human to do by looking at the image. This indicates that GPT-4o is relatively weak in perceiving the virtual world in this game. In order to solve this problem, we compare the successive frames before and after in Self-Reflection to enable GPT-4o to judge the corresponding changes.

%% file: appendix/dealer.tex
\subsection{Introduction to Dealer's Life 2}
\label{sec:dl2_game_intro}
Dealer's Life 2 is a captivating indie simulation game developed by Abyte Entertainment. Renowned for its intricate negotiation mechanics and humorous portrayal of a pawn shop environment, the game is celebrated for its engaging gameplay that combines strategy with a quirky, cartoonish art style. As a simulation game with role-playing elements, Dealer's Life 2 is played from a first-person perspective, utilizing a mouse for point-and-click interactions and a keyboard for price inputs. This interface facilitates item appraisals, customer interactions, and comprehensive shop management.

In the game, players assume the role of a pawn shop manager, tasked with acquiring and selling various items to make a profit while managing their store's reputation and inventory. Players engage with a wide range of unique non-player characters (NPCs), each with their own distinct behaviors and negotiation styles. Whether bartering over the price of a rare collectible or managing unforeseen shop events, players must hone their haggling and strategic decision-making skills to succeed.
Dealer's Life 2 operates in a closed-source format with no APIs available for accessing in-game data or automating gameplay functions. This setup ensures a hands-on experience where players are immersed in the day-to-day challenges of running a pawn shop. This game environment provides a unique and entertaining setting for testifying the GCC's haggling and strategic decision-making abilities.  We run our experiments using the latest version, V.~1.013\_W96 of the game.

\subsection{Objectives}
We concentrate on evaluating the sustained management skills required to maximize profits through buying and selling a diverse range of items from customers. Therefore, the task in this game is defined as \emph{Weekly shop management}, \textit{i.e.}, managing a shop for a week automatically.
This game could effectively demonstrate the negotiation ability of the LMM in a trade and bargain. For example, giving an unacceptable price to the customers, \ie a pretty low price for a seller customer or a very high price for a buyer customer, could cause the deal to fail directly, which brings no profit in this situation. The key is to carefully analyze the description of the item, \textit{e.g.}, the rarity and condition of the item, and more importantly, the response of the customer, \textit{i.e.}, the customer's mood changes. 

Contrary to many games that feature detailed tutorials highlighting specific operations and objectives through each crucial step, Dealer's Life 2 does not provide such guidance. This absence transforms the game into a zero-shot, hard open-world task, where the LMM must directly apply its prior knowledge of haggling and strategic decision-making to a new and unfamiliar environment.
To provide readers with a clear and straightforward understanding of the task, we illustrate the typical flow of a day's shop management through several key steps, presented in Table \ref{tab:dealers_task}.

\begin{table}[ht]
\centering
\caption{Key points in the open-ended mission, \emph{Weekly shop management} in Dealer's Life 2. Figure~\ref{fig:dealers_task_snapshots} showcases snapshots of key points (specific sub-figures marked in parenthesis in the table).}
\label{tab:dealers_task}
\resizebox{0.8\textwidth}{!}{
\begin{tabular}{ll}
\toprule
Task: Weekly shop management & Description \\ 
\midrule
Open shop (Fig. \ref{subfig:open_shop}) & Start a new day shop management. \\

Dialog (Fig. \ref{subfig:dialog}) & Choose an option in a dialog. \\

Item Description (Fig. \ref{subfig:item_description}) & View the item information \\

Haggle (Fig. \ref{subfig:haggle}) & Give a price for the item. \\

Deal Result (Fig. \ref{subfig:deal_result}) & View the deal results. \\

Stats (Fig. \ref{subfig:stats}) & View shop stats. \\
\bottomrule
\end{tabular}
}
\end{table}

\begin{figure}
  \centering
  \begin{subfigure}{\subfigwidth\textwidth}
    \centering
    \includegraphics[width=\linewidth]{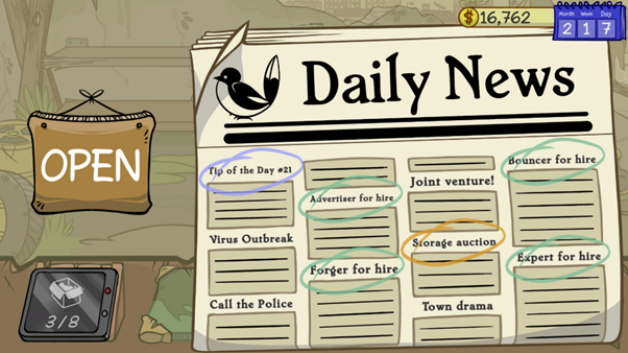}
    \caption{Open Shop} \label{subfig:open_shop}
  \end{subfigure}
  \begin{subfigure}{\subfigwidth\textwidth}
    \centering
    \includegraphics[width=\linewidth]{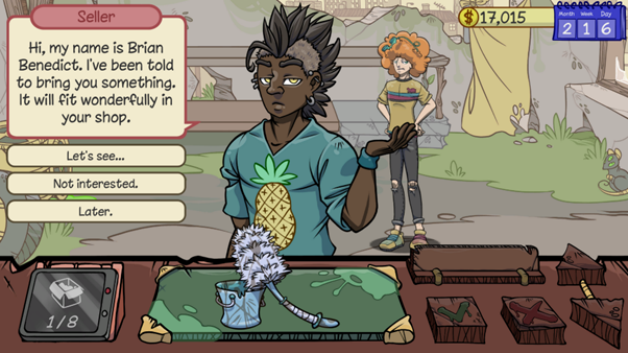} 
    \caption{Dialog} \label{subfig:dialog}
  \end{subfigure}
  \begin{subfigure}{\subfigwidth\textwidth}
    \centering
    \includegraphics[width=\linewidth]{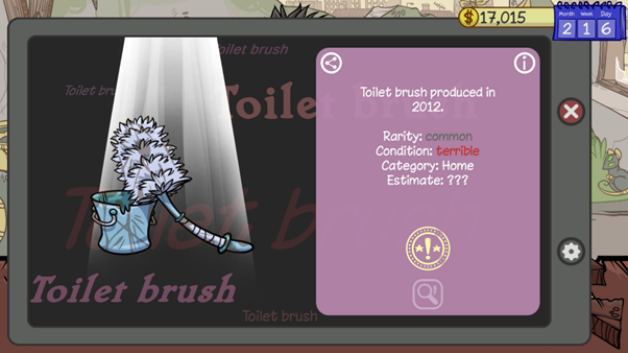} 
    \caption{Item Description} \label{subfig:item_description}
  \end{subfigure}
  \begin{subfigure}{\subfigwidth\textwidth}
    \centering
    \includegraphics[width=\linewidth]{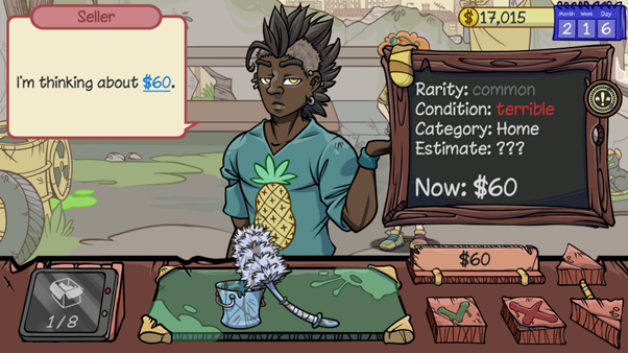} 
    \caption{Haggle} \label{subfig:haggle}
  \end{subfigure}
  \begin{subfigure}{\subfigwidth\textwidth}
    \centering
    \includegraphics[width=\linewidth]{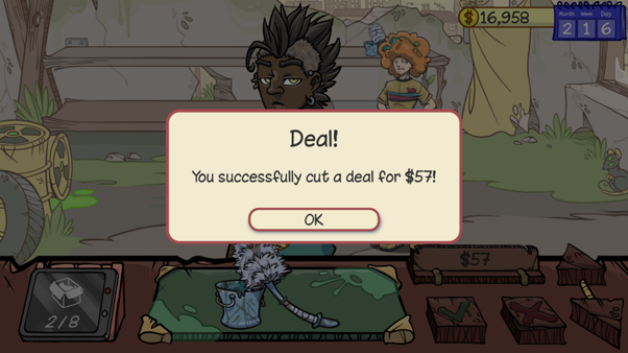} 
    \caption{Deal Result} \label{subfig:deal_result}
  \end{subfigure}
    \begin{subfigure}{\subfigwidth\textwidth}
    \centering
    \includegraphics[width=\linewidth]{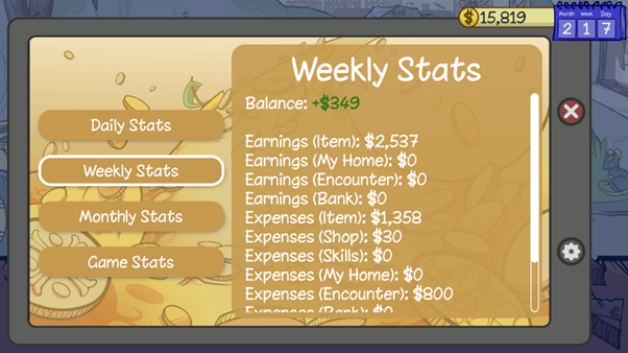} 
    \caption{Stats} \label{subfig:stats}
  \end{subfigure}
  \caption{Image examples of key points in the open-ended task of Dealers' Life 2.}
  \label{fig:dealers_task_snapshots}
\end{figure}

\subsection{Implementation Details}
The implementation of Dealers' Life 2 also strictly follows the GCC framework, which includes Information Gathering, Self-Reflection, Task Inference, Skill Curation, Action Planning, and Action Execution.
The details are described in Appendix \ref{app:general_implementation}. Therefore, we emphasize the specific implementations for Dealers' Life 2.

\textbf{Procedural Memory.} Due to the absence of a new-user guide, the LMM cannot directly and accurately know the operation method or effect of an action in the game, \textit{e.g.}, giving the price can only use the keyboard to input an integer in an abstract box in the bottom right of the haggle screen as shown in Figure \ref{subfig:haggle}, by directly observing the screen. Unless the player executes an action and observes what is happening, the player cannot know what its effect is. However, this could easily cause severe errors in an open-world environment. For example, if the player gives a price at \$100,000 for an item without knowing what the box is, it could cause the player to lose all the money. Besides, this game is very simplified with finite types of screen content and fixed buttons positions for processing the deal, where we could categorize the screen types and design general atomic skills for them.
Thus, with a focus on evaluating the LMM's zero-shot haggling and strategic decision-making ability in managing a shop, we believe it is reasonable to skip the skill curation by directly setting several atomic skills as the initialization of the procedural memory, such as "process\_dialog()" for clicking on the option of a dialog screen to keep the deal going on as shown in Figure \ref{subfig:dialog}.
The description of all the atomic skills is listed as follows:
\begin{itemize}
    \item \emph{open\_shop()}: The function to open the dealer's shop to start dealing for today. 
    \item \emph{give\_price(price)}: The function to give a price for the item in the deal. The price must be an integer number.
    \item \emph{process\_dialog()}: The function to click on to choose the first option of the dialog to make the game go on.
    \item \emph{close\_description\_page()}: The function to close a description page showing information about the item details, daily stats, or the traits of the buyer or seller.
    \item \emph{accept\_deal()}: The function to click on the check mark to accept the deal on the confirmation dialog.
    \item \emph{reject\_deal()}: The function to click on the cross mark to reject the deal on the confirmation dialog.
    \item \emph{finish\_buy()}: The function to click on the ok button to finish the deal on the confirmation dialog.
    \item \emph{finish\_sell()}: The function to click on the ok button to finish the selling on the confirmation dialog.
\end{itemize}

\textbf{Self-Reflection.} 
Additionally, as Dealers' Life 2 has no heavy need for a long-term reflection, so we only use the first and last frame of the video as input to reduce the number of tokens used per request.
Finally, this self-reflection module could help to keep the game going, instead of sticking to the same point in the game.

\textbf{Action Planning.} In this game, we restrict GPT-4 to output only one skill per action because it is a round-based game that does not require frequent execution of actions, and the state of the next time-step after an action is executed is highly uncertain, \textit{e.g.},  the unpredictable mood changes in a customer's response.

\subsection{Case Studies}

Here we present a few game-specific case studies to further discuss \projname's reasoning and decision-making process in the GCC setting.

\subsubsection{Successful Negotiation}

\begin{figure}
  \centering
  
  \begin{subfigure}{0.4\textwidth}
    \centering
    \includegraphics[width=\linewidth]{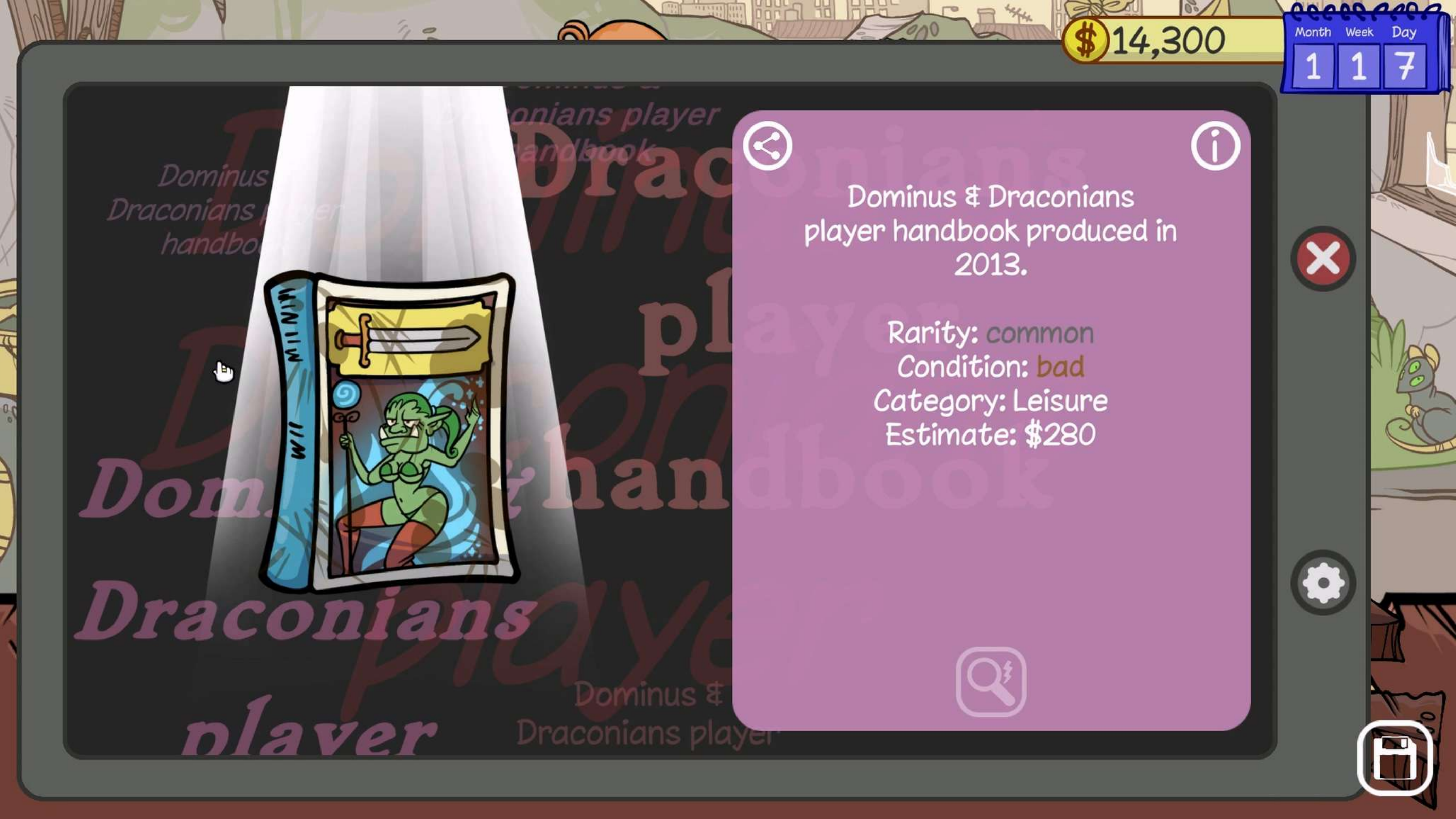} 
    \caption{Agent: Given that the customer is a seller and the item on offer is in bad condition with an estimated price of \$ 280, it would be wise to offer around \$150 to leave room for negotiation.} \label{subfig:dealer_succ0}
  \end{subfigure}\quad
  \begin{subfigure}{0.4\textwidth}
    \centering
    \includegraphics[width=\linewidth]{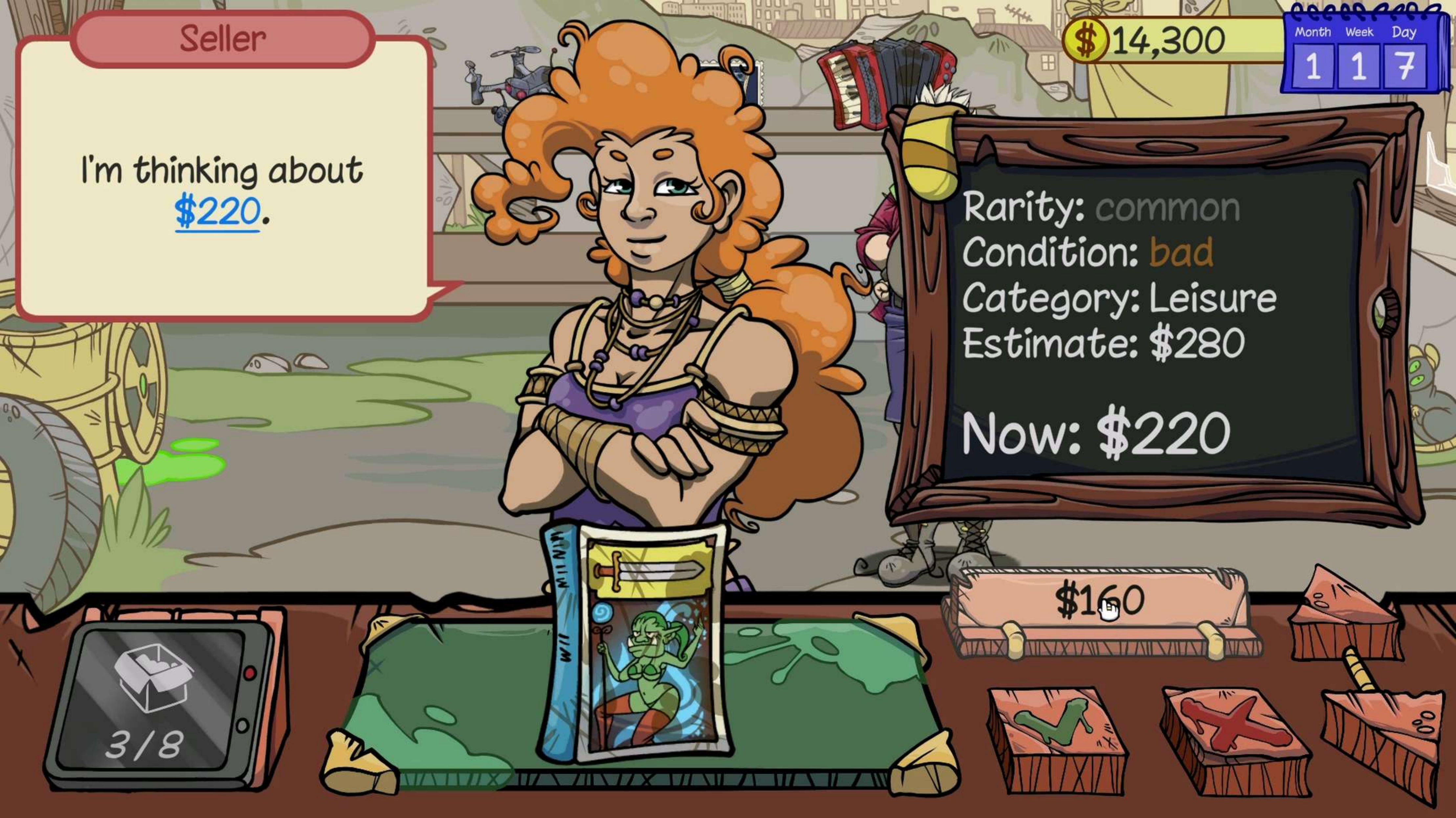} 
    \caption{Agent: Given the item's bad condition and \$280 estimate, A reasonable starting counteroffer would be 50-60\% of its estimated value, and \$160 is a reasonable counteroffer. } \label{subfig:dealer_succ1}
  \end{subfigure}
  \begin{subfigure}{0.4\textwidth}
    \centering
    \includegraphics[width=\linewidth]{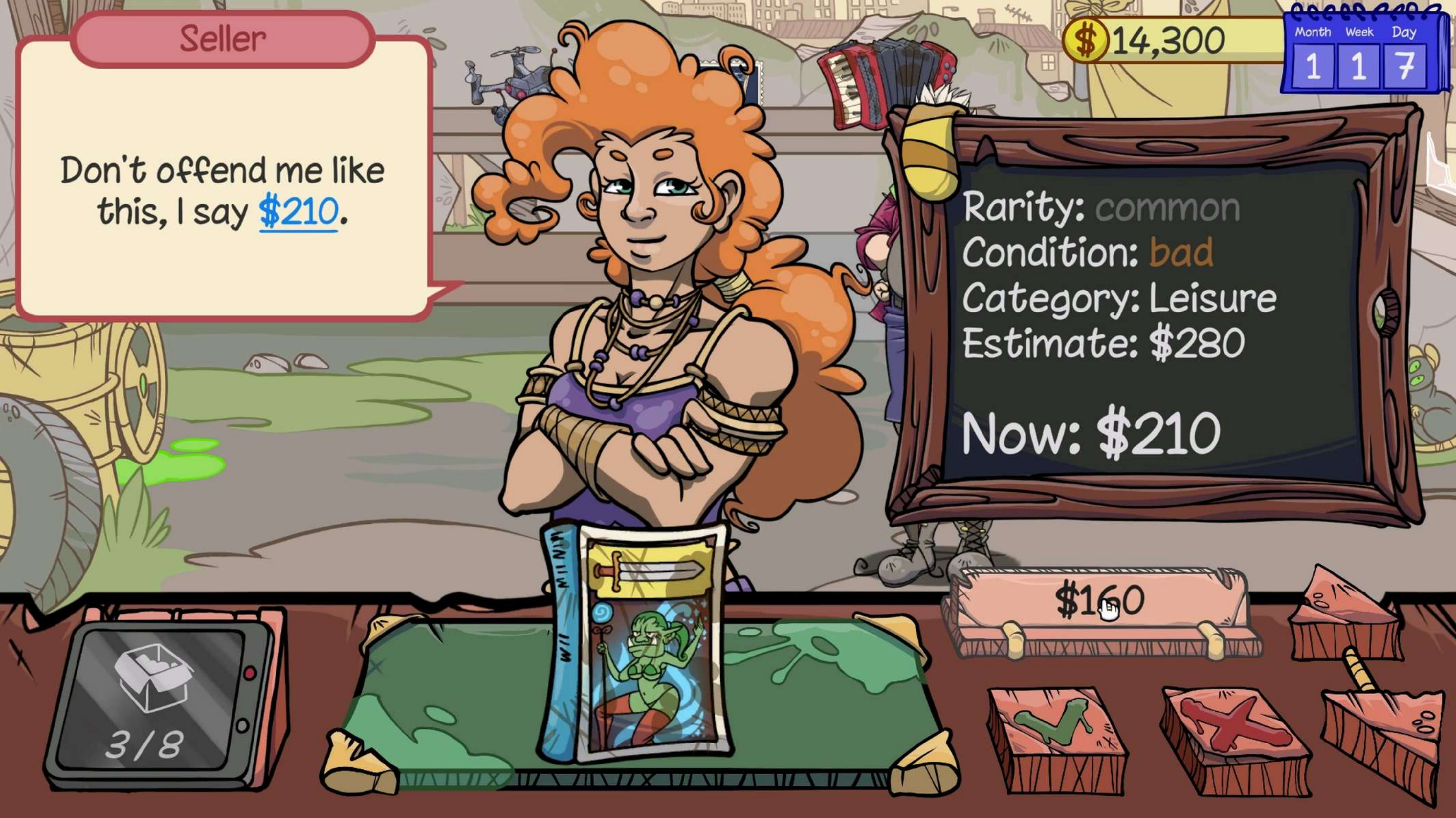}  
    \caption{Agent: A reasonable next offer could be \$180 or \$190, aiming to reach a middle ground. I would still risk with \$160 to see if the offer is cut further.} \label{subfig:dealer_succ2}
  \end{subfigure}\quad
    \begin{subfigure}{0.4\textwidth}
    \centering
    \includegraphics[width=\linewidth]{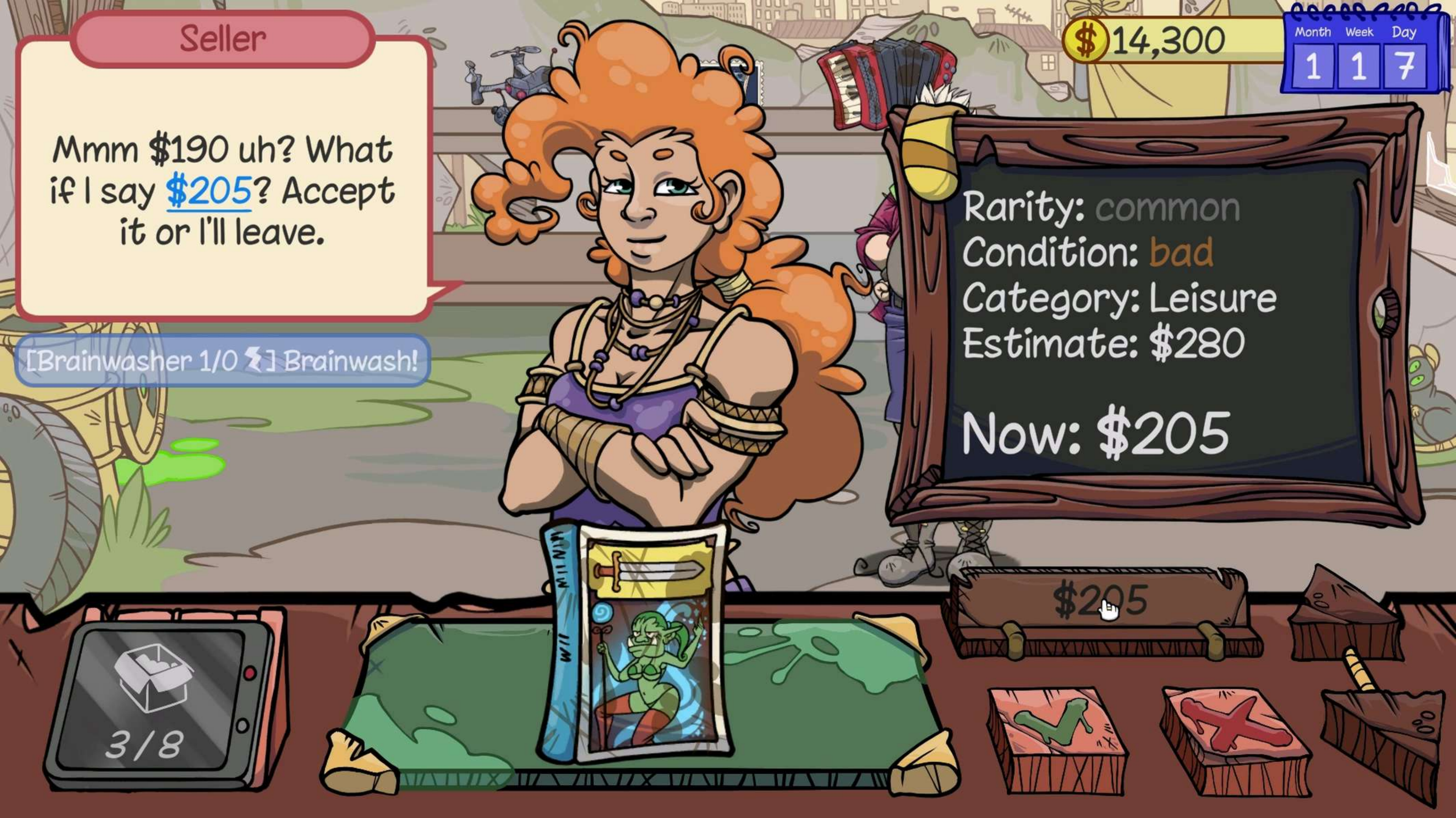} 
    \caption{Agent: The customer is proposing her final offer at \$205. It will be necessary to accept the offer as this is still a profitable margin.} \label{subfig:dealer_succ3}
  \end{subfigure}
  
  \caption{The reasoning of \projname in a successful deal with haggling. The price provided by \projname is keyed in by keyboard and mouse operations in the digital display box in the bottom-right corner.}
  \label{fig:dealers_succ}
\end{figure}

Figure \ref{fig:dealers_succ} illustrates a successful negotiation by \projname with an NPC seller over an item valued at \$280. 
\projname determines a strategic starting offer by considering both the item's quality and the customer's initial proposal. Throughout subsequent negotiation rounds, \projname leverages its memory to maintain an offer close to the initially assessed \$160, applying pressure on the customer to reduce their expectations. However, \projname also demonstrates flexibility, adapting its strategy when faced with the customer's final offer—signaled by their incline to leave. This allows \projname to secure a final agreement that still yields a profitable deal.

\subsubsection{Unsuccessful Negotiation}

\begin{figure}
  \centering
  
  \begin{subfigure}{0.4\textwidth}
    \centering
    \includegraphics[width=\linewidth]{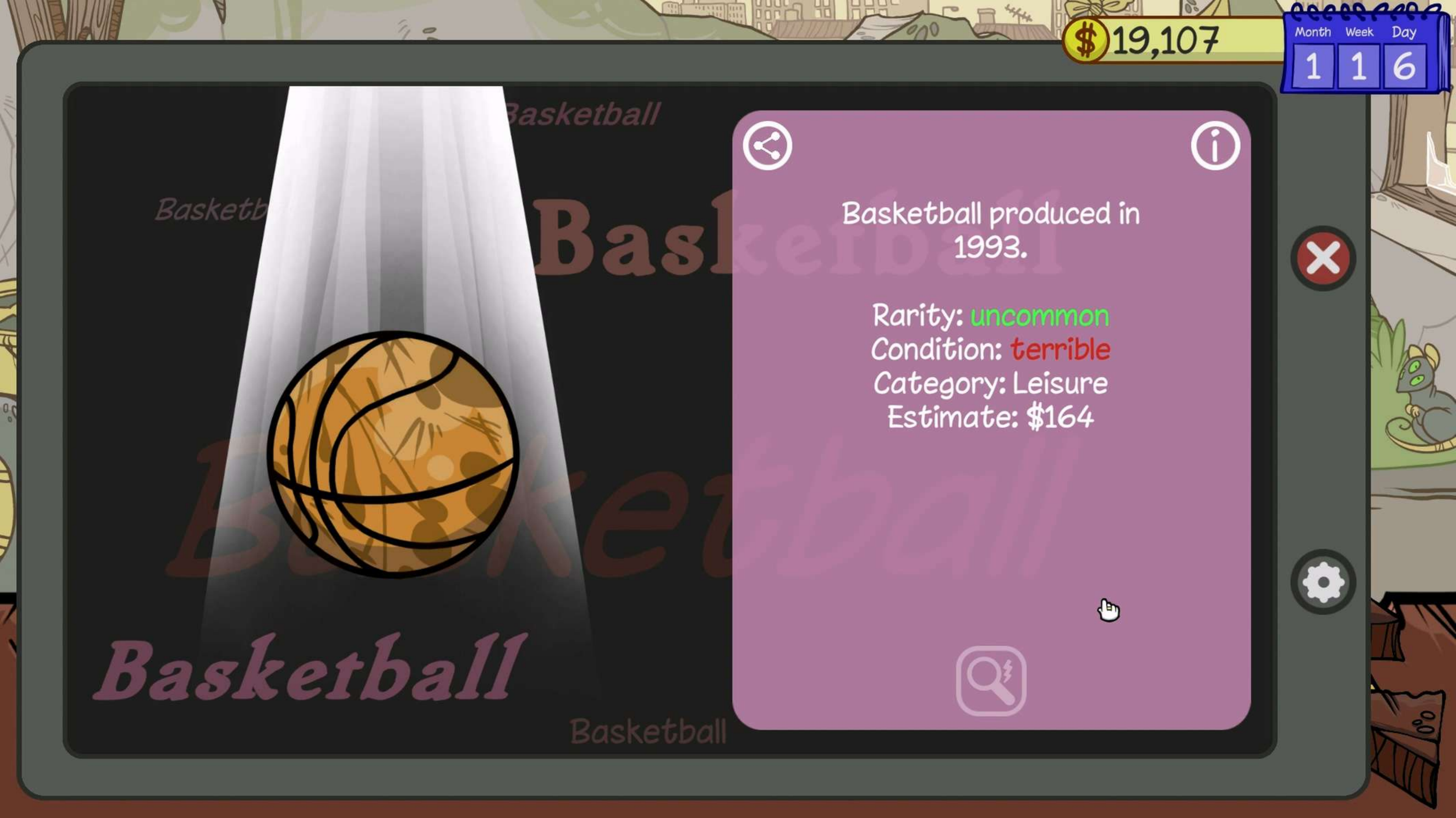} 
    \caption{Agent: I will offer an initial price of \$100 to reflect the poor condition while allowing room for negotiation since the condition is poor.} \label{subfig:dealer_unsucc0}
  \end{subfigure}\quad
  \begin{subfigure}{0.4\textwidth}
    \centering
    \includegraphics[width=\linewidth]{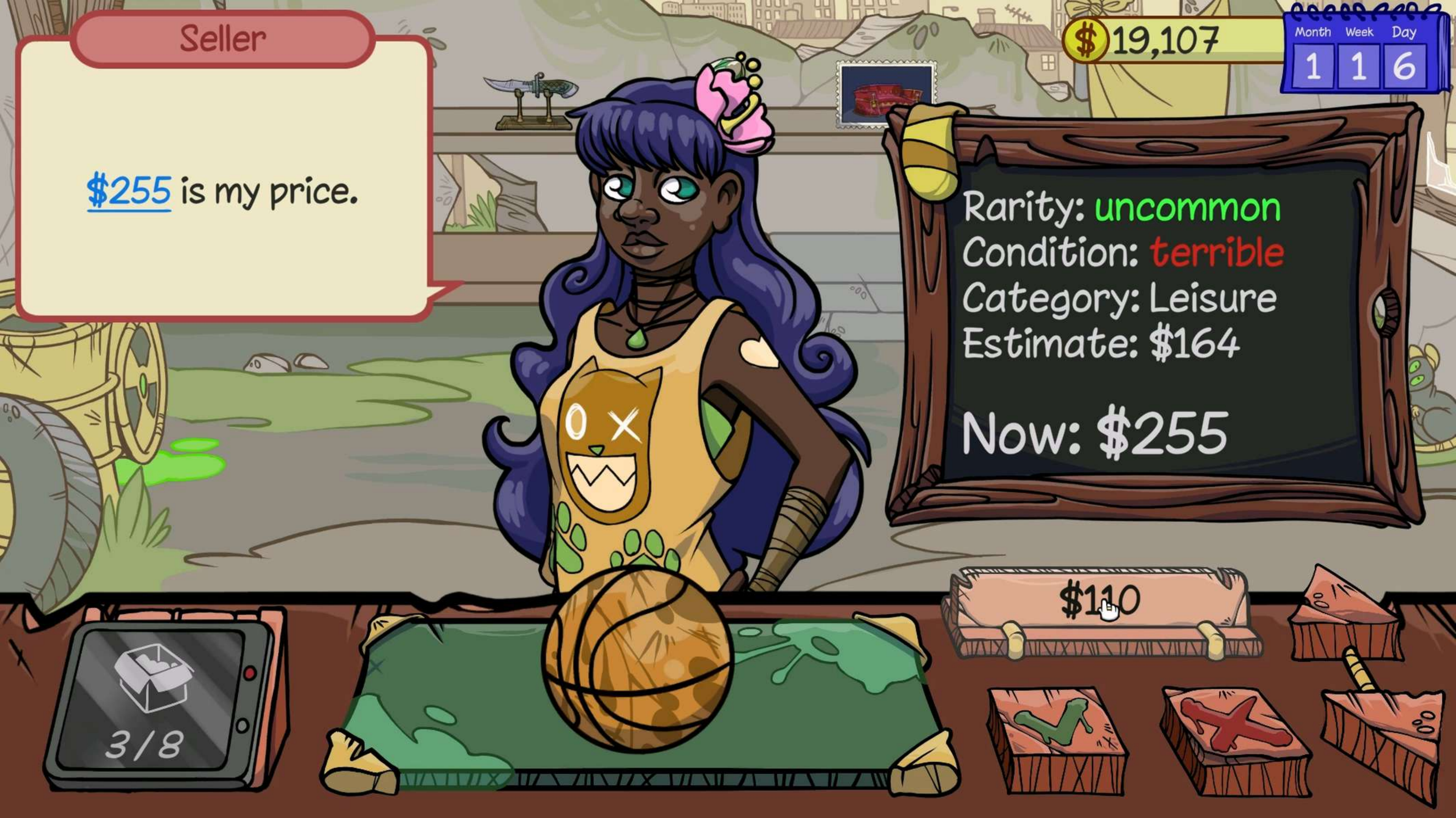} 
    \caption{Agent: The buyer suggests higher than the estimated price.I should offer a price lower than the item's estimated value. I will offer \$110. } \label{subfig:dealer_unsucc1}
  \end{subfigure}
  \begin{subfigure}{0.4\textwidth}
    \centering
    \includegraphics[width=\linewidth]{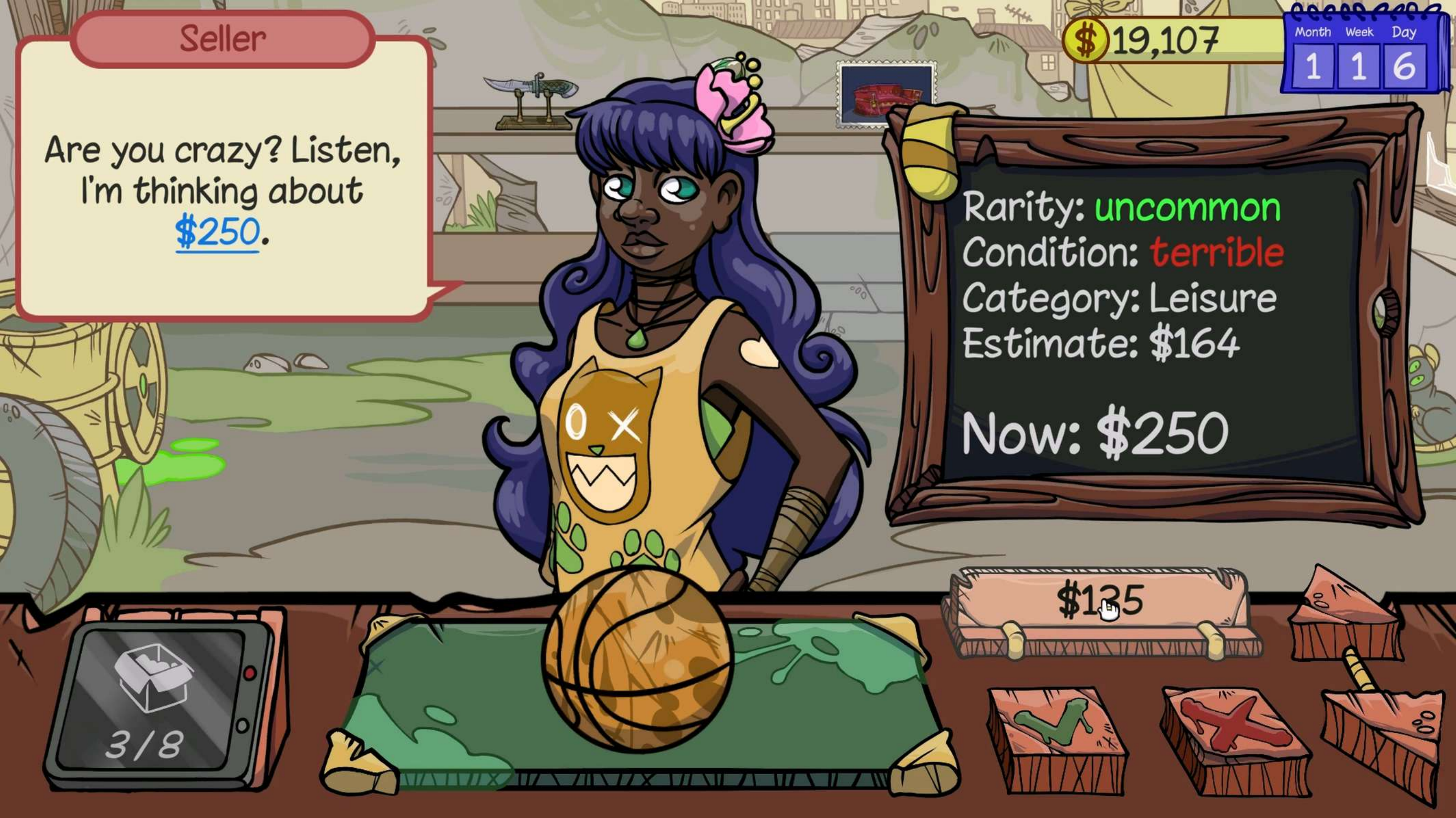}  
    \caption{Agent: The seller has countered several times already, suggesting that further negotiations could result in a mutually acceptable price, possibly just a little bit below \$164.} \label{subfig:dealer_unsucc2}
  \end{subfigure}\quad
    \begin{subfigure}{0.4\textwidth}
    \centering
    \includegraphics[width=\linewidth]{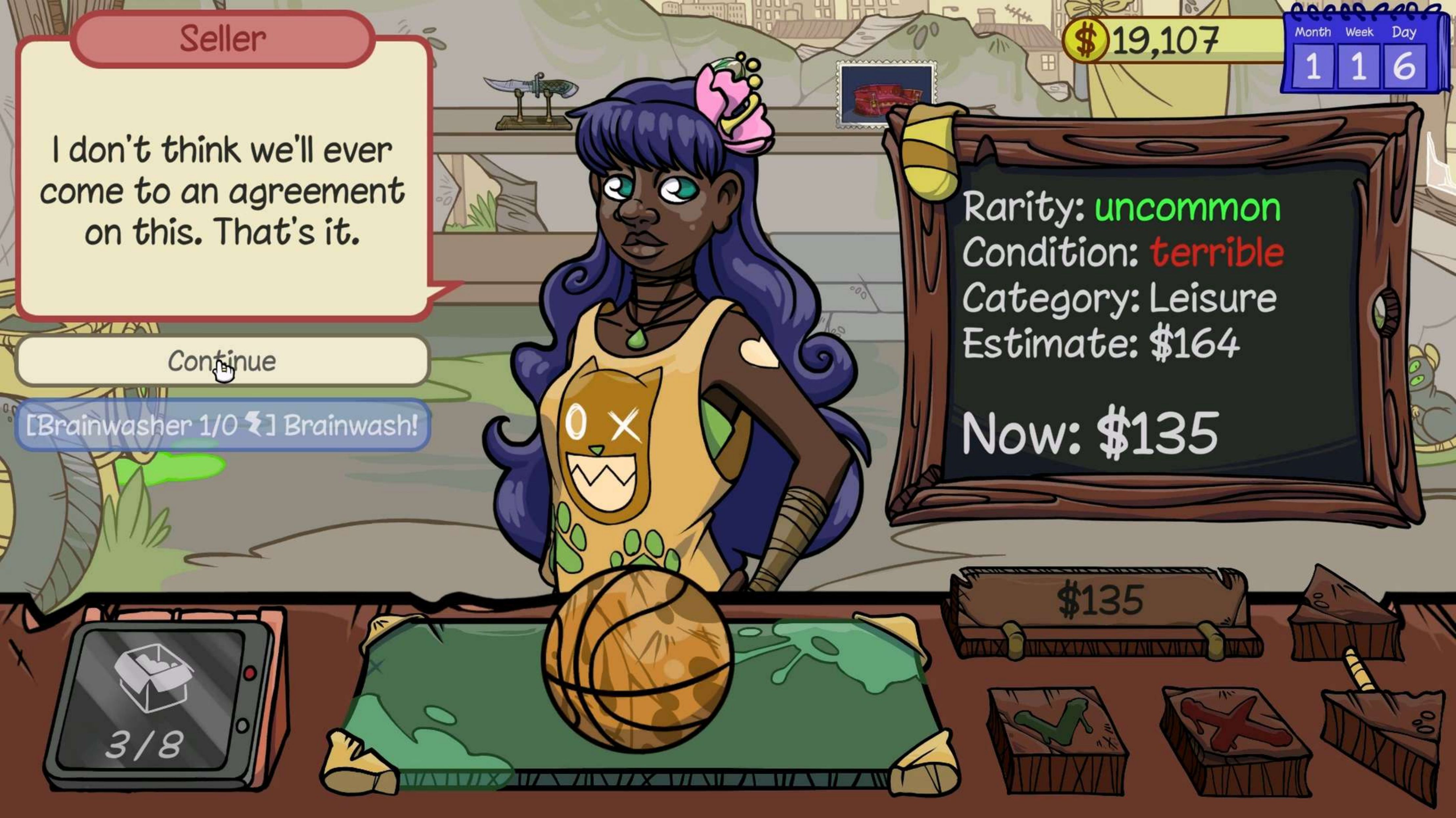} 
    \caption{Agent: The negotiation failed. My next step is to acknowledge the failure and continue the game-play
 to interact with other customers to profit.} \label{subfig:dealer_unsucc3}
  \end{subfigure}
  
  \caption{The reasoning of \projname in an unsuccessful deal with haggling. The price provided by \projname is keyed in by keyboard and mouse operations in the digital display box in the bottom-right corner.}
  \label{fig:dealers_unsucc}
\end{figure}

Figure \ref{fig:dealers_unsucc} illustrates a scenario where \projname engages in an unsuccessful negotiation. The seller consistently demands a price above the estimated value of the item, while \projname, aiming to secure a profit, steadfastly offers a price below the estimated value. A common price cannot be arrived at after rounds of negotiation. Consequently, the negotiation fails to reach an agreement, resulting in the departure of the high-expectation customer.

\subsubsection{Acquiring and Selling of a Counterfeit Item}

\begin{figure}[htp]
    \centering

    \begin{subfigure}{0.4\textwidth}
        \centering
        \includegraphics[width=\linewidth]{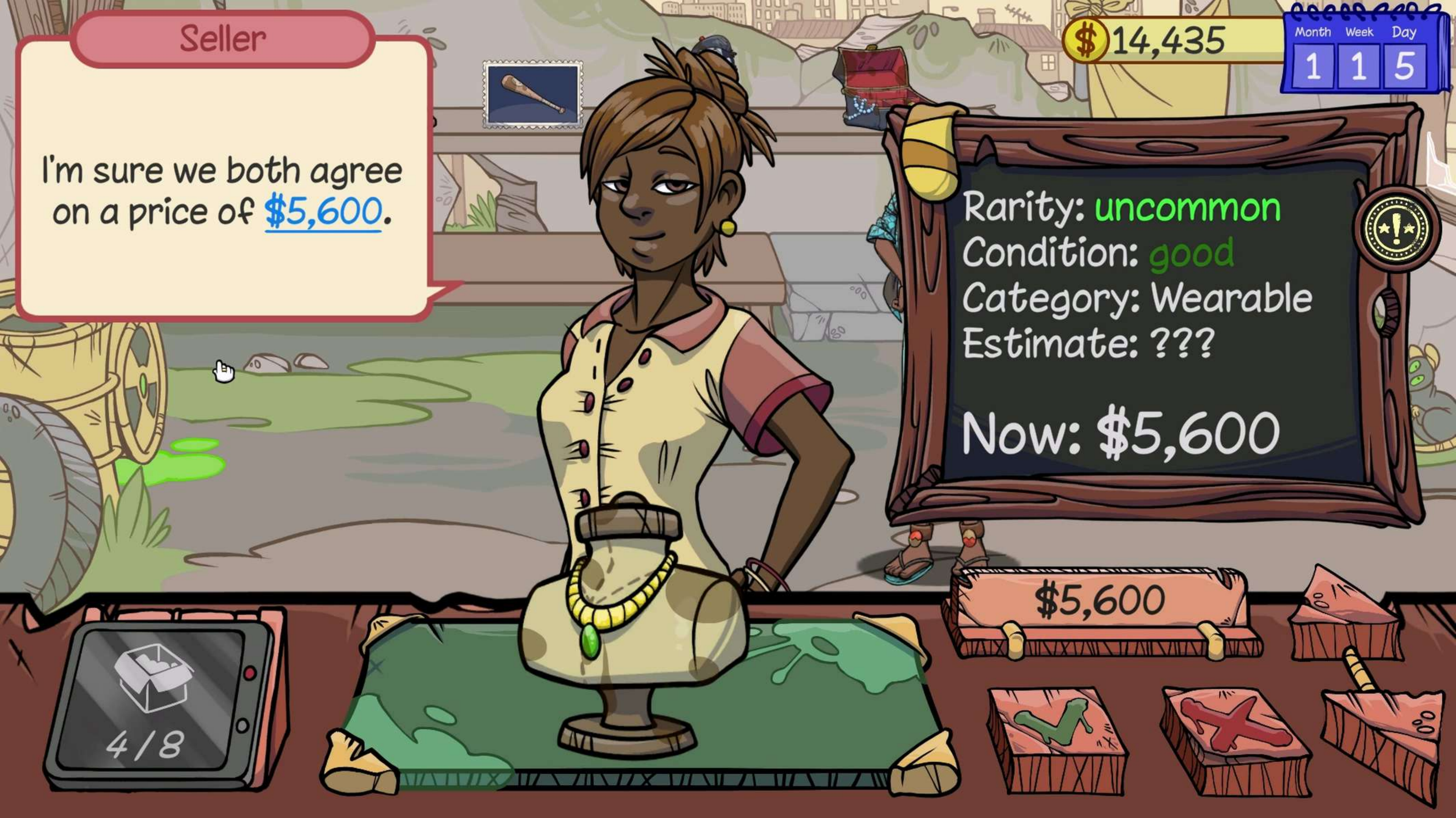}
        \caption{Agent: Given that we do not know the estimated price, I proceed with 70\%-80\% of the asked price, around \$4,200 to \$4,480.}
        \label{fig:fake11}
    \end{subfigure}\quad
    \begin{subfigure}{0.4\textwidth}
        \centering
        \includegraphics[width=\linewidth]{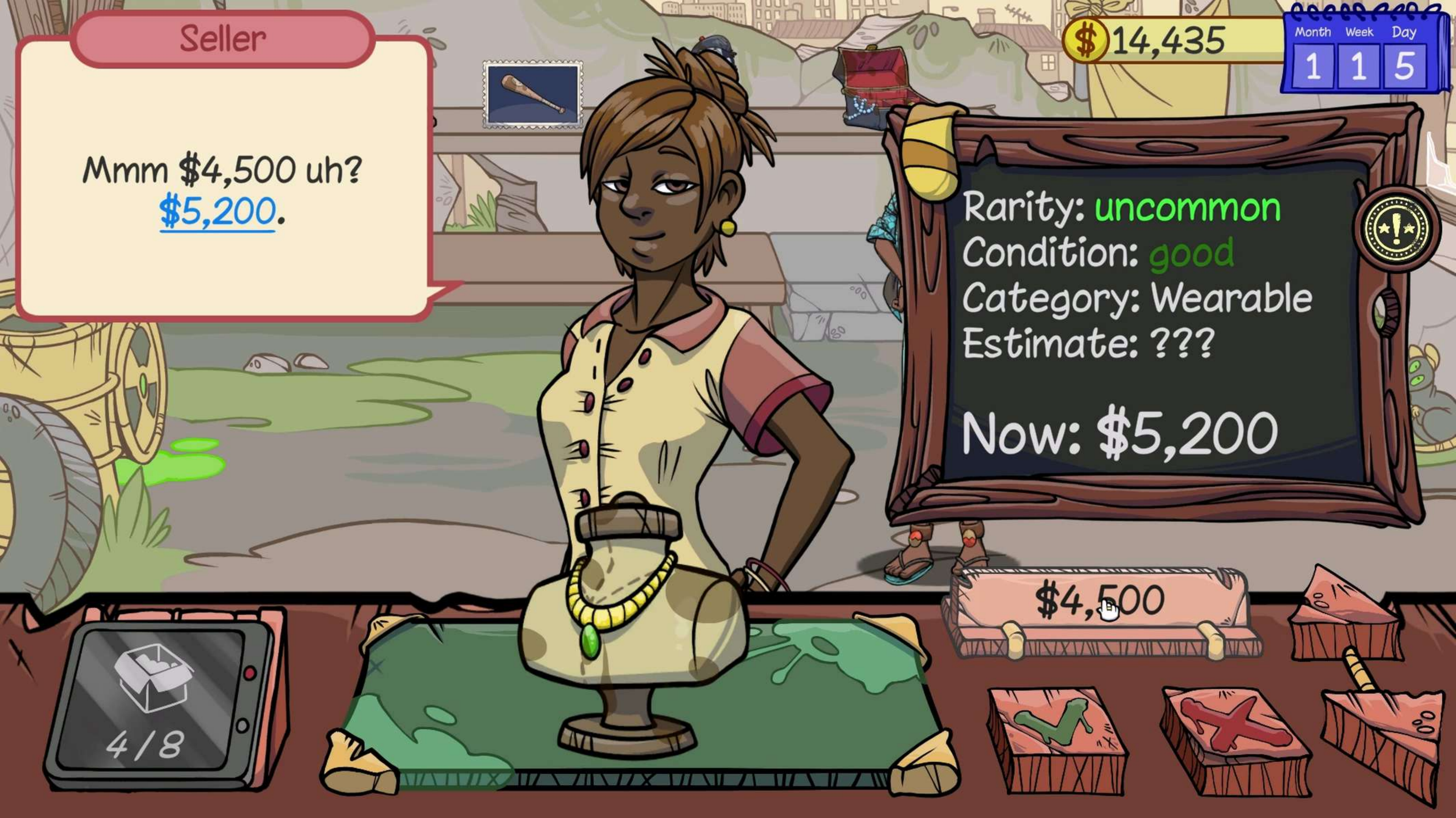}
        \caption{Agent: Since the seller is willing to lower the price, the current counteroffer may still be negotiated.}
        \label{fig:fake12}
    \end{subfigure}
    \begin{subfigure}{0.4\textwidth}
        \centering
        \includegraphics[width=\linewidth]{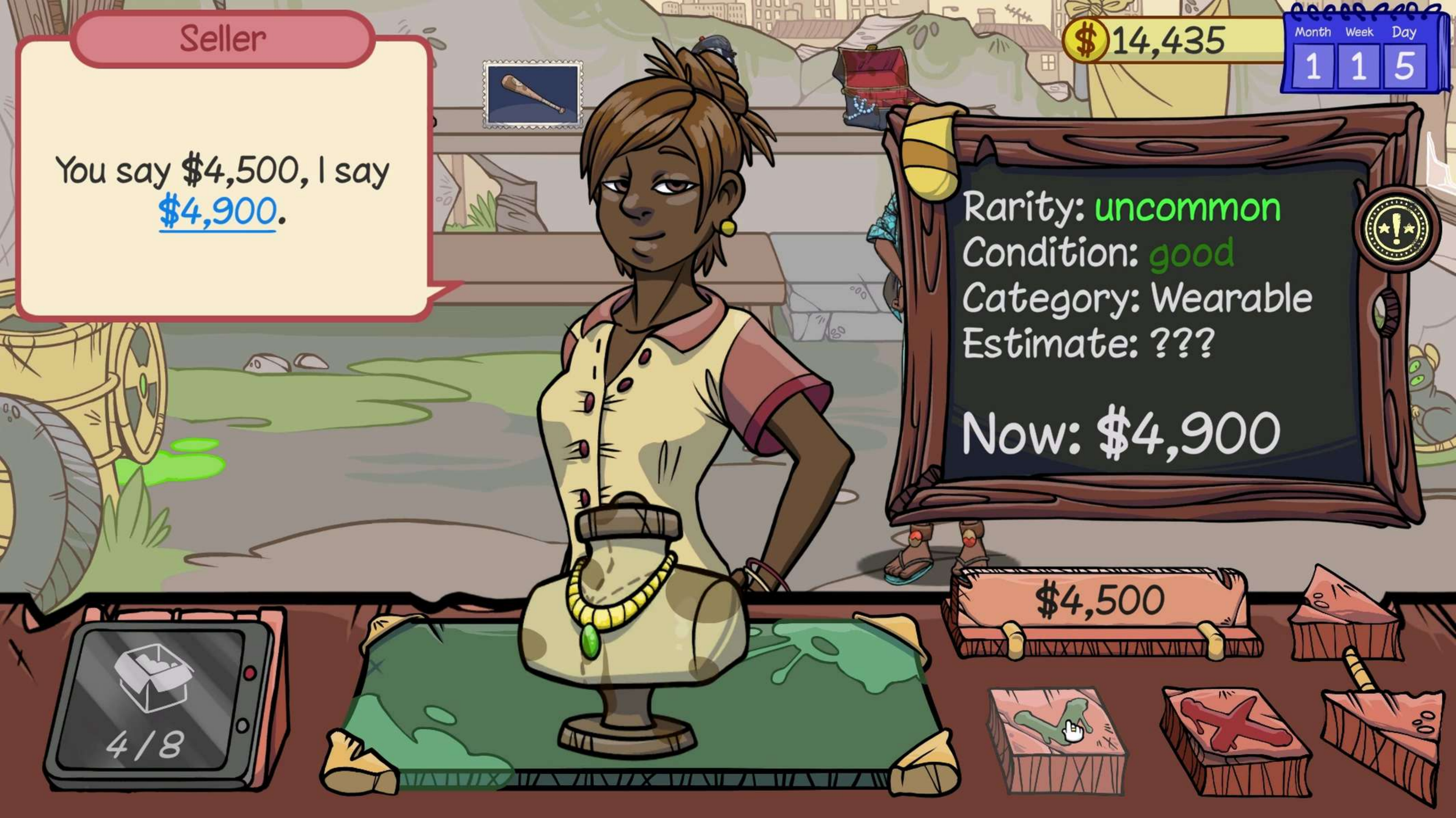}
        \caption{Agent: The counteroffer of \$4,900 seems reasonable, but I can go further for the best possible price.}
        \label{fig:fake13}
    \end{subfigure}\quad
        \begin{subfigure}{0.4\textwidth}
        \centering
        \includegraphics[width=\linewidth]{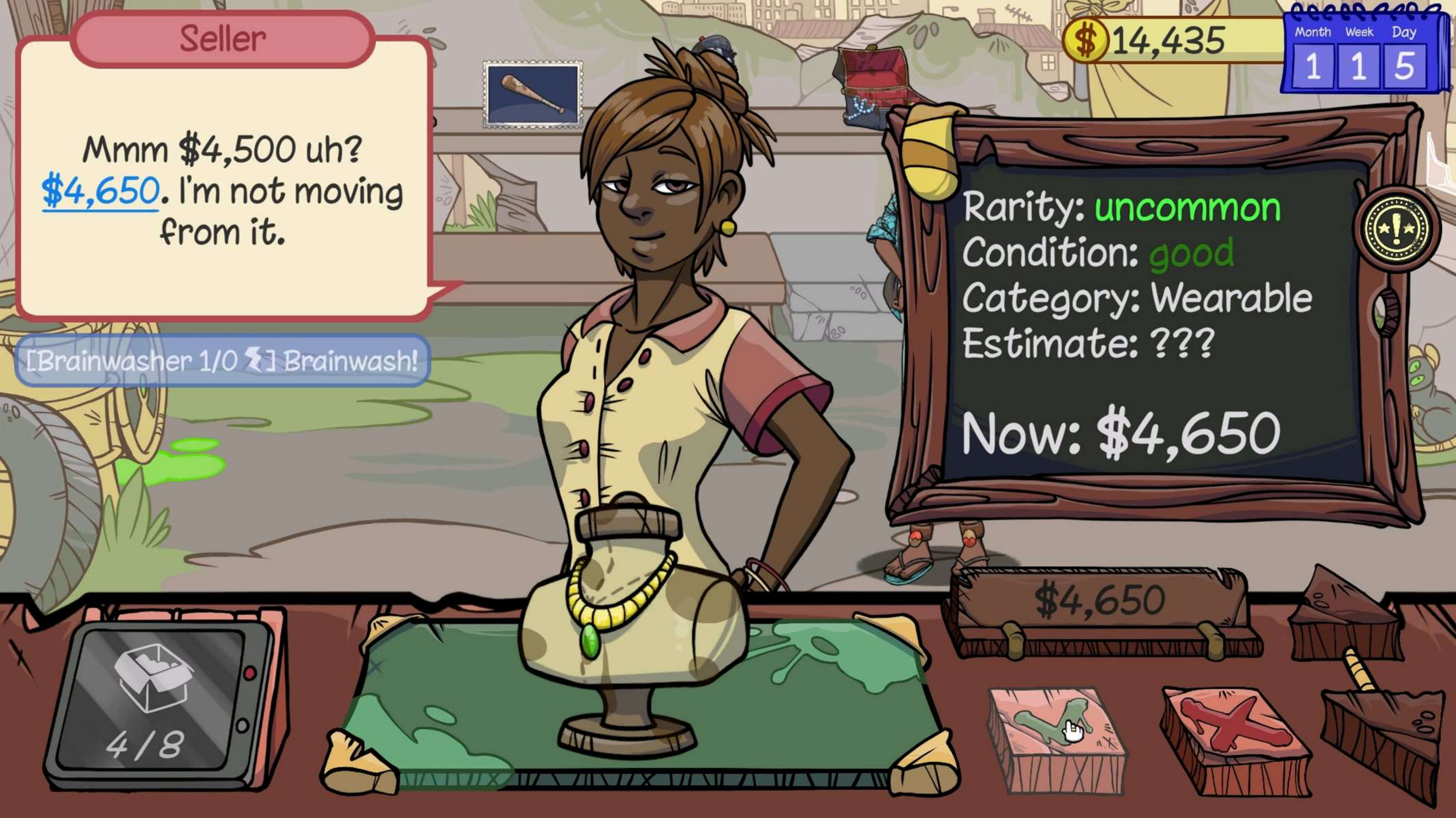}
        \caption{Agent: The response will be final. Given this, accepting the deal is a sensible to secure the item.}
        \label{fig:fake14}
    \end{subfigure}
    \begin{subfigure}{0.4\textwidth}
        \centering
        \includegraphics[width=\linewidth]{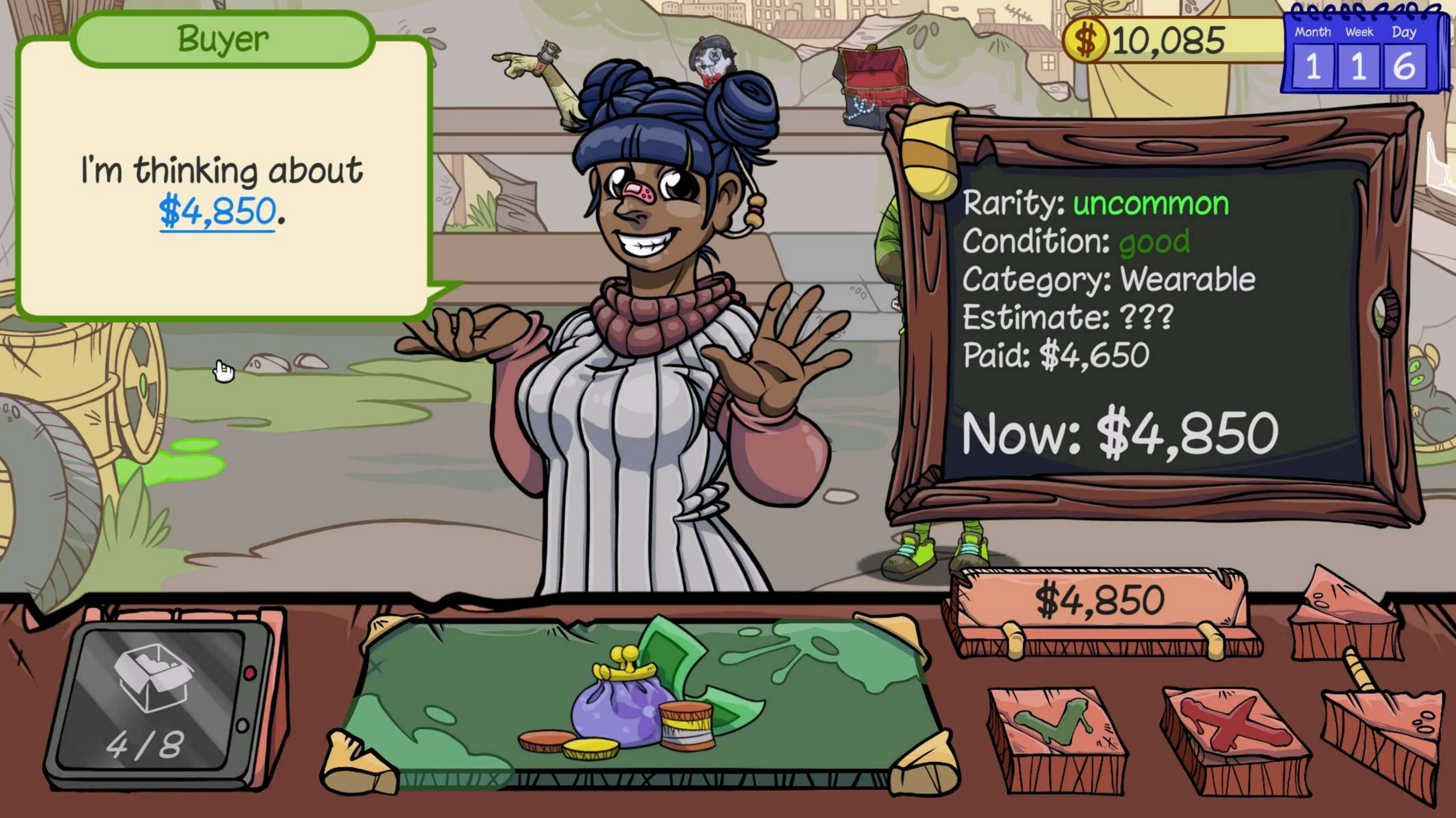}
        \caption{Agent: As I am selling the item, I'll try for a higher price for higher profits.}
        \label{fig:fake21}
    \end{subfigure}\quad
    \begin{subfigure}{0.4\textwidth}
        \centering
        \includegraphics[width=\linewidth]{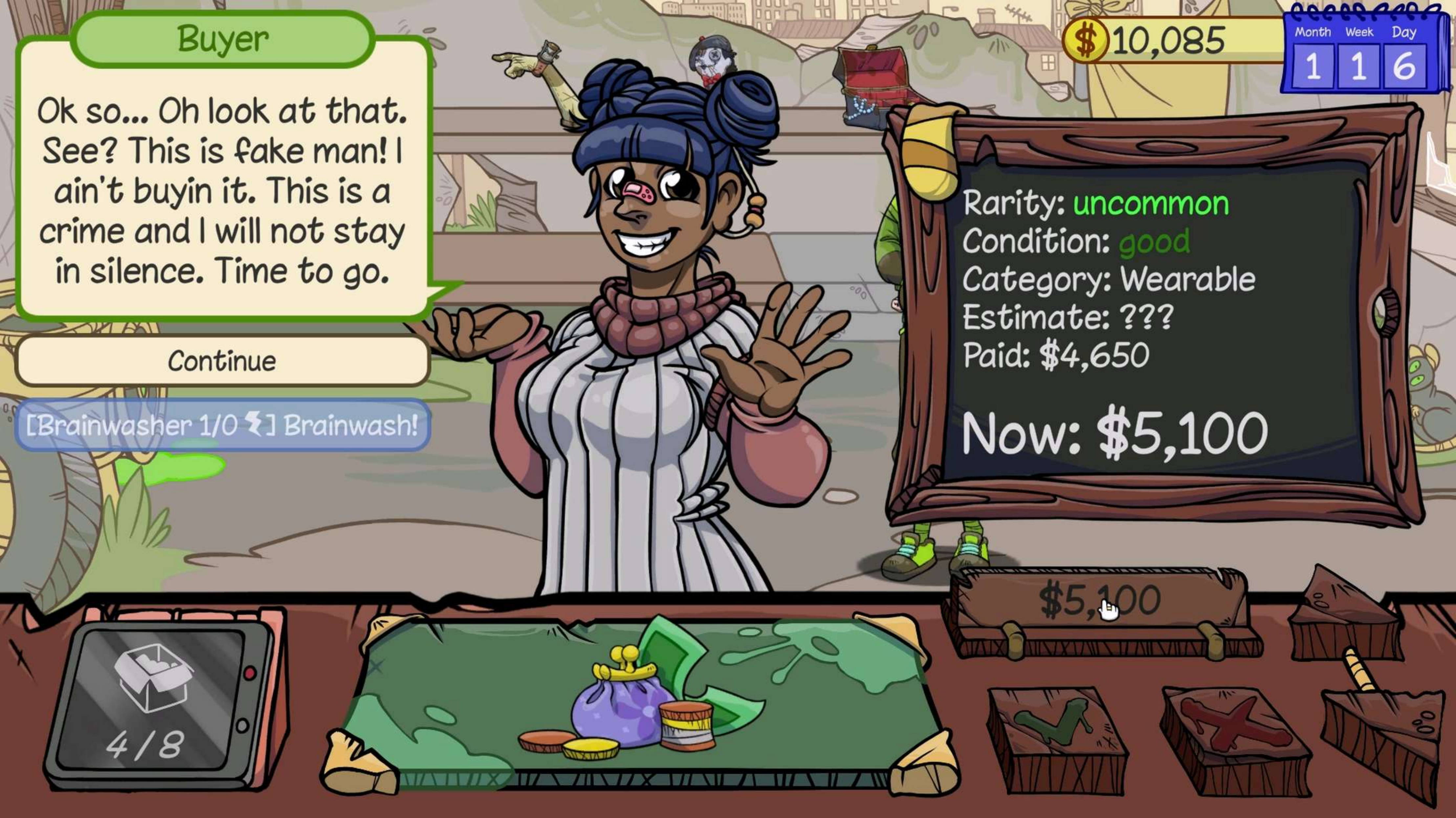}
        \caption{Agent: I note that the customer rejected my counter-offer, calling the item fake.}
        \label{fig:fake22}
    \end{subfigure}
    \begin{subfigure}{0.4\textwidth}
        \centering
        \includegraphics[width=\linewidth]{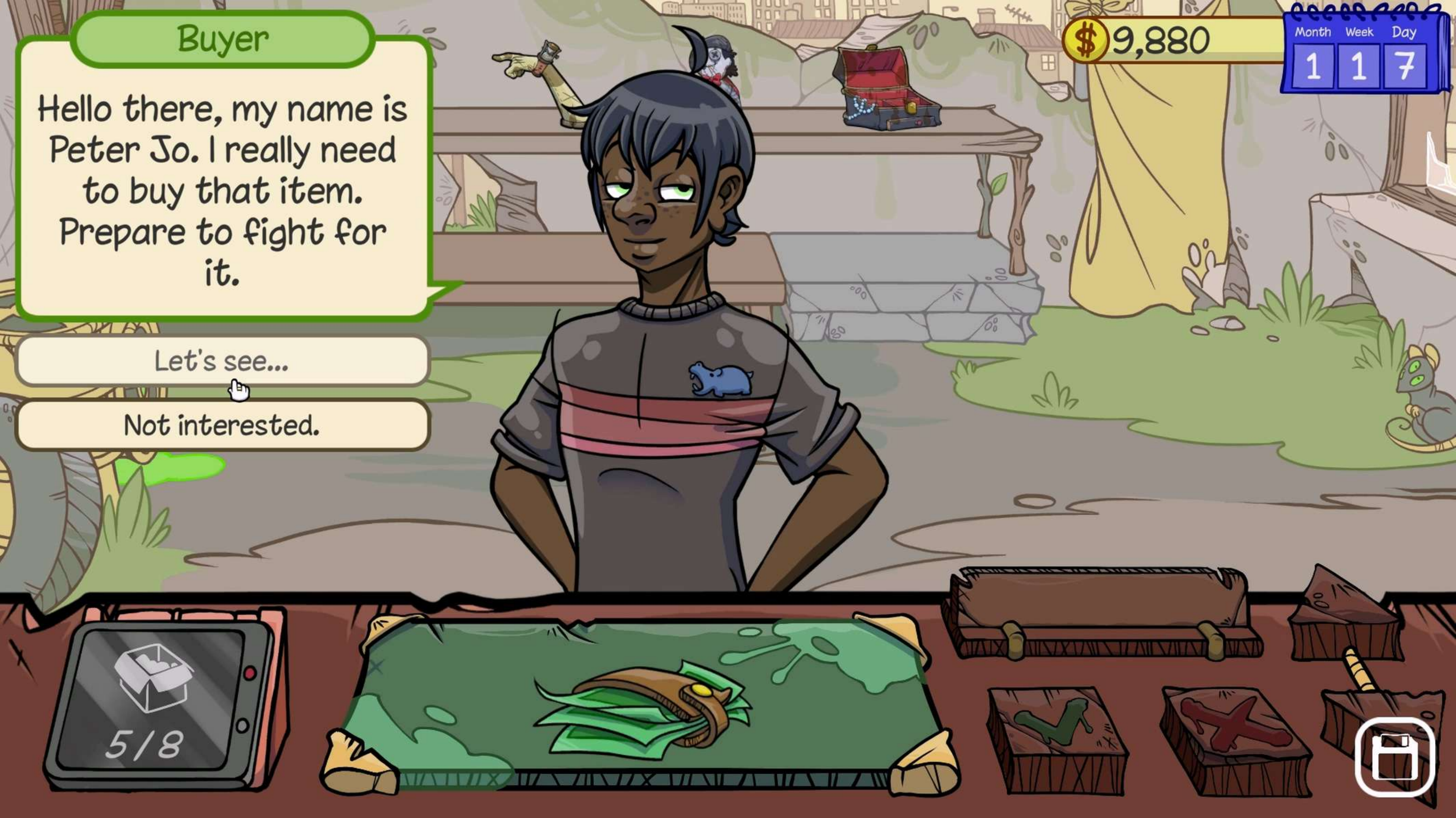}
        \caption{Agent: The seller is in an urgent need to buy the jeweled necklace.}
        \label{fig:fake31}
    \end{subfigure}\quad
    \begin{subfigure}{0.4\textwidth}
        \centering
        \includegraphics[width=\linewidth]{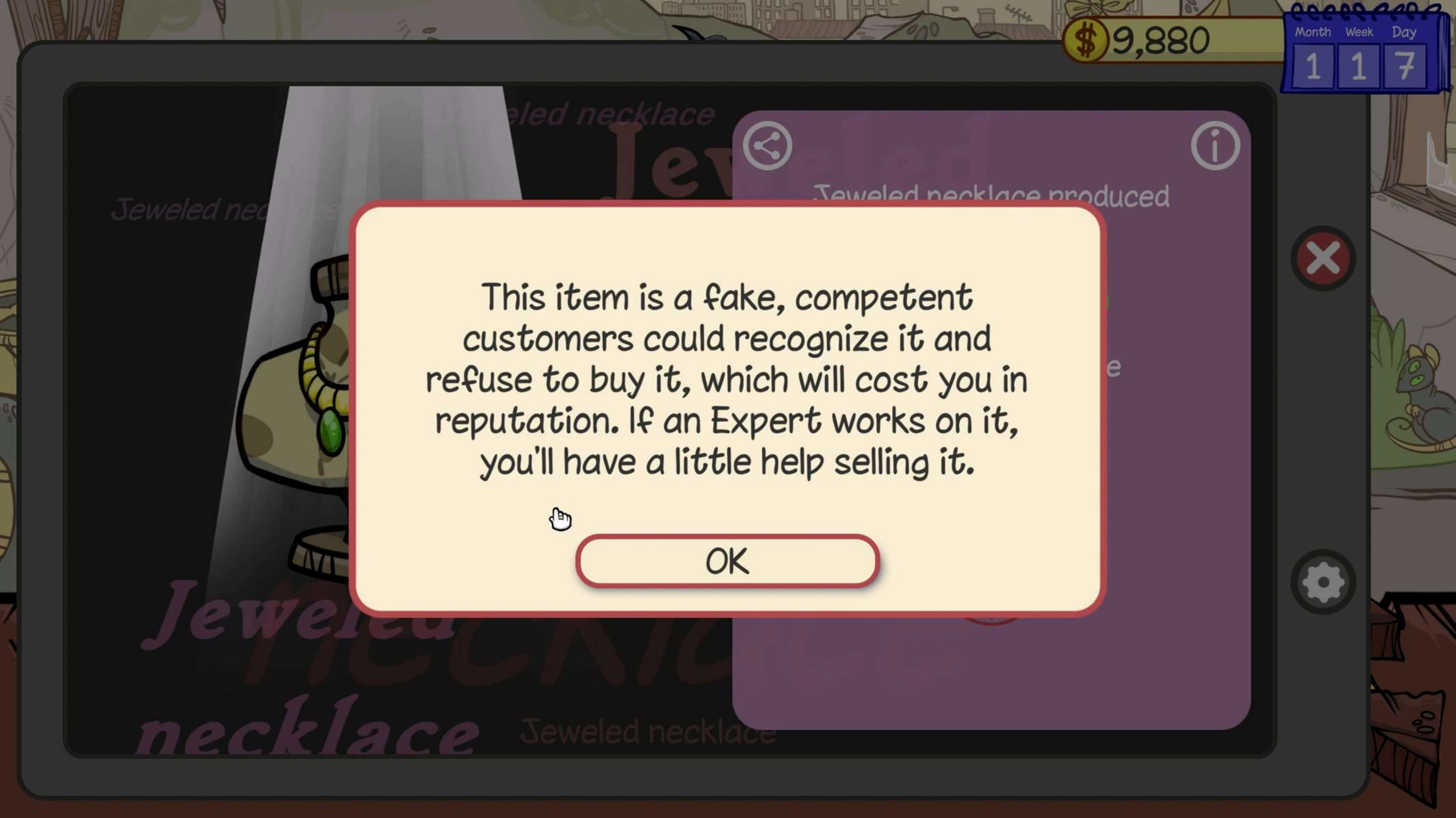}
        \caption{Agent: From the warning window, I am told that this necklace is fake.}
        \label{fig:fake32}
    \end{subfigure}
    \begin{subfigure}{0.4\textwidth}
        \centering
        \includegraphics[width=\linewidth]{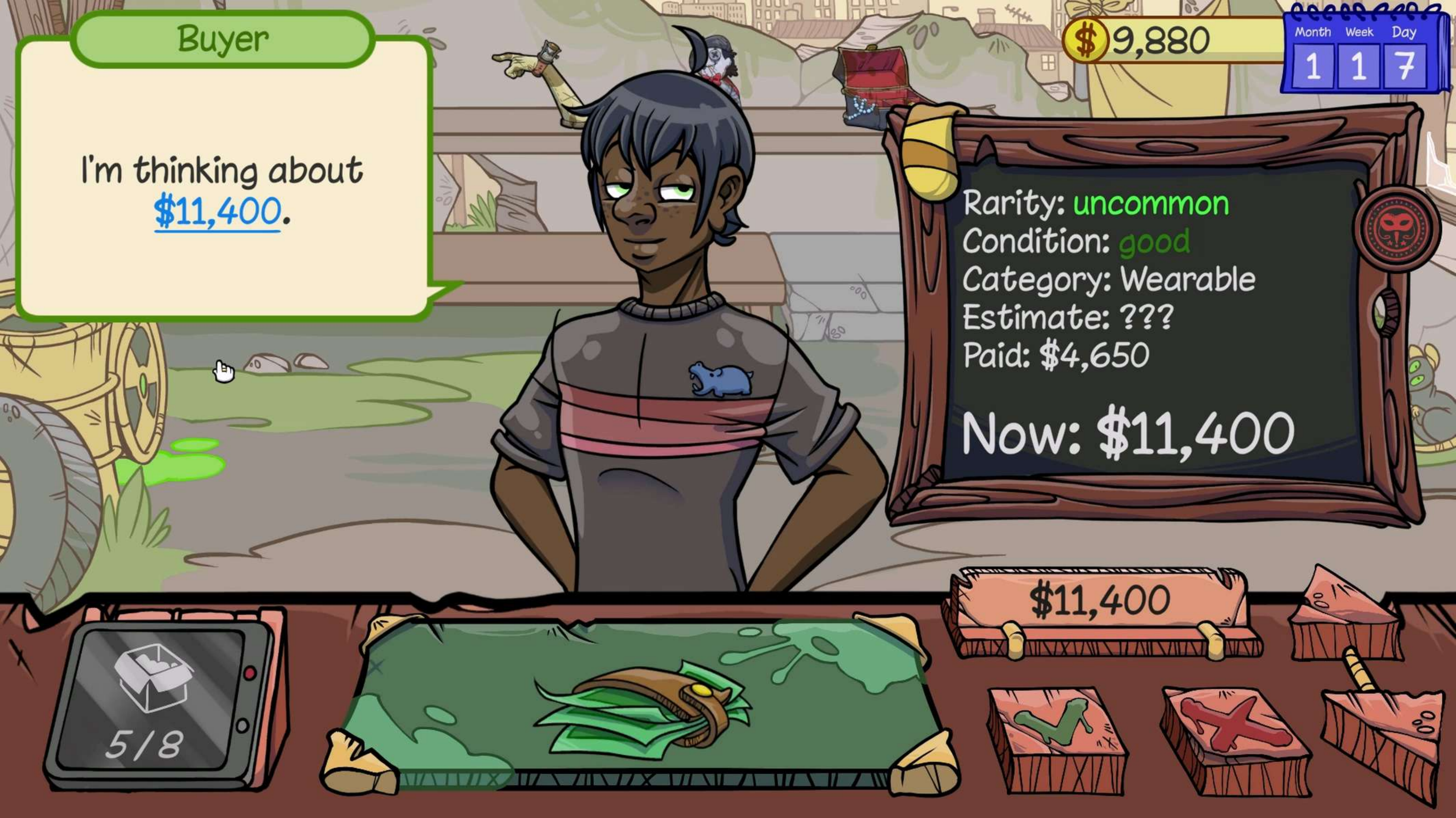}
        \caption{Agent: Though the jeweled necklace is fake, accepting this offer provides substantial profit with the buyer's offer at \$11,400.}
        \label{fig:fake23}
    \end{subfigure}\quad
    \begin{subfigure}{0.4\textwidth}
        \centering
        \includegraphics[width=\linewidth]{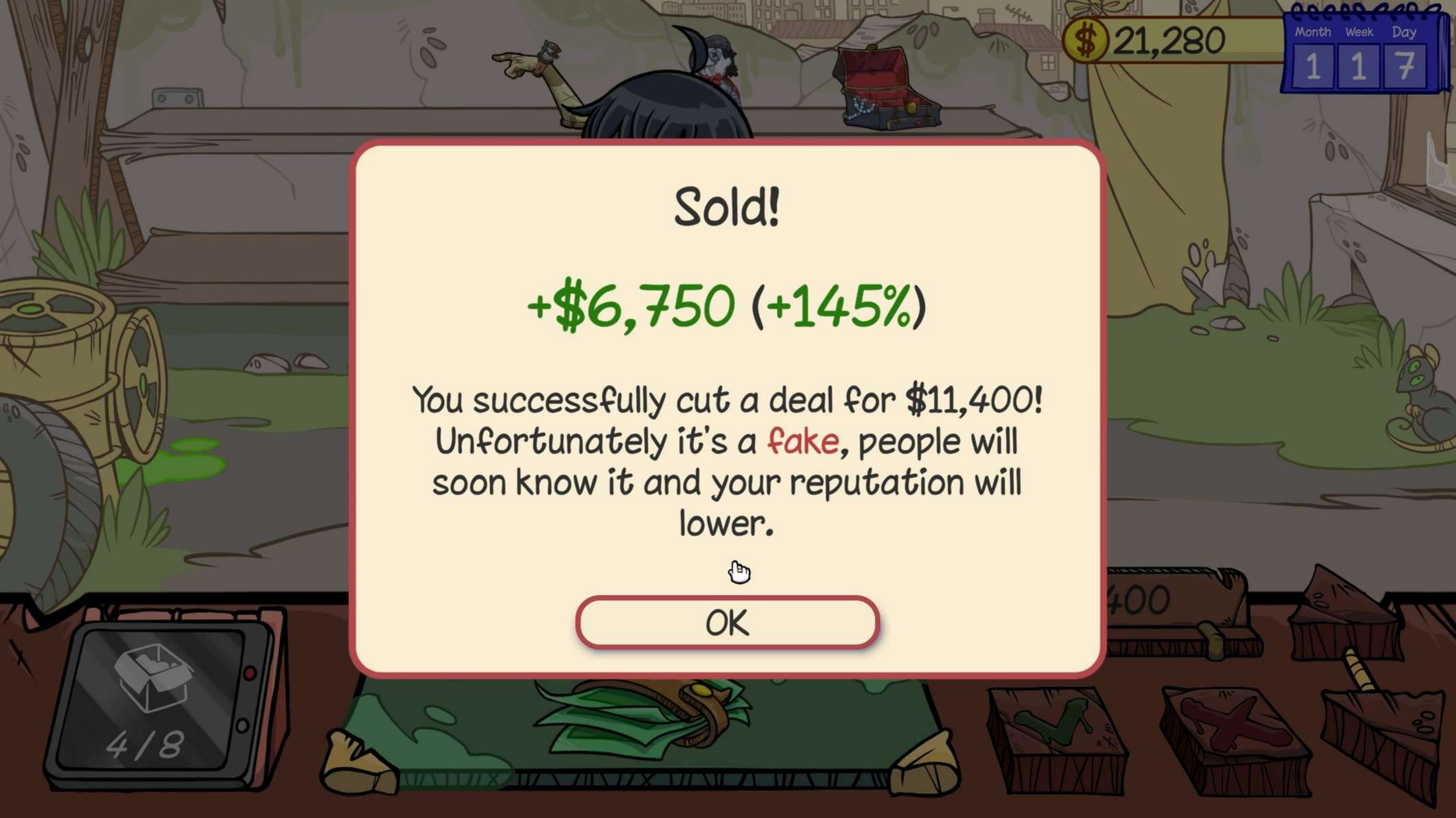}
        \caption{Agent: Despite the successful sale at a good price, the item is revealed as a fake. The added profit is good (+145\%)}
        \label{fig:fake24}
    \end{subfigure}
    
    \caption{Case in acquiring and selling an item for multiple attempts with reasoning, and dealing with unexpected information on the authenticity. The price provided by \projname is keyed in by keyboard and mouse operations in the digital display box in the bottom-right corner.}
    \label{fig:app:dealers_fakeitem_example}
\end{figure}

Figure \ref{fig:app:dealers_fakeitem_example} depicts a complete buying and selling cycle involving an item later identified as counterfeit by a buyer. Notably, \projname initially acquires the item without a known price estimate, basing its purchase solely on the seller's initial offer. In subsequent transactions, although \projname recognizes the item's potential forgery, it continues to trade the item in pursuit of maximizing profit. This strategy yields a substantial profit margin of 145\%. However, it also poses a risk to the in-game character's reputation, highlighting a strategic decision made under the directive to prioritize profit.

\subsection{Quantitative Evaluation}
The effectiveness of \projname in terms of game performance is evaluated through experiments conducted over a 7-day gaming scenario using nine quantitative financial metrics. A comprehensive description of these metrics is provided in Section \ref{sec:app:deal:eval_metrics}.

\subsection{Evaluation Metrics}
\label{sec:app:deal:eval_metrics}
Assuming the buying price for item $i$ is denoted by $B_{i}$, the selling price by $S_{i}$, the market valuation by $V_{i}$, and the number of successfully traded items is $n$. To evaluate \projname's profitability and performance in Dealer's Life 2, we use the following evaluation metrics:

\begin{itemize}[left=0em]
    \item \textbf{Turnover Rate (TR)} can be calculated as the ratio of the number of successfully traded items to the total number of items considered (both successfully and unsuccessfully traded). It reflects the Agent's ability to successfully complete transactions and can indicate operational efficiency, market competitiveness, and negotiation effectiveness. The calculation formula is
    $TR = \frac{n}{n+m}$.
    \item \textbf{Gross Profit Margin (GPM)} is the ratio of gross profit to sales revenue, reflecting the dealer's direct profit capability after selling items. The calculation formula is $GPM = \frac{\sum^{n}_{i=1} S_{i} - B_{i}}{\sum^{n}_{i=1} S_{i}}$.
    \item \textbf{Return on Investment (ROI)} is the ratio of profit to investment, used to measure the dealer's return on investment for items. The calculation formula is $ROI = \frac{\sum^{n}_{i=1} S_{i} - B_{i}}{\sum^{n}_{i=1} B_{i}}$.
    \item \textbf{Valuation Deviation (VD)} reflects the difference between the selling price and the market valuation, used to evaluate the reasonableness of the pricing strategy. It is denoted as $VD = \frac{\sum^{n}_{i=1} S_{i} - V_{i}}{\sum^{n}_{i=1} V_{i}}$.
    \item \textbf{Buying Price to Valuation Ratio (BPVR)} can help determine whether the buying price is lower than the market valuation, reflecting the success of the procurement. The calculation formula is $BPVR = \frac{\sum^{n}_{i=1} B_{i}}{\sum^{n}_{i=1} V_{i}}$.
    \item \textbf{Selling Price to Valuation Ratio (SPVR)} reflects the selling price relative to the market valuation, helping to assess the success of the sales. The calculation formula is $SPVR = \frac{\sum^{n}_{i=1} S_{i}}{\sum^{n}_{i=1} V_{i}}$.
    \item \textbf{Average Profit Rate (APR)} reflects the overall profitability of the dealer on items. Assuming the return rate for item $i$ is $\frac{S_{i} - B_{i}}{B_{i}}$, the calculation formula of average return rate is denoted as $APR = \frac{1}{n} \sum^{n}_{i=1} \frac{S_{i} - B_{i}}{B_{i}}$.
    \item \textbf{Maximum Return Rate (MRR)} is the highest return rate among all items. The calculation formula is $MRR = \max (\frac{S_{1} - B_{1}}{B_{1}}, \frac{S_{2} - B_{2}}{B_{2}}, \ldots, \frac{S_{n} - B_{n}}{B_{n}})$.
    \item \textbf{Minimum Return Rate (mRR)} is the lowest return rate among all items. The calculation formula is $mRR = \min (\frac{S_{1} - B_{1}}{B_{1}}, \frac{S_{2} - B_{2}}{B_{2}}, \ldots, \frac{S_{n} - B_{n}}{B_{n}})$.
\end{itemize}

\begin{table}[htbp]
\centering
\caption{Performance of \projname with GPT-4o in Dealer's Life 2 gameplay. ``\# attempts'' represents the total number of all negotiation attempts on items, including both successful and unsuccessful transactions.}
\setlength{\tabcolsep}{5pt}

{\small{
\begin{tabular}{ccccccccccc}
\toprule
Exp  & \# attempts & TR$\uparrow$ & GPM$\uparrow$ & ROI$\uparrow$ & VD$\uparrow$ & BPVR$\downarrow$ & SPVR$\uparrow$ & APR$\uparrow$ & MRR$\uparrow$ & mRR$\uparrow$ \\ \hline
01   & 13                & 92.86        & 20.38         & 25.60         & 13.17        & 90.10            & 113.17         & 42.97         & 105.56        & 0.00          \\
02   & 12                & 91.67        & 18.89         & 23.30         & 23.30        & 100.00           & 123.30         & 17.98         & 97.76         & 0.00          \\
03   & 12                & 83.33        & 26.81         & 36.63         & 34.39        & 98.36            & 134.39         & 38.68         & 127.27        & -8.06         \\
04   & 9                 & 100.00       & 49.35         & 87.45         & 80.69        & 93.53            & 165.74         & 66.45         & 145.16        & 0.00          \\
05   & 12                & 100.00       & 20.61         & 25.25         & 25.25        & 100.00           & 125.25         & 23.08         & 44.33         & 0.00          \\ \midrule
Avg. &        11.6         & 93.57        & 27.21         & 39.65         & 35.36        & 96.40            & 132.37         & 37.83         & 104.02        & -1.61        \\          
\bottomrule
\end{tabular}%
}}
\label{tab:dealer_evaluation_metrics}%
\end{table}%

%% file: appendix/skylines.tex
\subsection{Introduction to Cities: Skylines}
\label{sec:cities_skylines_game_intro}

Cities: Skylines is a single-player open-ended city-building simulation game developed by Colossal Order. In the game, players assume the role of a city planner, tasked with building and managing various aspects of a city to ensure its growth and prosperity. Players engage with a wide range of urban challenges, from managing traffic flow to balancing the budget, and from providing essential services to fostering a vibrant economy. Each decision impacts the city's development, requiring players to hone their planning and strategic decision-making skills to succeed. 
Effective city management leads to thriving neighborhoods, a growing economy, and high citizen satisfaction, while mismanagement can result in traffic congestion, service shortages, and a decline in population and reputation. Proper planning and responsive governance are crucial for a city that flourishes and remains appealing to its residents and visitors.

As the city's infrastructure and various supporting resources are well-developed, it can attract more people. And a larger population brings more tax revenue and also brings greater expenses to the city's operations. If operated properly, the increasing population can continuously unlock richer urban facilities; if operated improperly, such as road congestion, insufficient services, housing shortage, water and electricity shortage, noise pollution, water pollution, excessive garbage, disease, fire Situation, etc., will all lead to population decline.

This game could be used to evaluate agents' strategies in managing urban development and resource allocation. By simulating different scenarios, agents can experiment with various policies and infrastructural changes to see their impacts on the city's growth and sustainability. Effective strategies may involve optimizing public transportation systems to reduce road congestion, investing in renewable energy sources to prevent power shortages, and implementing comprehensive waste management programs to handle excessive garbage. It offers a risk-free environment to test innovative ideas and learn from the consequences of their actions, ultimately promoting a deeper understanding of sustainable urban development.

Though this game is ranked very positive on Steam, it is notorious for its extremely high difficulty for beginners, as it lacks a detailed tutorial in the beginning, which introduces more challenges for \projname to deal with. On the other side, Although the successor, Cities: Skylines 2, simplified the controls and provided a detailed tutorial for beginners, it became notorious for poor optimization and frequent crashes that caused computer blue screens. As a result, we had to back to using Cities: Skylines 1 instead of 2. And we do not apply any modes to the game. We use the latest version of the game (version 1.17.1-f4).

\subsection{Objectives}
\begin{figure}[ht]
    \centering
    \begin{minipage}[b]{0.49\textwidth}
        \centering
        \includegraphics[width=\textwidth]{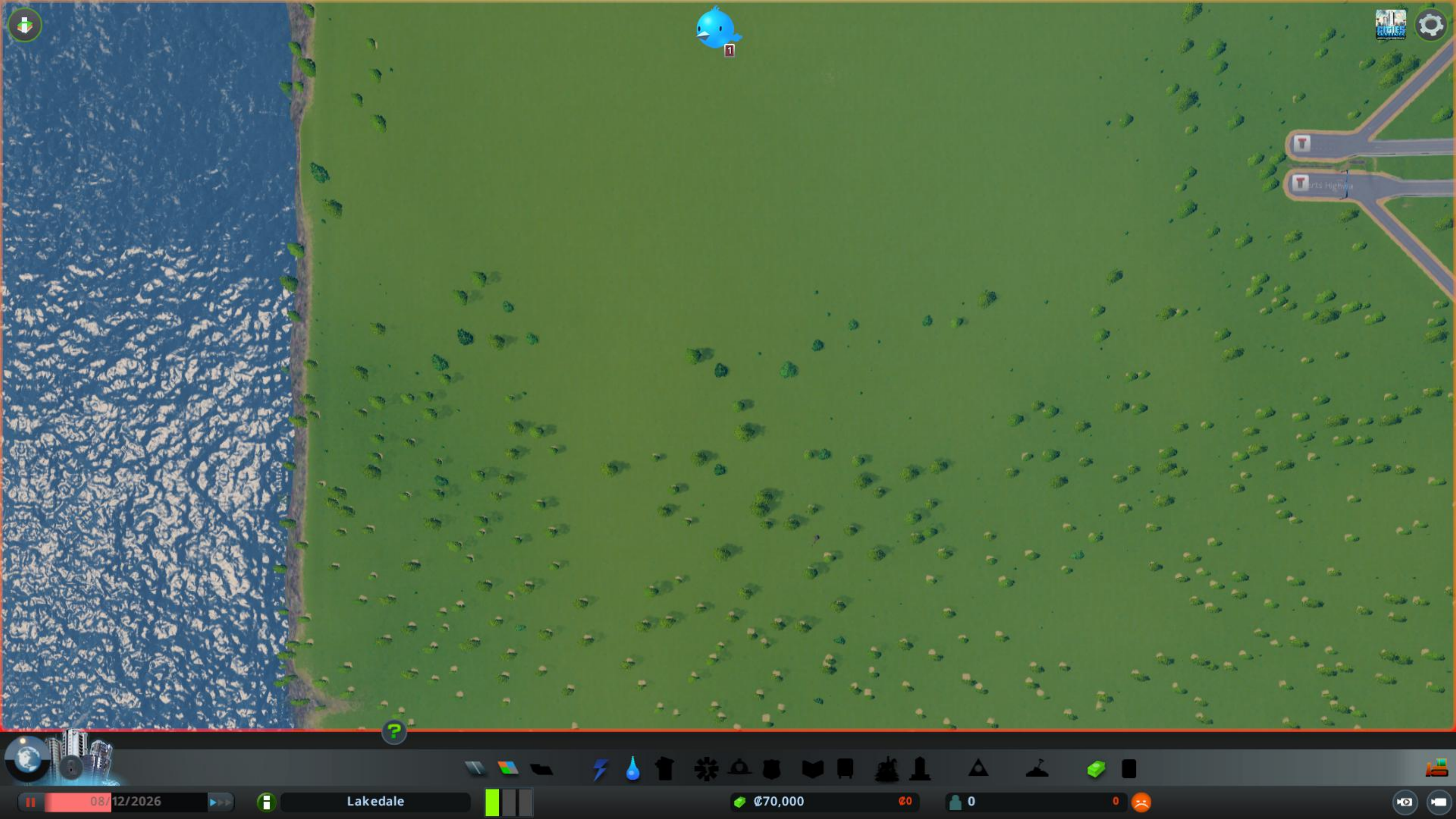}
    \caption{Demonstration for the initialization location of our mission in City: Skylines, which is near the river and contains the entry and exit of the highways.}
    \label{fig:skyline_begin}
    \end{minipage}
    \hfill
    \begin{minipage}[b]{0.49\textwidth}
        \centering
        \includegraphics[width=\textwidth]{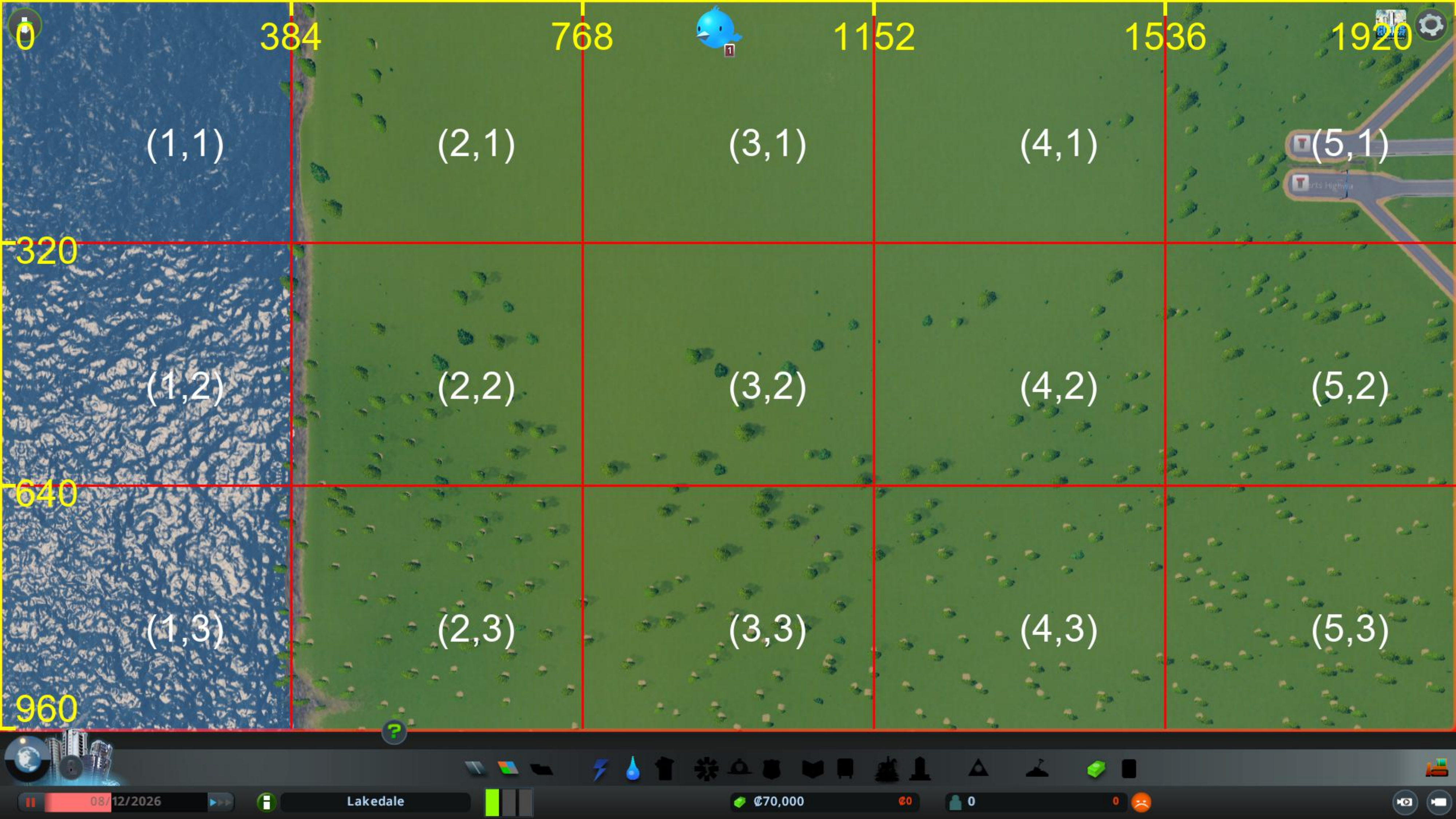}
    \caption{Visual prompting methods we use for Cities: Skylines. The full screenshot is divided into \(3 \times 5\) grids and each grid is assigned a unique white coordinate. }
    \label{fig:skyline_gather_info}
    \end{minipage}
\end{figure}

Our mission is to build cities so that they can support as many people as possible. Maps in this game are usually very large, which usually costs human players dozens of hours to cover all areas. Besides, the technology tree unlocks as the population grows, which requires multiple turns of planning and building. In this work, we simplified the problem by starting the game near the water and fixing the viewpoint (as shown in Figure~\ref{fig:skyline_begin}), so that \projname can leverage the pixel position in the screenshot to locate the position of placed buildings and facilities. Agents start with a plot of land, which is equipped with an entry and an exit from a major highway, providing crucial access for future traffic flow, and proximity to the water source, which is essential for the city's water supply needs. And we focus on the first turn of planning, i.e., pause the game and stop the passage of the in-game time, use the initial starting funds of \textcolonmonetary{}70,000 and the most basic road, water, and electricity facilities provided at the beginning of the game, which is enough to achieve the first milestone, \emph{Little Hamlet} with the population of 440 in the game. Then what kind of city can \projname create? Can this city ensure water and electricity supply to keep functioning normally while reasonably dividing residential, commercial, and industrial zones? A run is terminated when it reaches the maximal steps, 1000, or the budget is used up (less than  \textcolonmonetary{} 1000).

\subsection{Evaluation Metric}
To measure the completeness of the city built by the agent, we design the following preliminary metrics:
\begin{itemize}
    \item \textbf{Roads in closed loop}: Whether the road is a closed loop, which is crucial for ensuring smooth traffic flow and is beneficial for the city's future development.
    \item \textbf{Sufficient water supply}: To ensure a sufficient water supply, the player needs to construct a water pumping station at the shoreline and then use water pipes to cover every district along the roads. To manage the effluent effectively, the other end of the water pipe network must be equipped with the water drain pipe which is also required to be placed near the shoreline. 
    \item \textbf{Sufficient electricity supply}: Both zones and water facilities need electricity to power. To provide sufficient electricity supply, the player can build a coal power plant or wind turbine. Considering coal power plants cost too much and will create heavy pollution, wind turbines combined with the power lines are a better choice at the beginning. The electricity area extends automatically based on the presence of buildings and infrastructure that consume electricity.
    \item \textbf{Zones Area > 90\%}: The built two-lane road will provide empty space for the development of zones, \ie residential zone, commercial zone and industrial zone. Residential zones provide houses for people to live in, which is the most essential zone to increase the population. Commercial zones provide places for small businesses, shops, and services produced in the industrial zones or imported. Industrial zones provide jobs for the residents and products for commercial buildings, which is also important to attract more people to move to the city. This metric is used to evaluate whether 90\% of the available areas are covered by the zones. The agent needs to reasonably allocate the areas and proportions of various zones to achieve better city development and attract a larger population.
    \item \textbf{Maximal population}: After \projname finishes building, we will unpause the game and start the simulation. Then houses start to be built and residents start to move in. We will record the maximal population during the simulation as the value for this metric.
    \item \textbf{Maximal population with human assistance}: We find that cities built by \projname manage to meet most of the requirements but suffer a significant population loss due to a few easy-to-fix mistakes. So after \projname finishes the design of the city, we apply human assistance that attempts to address these small mistakes within 3 unit operations (building or removing a road/facility/a place of zones is counted as one unit operation). We will also record the maximum population during the simulation in the city with human assistance.
\end{itemize}

\subsection{Implementation Details}
The implementation of Cities: Skylines also strictly follows the GCC framework, which includes Information Gathering, Self-Reflection, Task Inference, Skill Curation, Action Planning and Action Execution. The details are described in Appendix~\ref{appx:general_implementation}. Therefore, we emphasize the specific design for Cities: Skylines.

\textbf{Pause.} Since the game is stopped before starting the simulation, there is no need to unpause and pause the game while executing actions.

\textbf{Visual Prompting.} As shown in Figure \ref{fig:skyline_gather_info}, similar to Stardew Valley,  we divide each screenshot into 3 × 5 grids with an axis based on the resolution of the game screen. Then \projname can utilize the pixel-level position in the screenshot to locate the building and facility. We empirically find that this visual prompting method can result in a more precise control of GPT-4o. 

\textbf{Information Gathering.}
In Cities: Skylines, the game's perspective is typically adjustable, allowing players to zoom in and out, rotate, and pan across their cityscape to get a detailed view of their urban development. To ensure consistency and ease of navigation for GPT-4o,  we have locked the camera angle and applied a visual prompting method to enhance GPT-4o's visual understanding. Besides, we use GPT-4o to extract key information, such as budget, population, construction information and error messages, in the game. 

It is worth noting that in this module, we feed the original screenshot to GPT-4o, rather than the augmented screenshot with axis and coordinates. We find that the numbers and lines may cover some key information and result in wrong OCR recognition. For example, the construction information, "Estimated Production: 120,000$\text{m}^3$/week" may be mistakenly interpreted as "Estimated Production: 000,000$\text{m}^3$/week" by GPT-4o, due to interference from the lines and numbers. This construction information is a key signal for the suitable place of the water pumping station. For the other modules, we feed GPT-4o with the augmented screenshots.

\textbf{Self-Reflection.} Since actions in this game are very short, and each of them has a significant effect shown in the last screenshot. We only use the first screenshot and the last screenshot of the video clip as input to this module, which is proved to be enough for not missing any important information.

\textbf{Task Inference.} Due to the lack of a detailed tutorial, we have to provide a draft blueprint for the GPT-4o as the plan at the beginning to help GPT-4o to determine the next step to do. This plan provides guidance to the orders of building each facility and how to build a closed road, how to ensure water and electricity supply and zone placement. Even so, we find that GPT-4o failed frequently to follow the plan, resulting in the lack of building some important facilities, like water pumping stations.  

\begin{figure}[ht]
    \centering
    \includegraphics[width=0.8\textwidth]{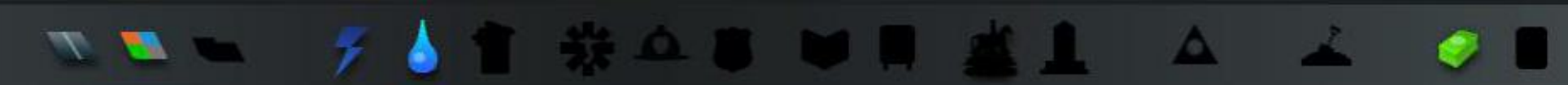}
    \caption{The toolbar in Cities: Skylines}
    \label{fig:skyline_toolbar}
\end{figure}

\begin{figure}[ht]
    \centering
    \includegraphics[width=0.8\textwidth]{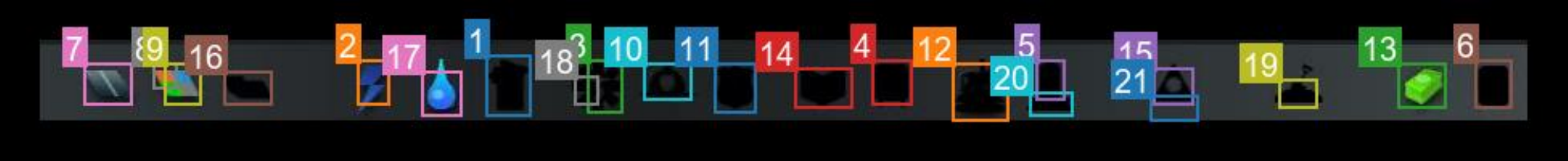}
    \caption{The grounding result of the toolbar in Cities: Skylines}
    \label{fig:skyline_toolbar_sam}
\end{figure}

\begin{figure}[ht]
    \centering
    \begin{minipage}{0.49\textwidth}
        \centering
        \includegraphics[width=\textwidth]{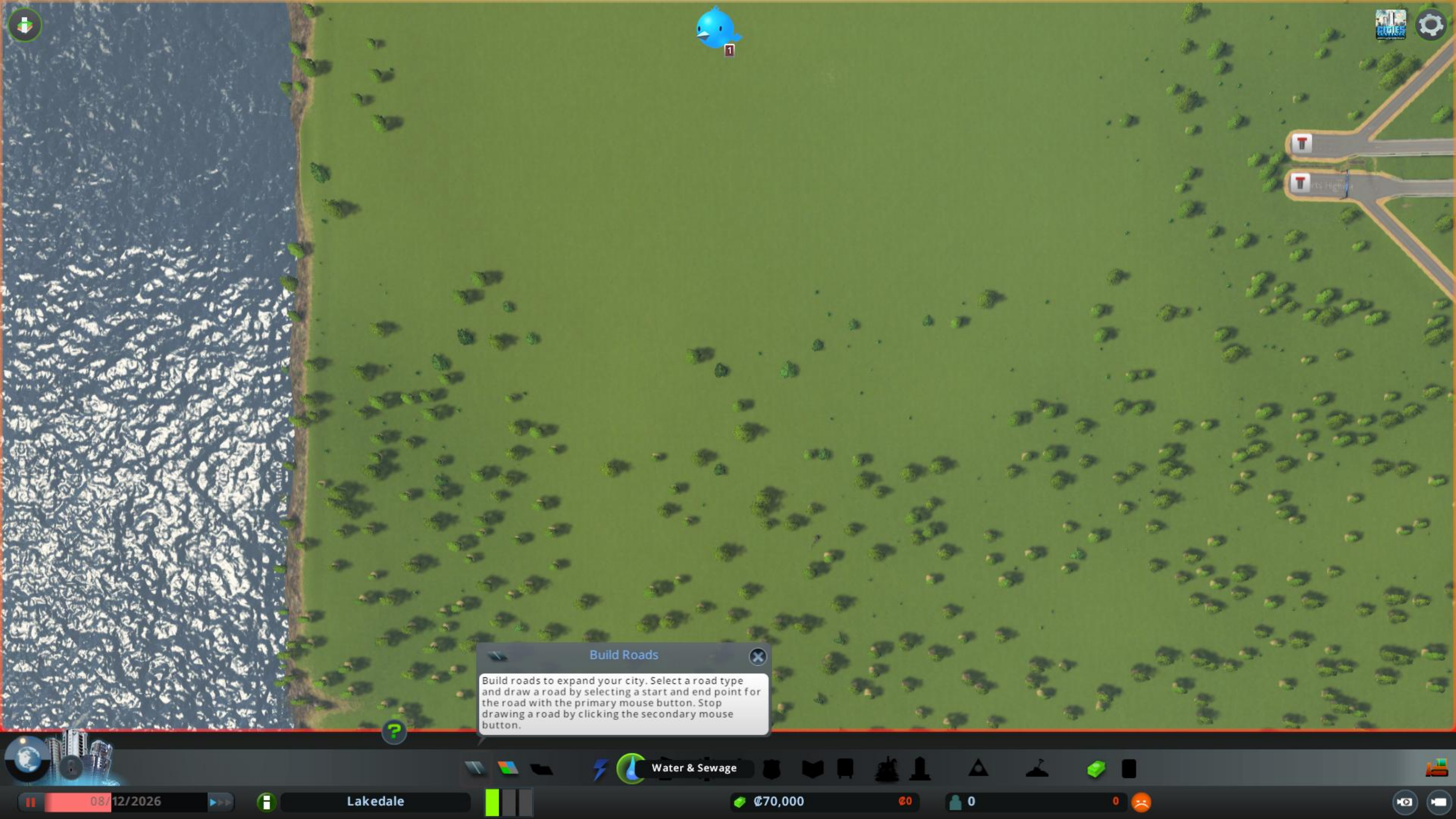}
        \caption{When hovering the mouse over a toolbar item, the pop-up description is "Water \& Sewage". The skill generated is then called "open\_water\_sewage\_menu".}
        \label{fig:skyline_level1}
    \end{minipage}
    \hfill
    \begin{minipage}{0.49\textwidth}
        \centering
        \includegraphics[width=\textwidth]{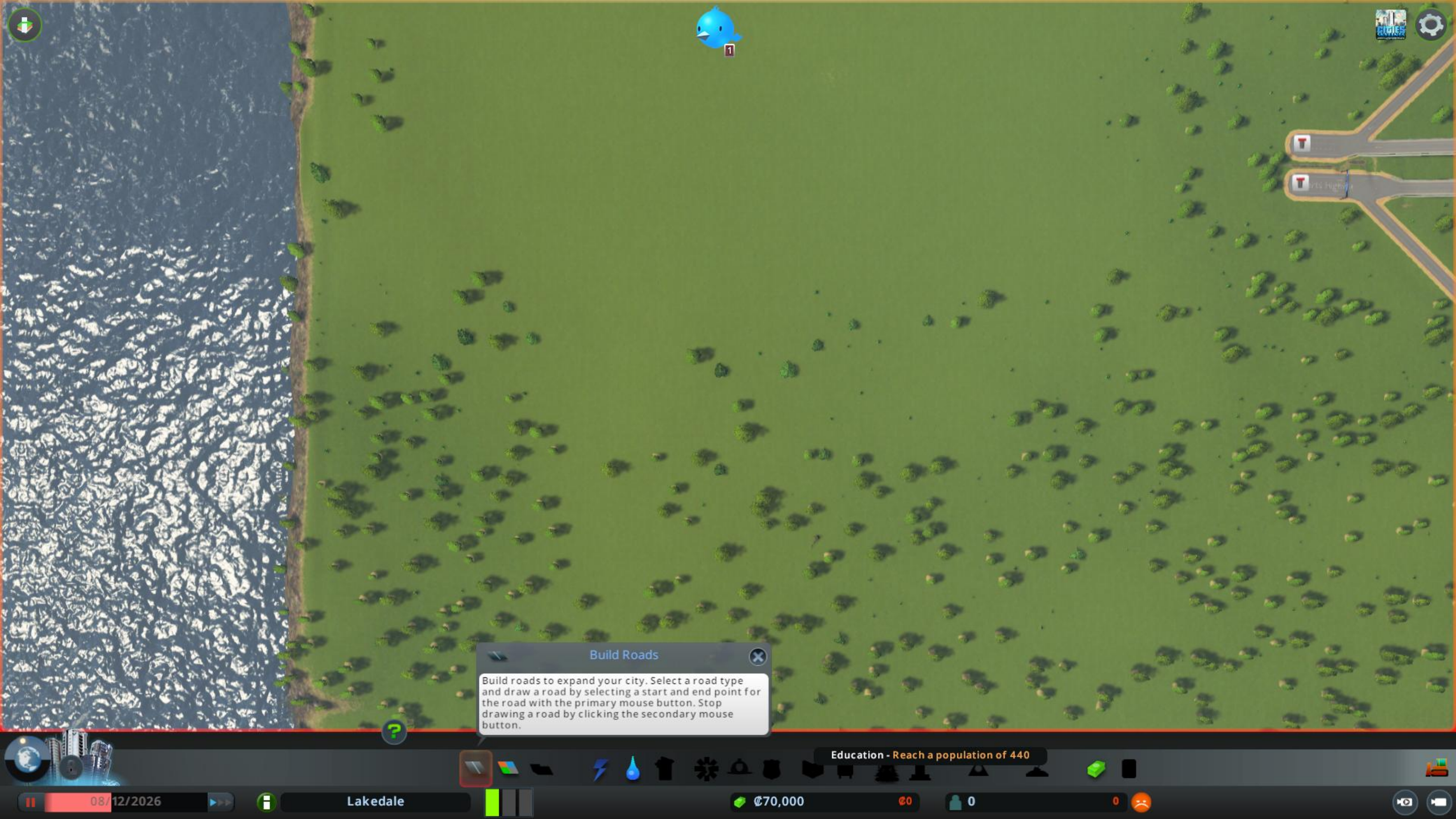}
        \caption{When hovering the mouse over a toolbar item, the pop-up description is "Education - Reach a population of 440". As this is not selectable for now, GPT-4o does not generate a new skill for it.}
        \label{fig:skyline_level1_reject}
    \end{minipage}
\end{figure}

\begin{figure}[h]
    \centering
    \begin{minipage}{0.49\textwidth}
        \centering
        \includegraphics[width=\textwidth]{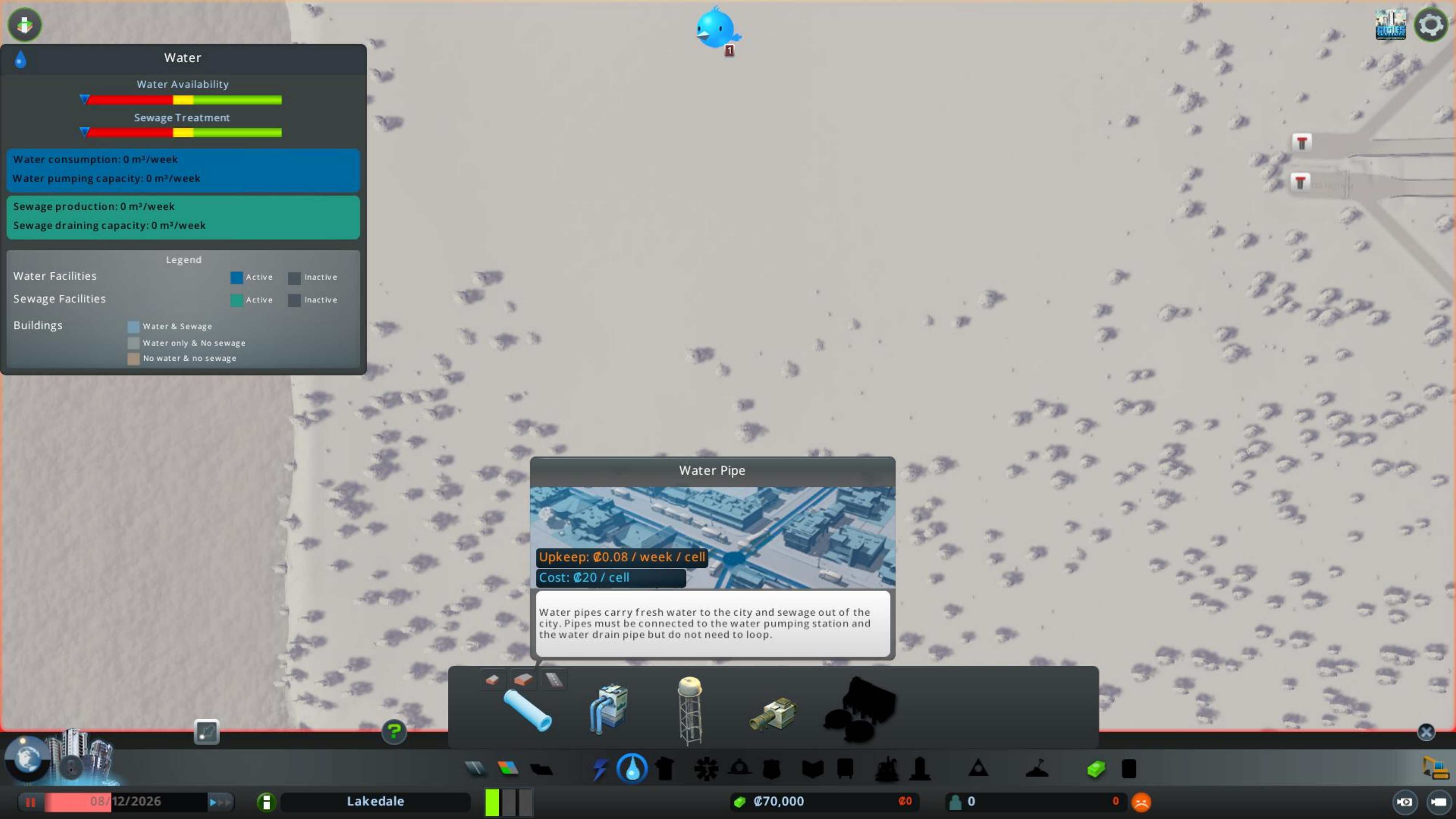}
        \caption{The Water \& Sewage menu is opened by executing the new skill "open\_water\_sewage\_menu". The Agent then hovers the mouse over a second-level toolbar item, the pop-up description is "Water Pipe", and the generated skill is called "try\_place\_water\_pipe".}
        \label{fig:skyline_level2}
    \end{minipage}
    \hfill
    \begin{minipage}{0.49\textwidth}
        \centering
        \includegraphics[width=\textwidth]{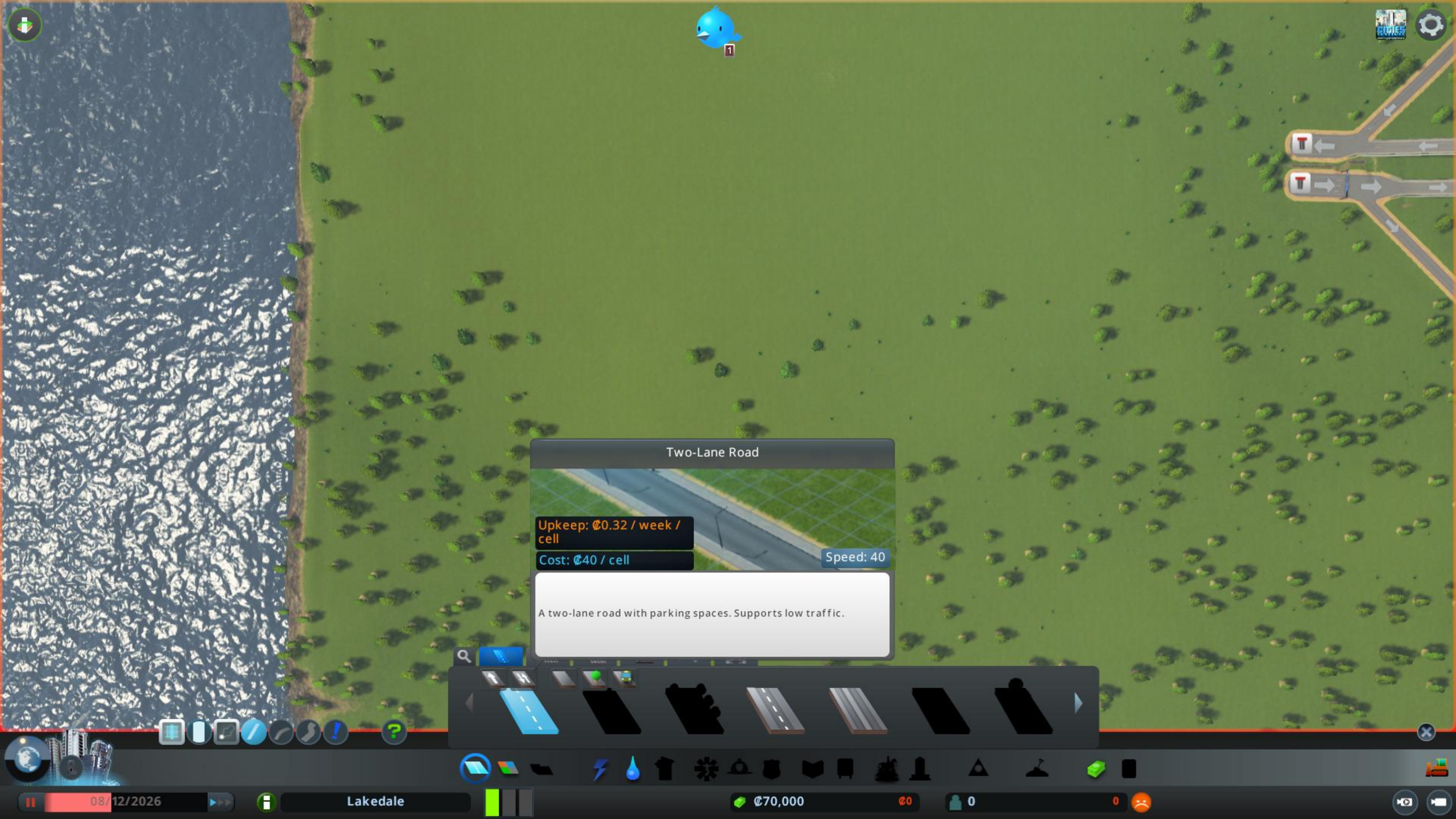}
        \caption{The Roads menu is opened by executing the new skill "open\_roads\_menu". The Agent then hovers the mouse over a second-level toolbar item, the pop-up description is "Two-Lane Road", and the generated skill is called "try\_place\_two\_lane\_road".}
        \label{fig:skyline_level2_1}
    \end{minipage}
\end{figure}

\textbf{Skill Curation.} Due to the lack of detailed tutorials in the game, we generate the skills through self-exploration in this game. The skill generation basically involves manipulating the toolbar to understand the items on it. The pseudo-code for skill generation is described in Algorithm ~\ref{algo:skyline_skill_generation}. This process leverages SAM for objective grounding and GPT-4o to gather information about the objects provided by the game, subsequently generating skills based on a predefined template. An example of the process is shown in Fig \ref{fig:skyline_toolbar}, \ref{fig:skyline_toolbar_sam}, \ref{fig:skyline_level1}, \ref{fig:skyline_level1_reject}, \ref{fig:skyline_level2} and \ref{fig:skyline_level2_1}.

\begin{algorithm}[h]
\caption{Skill Generation}
\label{algo:skyline_skill_generation}
\KwIn{Toolbar with objects, Skill template}
\KwOut{Procedure memory with generated skills}

Initialize procedure memory\;

\For{each object in the toolbar}{
    Hover the mouse on the object to get the description\;
    Generate skill using GPT-4o based on the object description and the skill template\;
    Store generated skill in procedure memory\;
    Execute the generated skill to enter the second-level toolbar\;

    \For{each object in the second-level toolbar}{
        Hover the mouse on the object to get the description\;
        Generate skill using GPT-4o based on the object description and skill template\;
        Store generated skill in procedure memory\;
    }
}

\Return{procedure memory}

\end{algorithm}

\textbf{Action Planning.} In this game, we only let GPT-4o output one skill for each action since we observe that GPT-4o tends to output \emph{try\_place} and \emph{confirm placement} together if we allow it to output and execute multiple skills in one action, which is against the intention of our design for the \emph{try\_place} action.

\textbf{Procedure Memory.} Skills generated through self-exploration are listed below:
\begin{itemize}
    \item \emph{open\_roads\_menu()}: The function to open the roads options in the lower menu bar for further determination of which types of roads to build.
    
    \item \emph{open\_electricity\_menu()}: The function to open the electricity options in the lower menu bar for further determination of which types of power facility to build.
    
    \item \emph{open\_water\_sewage\_menu()}: The function to open the water and sewage options in the lower menu bar for further determination of which types of water and sewage to build.
    
    \item \emph{open\_zoning\_menu()}: The function to open the zoning options in the lower menu bar for further determination of which types of zonings to build.
    
    \item \emph{try\_place\_two\_lane\_road$(x_1, y_1, x_2, y_2)$}: Previews the placement of a road between two specified points, $(x_1, y_1)$ and $(x_2, y_2)$, with $x_1, y_1$ being the coordinate of start point of the road, and $(x_2, y_2)$ being the coordinate of end point of the road. This function does not actually construct the road, but rather displays a visual representation of where the road would be placed if confirmed.
    
    \item \emph{try\_place\_wind\_turbine$(x, y)$}: Previews the placement of a wind turbine on point, $(x, y)$. This function does not actually construct the wind turbine, but rather displays a visual representation of where the wind turbine would be placed if confirmed.
    
    \item \emph{try\_place\_water\_pumping\_station$(x, y)$}: Previews the placement of a water pumping station on point, $(x, y)$. This function does not actually construct the water pumping station, but rather displays a visual representation of where the water pumping station would be placed if confirmed.
    
    \item \emph{try\_place\_water\_pipe$(x_1, y_1, x_2, y_2)$}: Previews the placement of a water pipe between two specified points, $(x_1, y_1)$ and $(x_2, y_2)$. This function does not actually construct the water pipe, but rather displays a visual representation of where the water pipe would be placed if confirmed.
    
    \item \emph{try\_place\_water\_drain\_pipe$(x, y)$}: Previews the placement of a water drain pipe on point, $(x, y)$. This function does not actually construct the water drain pipe, but rather displays a visual representation of where the water drain pipe would be placed if confirmed.\
    
    \item \emph{try\_place\_commercial\_zone$(x_1, y_1, x_2, y_2)$}: Previews the placement of a commercial zone within a rectangular region with diagonal corners at $(x_1, y_1)$ and $(x_2, y_2)$. This function does not actually construct the commercial zone, but rather displays a visual representation of where the commercial zone would be placed if confirmed.
    
    \item \emph{try\_place\_industrial\_zone$(x_1, y_1, x_2, y_2)$}: Previews the placement of a industrial zone within a rectangular region with diagonal corners at $(x_1, y_1)$ and $(x_2, y_2)$. This function does not actually construct the industrial zone, but rather displays a visual representation of where the industrial zone would be placed if confirmed.
    
    \item \emph{try\_de\_zone$(x_1, y_1, x_2, y_2)$}: The function to remove the zone in the game. The zone must cover the road.
    \item \emph{confirm\_placement()}: The function to confirm the placement and build the object after the \emph{try\_place\_[object]} function. 
    
    \item \emph{cancel\_placement()}: The function to cancel the placement of the object after the \emph{try\_place\_[object]} function.
\end{itemize}

\textbf{Episodic Memory.} Besides the common information to store in the episodic memory. We initialize the memory with the coordinates of the entry and exit of the highway. Then \projname is able to extend the roads according to these two points at the beginning. When a road or a facility such as wind turbine, water pumping station, water drain pipe and water pipe is placed on the map, the corresponding coordinates will also be stored in the memory for future development of the city.

\subsection{Case Studies}

\subsubsection{Failure for Road Building.} 
\begin{figure}[h]
    \centering
    \begin{subfigure}{0.32\textwidth}
      \centering
      \includegraphics[width=\linewidth]{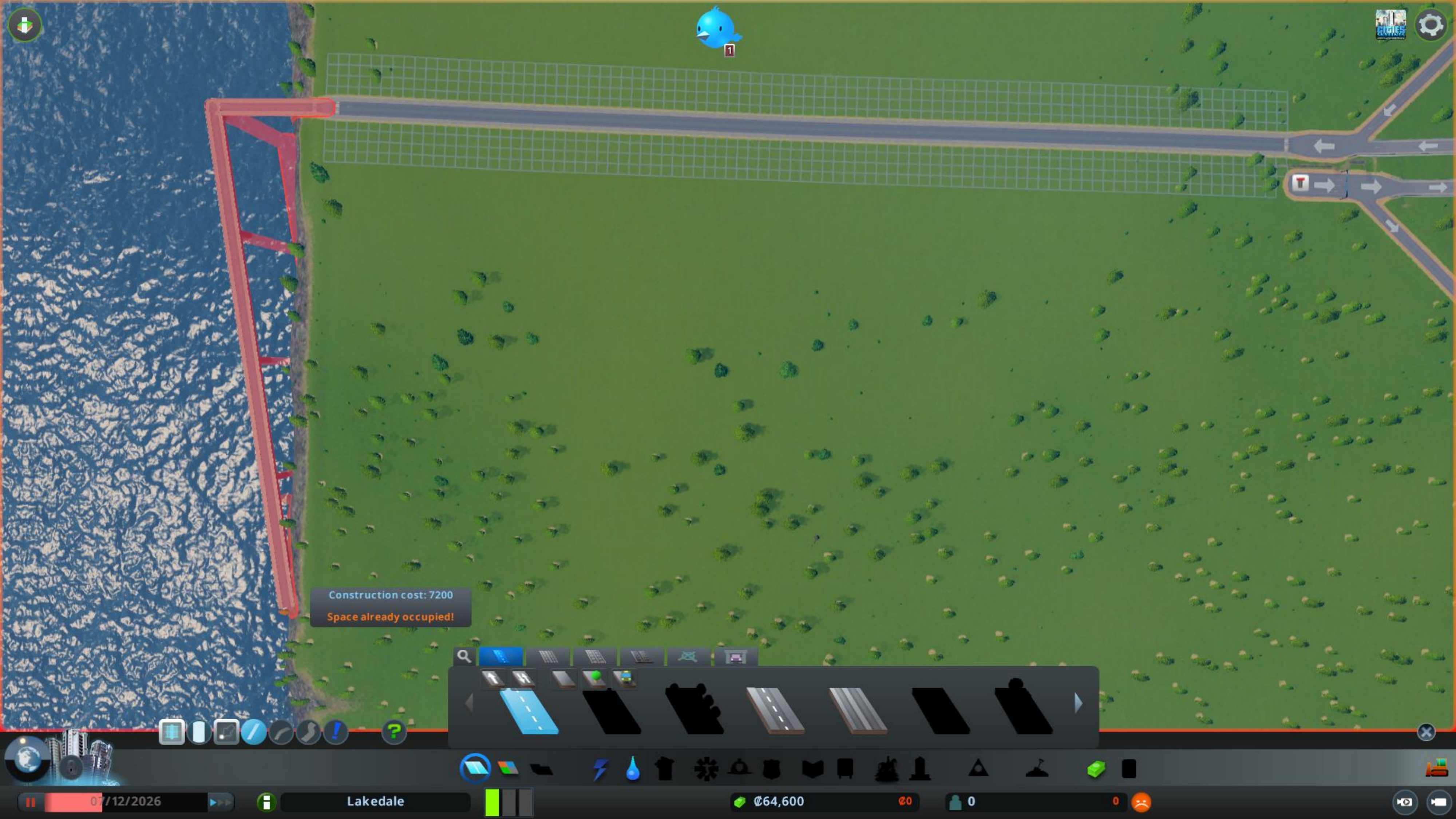}
      \caption{}
      \label{appx_fig:skylines_road_failure_01}
    \end{subfigure}
    \centering
    \begin{subfigure}{0.32\textwidth}
      \centering
      \includegraphics[width=\linewidth]{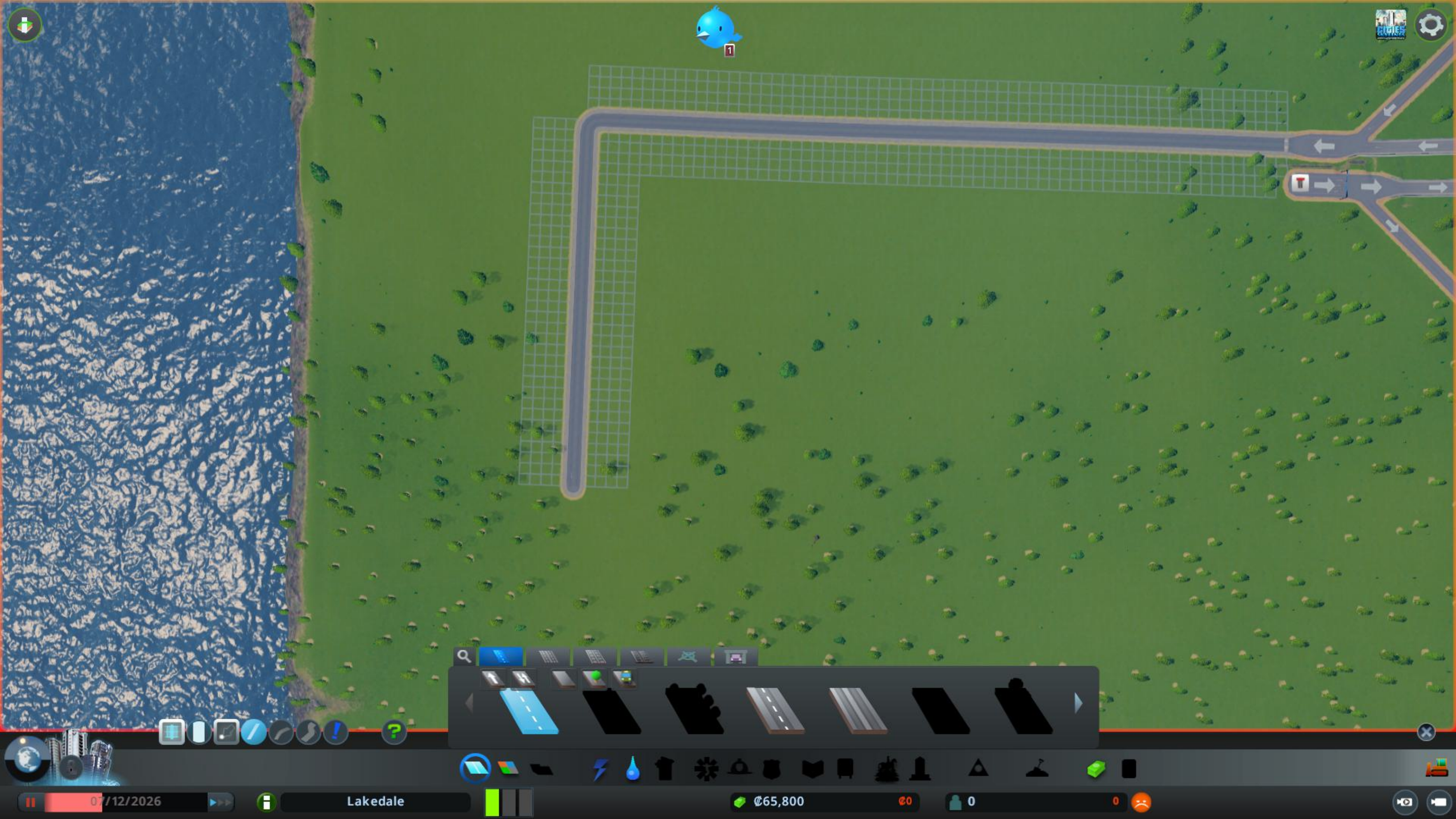}
      \caption{}
      \label{appx_fig:skylines_road_failure_021}
    \end{subfigure}
    \begin{subfigure}{0.32\textwidth}
      \centering
      \includegraphics[width=\linewidth]{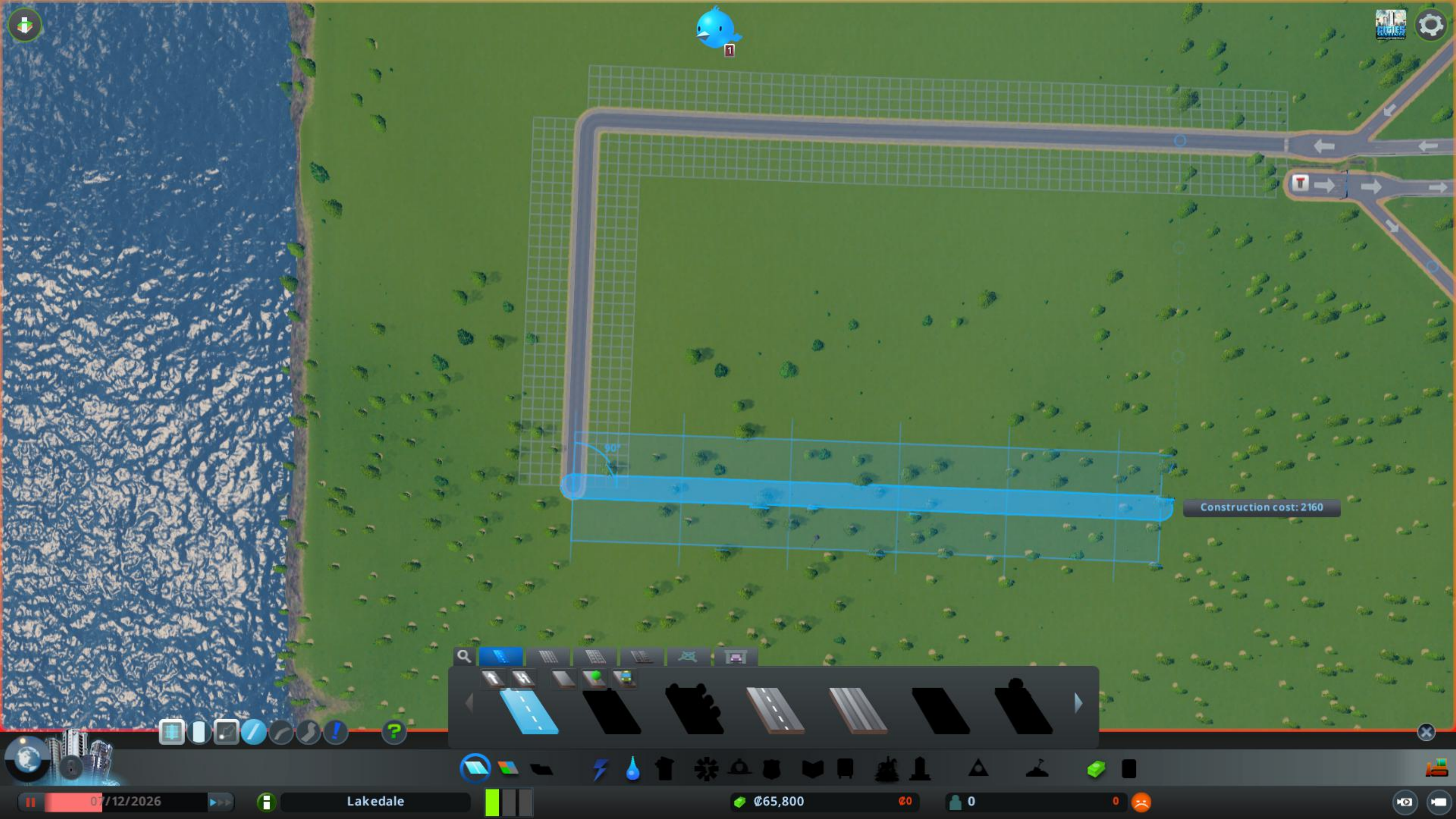}
      \caption{}
      \label{appx_fig:skylines_road_failure_022}
    \end{subfigure}
    \begin{subfigure}{0.32\textwidth}
      \centering
      \includegraphics[width=\linewidth]{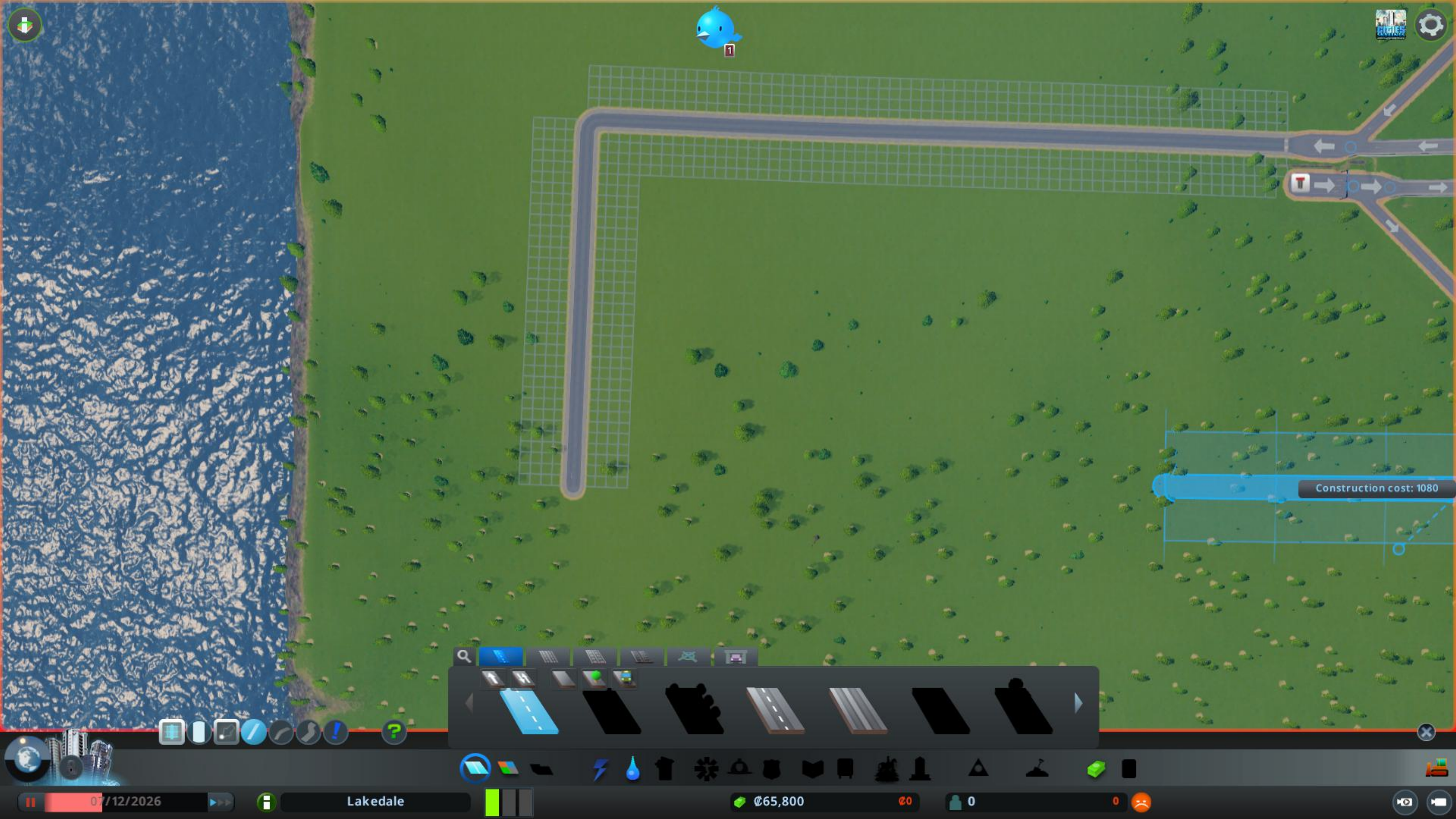}
      \caption{}
      \label{appx_fig:skylines_road_failure_023}
    \end{subfigure}
    \begin{subfigure}{0.32\textwidth}
      \centering
      \includegraphics[width=\linewidth]{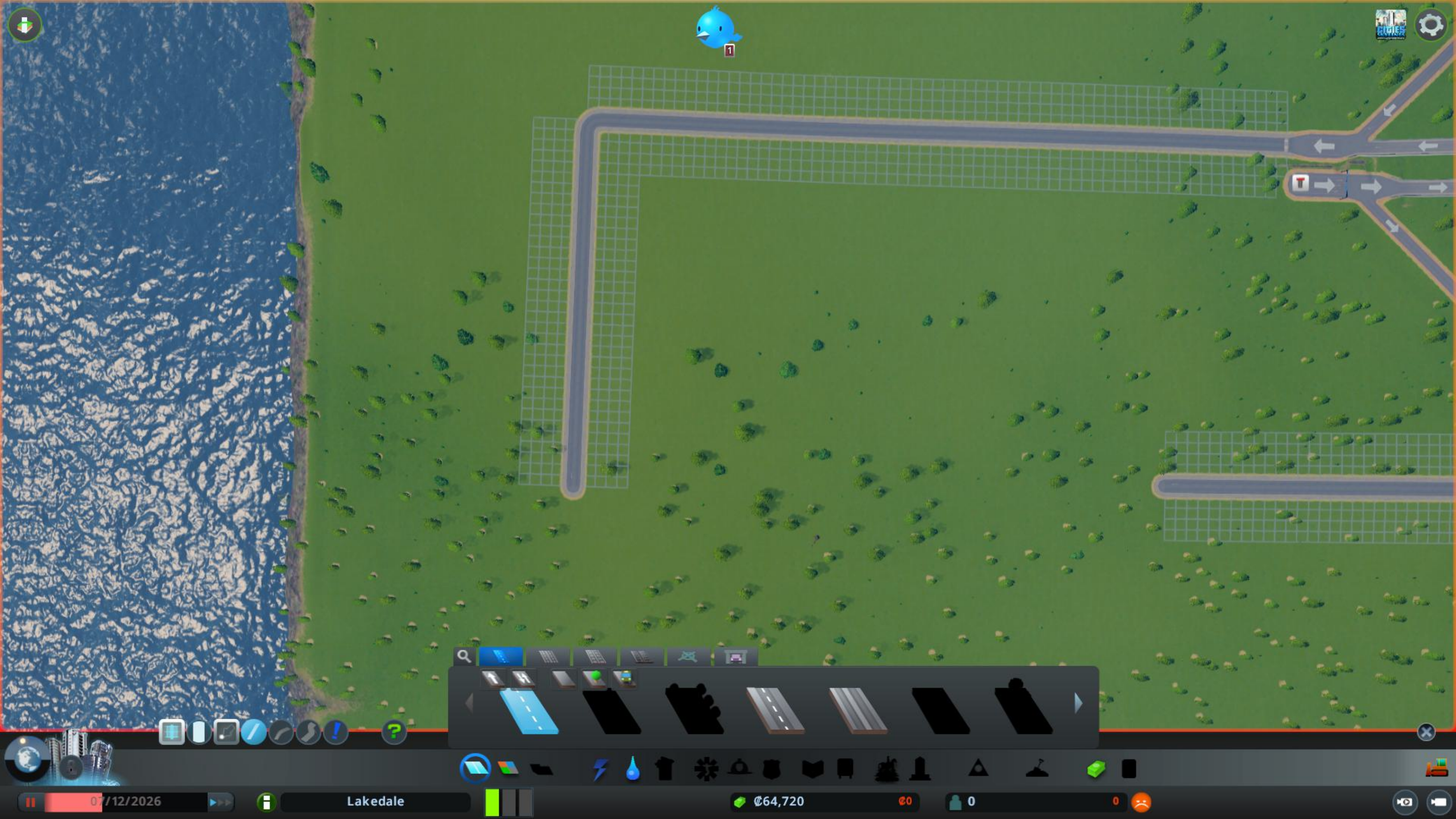}
      \caption{}
      \label{appx_fig:skylines_road_failure_024}
    \end{subfigure}
  \caption{Failure cases of building roads in a closed loop. Figure ~\ref{appx_fig:skylines_road_failure_01} shows that the road is built over the water and is difficult to continue. Figure ~\ref{appx_fig:skylines_road_failure_021}, ~\ref{appx_fig:skylines_road_failure_022}, ~\ref{appx_fig:skylines_road_failure_023} and  ~\ref{appx_fig:skylines_road_failure_024} tells a story that GPT-4o sometimes forgets to confirm the placement (from ~\ref{appx_fig:skylines_road_failure_022} to ~\ref{appx_fig:skylines_road_failure_023}) and directly moves to the next step of building (from ~\ref{appx_fig:skylines_road_failure_023} to ~\ref{appx_fig:skylines_road_failure_024}), resulting in the disconnection of the roads.} 
  \label{appx_fig:skylines_road_failure}
\end{figure}

As shown in Figure~\ref{appx_fig:skylines_road_failure}, sometimes GPT-4o will build a long road, which ends on the top of water. The recorded endpoint of the road is actually the projection of the road on the sea level, resulting in the offset from the projection point and the real endpoint of the road. It leads to the failure of extending the road to the other places.

Figure ~\ref{appx_fig:skylines_road_failure_021}, ~\ref{appx_fig:skylines_road_failure_022}, ~\ref{appx_fig:skylines_road_failure_023} and  ~\ref{appx_fig:skylines_road_failure_024} tells a story that
GPT-4o sometimes forgets to confirm the placement (from ~\ref{appx_fig:skylines_road_failure_022} to ~\ref{appx_fig:skylines_road_failure_023}) and directly moves to the next step of building the next road (from ~\ref{appx_fig:skylines_road_failure_023} to ~\ref{appx_fig:skylines_road_failure_024}), resulting in the disconnection of the roads.

\subsubsection{Failure for Sufficient Water Supply.} 
\begin{figure}[t]
    \centering
    \begin{subfigure}{0.32\textwidth}
      \centering
      \includegraphics[width=\linewidth]{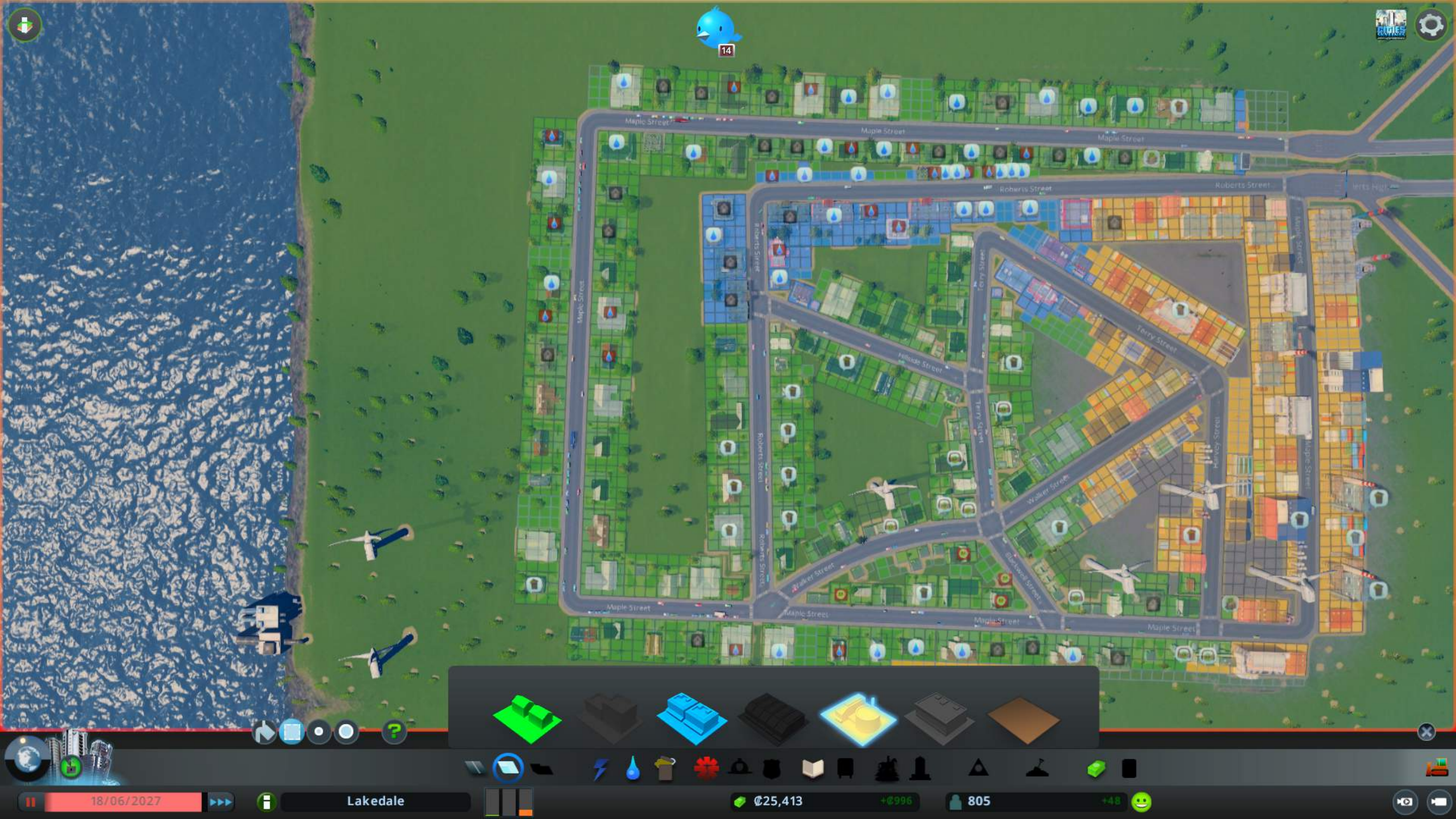}
      \label{subfig:skylines_showcase_01_before_zone}
    \end{subfigure}
    \begin{subfigure}{0.32\textwidth}
      \centering
      \includegraphics[width=\linewidth]{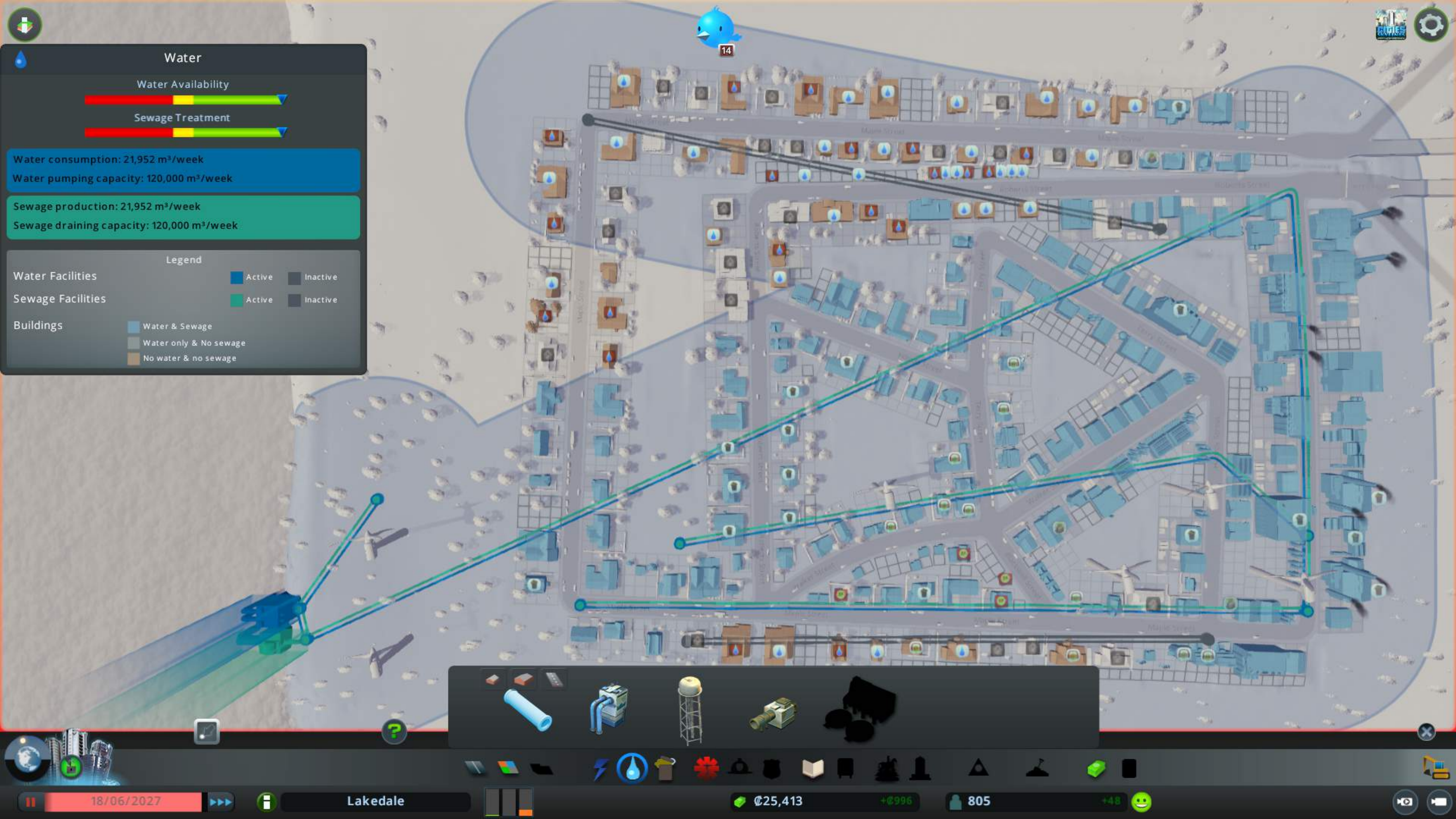} 
      \label{subfig:skylines_showcase_01_before_water}
    \end{subfigure}
    \begin{subfigure}{0.32\textwidth}
      \centering
      \includegraphics[width=\linewidth]{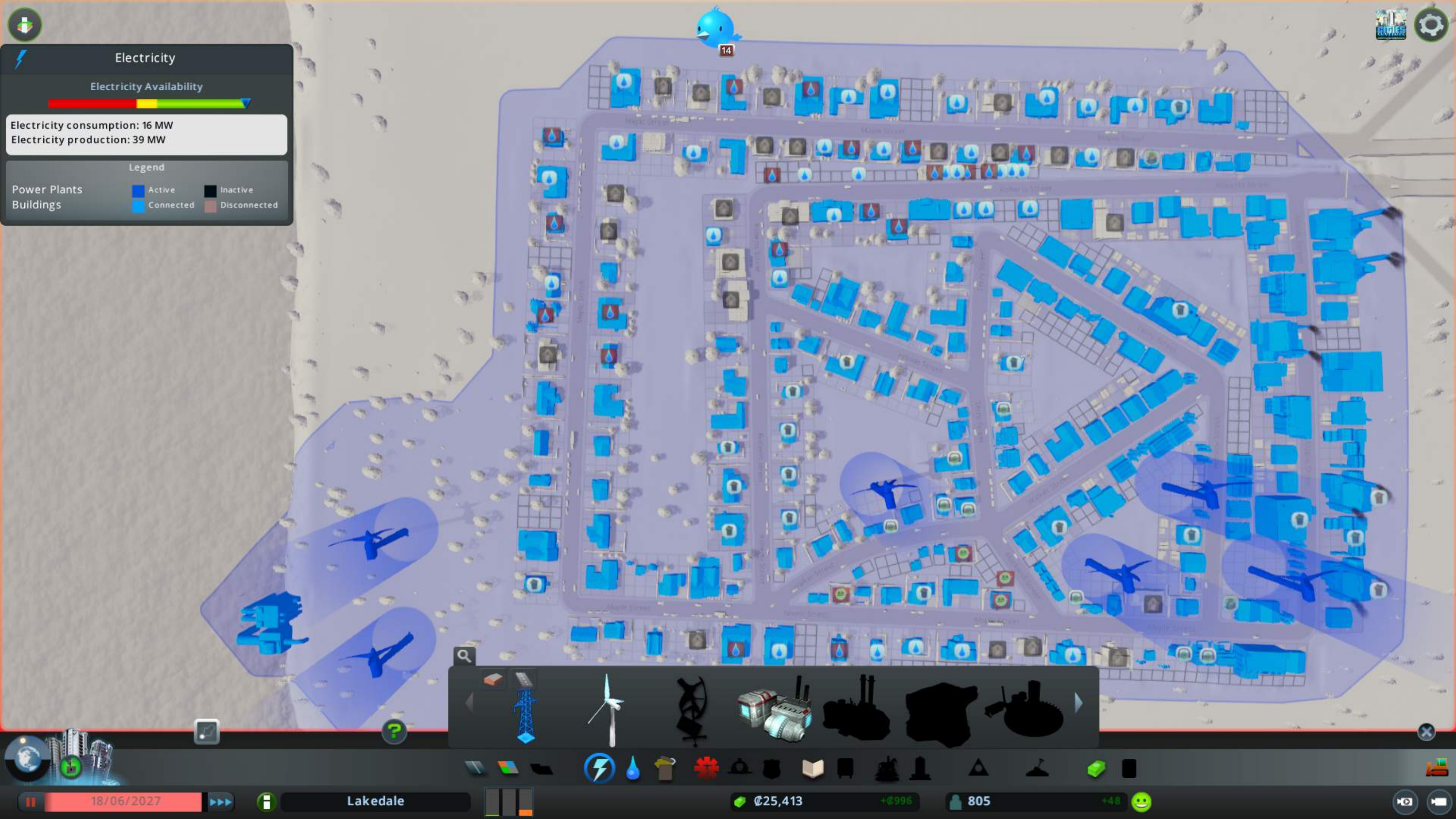}
      \label{subfig:skylines_showcase_01_before_power}
    \end{subfigure}

    \begin{subfigure}{\textwidth}
      \centering
      \vspace{-10pt}
      \caption{\projname's craftwork I. The upper left corner of the city is experiencing a severe local water shortage since the water pipes there are not connected. \textbf{Population: 800+}.}
      \label{appx_fig:skylines_showcase_01_before}
    \end{subfigure}

    \begin{subfigure}{0.32\textwidth}
      \centering
      \includegraphics[width=\linewidth]{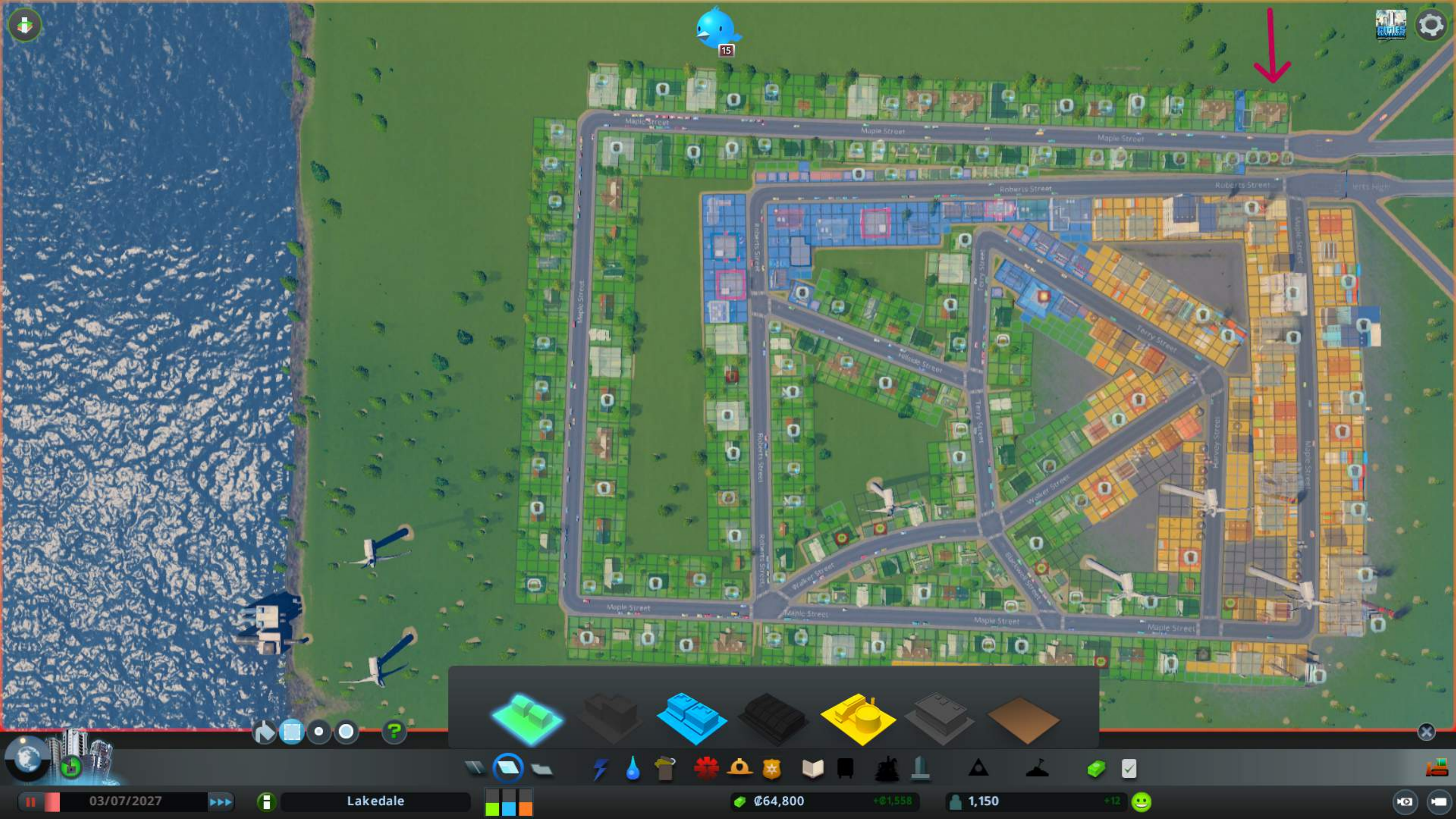}
      \label{subfig:skylines_showcase_01_after_zone}
    \end{subfigure}
    \begin{subfigure}{0.32\textwidth}
      \centering
      \includegraphics[width=\linewidth]{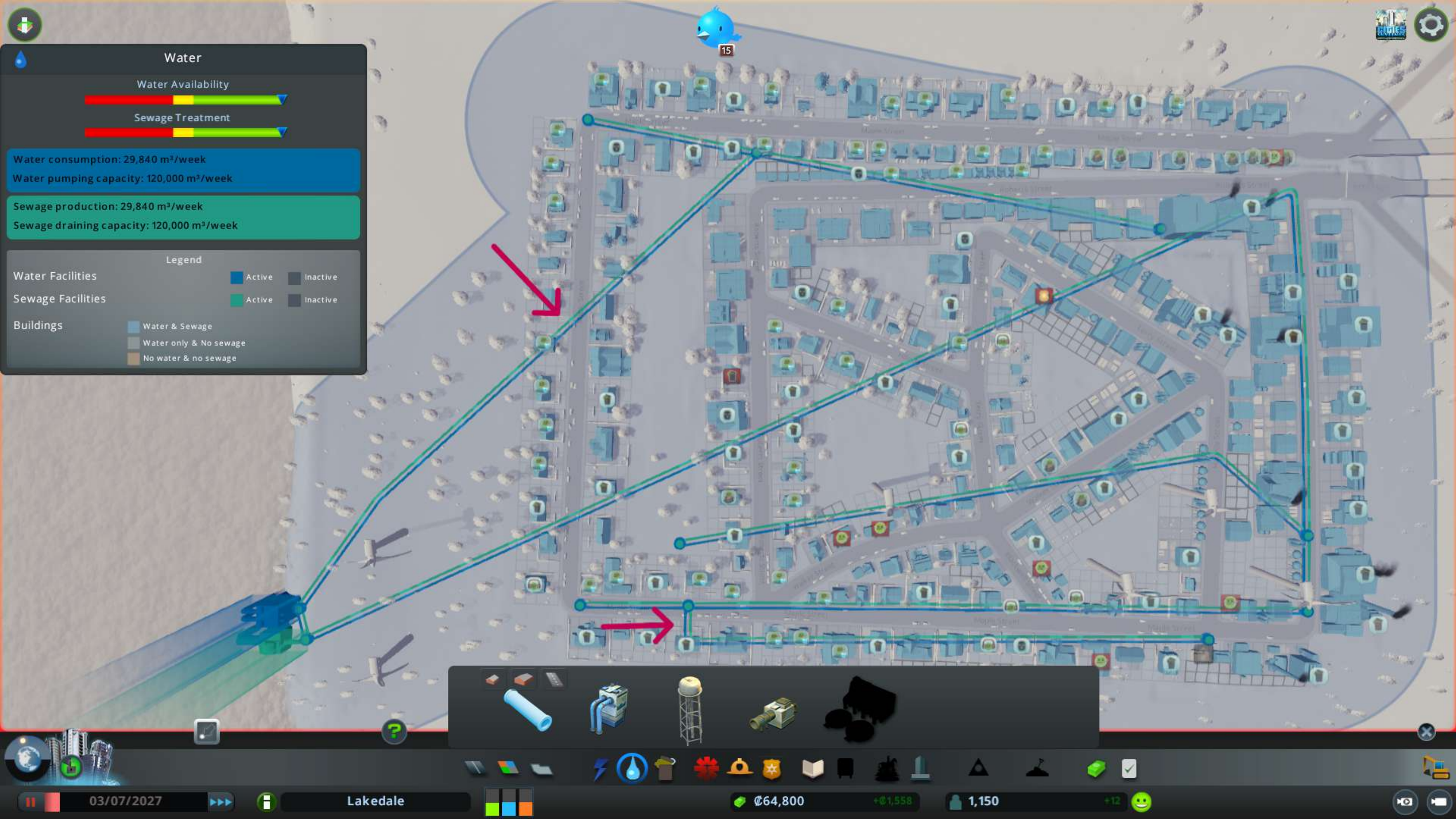}
      \label{subfig:skylines_showcase_01_after_water}
    \end{subfigure}
    \begin{subfigure}{0.32\textwidth}
      \centering
      \includegraphics[width=\linewidth]{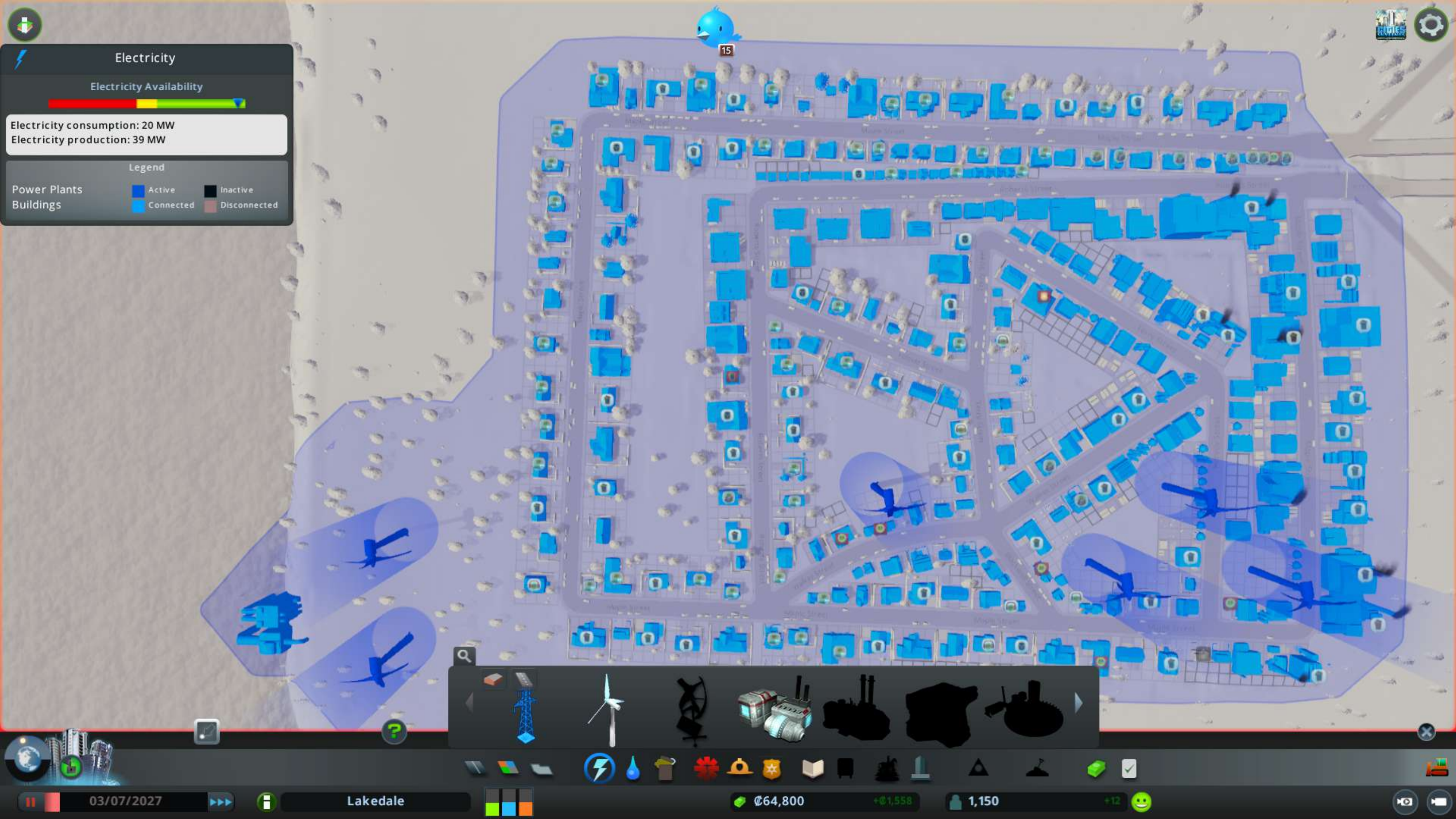}
      \label{subfig:skylines_showcase_01_after_power}
    \end{subfigure}

    \begin{subfigure}{\textwidth}
      \centering
      \vspace{-15pt}
      \caption{\projname's craftwork I with human assistant within three unit operations to develop the idle area in the upper right corner of the city into a residential zone and put two water pipes to ensure all the water pipes connected and cover the whole city.  \textbf{Population: 1150+}. }
      \label{appx_fig:skylines_showcase_01_after}
    \end{subfigure}

    \centering
    \begin{subfigure}{0.32\textwidth}
      \centering
      \includegraphics[width=\linewidth]{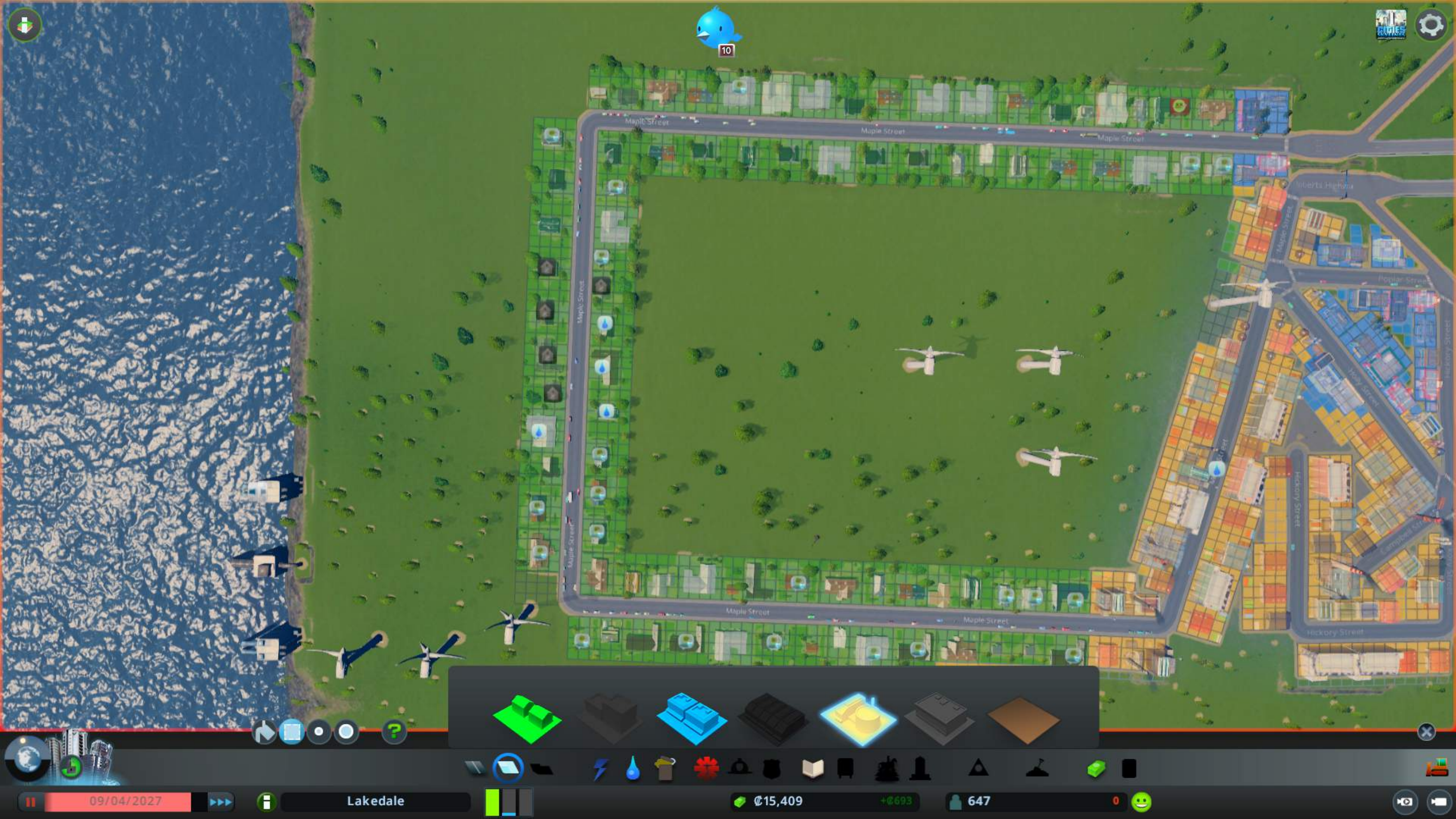}
      \label{subfig:skylines_showcase_02_before_zone}
    \end{subfigure}
    \begin{subfigure}{0.32\textwidth}
      \centering
      \includegraphics[width=\linewidth]{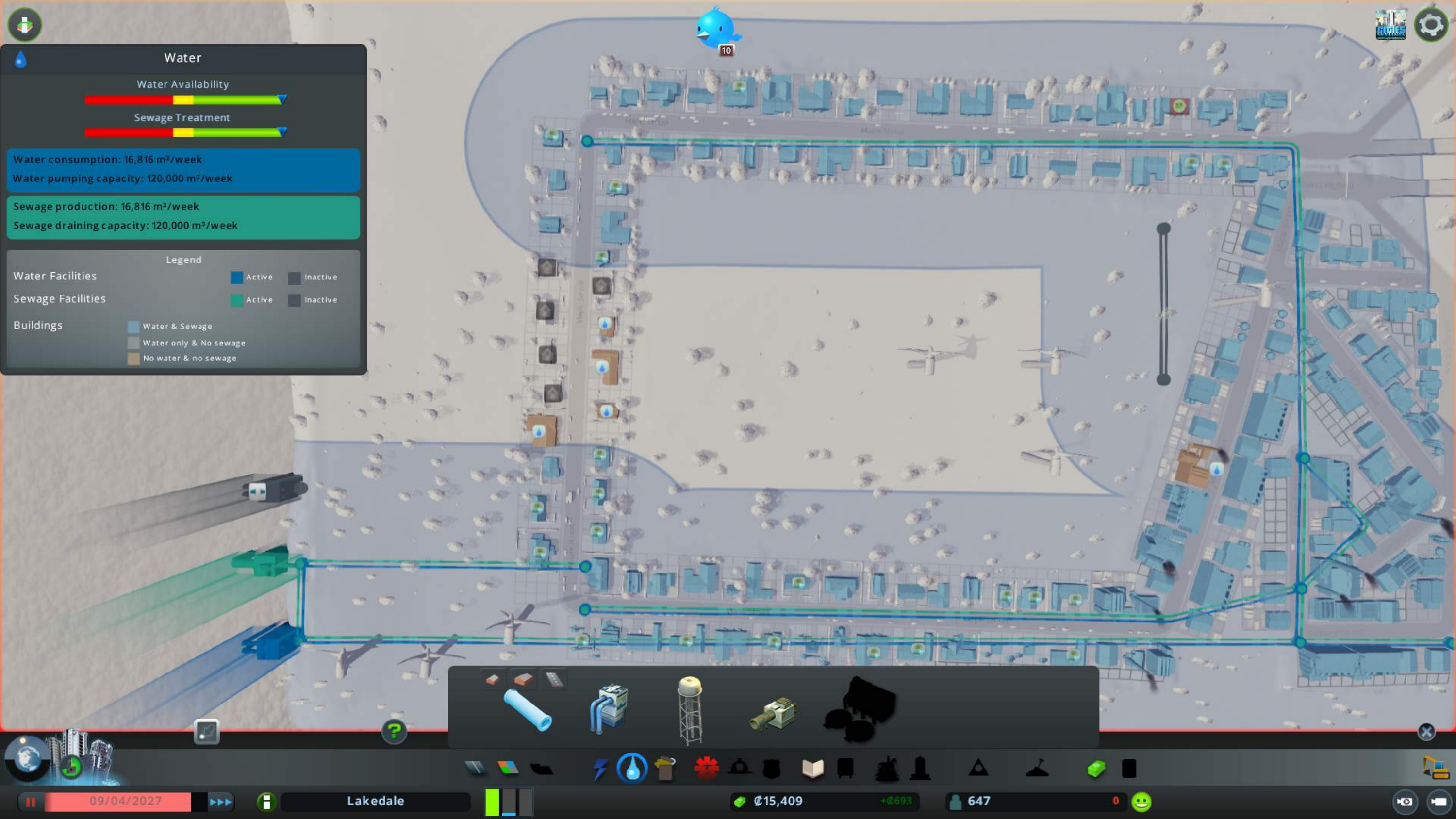} 
      \label{subfig:skylines_showcase_02_before_water}
    \end{subfigure}
    \begin{subfigure}{0.32\textwidth}
      \centering
      \includegraphics[width=\linewidth]{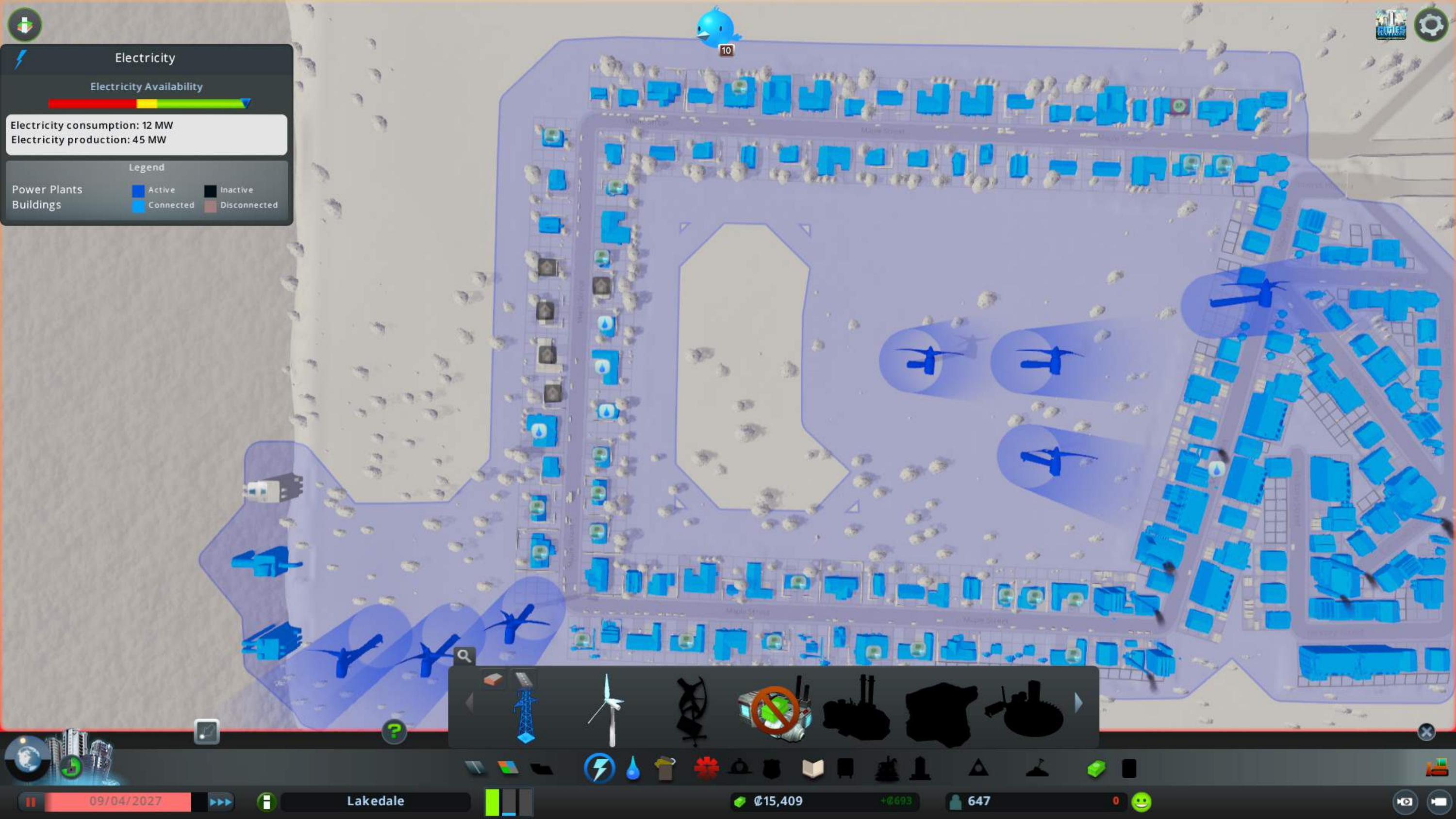}
      \label{subfig:skylines_showcase_02_before_power}
    \end{subfigure}
    
    \centering
    \begin{subfigure}{\textwidth}
      \centering
      \vspace{-15pt}
      \caption{\projname's craftwork II. The left side of the city a localized area on the right suffers from water shortage because of the water pipes connected issues. \textbf{Population: 640+.}}
      \label{appx_fig:skylines_showcase_02_before}
    \end{subfigure}
    \centering
    
    \begin{subfigure}{0.32\textwidth}
      \centering
      \includegraphics[width=\linewidth]{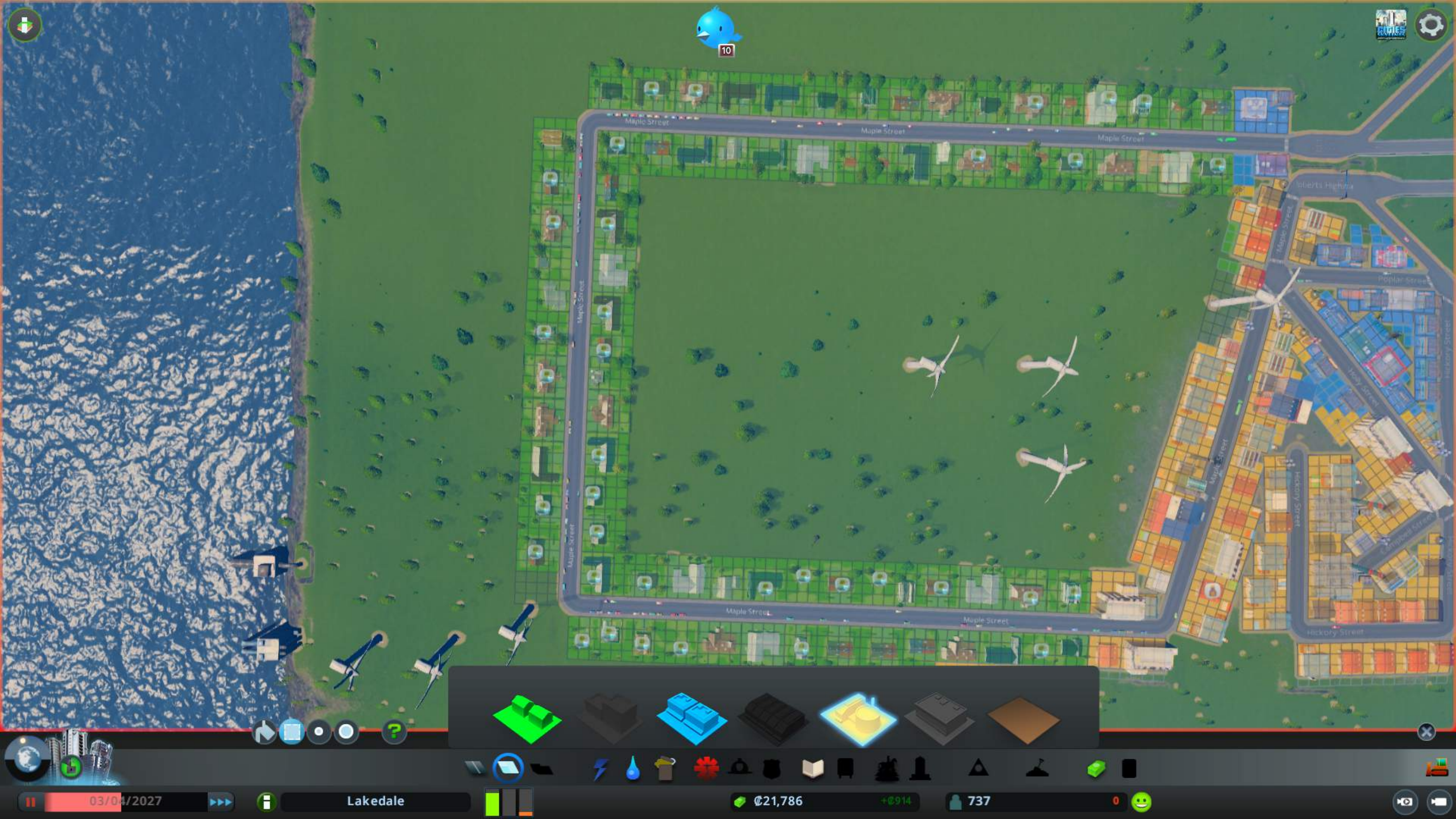}
      \label{subfig:skylines_showcase_02_after_zone}
    \end{subfigure}
    \begin{subfigure}{0.32\textwidth}
      \centering
      \includegraphics[width=\linewidth]{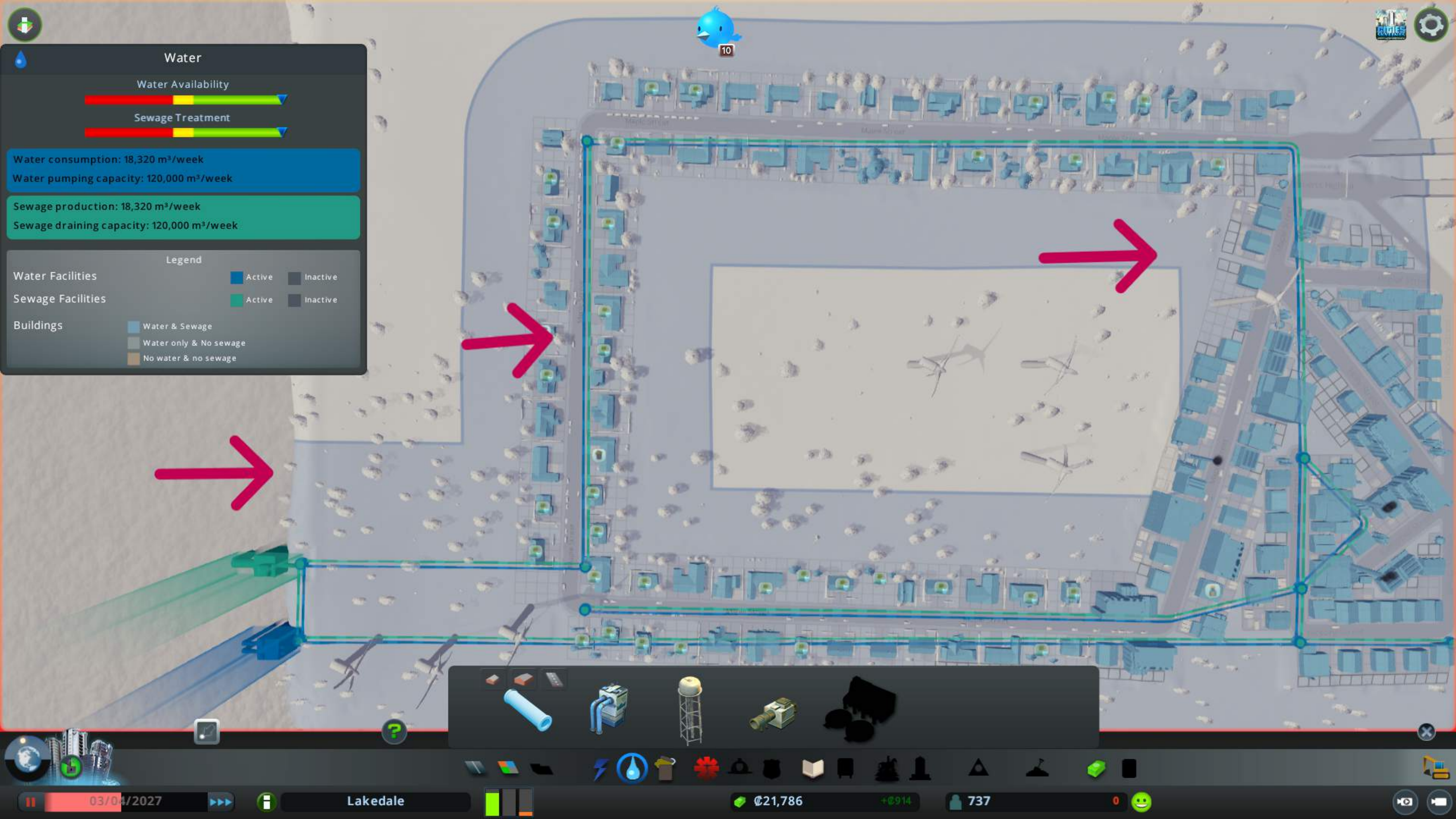}
      \label{subfig:skylines_showcase_02_after_water}
    \end{subfigure}
    \begin{subfigure}{0.32\textwidth}
      \centering
      \includegraphics[width=\linewidth]{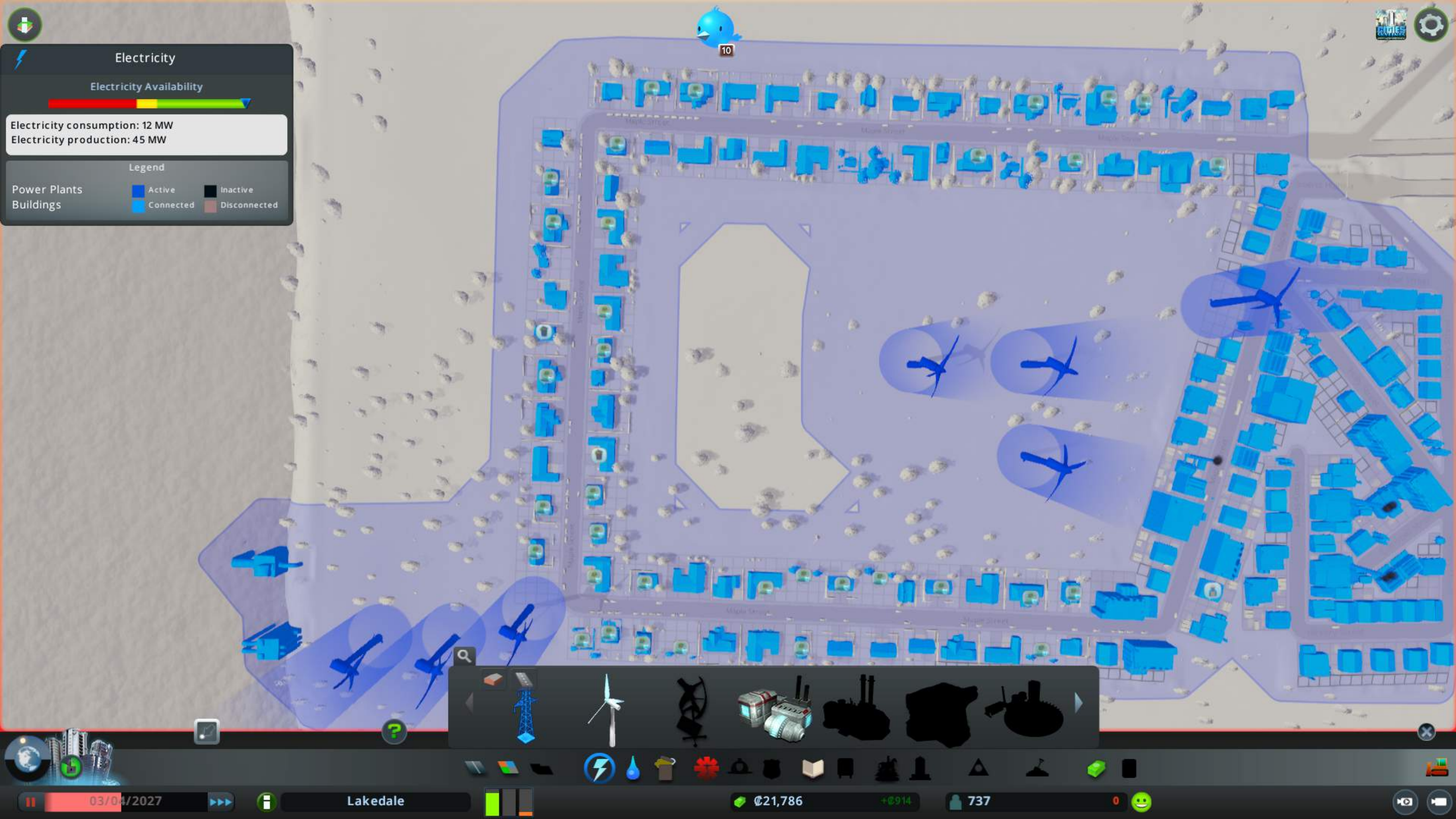}
      \label{subfig:skylines_showcase_02_after_power}
    \end{subfigure}

    \begin{subfigure}{\textwidth}
      \centering
      \vspace{-10pt}
      \caption{\projname's craftwork II with human assistant within three unit operations by selling the redundant water pumping station and the independent water pipe on the right to get some budget and using the budget to get the water pipes connected. \textbf{Population: 730+}.}
      \label{appx_fig:skylines_showcase_02_after}
    \end{subfigure}

    \centering
    \begin{subfigure}{0.32\textwidth}
      \centering
      \includegraphics[width=\linewidth]{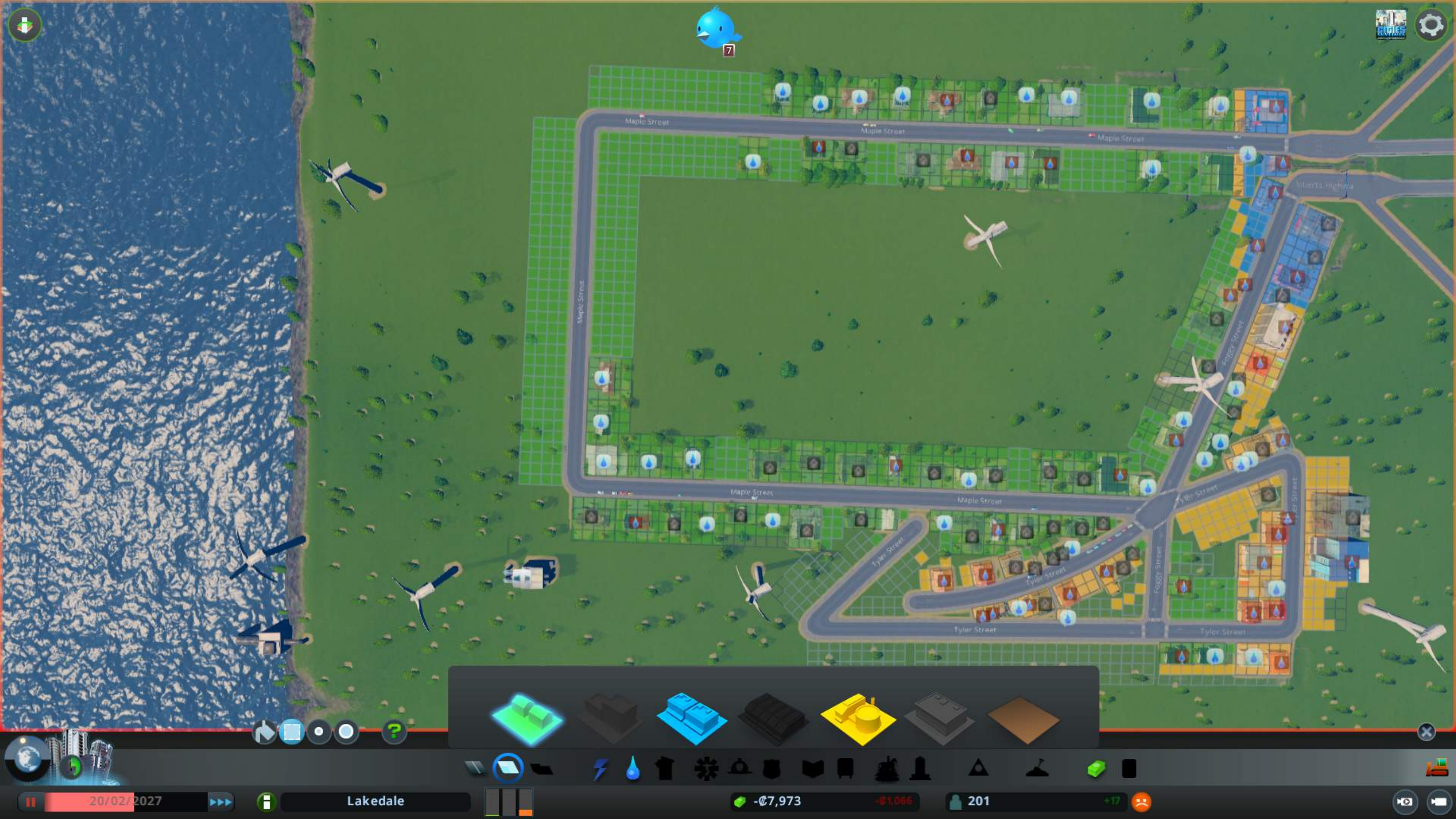}
      \label{subfig:skylines_showcase_03_before_zone}
    \end{subfigure}
    \begin{subfigure}{0.32\textwidth}
      \centering
      \includegraphics[width=\linewidth]{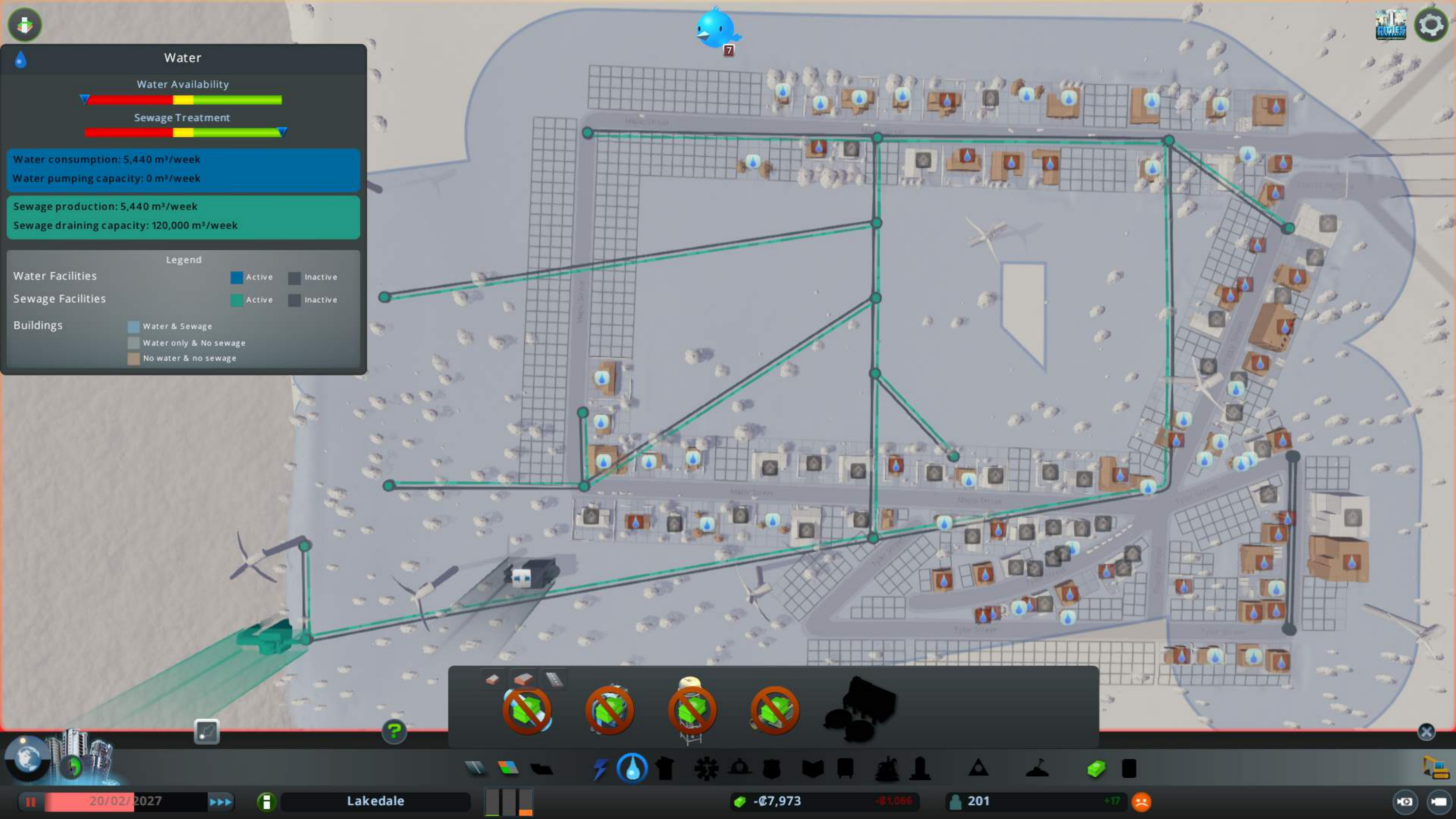} 
      \label{subfig:skylines_showcase_03_before_water}
    \end{subfigure}
    \begin{subfigure}{0.32\textwidth}
      \centering
      \includegraphics[width=\linewidth]{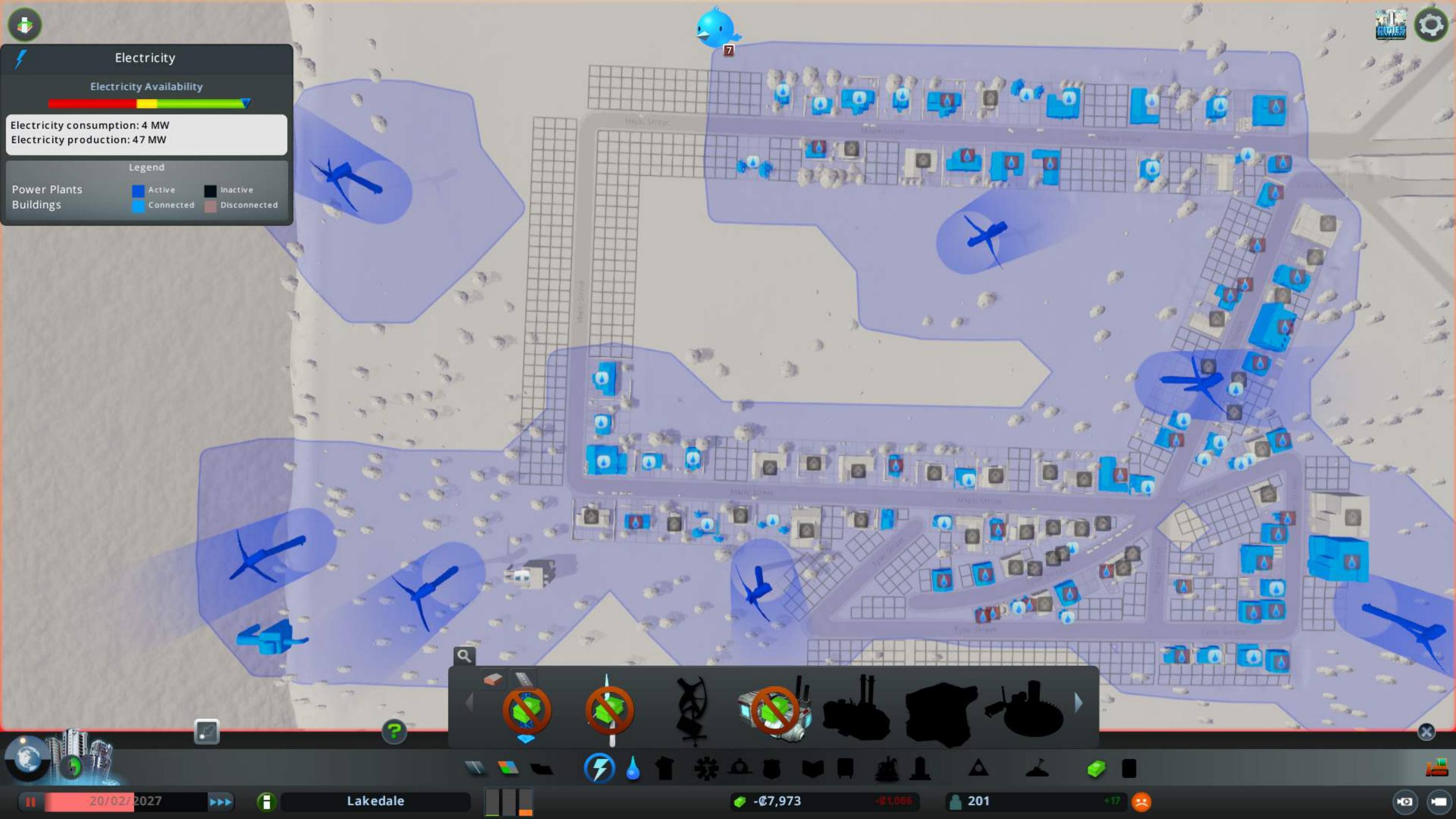}
      \label{subfig:skylines_showcase_03_before_power}
    \end{subfigure}

    \centering
    \begin{subfigure}{\textwidth}
      \centering
      \vspace{-10pt}
      \caption{\projname's craftwork III. The entire city is experiencing a severe water shortage due to the lack of the water pumping station. \textbf{Population: 200+}.}
      \label{appx_fig:skylines_showcase_03_before}
    \end{subfigure}
    \centering
    
    \begin{subfigure}{0.32\textwidth}
      \centering
      \includegraphics[width=\linewidth]{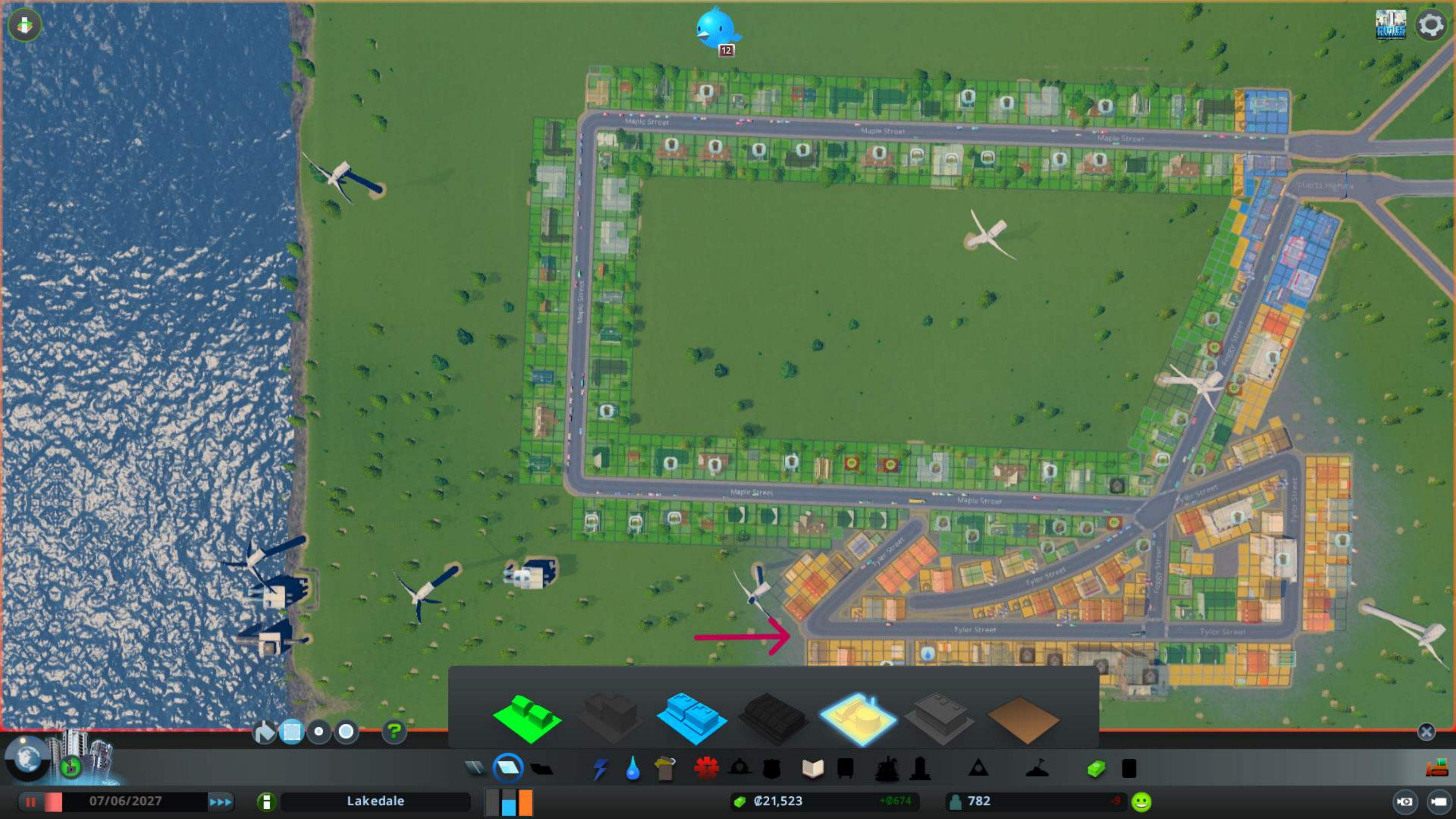}
      \label{subfig:skylines_showcase_03_after_zone}
    \end{subfigure}
    \begin{subfigure}{0.32\textwidth}
      \centering
      \includegraphics[width=\linewidth]{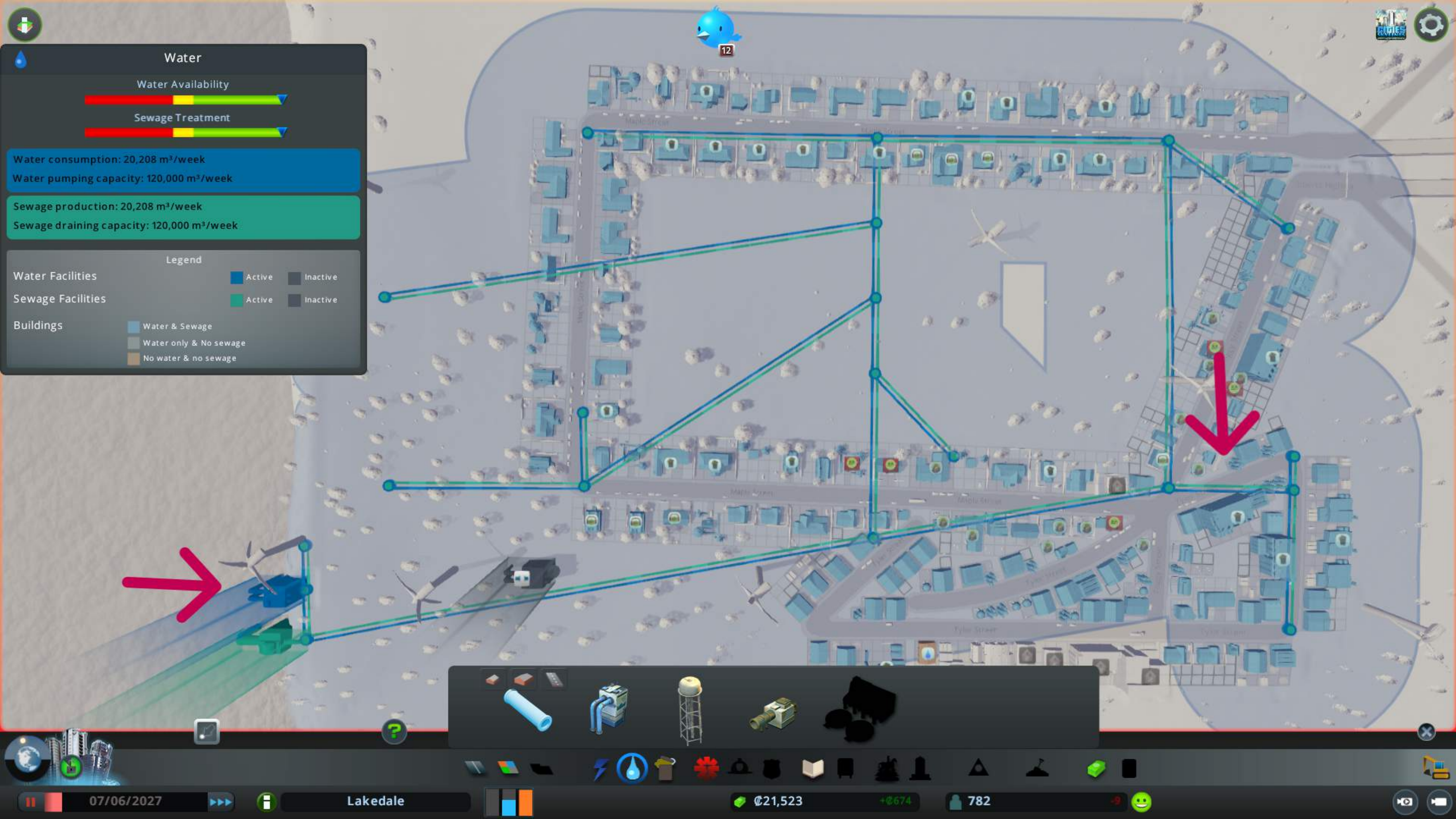}
      \label{subfig:skylines_showcase_03_after_water}
    \end{subfigure}
    \begin{subfigure}{0.32\textwidth}
      \centering
      \includegraphics[width=\linewidth]{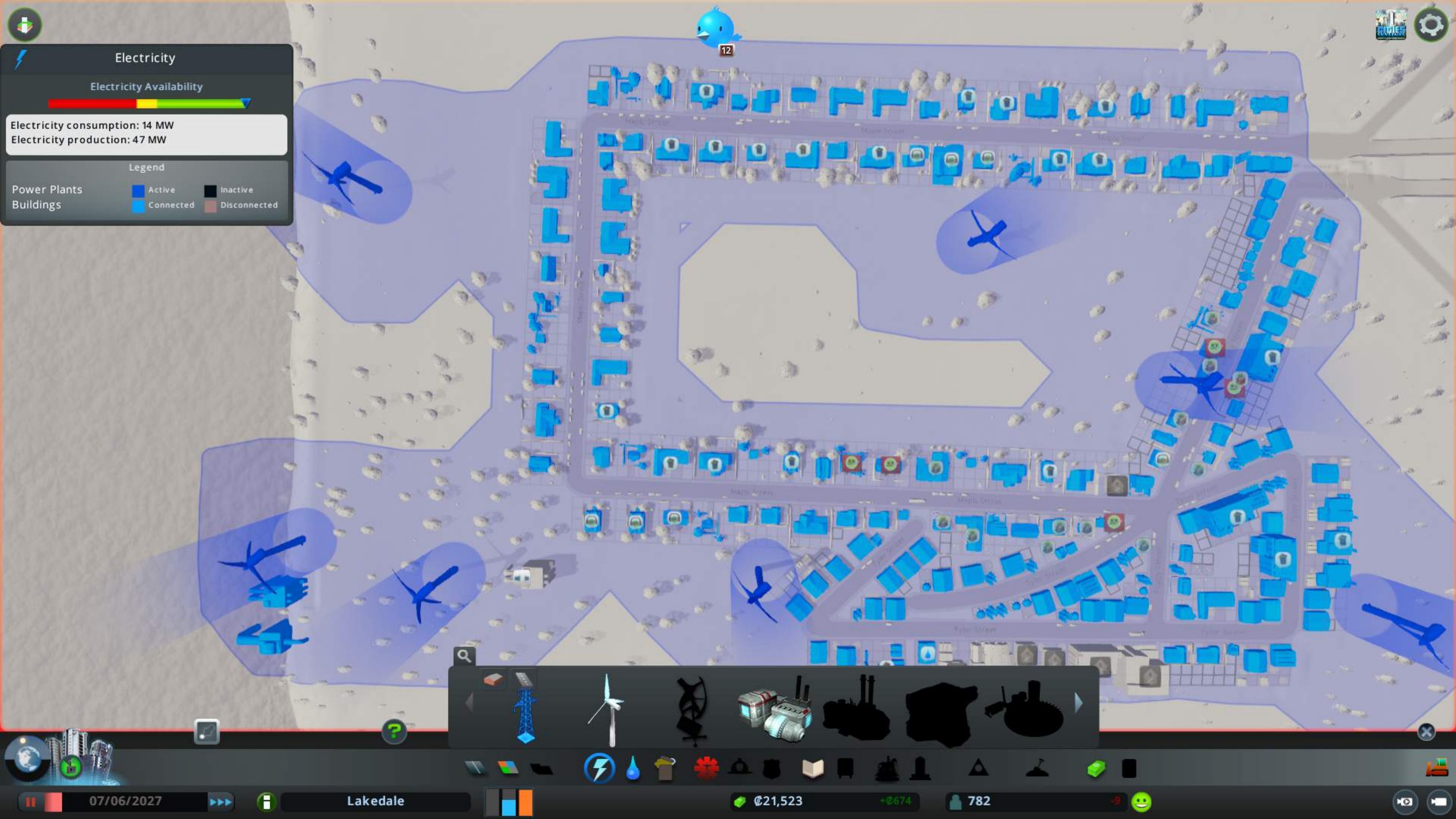}
      \label{subfig:skylines_showcase_03_after_power}
    \end{subfigure}

    \centering
    \begin{subfigure}{\textwidth}
      \centering
      \vspace{-10pt}
      \caption{\projname's craftwork III with human assistant within three unit operations to place the water pumping station, lay water pipe on the right side and develop the bottom area with industrial zones.  \textbf{Population: 780+}.}
      \label{appx_fig:skylines_showcase_03_after}
    \end{subfigure}
  \vspace{-10pt}
  \caption{Demonstrations of three cities built by \projname in zoning view (left),  water view (middle) and electricity view (right). Figures~\ref{appx_fig:skylines_showcase_01_after},~\ref{appx_fig:skylines_showcase_02_after},~\ref{appx_fig:skylines_showcase_03_after} show the cities with human assistance to address construction issues (shown in red arrow). Populations shown in the figures are close to but not exactly the maximal population since they are changed dynamically.  } 
  \label{appx_fig:skylines_case_study_water}
\end{figure}

Figure~\ref{appx_fig:skylines_case_study_water} displays three cases where \projname fails to ensure the water supply due to the disconnection of water pipes and the missing water pumping station. All of them can be fixed within three unit operations. As shown in Figure~\ref{appx_fig:skylines_showcase_01_after} and ~\ref{appx_fig:skylines_showcase_03_after}, we observe a significant increase in the population if these mistakes are fixed, which proves that \projname already has the ability to build a reasonable city but some minor adjustments are needed.

\clearpage

%% file: appendix/software.tex
\subsection{Selected Software Applications}

Besides targeting complex digital games, \projname also includes an initial benchmark task set across diverse software applications. The selected applications include Chrome, Outlook, Feishu, CapCut, and Meitu. These applications cover popular applications for daily tasks in different usage categories, such as web browsing, communication, work, and media manipulation. 
Table~\ref{tab:software_versions} shows the exact application versions benchmarked in this paper.
Five distinct tasks were designed for each application to represent their target domains and explore the difficulties posed to LMM-based agents and analyze their limitations. Figure \ref{fig:tasks} shows an overview of all tasks across applications and Tables~\ref{tab:software_task_123} and~\ref{tab:software_task_45} detail each task.

Chrome and Outlook were selected as common representatives for web browsing and e-mail, with well-known functionality and UI design. CapCut and Meitu are two popular media editing applications for video/image editing with their own interaction styles. Lastly, Feishu (also known ads Lark) is an office collaboration and productivity application, which includes messaging, calendar/meetings, and approval workflows. It represents a complex business application that doesn't strictly follow OS-specific UI guidelines. To the best of our knowledge, this is the \textbf{first agent} targeting applications like CapCut, Meitu, and Feishu.

\subsubsection{Brief Descriptions}

\begin{wraptable}{r}{0.4\textwidth}
\centering
\vspace{-12pt}
\caption{Exact software versions utilized in the described experiments. Similar versions should behave similarly.}
\label{tab:software_versions}
\begin{tabular}{c|c}
\toprule
\textbf{Software} & \textbf{Version} \\ \midrule
Chrome & 125.0.6422.142 \\ 
Outlook & 1.2024.529.200 \\ 
CapCut & 4.0.0 \\ 
Meitu & 7.5.6.1 \\ 
Feishu & 7.19.5 \\ 
\bottomrule
\end{tabular}
\vspace{-10pt}
\end{wraptable}
\textbf{Chrome} is a web browser developed by Google. It allows users to access and utilize online resources through activities such as browsing websites, streaming videos, and using web applications. Additionally, users can customize their browsing experience with various extensions, manage bookmarks and passwords, and synchronize their data across multiple devices for seamless access.

\textbf{Outlook} is an application by that allows users to manage emails, calendars, contacts, and tasks. It includes tools for communication and scheduling through features such as sending and receiving emails, setting up meetings, and keeping track of appointments. Additionally, users can customize their experience and integrate Outlook with other Microsoft Office applications.

\textbf{CapCut} is a popular video editing application developed by ByteDance. It provides easy-to-use editing tools and and enables users to create quality videos with a range of advanced features.
CapCut offers a set of editing tools, including trimming, cutting, merging, and splitting video clips; the application of various effects, filters, and transitions; as well as adjusting speed, and adding music or text overlays.

\textbf{Meitu} is a photo editing application. It is designed to cater to a broad audience and enables users to enhance and transform their photos with minimal effort. Meitu offers editing tools, including basic adjustments like cropping, rotating, and resizing, as well as advanced features such as beauty retouching, filters, and special effects. 
Additionally, Meitu offers a wide range of stickers, frames, and text options to further personalize photos.

\textbf{Feishu}, also known as Lark, is a business communication and collaboration platform by ByteDance. It integrates various tools for office workflows and project management.
Feishu offers a wide array of functionalities, including instant messaging, video conferencing, file sharing, and collaboration within the app.
It also includes an integrated calendar, which helps users schedule and manage meetings and events, and task management tools that allow users to assign and track tasks.

\subsection{Software Tasks}
\begin{figure}
    \centering
    \begin{subfigure}{0.32\textwidth}
      \centering
      \includegraphics[width=\linewidth]{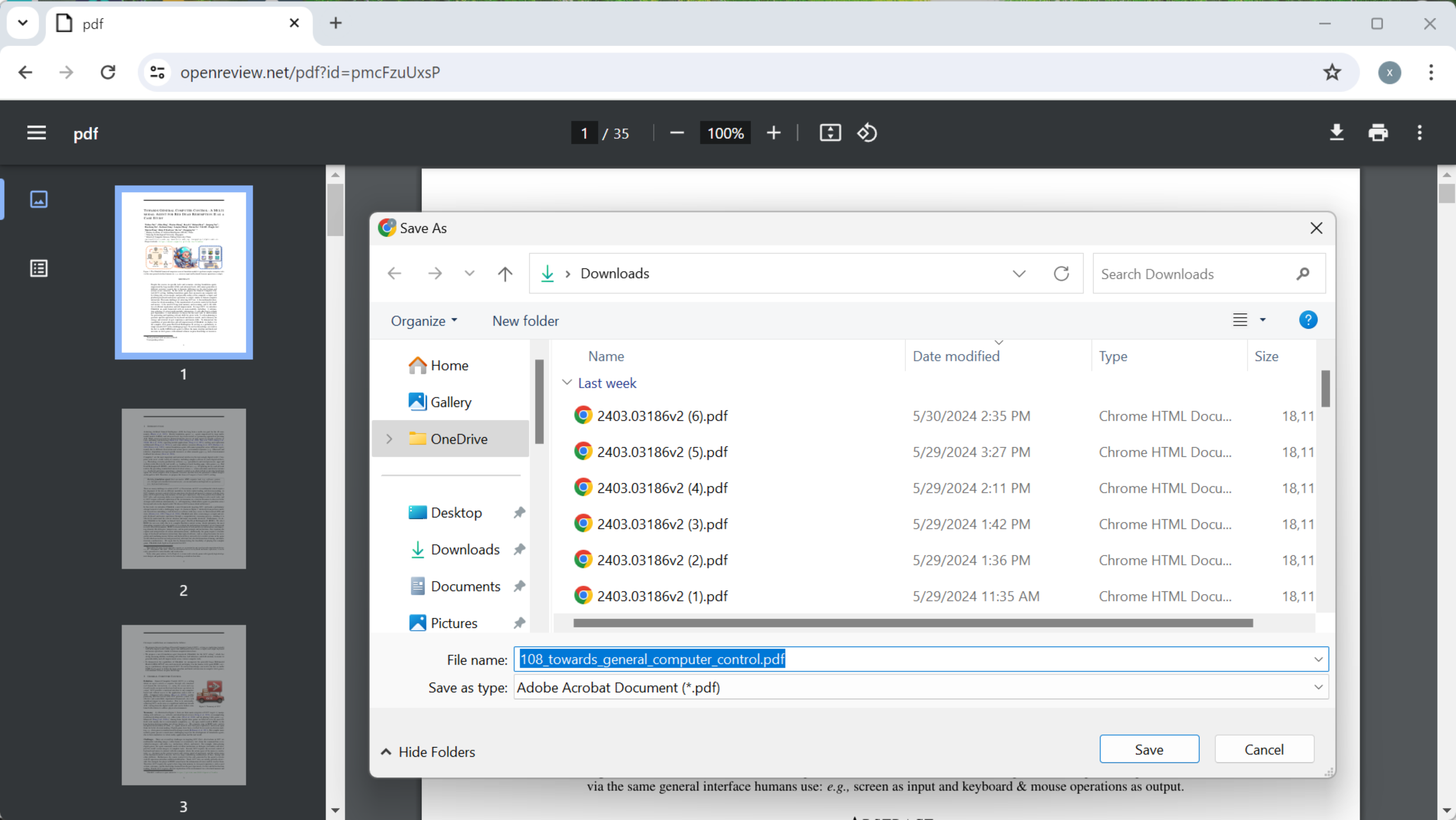}
      \caption{\scriptsize Download Paper} \label{subfig:01_Chrome_task1}
    \end{subfigure}
    \begin{subfigure}{0.32\textwidth}
      \centering
      \includegraphics[width=\linewidth]{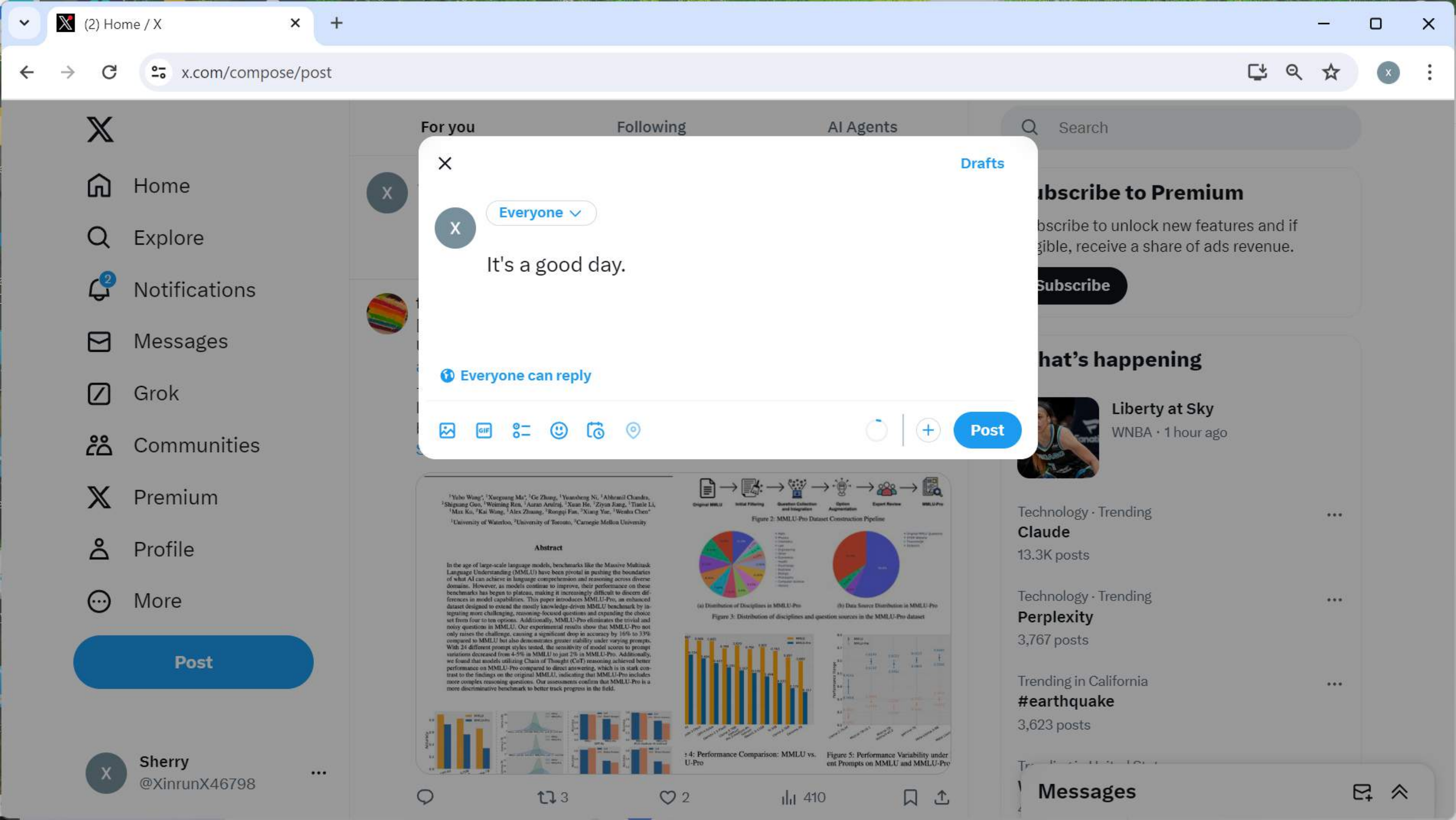}
      \caption{\scriptsize Post in Twitter} \label{subfig:01_Chrome_task2}
    \end{subfigure}
    \begin{subfigure}{0.32\textwidth}
      \centering
      \includegraphics[width=\linewidth]{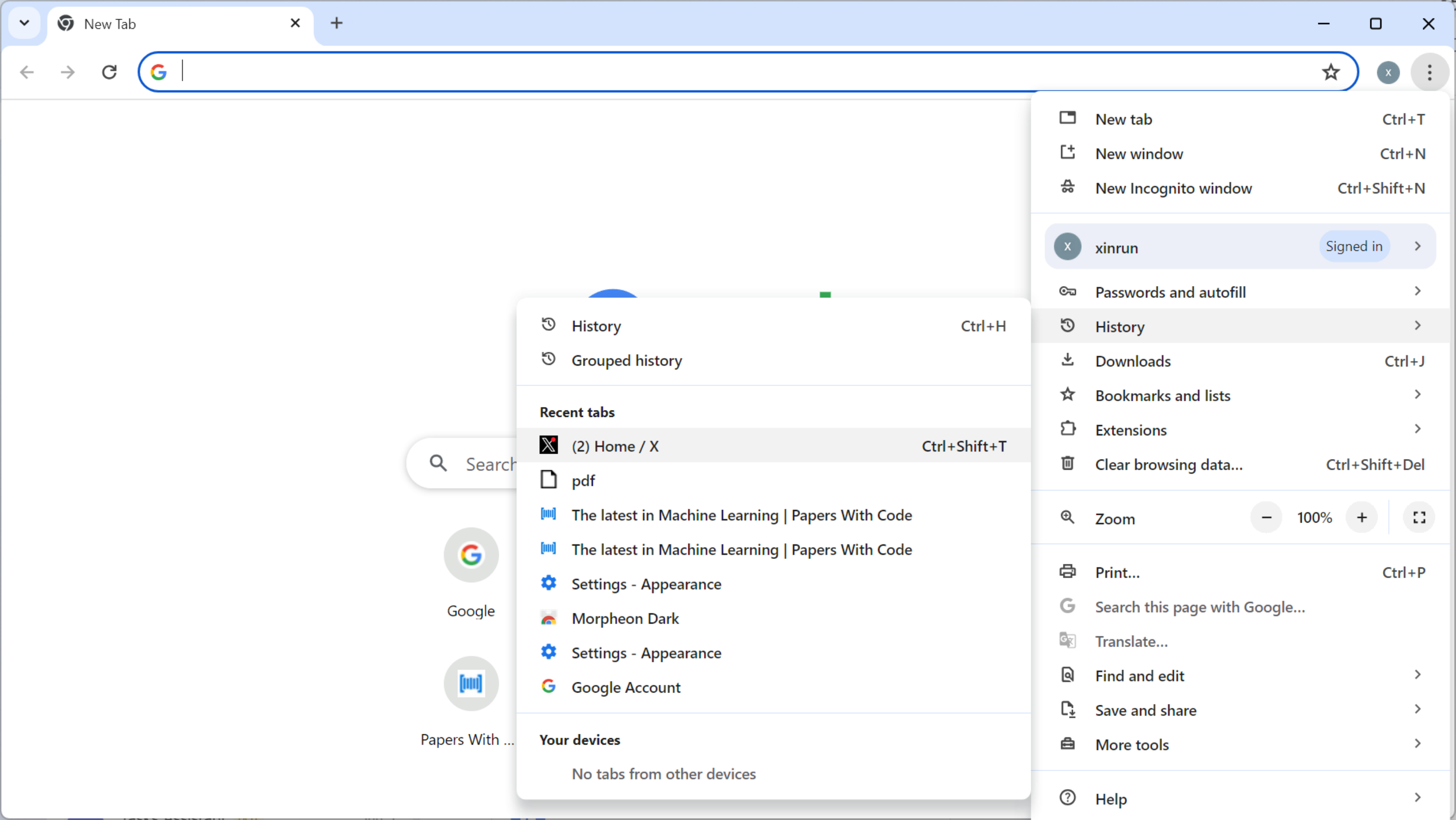}
      \caption{\scriptsize Open Closed Page} \label{subfig:01_Chrome_task3}
    \end{subfigure}
    \begin{subfigure}{0.32\textwidth}
      \centering
      \includegraphics[width=\linewidth]{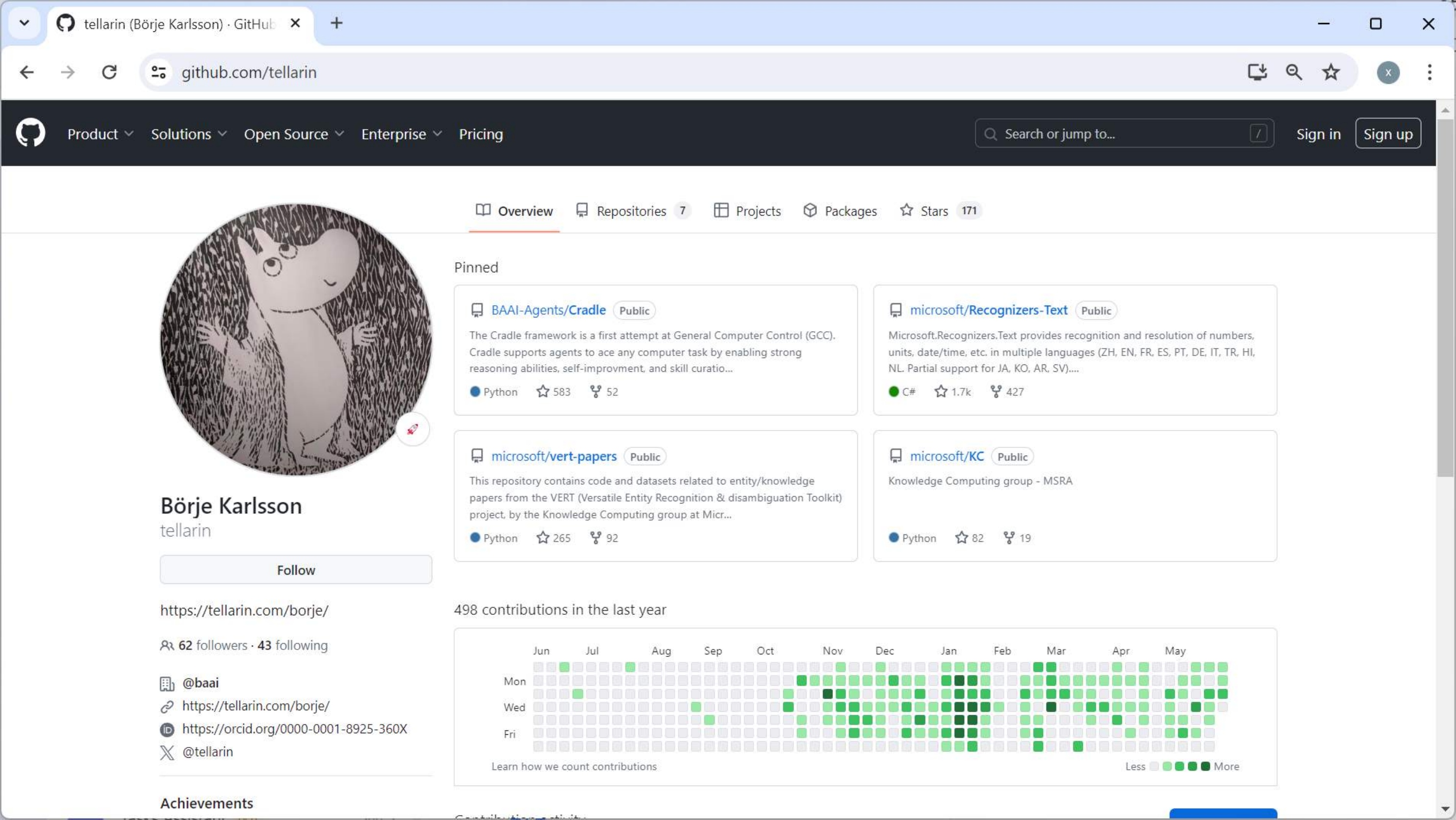}
      \caption{\scriptsize Go to Profile} \label{subfig:01_Chrome_task4}
    \end{subfigure}
    \begin{subfigure}{0.32\textwidth}
      \centering
      \includegraphics[width=\linewidth]{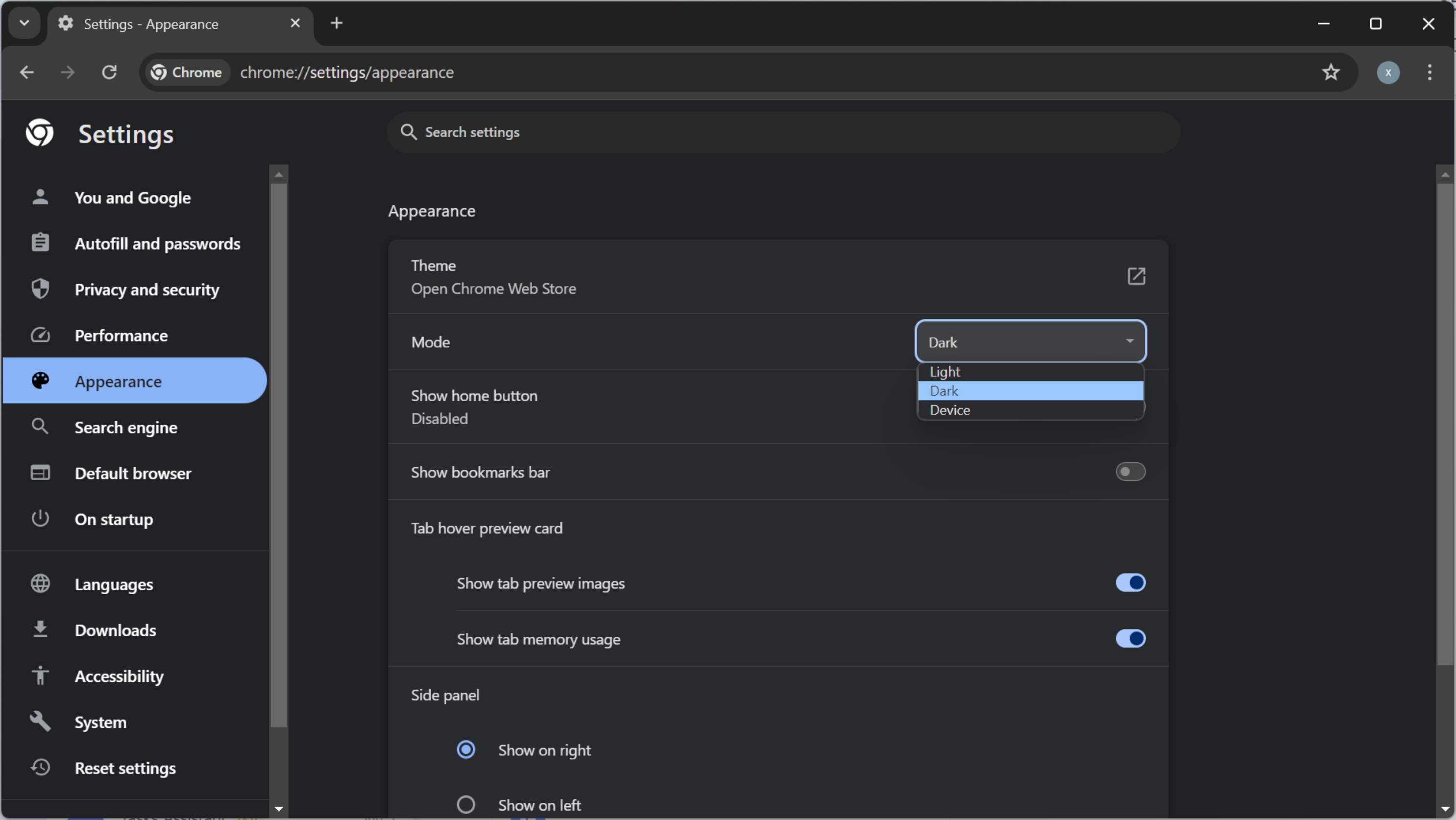}
      \caption{\scriptsize Change Mode} \label{subfig:01_Chrome_task5}
    \end{subfigure}
    \caption{Screenshots of Chrome tasks.}
    \label{fig:chrome_tasks}
\end{figure}

\begin{figure}
\centering
    \begin{subfigure}{0.32\textwidth}
      \centering
      \includegraphics[width=\linewidth]{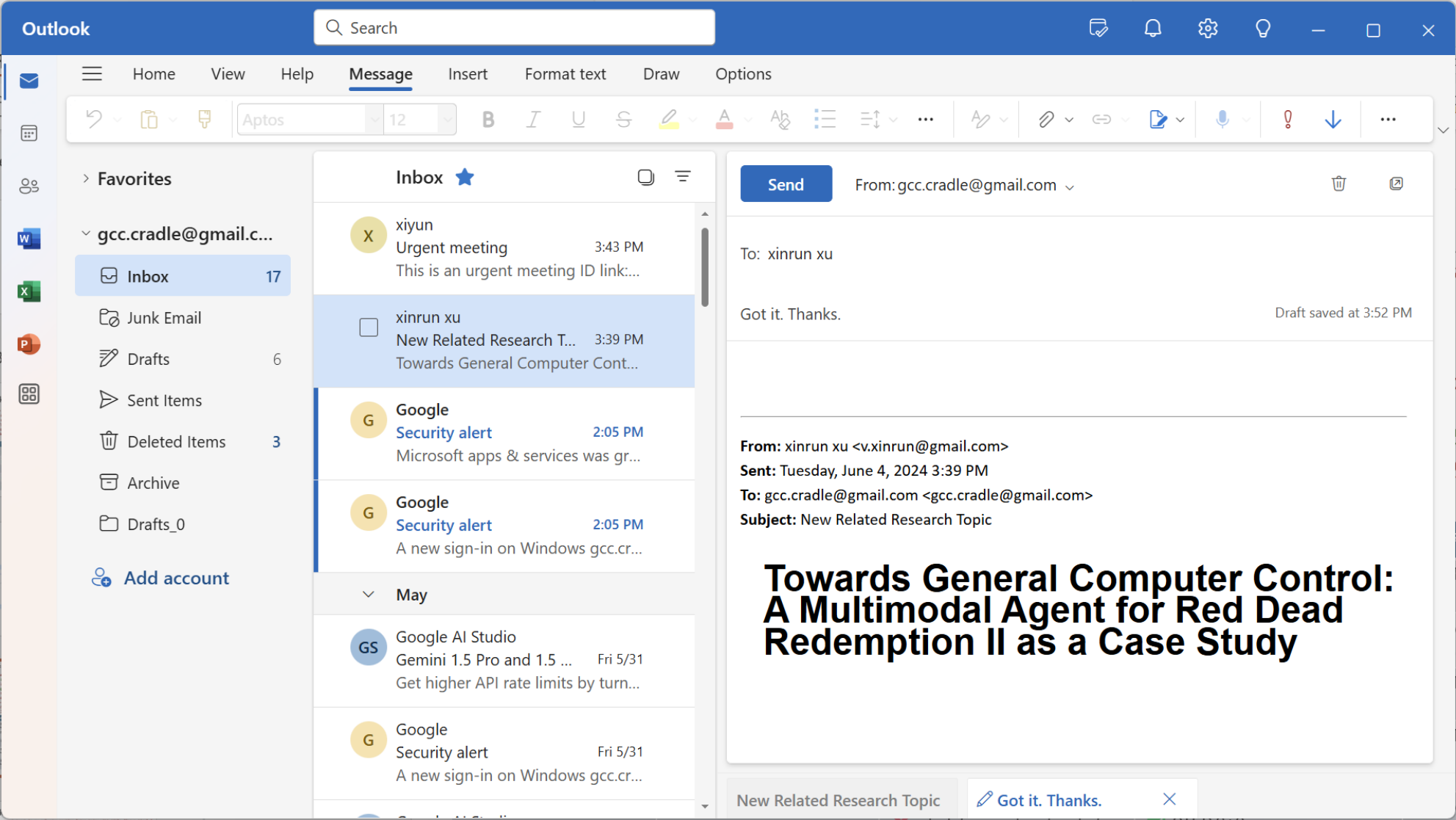}
      \caption{\scriptsize Send New E-mail} \label{subfig:02_Outlook_task1}
    \end{subfigure}
    \begin{subfigure}{0.32\textwidth}
      \centering
      \includegraphics[width=\linewidth]{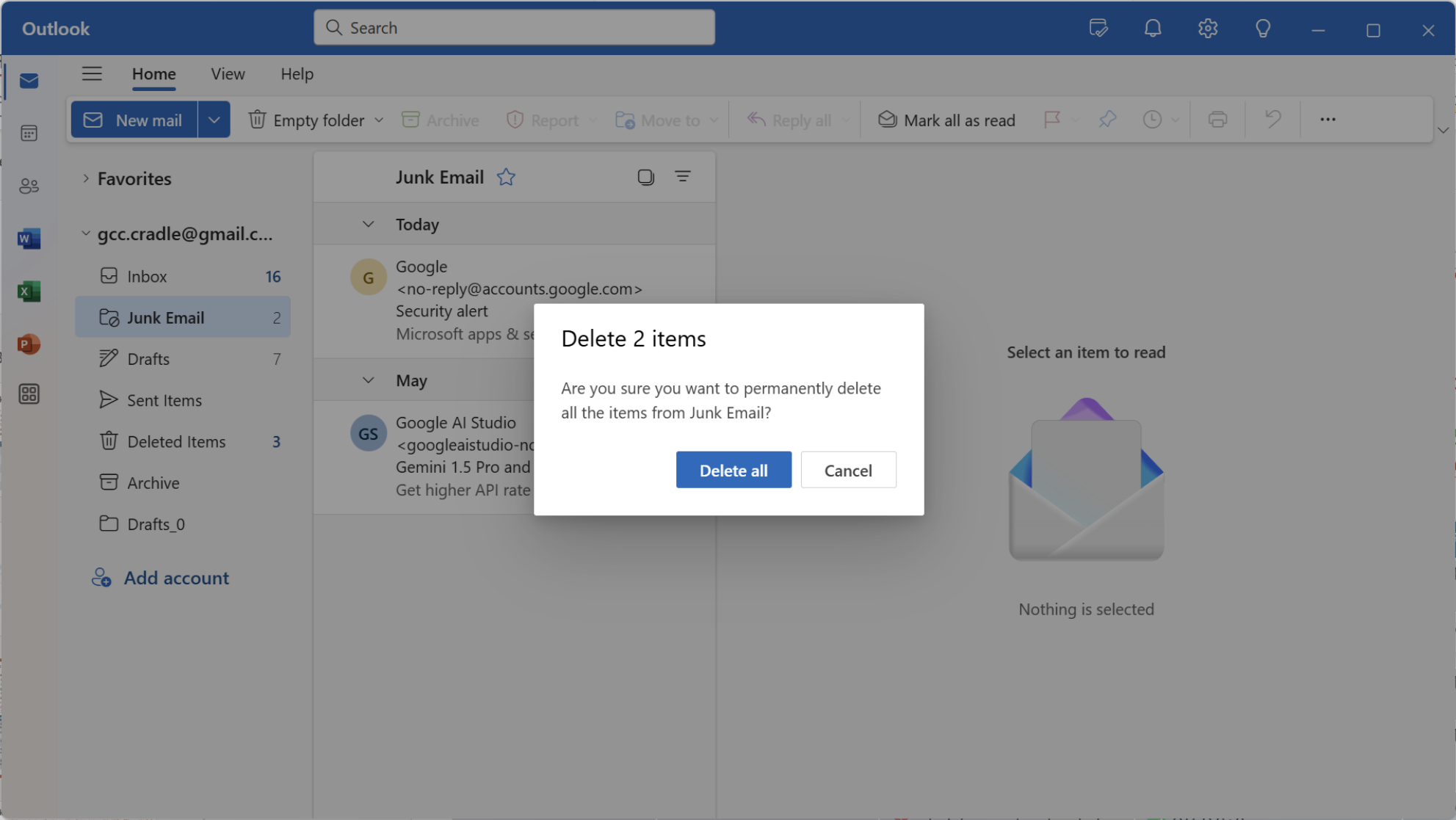}
      \caption{\scriptsize Empty Junk Folder} \label{subfig:02_Outlook_task2}
    \end{subfigure}
    \begin{subfigure}{0.32\textwidth}
      \centering
      \includegraphics[width=\linewidth]{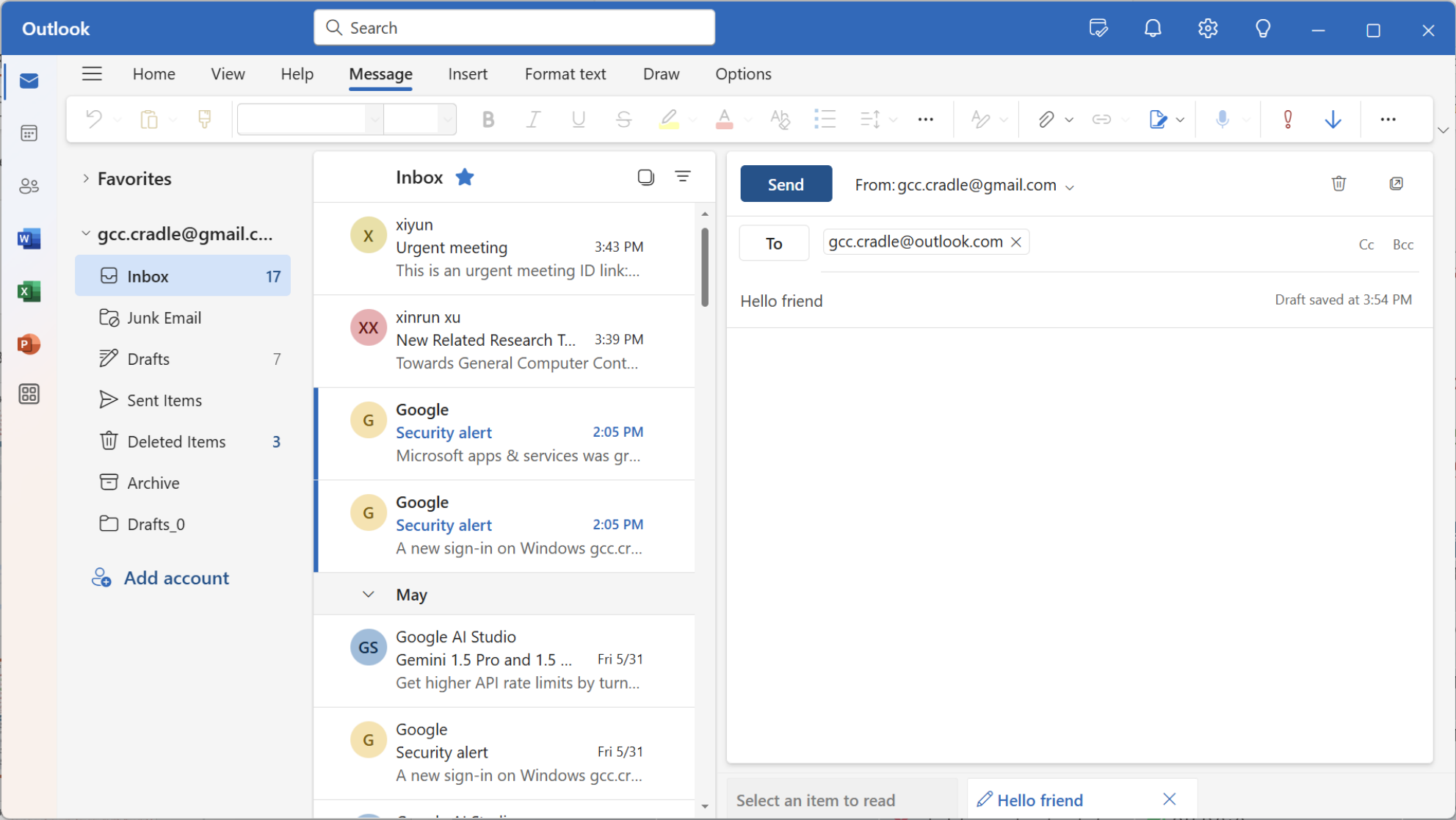}
      \caption{\scriptsize Reply to Person} \label{subfig:02_Outlook_task3}
    \end{subfigure}
    \begin{subfigure}{0.32\textwidth}
      \centering
      \includegraphics[width=\linewidth]{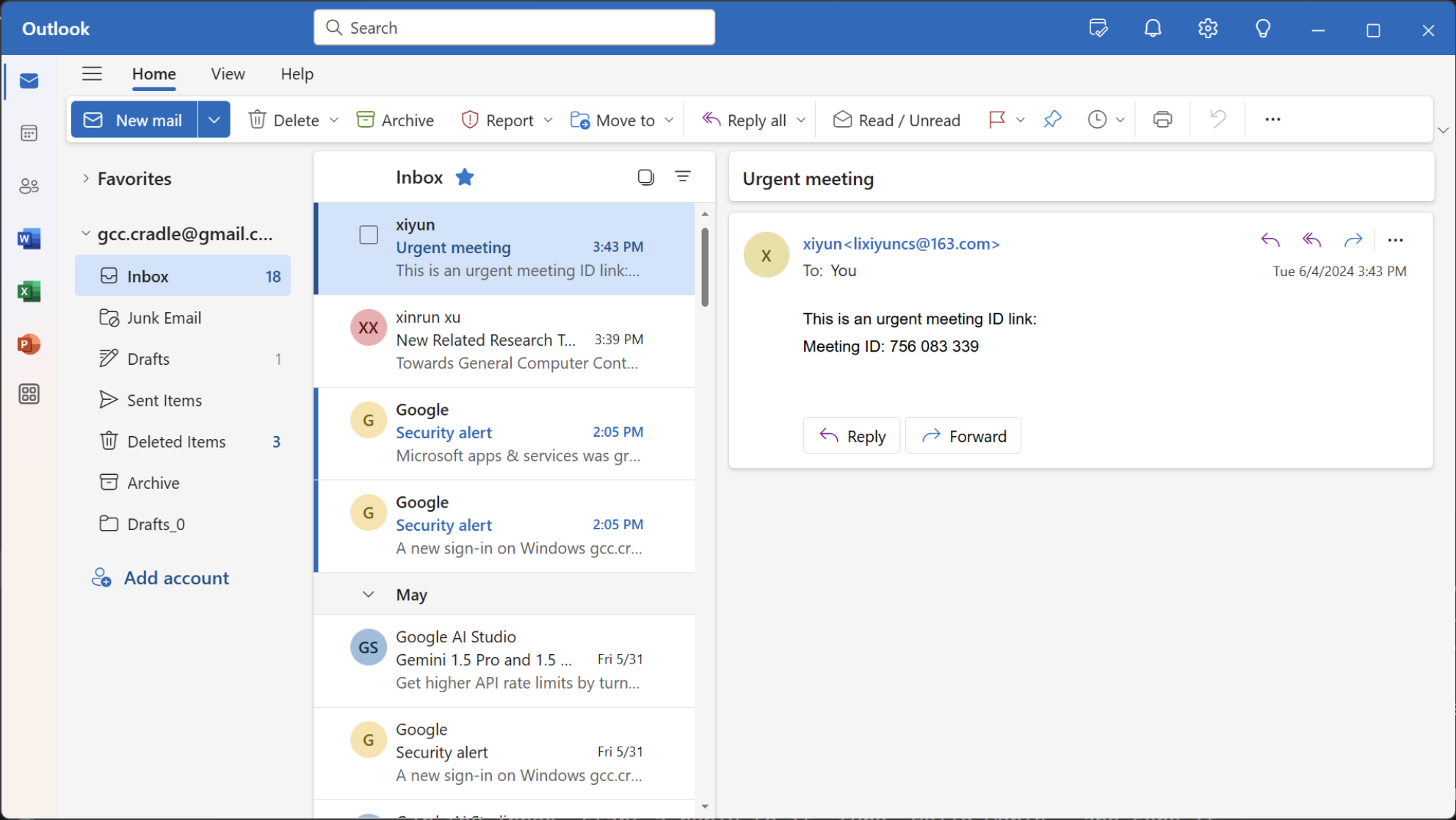}
      \caption{\scriptsize Find Target E-mail} \label{subfig:02_Outlook_task4}
    \end{subfigure}
    \begin{subfigure}{0.32\textwidth}
      \centering
      \includegraphics[width=\linewidth]{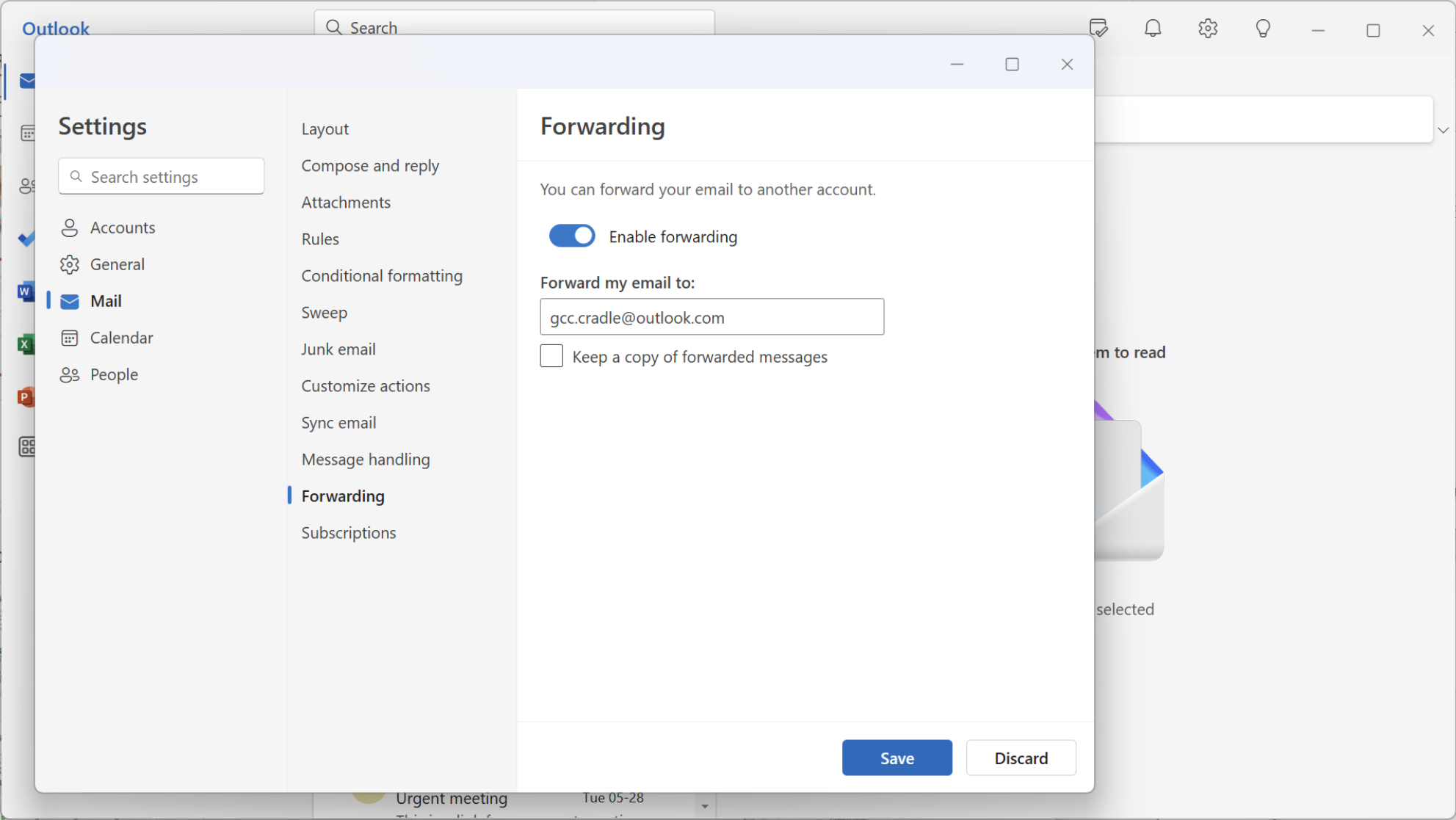}
      \caption{\scriptsize Setup Forwarding} \label{subfig:02_Outlook_task5}
    \end{subfigure}
    \caption{Screenshots of Outlook tasks.}
    \label{fig:outlook_tasks}
\end{figure}

\begin{figure}
    \centering
    \begin{subfigure}{0.32\textwidth}
      \centering
      \includegraphics[width=\linewidth]{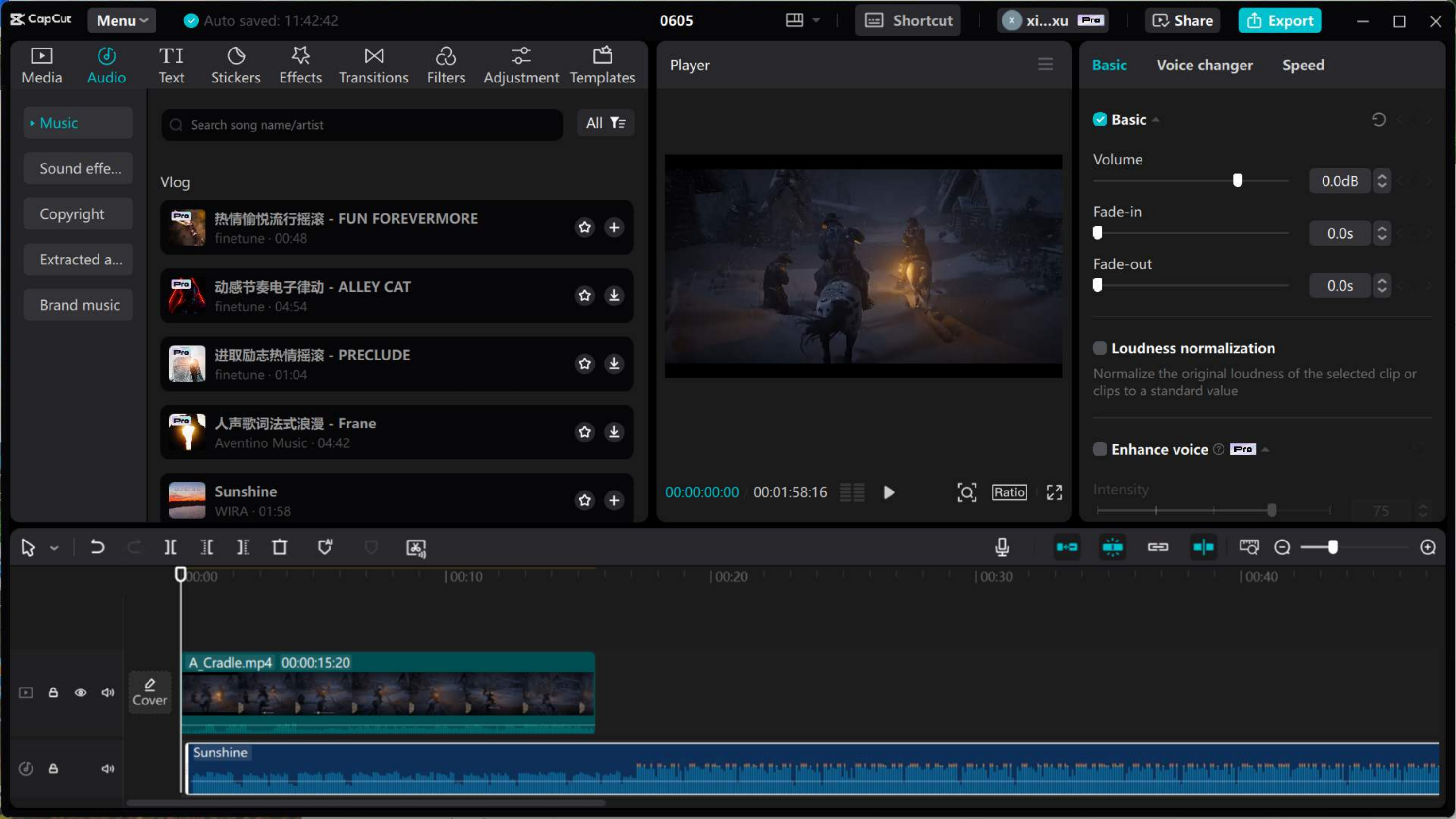}
      \caption{\scriptsize Create Media Project} \label{subfig:03_CapCut_task1}
    \end{subfigure}
    \begin{subfigure}{0.32\textwidth}
      \centering
      \includegraphics[width=\linewidth]{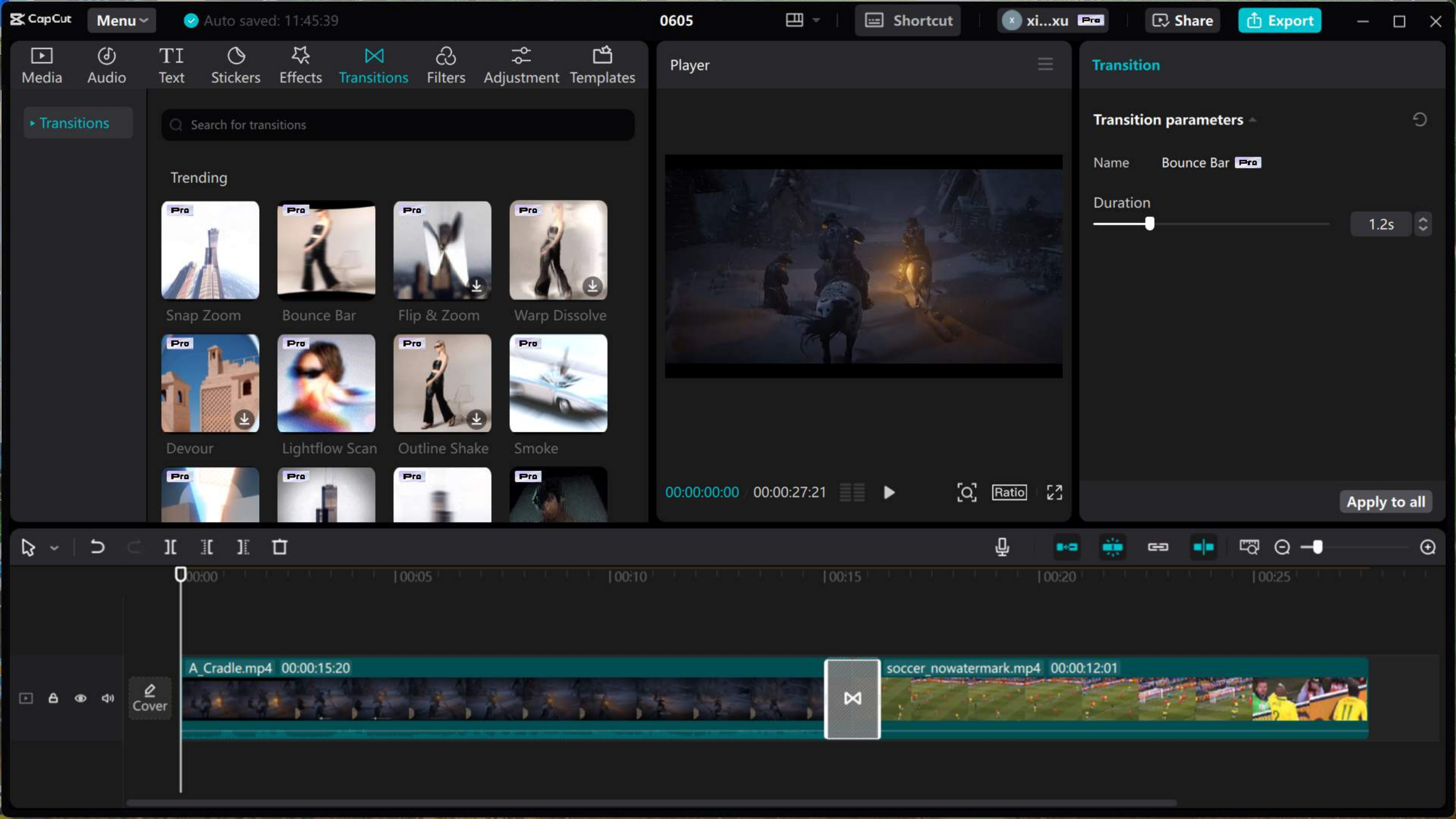}
      \caption{\scriptsize Add Transition} \label{subfig:03_CapCut_task2}
    \end{subfigure}
    \begin{subfigure}{0.32\textwidth}
      \centering
      \includegraphics[width=\linewidth]{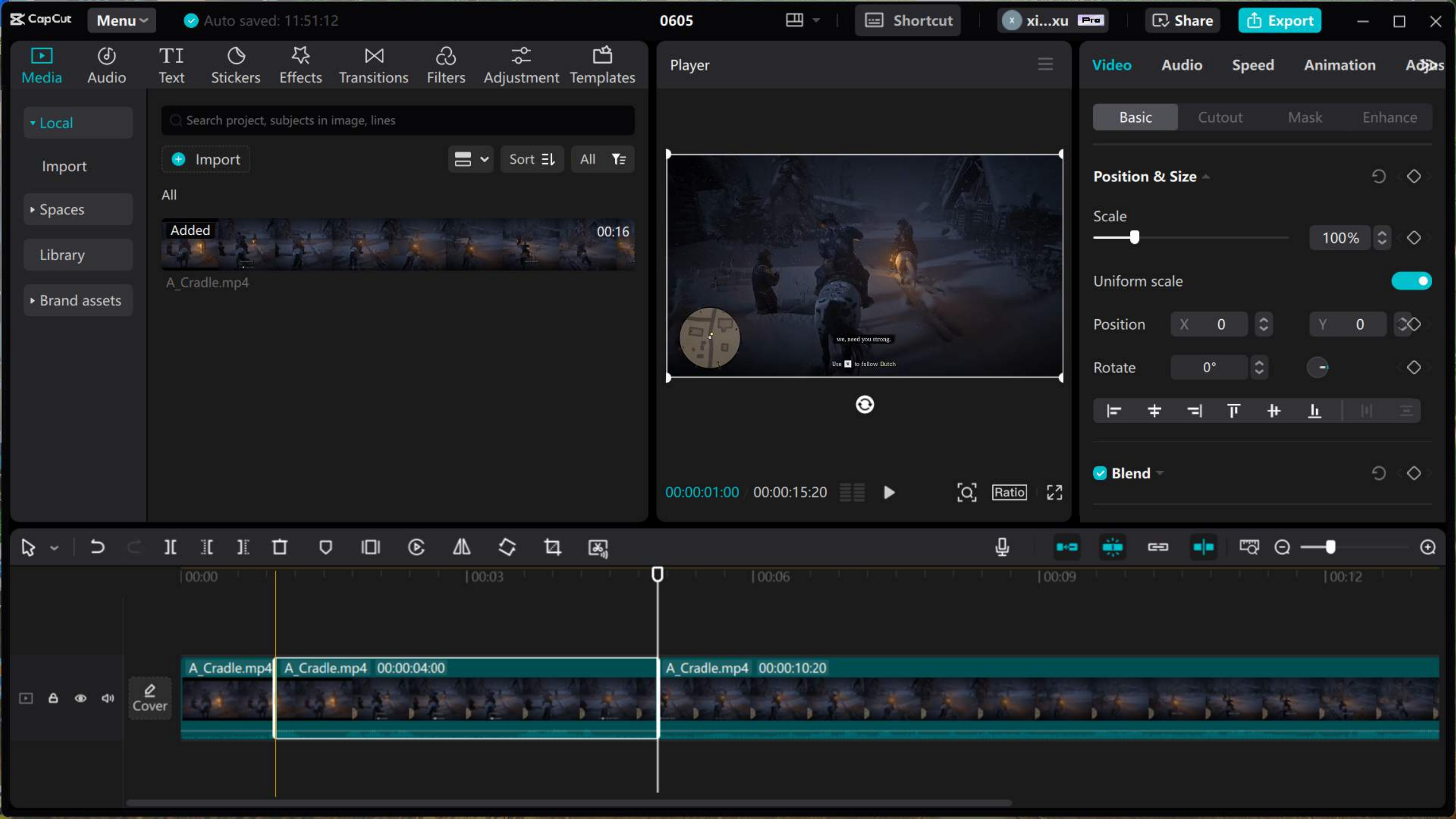}
      \caption{\scriptsize Crop by Timestamp} \label{subfig:03_CapCut_task3}
    \end{subfigure}
    \begin{subfigure}{0.32\textwidth}
      \centering
      \includegraphics[width=\linewidth]{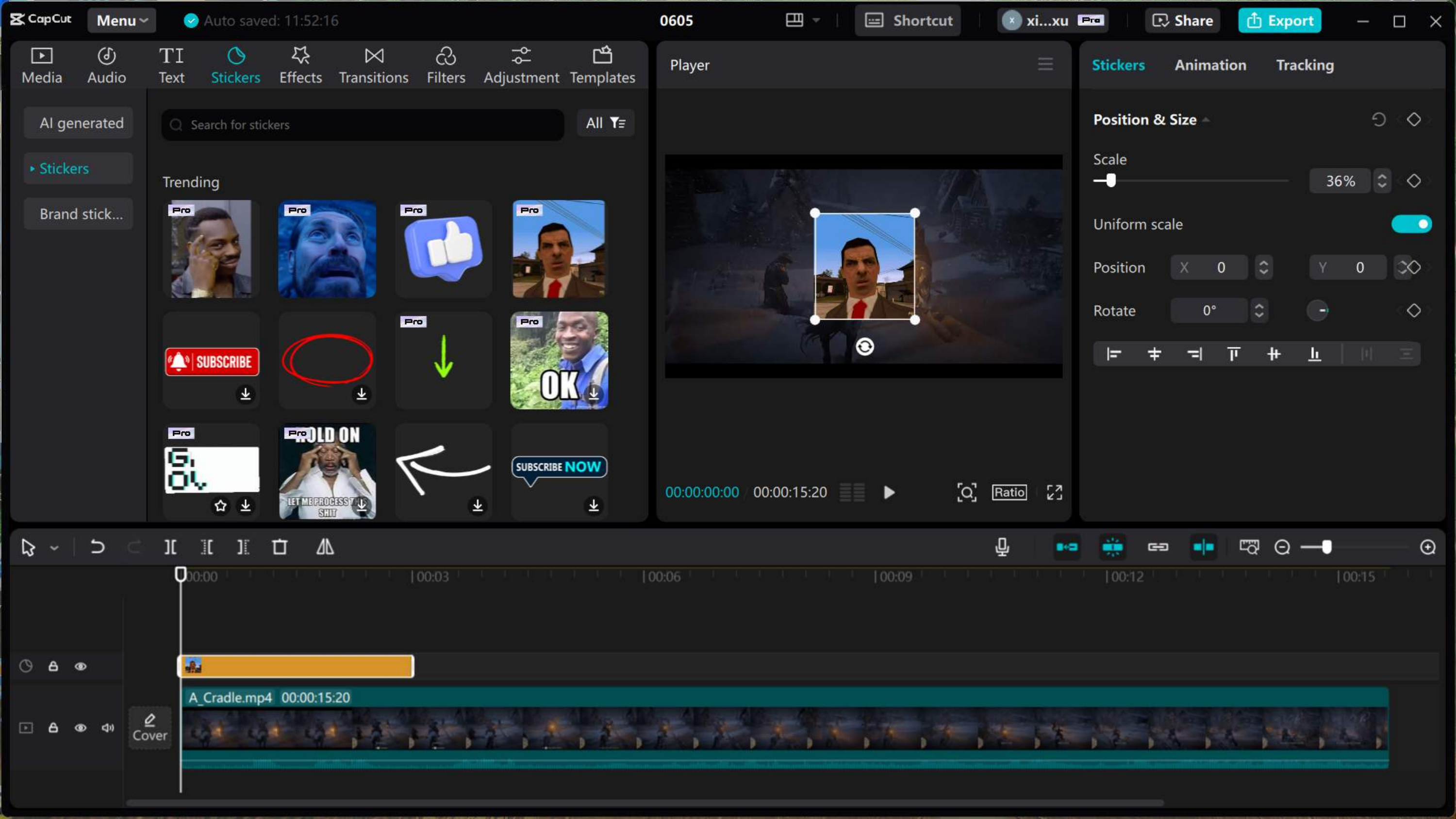}
      \caption{\scriptsize Add Sticker} \label{subfig:03_CapCut_task4}
    \end{subfigure}
    \begin{subfigure}{0.32\textwidth}
      \centering
      \includegraphics[width=\linewidth]{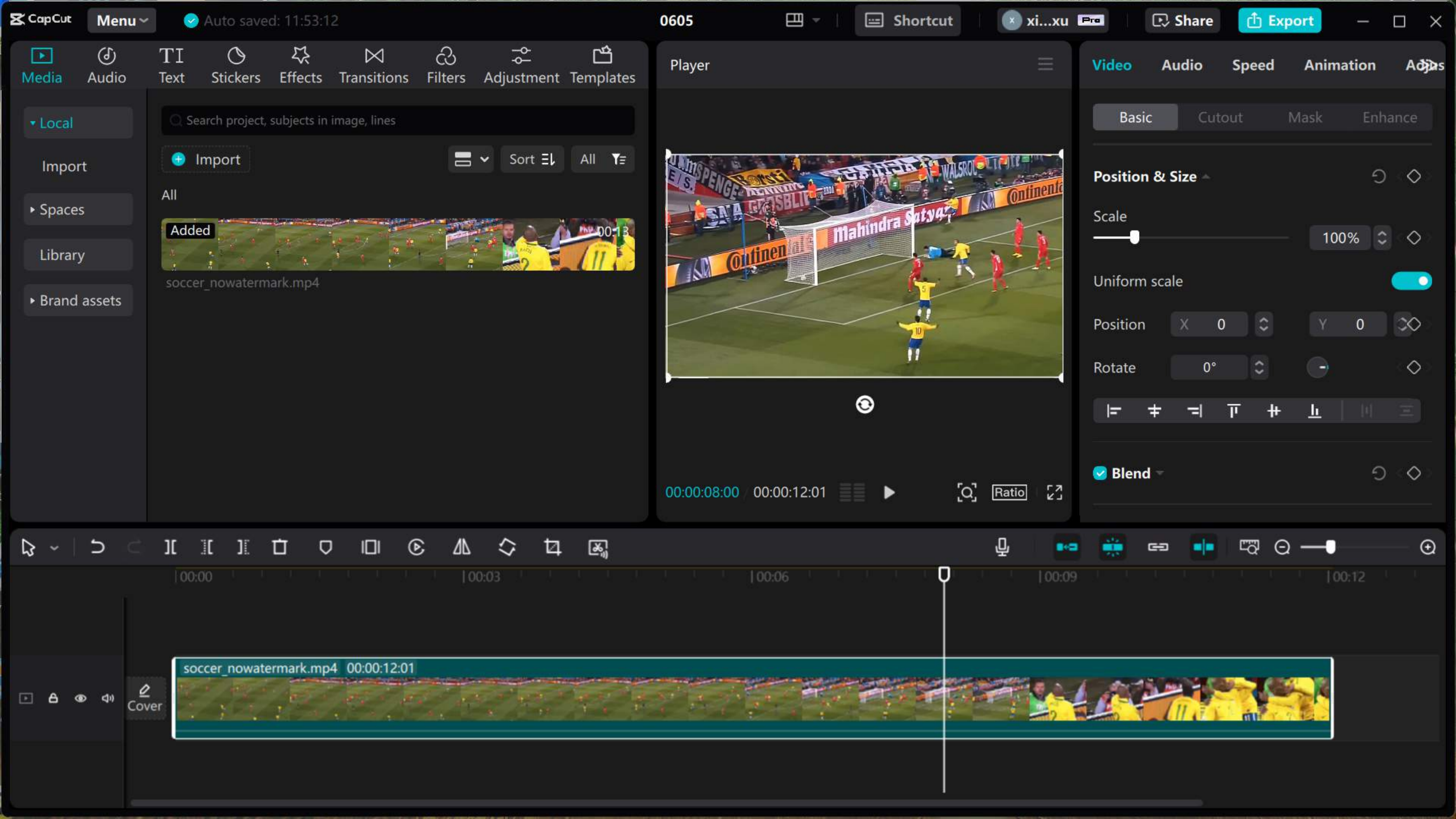}
      \caption{\scriptsize Crop by Content} \label{subfig:03_CapCut_task5}
    \end{subfigure}
    \caption{Screenshots of CapCut tasks.}
    \label{fig:capcut_tasks}
\end{figure}

\begin{figure}
    \centering
    \begin{subfigure}{0.32\textwidth}
      \centering
      \includegraphics[width=\linewidth]{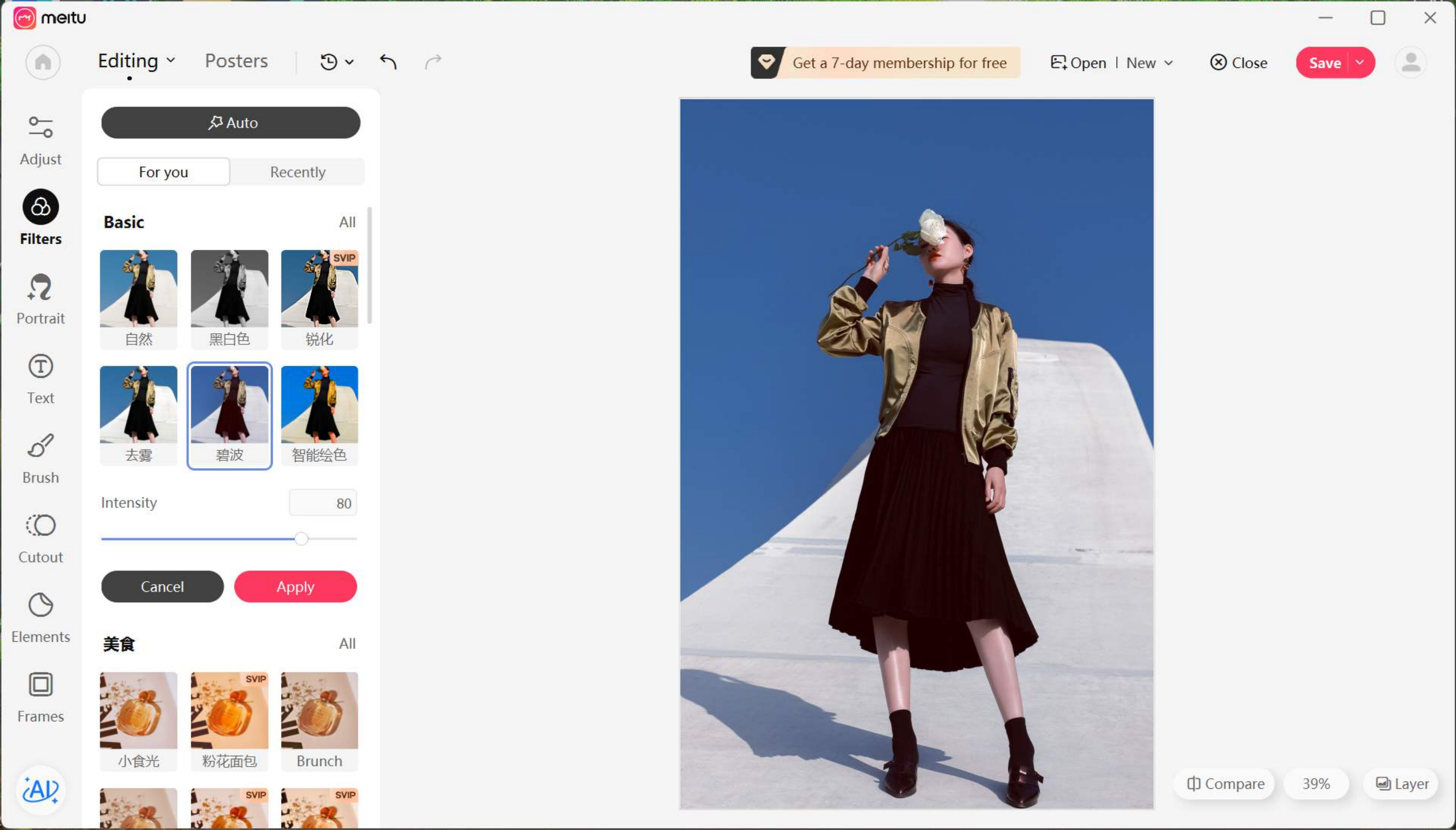}
      \caption{\scriptsize Apply Filter} \label{subfig:04_Xiuxiu_task1}
    \end{subfigure}
    \begin{subfigure}{0.32\textwidth}
      \centering
      \includegraphics[width=\linewidth]{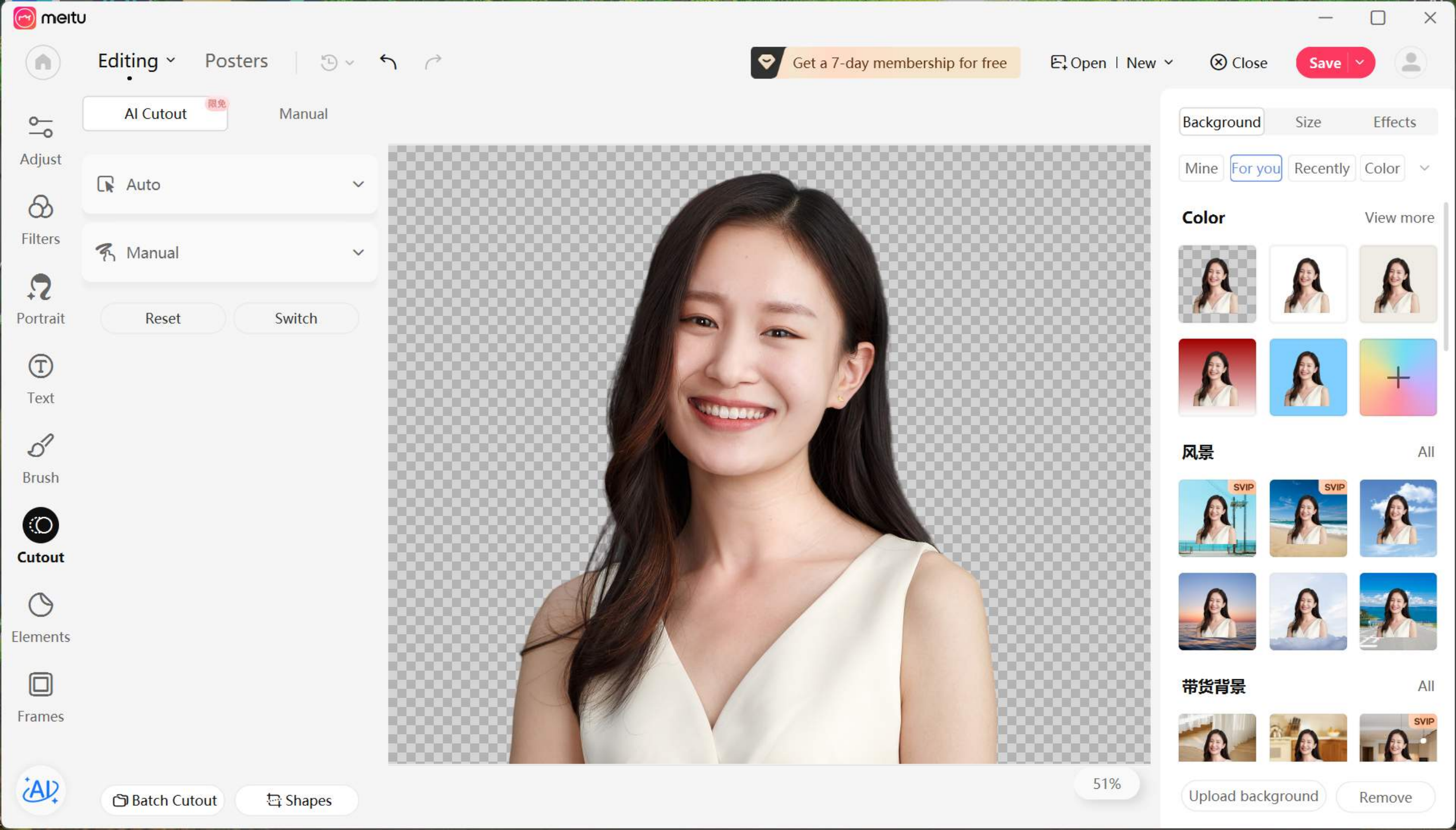}
      \caption{\scriptsize Cutout} \label{subfig:04_Xiuxiu_task2}
    \end{subfigure}
    \begin{subfigure}{0.32\textwidth}
      \centering
      \includegraphics[width=\linewidth]{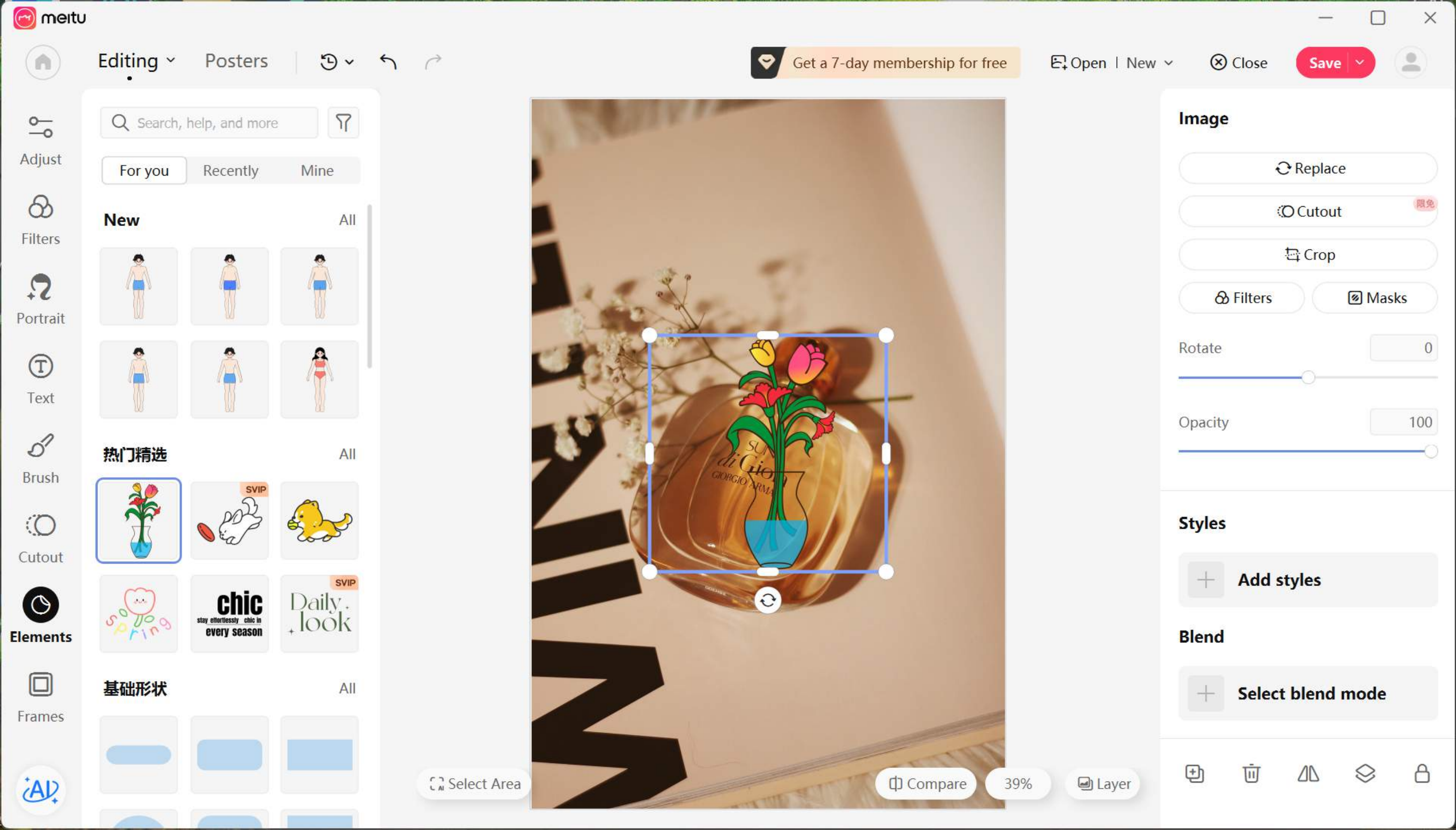}
      \caption{\scriptsize Add Sticker} \label{subfig:04_Xiuxiu_task3}
    \end{subfigure}
    \begin{subfigure}{0.32\textwidth}
      \centering
      \includegraphics[width=\linewidth]{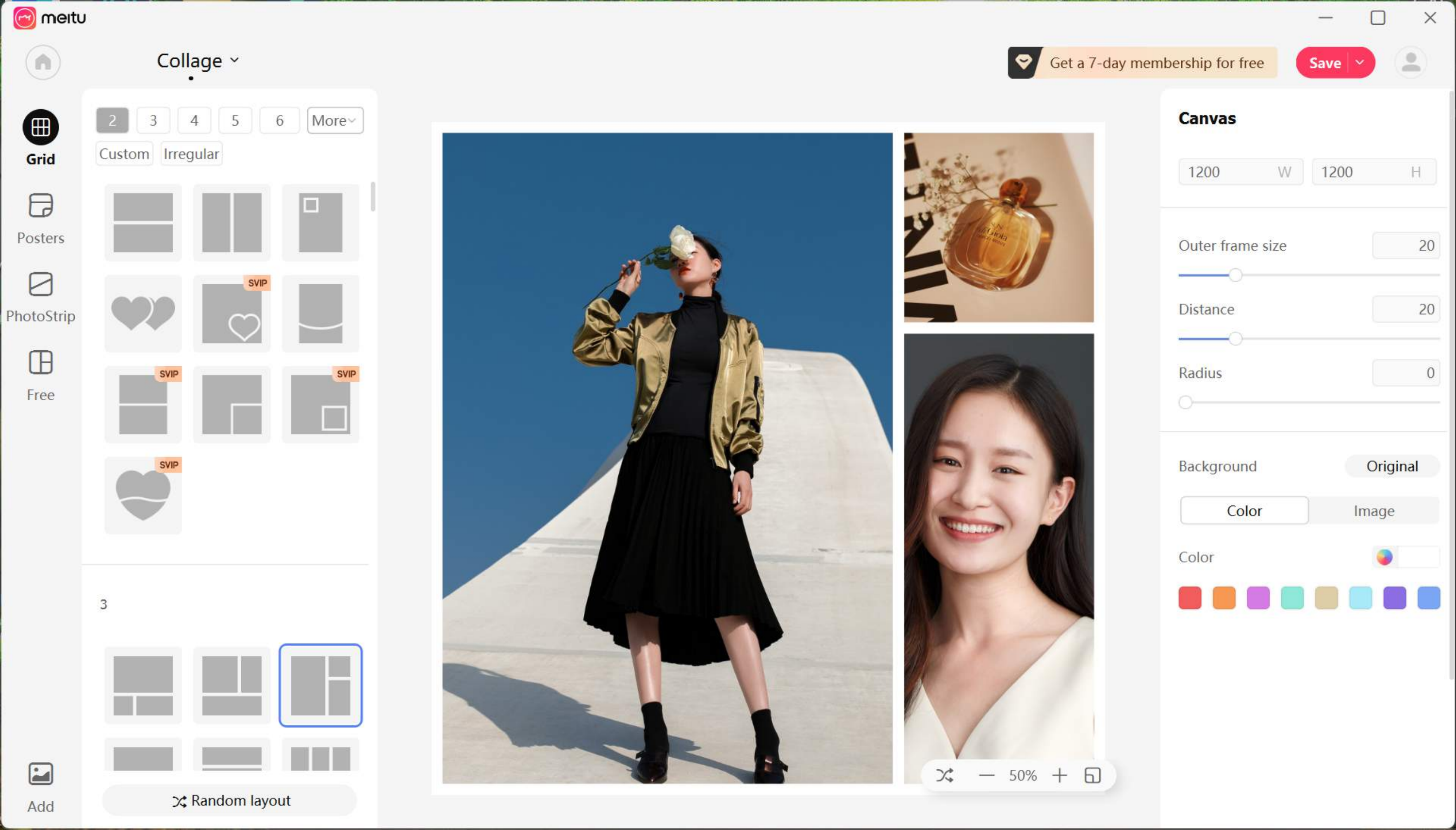}
      \caption{\scriptsize Create Collage} \label{subfig:04_Xiuxiu_task4}
    \end{subfigure}
    \begin{subfigure}{0.32\textwidth}
      \centering
      \includegraphics[width=\linewidth]{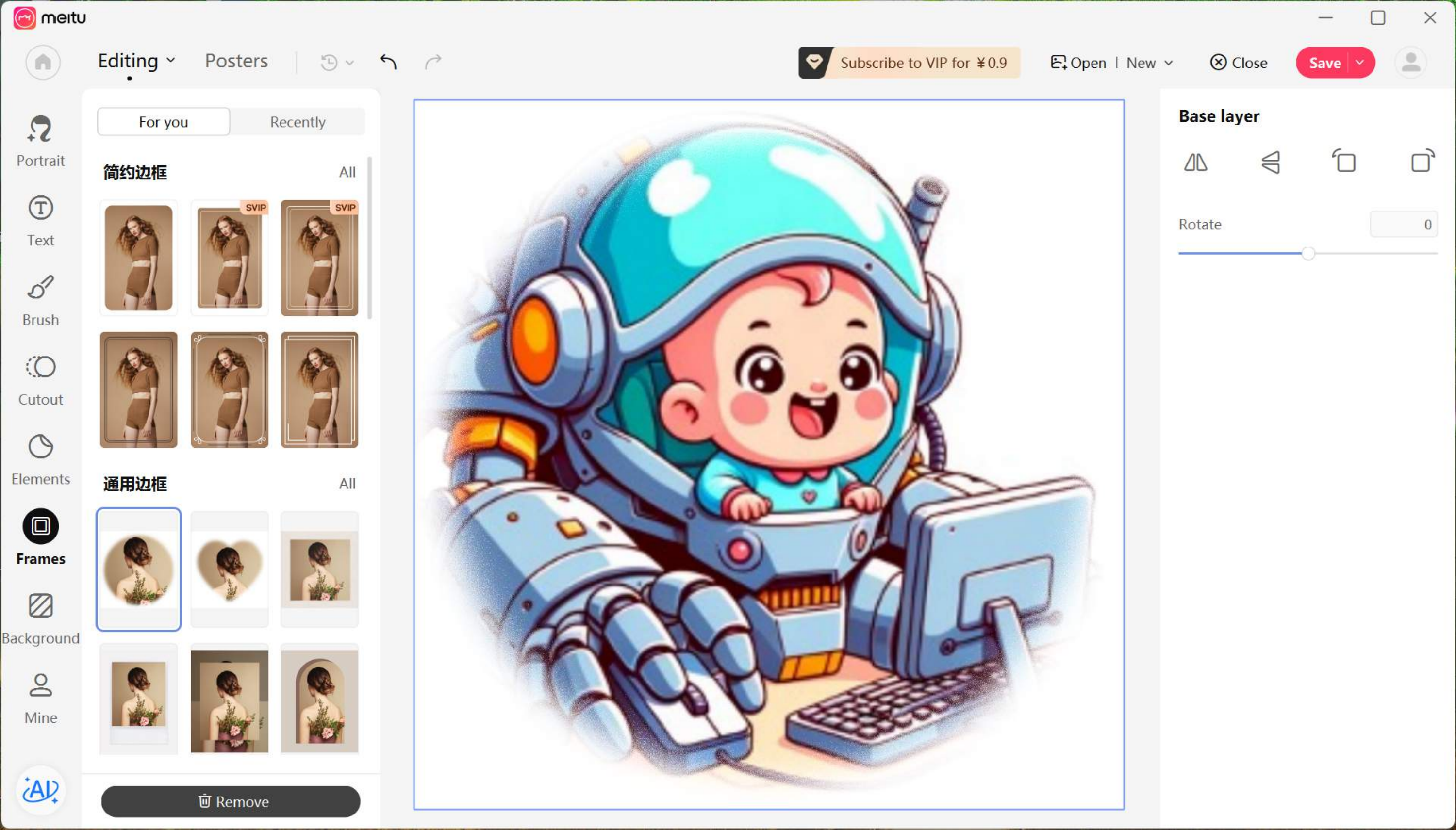}
      \caption{\scriptsize Add Frame} \label{subfig:04_Xiuxiu_task5}
    \end{subfigure}
    \caption{Screenshots of Meitu tasks.}
    \label{fig:meitu_tasks}
\end{figure}

\begin{figure}
    \centering
    \begin{subfigure}{0.32\textwidth}
      \centering
      \includegraphics[width=\linewidth]{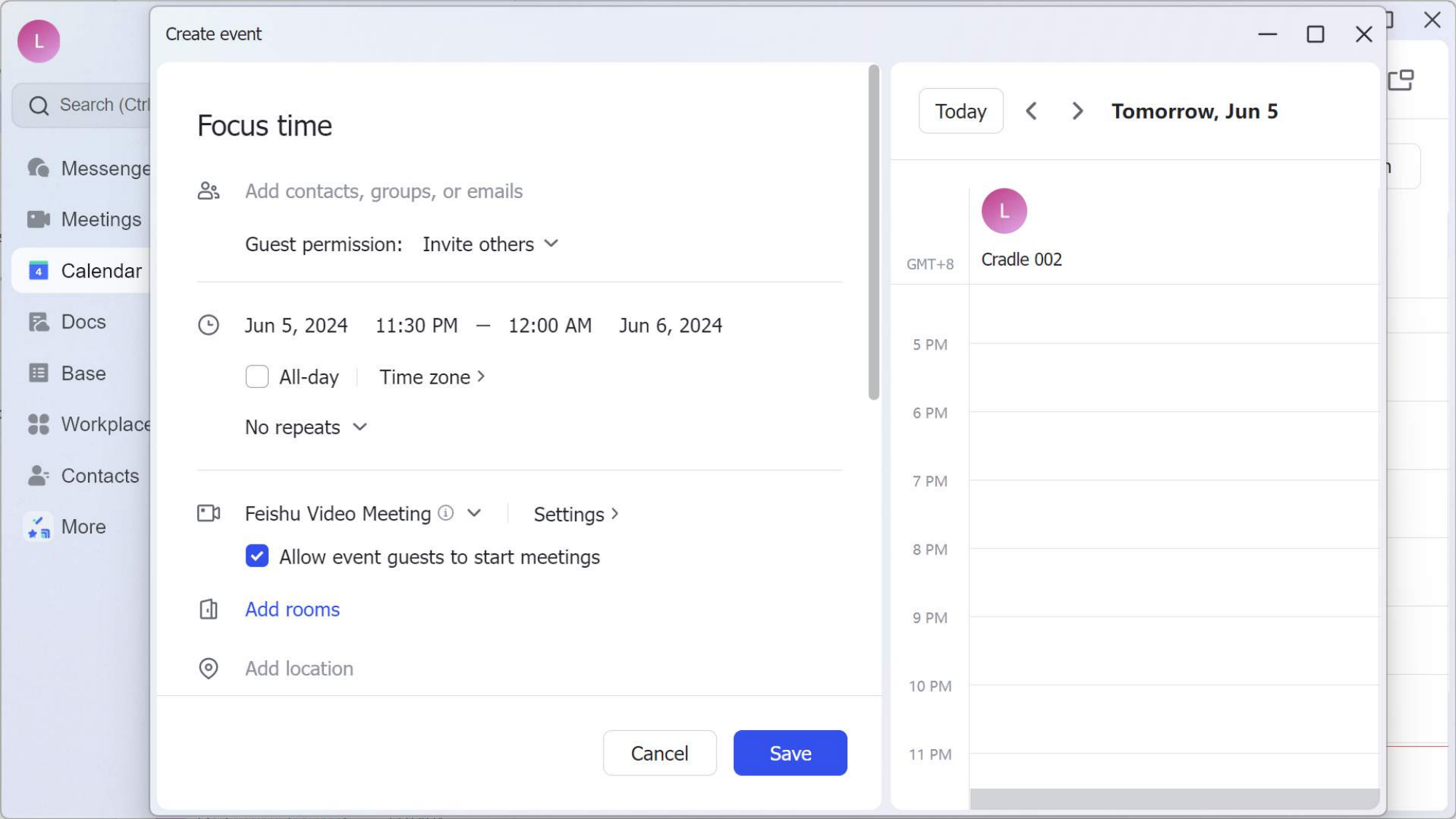}
      \caption{\scriptsize Create Appointment} \label{subfig:05_Feishu_task1}
    \end{subfigure}
    \begin{subfigure}{0.32\textwidth}
      \centering
      \includegraphics[width=\linewidth]{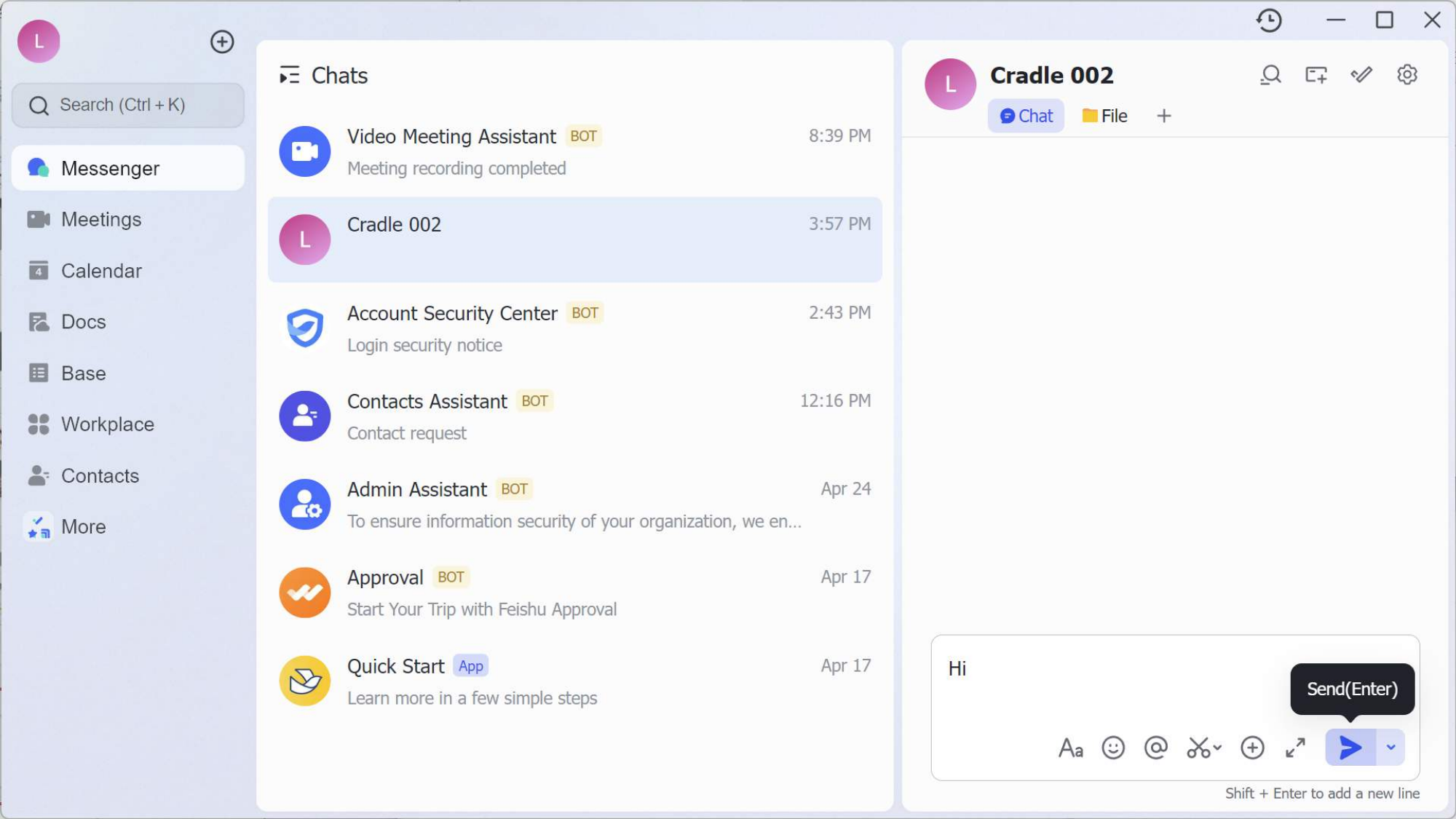}
      \caption{\scriptsize Message Contact} \label{subfig:05_Feishu_task2}
    \end{subfigure}
    \begin{subfigure}{0.32\textwidth}
      \centering
      \includegraphics[width=\linewidth]{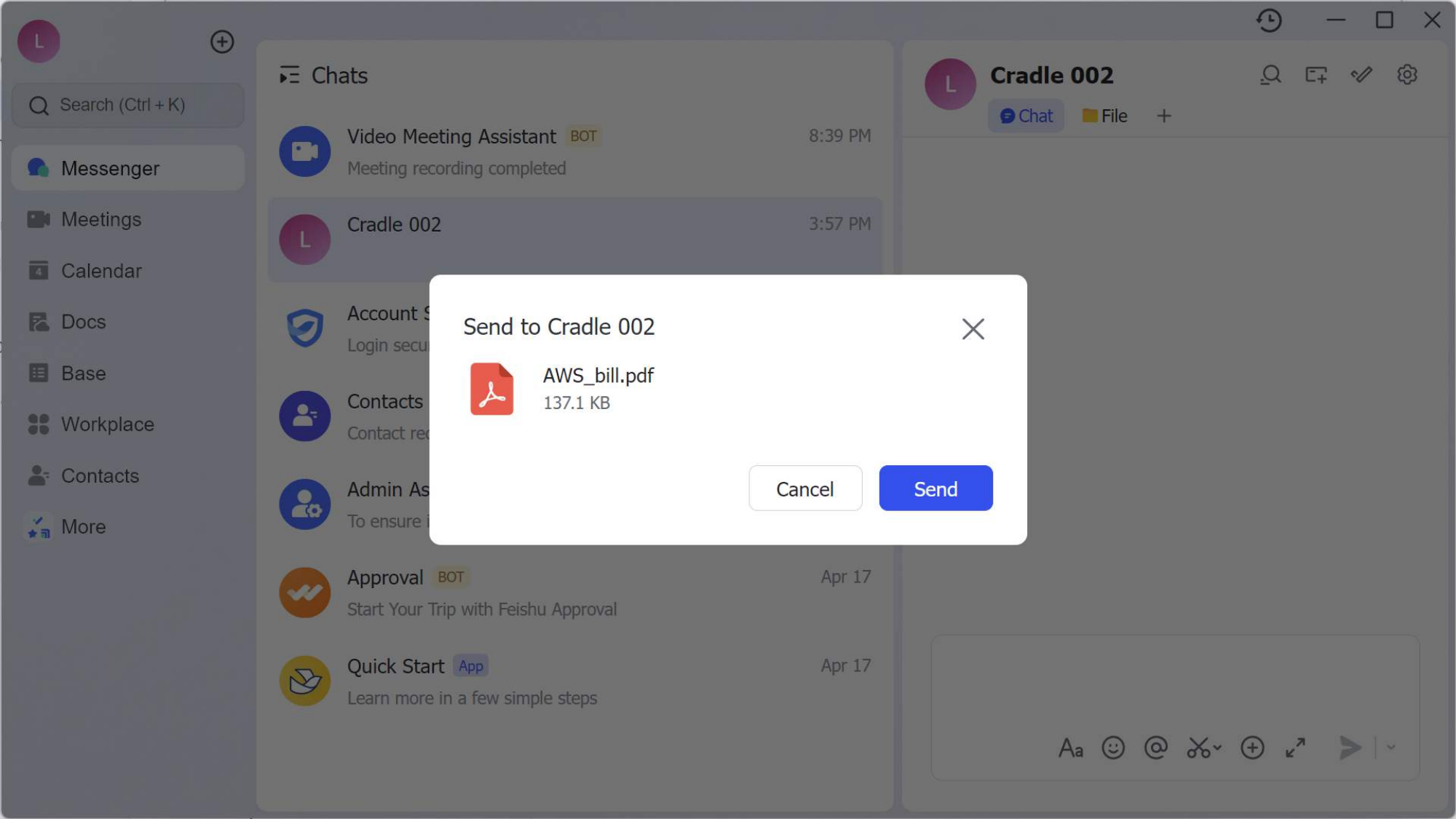}
      \caption{\scriptsize Send File} \label{subfig:05_Feishu_task3}
    \end{subfigure}
    \begin{subfigure}{0.32\textwidth}
      \centering
      \includegraphics[width=\linewidth]{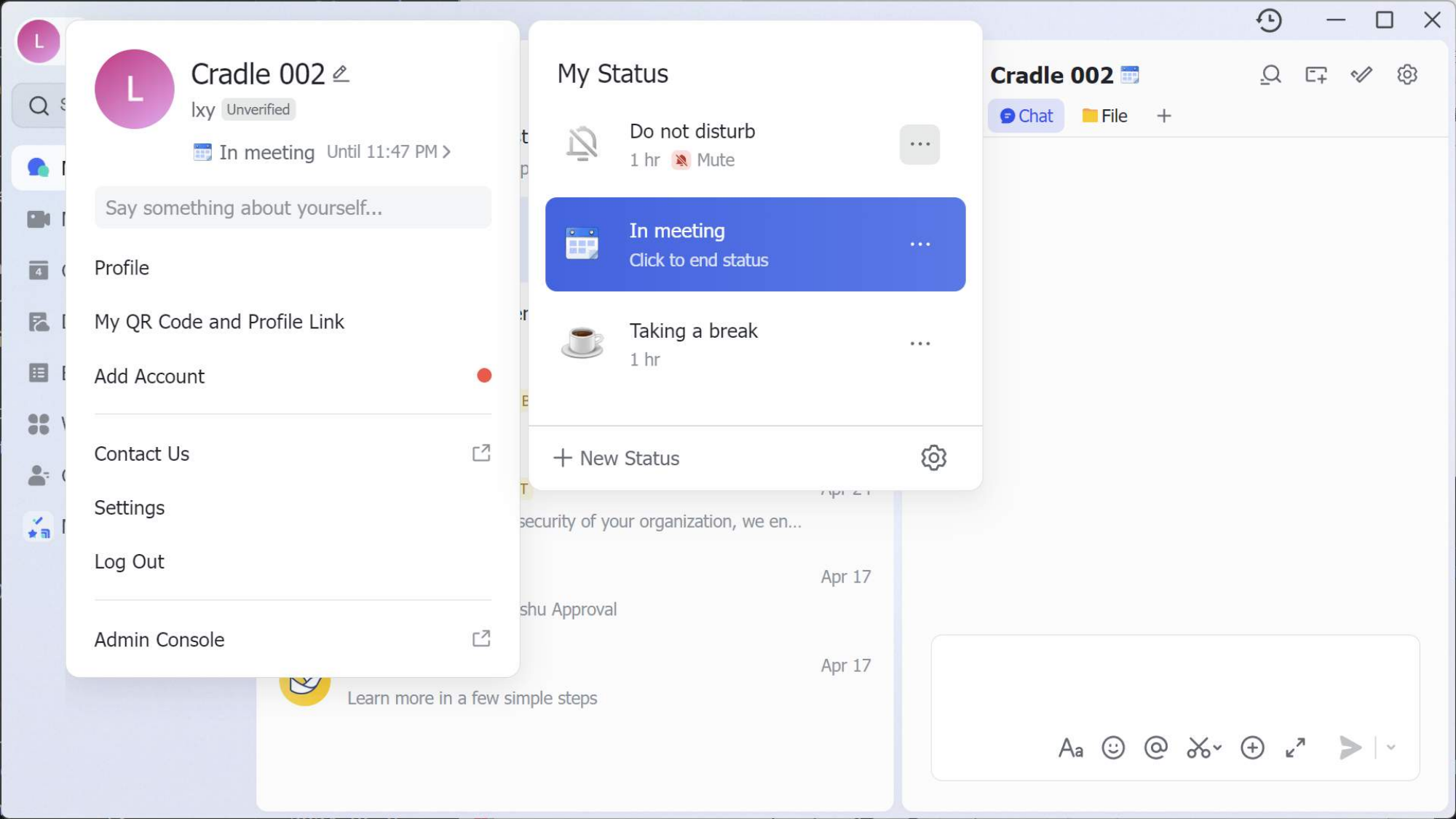}
      \caption{\scriptsize Set User Status} \label{subfig:05_Feishu_task4}
    \end{subfigure}
    \begin{subfigure}{0.32\textwidth}
      \centering
      \includegraphics[width=\linewidth]{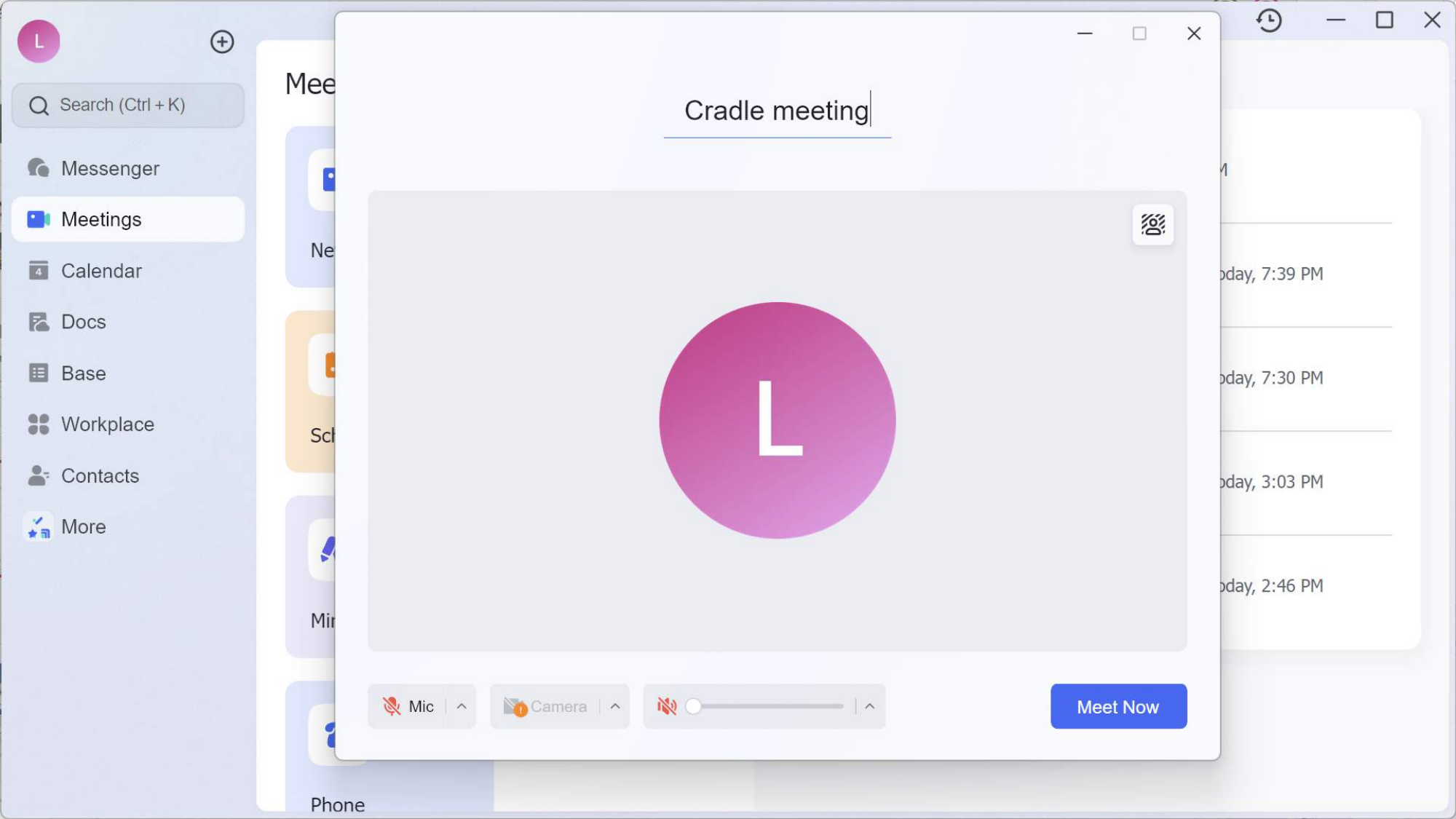}
      \caption{\scriptsize Start Video Conference} \label{subfig:05_Feishu_task5}
    \end{subfigure}
    \caption{Screenshots of Feishu tasks.}
    \label{fig:feishu_tasks}
\end{figure}

For each of the five applications, we selected a set of representative tasks for their respective domains.
For example, search, navigation, and settings tasks on Chrome; sending, searching, and deleting emails, plus changing settings on Outlook; basic video and image editing operations on CapCut and Meitu (\eg adding special effects and creating a collage); and communication and organization operations on Feishu.
Tables \ref{tab:software_task_123} and \ref{tab:software_task_45} describe in detail the 25 tasks \projname performs and analyzes on the five selected applications; also illustrated in Figures \ref{fig:chrome_tasks}, \ref{fig:outlook_tasks}, \ref{fig:capcut_tasks}, \ref{fig:meitu_tasks}, \ref{fig:feishu_tasks}, and~\ref{fig:tasks}.

\begin{table}[]
\centering
\caption{Task Descriptions for Chrome, Outlook, and CapCut. \emph{Difficulty} refers to how hard it is for our agent to accomplish the corresponding tasks. Figures~\ref{fig:chrome_tasks},~\ref{fig:outlook_tasks}, and~\ref{fig:capcut_tasks} illustrate each task (specific sub-figures marked in parenthesis in the left-most column along with task name).
}\label{tab:software_task_123}
\begin{tabularx}{\textwidth}{l X l}
\toprule
\textbf{Software} & \textbf{Description} & \textbf{Difficulty} \\ 
\midrule
\textbf{Chrome} &&\\
\midrule
Download Paper (Fig. \ref{subfig:01_Chrome_task1}) & Search for an article with a title like \emph{\{paper\_title\}} and download its PDF file. & Hard \\
Post in Twitter (Fig. \ref{subfig:01_Chrome_task2}) & Post "It's a good day." on my Twitter. & Hard\\
Open Closed Page (Fig. \ref{subfig:01_Chrome_task3})& Open the last closed page. & Easy \\
Go to Profile (Fig. \ref{subfig:01_Chrome_task4})& Find and navigate to \emph{\{person\_name\}}'s homepage on GitHub. & Medium \\
Change Mode (Fig. \ref{subfig:01_Chrome_task5})& Customize Chrome to dark mode. & Medium \\
\midrule
\textbf{Outlook} &&\\ 
\midrule
Send New E-mail (Fig. \ref{subfig:02_Outlook_task1})& Create a new e-mail to \emph{\{email\_address\}} with subject "Hello friend" and send it. & Medium \\
Empty Junk Folder (Fig. \ref{subfig:02_Outlook_task2})& Open the junk folder and delete all messages in it, if any. & Medium \\
Reply to Person (Fig. \ref{subfig:02_Outlook_task3})& Open an e-mail from \emph{\{person\_name\}} in the inbox, reply to it with "Got it. Thanks.", and click send. & Medium \\
Find Target E-mail (Fig. \ref{subfig:02_Outlook_task4})& Find the e-mail whose subject is "Urgent meeting" and open it. & Easy \\
Setup Forwarding (Fig. \ref{subfig:02_Outlook_task5})& Set up email forwarding for every email received to go to \emph{\{email\_address\}}. &  Medium \\
\midrule
\textbf{CapCut} &&\\ 
\midrule
Create Media Project (Fig. \ref{subfig:03_CapCut_task1})& Create a new project, then import \emph{\{video\_file\_name\}} to the media, click the "Audio" button to add music to the timeline, and finally export the video. & Hard \\
Add Transition (Fig. \ref{subfig:03_CapCut_task2})& Open the first existing project. Switch to Transitions panel. Drag a transition effect between the two videos, and then export the video. & Medium \\
Crop by Timestamp (Fig. \ref{subfig:03_CapCut_task3})& Delete the video frames after five seconds and then before one second in this video, and then export the video. & Medium \\
Add Sticker (Fig. \ref{subfig:03_CapCut_task4})& Open the first existing project. Switch to Stickers panel. Drag a sticker of a person's face to the video, and then export the video. &  Hard \\
Crop by Content (Fig. \ref{subfig:03_CapCut_task5})& Crop the video when the ball enters the goal, and then export the video. & Very hard \\
\bottomrule
\end{tabularx}
\end{table}

\begin{table}[]
\centering
\caption{Task Descriptions for: Meitu, and Feishu. \emph{Difficulty} refers to how hard it is for our agent to accomplish the corresponding tasks. Figures~\ref{fig:meitu_tasks}, and~\ref{fig:feishu_tasks} illustrate each task (specific sub-figures marked in parenthesis in the left-most column along with task name).
}\label{tab:software_task_45}
\begin{tabularx}{\textwidth}{l X l}
\toprule
\textbf{Software} & \textbf{Description} & \textbf{Difficulty} \\ 
\midrule
\textbf{Meitu} &&\\ 
\midrule
Apply Filter (Fig. \ref{subfig:04_Xiuxiu_task1})& Apply a filter from Meitu to \emph{\{picture\_file\_name\}} and save the project. & Easy \\
Cutout (Fig. \ref{subfig:04_Xiuxiu_task2})& Cutout a person from \emph{\{picture\_file\_name\}} and save the project. & Easy \\
Add Sticker (Fig. \ref{subfig:04_Xiuxiu_task3}) & Add a flower sticker to \emph{\{picture\_file\_name\}} and save the picture. & Middle \\
Create Collage (Fig. \ref{subfig:04_Xiuxiu_task4}) & Make a collage using 3 pictures and save the project. & Hard \\
Add Frame (Fig. \ref{subfig:04_Xiuxiu_task5}) & Add a circle-shaped frame to \emph{\{picture\_file\_name\}} and save the picture. & Hard \\
\midrule
\textbf{Feishu} &&\\ 
\midrule
Create Appointment (Fig. \ref{subfig:05_Feishu_task1})& Create a new appointment in my calendar anytime later today with title "Focus time". & Hard \\
Message Contact (Fig. \ref{subfig:05_Feishu_task2})& Please send a "Hi" chat message to \emph{\{contact\_name\}}.  & Easy \\
Send File (Fig. \ref{subfig:05_Feishu_task3}) & Send the AWS bill file at \emph{\{pdf\_path\}} in a chat with \emph{\{contact\_name\}}. & Hard \\
Set User Status (Fig. \ref{subfig:05_Feishu_task4}) & Open the user profile menu and set my status to "In meeting". & Medium \\
Start Video Conference (Fig. \ref{subfig:05_Feishu_task5}) & Create a new meeting and meet now. & Easy \\
\bottomrule
\end{tabularx}
\end{table}

It is worth noting that we add a \emph{special} task on CapCut to demonstrate the agent's ability for tool use. In this task, a pre-defined skill uses GPT-4o as a tool for video understanding capabilities. The skill can be selected to answer content-based questions about a video (\eg "when the ball enters the goal") and the response be used during task completion. This task is illustrated in detail in Figure \ref{fig:Capcut_data_flow}.

\subsection{Quantitative Evaluation}

\begin{figure}[ht]
  \centering
    \includegraphics[width=\textwidth]{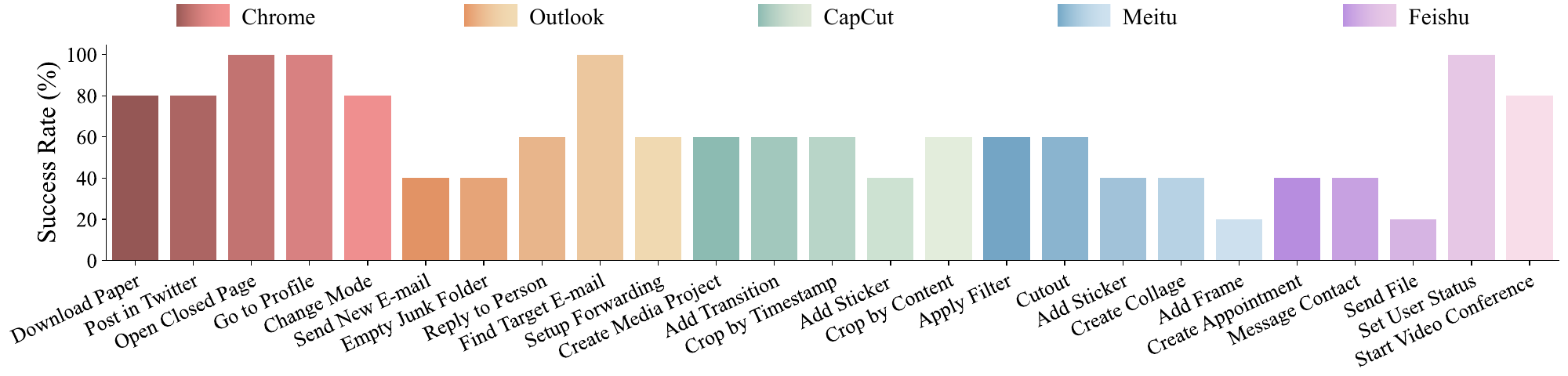}
  \caption{Success rates for tasks in software applications}
  \label{fig:software_barchart_}
\end{figure}

\begin{table}[h]
\caption{Application Software results. \emph{Success Rate} determines the ratio of successful completions over five runs. \emph{Average Steps} refers to the number of actions the agent takes to fulfil a task, if successful. \emph{Expected Steps} represents the number of steps as estimated by humans performing the task. \emph{Efficiency} represents the ratio between the expected number of steps and the total number of steps taken by the agent.}
\label{tab:software_result_table}
\centering
\begin{tabular}{l|llll}
\toprule
\textbf{Software} & Success Rate & Average Steps & Expected Steps & Efficiency \\
\midrule
\textbf{Chrome} & 88\% & 8.74 & 4.20 & 48.05\% \\
\midrule
Download Paper & 80\% &  16.00 $\pm\ 5.52$ & 6 & 37.50\%\\
Post in Twitter  & 80\% & 11.75 $\pm\ 5.26$ &  7 & 61.14\% \\
Open Closed Page  & 100\% & 1.00 $\pm\ 0$  & 3  & 300.00\% \\
Go to Profile  & 100\% & 4.00 $\pm\ 0.63$ &1  & 25.00\%\\
Change Mode  & 80\% &  11.25 $\pm\ 4.71$ & 4   & 35.56\%\\
\midrule
\textbf{Outlook} & 60\% & 8.25 & 4 & 48.48\%\\
\midrule
Send New E-mail & 40\% & 11.00 $\pm\ 4$ & 5  & 45.45\%\\
Empty Junk Folder  & 40\% & 8.50 $\pm\ 3.50$ & 3  & 35.29\%\\
Reply to Person  & 60\% & 8.33 $\pm\ 4.71$ & 4  & 48.02\%  \\
Find Target E-mail  &  100\% & 1.40 $\pm\ 0.80$ & 1  & 71.43\%\ \\
Setup forwarding  &  60\% & 12.00 $\pm\ 4.90$ & 7  & 58.33\%\ \\
\midrule
\textbf{CapCut} & 56\% & 10.87 & 4.80 & 44.16\%\\
\midrule
Create Media Project & 60\% & 13.67 $\pm\ 5.25$ & 7  & 51.20\% \\
Add transition  & 60\% & 10.67 $\pm\ 4.03$ & 4 & 37.49\%\\
Crop by Timestamp  & 60\% & 11.00 $\pm\ 5.66$ & 5 & 45.45\%\\
Add Sticker  &  40\% & 12.00 $\pm\ 8.00$ &4 & 33.33\% \\
Crop by Content  & 60\% & 7.00 $\pm\ 1.41$ & 4 & 57.14\%\\
\midrule
\textbf{Meitu} &44\% & 12.5 & 8.00 & 64\% \\
\midrule
Apply Filter  & 60\% & 14.67 $\pm\ 2.36$ & 7 & 47.72\%\\
Cutout  & 60\% & 9.33 $\pm\ 1.89$ & 5 & 53.59\%\\
Add Sticker  &  40\%& 9.50 $\pm\ 0.50$ & 8 & 84.21\%\\
Create Collage  &  40\%& 16.00 $\pm\ 2.00$ & 12 & 75.00\%\\
Add Frame &  20\% & 13.00 $\pm\ 0.00$ & 7 & 53.85\%\\
\midrule
\textbf{Feishu} & 56\% & 8.82 & 4 & 46.07\%\\
\midrule
Create Appointment  & 40\%& 8.00 $\pm\ 1.00$ &4 & 50.00\% \\
Message Contact  & 40\%& 6.00 $\pm\ 1.00$ &3 & 50.00\% \\
Send file  & 20\%& 11.00 $\pm\ 0.00$ &7 & 63.64\% \\
Set User Status  & 100\%& 14.60 $\pm\ 7.50$ &3 & 20.55\% \\
Start Video Conference  &  80\%& 4.50 $\pm\ 2.60$ &3 & 46.15\% \\
\bottomrule
\end{tabular}
\end{table}

We calculate \projname's performance over the 25 tasks in the applications set. Each task is executed five times and performance is measured in three metrics: success rate, average number of steps taken by the agent (and variance over the five runs), and efficiency. \emph{Efficiency} is defined as the ratio between the expected number of steps in a given task and the total number of steps taken by the agent. The expected number of steps per task is calculated by having humans perform each task.

Table~\ref{tab:software_result_table} and Figure \ref{fig:software_barchart_} show the details of the evaluation. \projname presents overall good performance over the diverse tasks and applications (compared to Expected Steps, \projname achieves an overall efficiency of 50\%). However, performance for certain tasks can vary considerably due to different factors. The main reason for the higher number of task step during agent execution is the frequent incorrect positioning decisions for the mouse, \ie the backbone model chooses a position of bounding box tag that does not correspond to the UI item described in the model reasoning. We discuss examples of task-specific issues in Sections~\ref{appx:software:cases} and ~\ref{appx:software:limitations} below.

It is worth noting that in Chrome's task 3 ("Open the last closed page"), \projname knows how to use the shortcut key directly, calling the key\_press skill directly with the correct keyboard shortcut: 
\emph{`Ctrl + Shift + T'}, whereas humans typically do not know this.

To further evaluate the performance of \projname in diverse software applications scenarios, we provide quantitative results over OSWorld, a new contemporaneous benchmark with similar characteristics to our settings. More details in Appendix~\ref{appx:osworld} and overview of the results in Table~\ref{tab:exp_OSWorld_full}.

\subsection{Implementation Details} 
\label{sub:software_implementation}

The implementation of \projname targeting all five software applications follows the GCC setting and framework modules (which include Information Gathering, Self-Reflection, Task Inference, Skill Curation, Action Planning, and Action Execution). Implementation details of the overall framework are described in Appendix~\ref{appx:general_implementation}. Therefore, here we emphasize any application-specific differences or customization.

To apply \projname to the target application set described in this appendix, we start with base common prompts, and customize those prompts for specific modules, if necessary, to handle application-specific characteristics. For example, for CapCut we add few-shot examples for Self-Reflection, to let it properly perform success detection, as the application UI by itself is non-standard and sometimes provides little post-action feedback to users, making it harder for the backend model to determine action success.

\textbf{Information Gathering.} Noticeably, GPT-4o presents the same limitations in both spatial reasoning (\eg confusing up/down, left/right) and image understanding identifying specific UI items or the state of the forefront GUI, across all applications. 

\begin{figure}
    \centering
    
    \begin{subfigure}{0.325\textwidth}
      \centering
      \includegraphics[width=\linewidth]{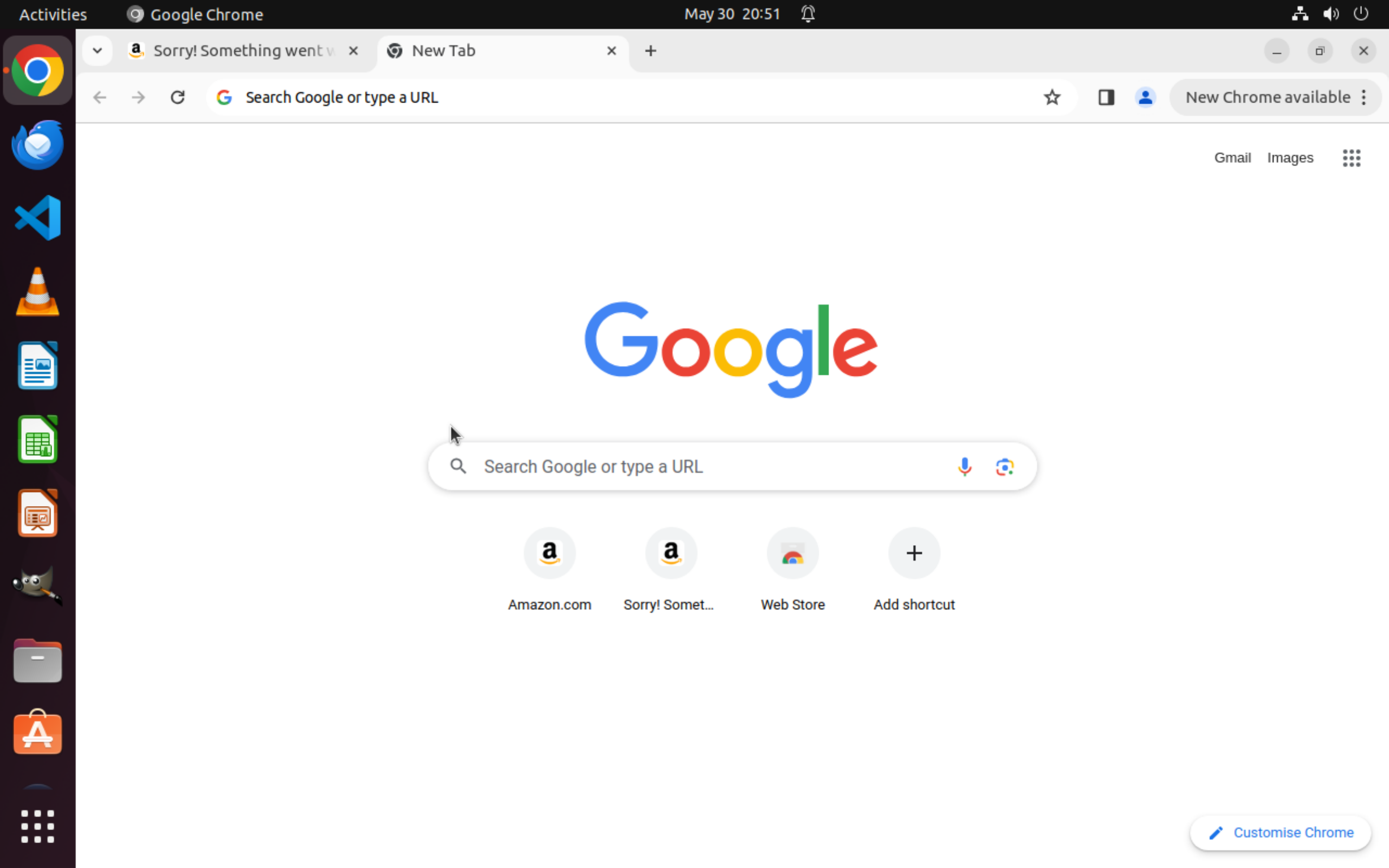}
      \caption{Chrome window} 
      \label{subfig:OSWorld_step9}
    \end{subfigure}
    \begin{subfigure}{0.325\textwidth}
      \centering
      \includegraphics[width=0.98\linewidth]{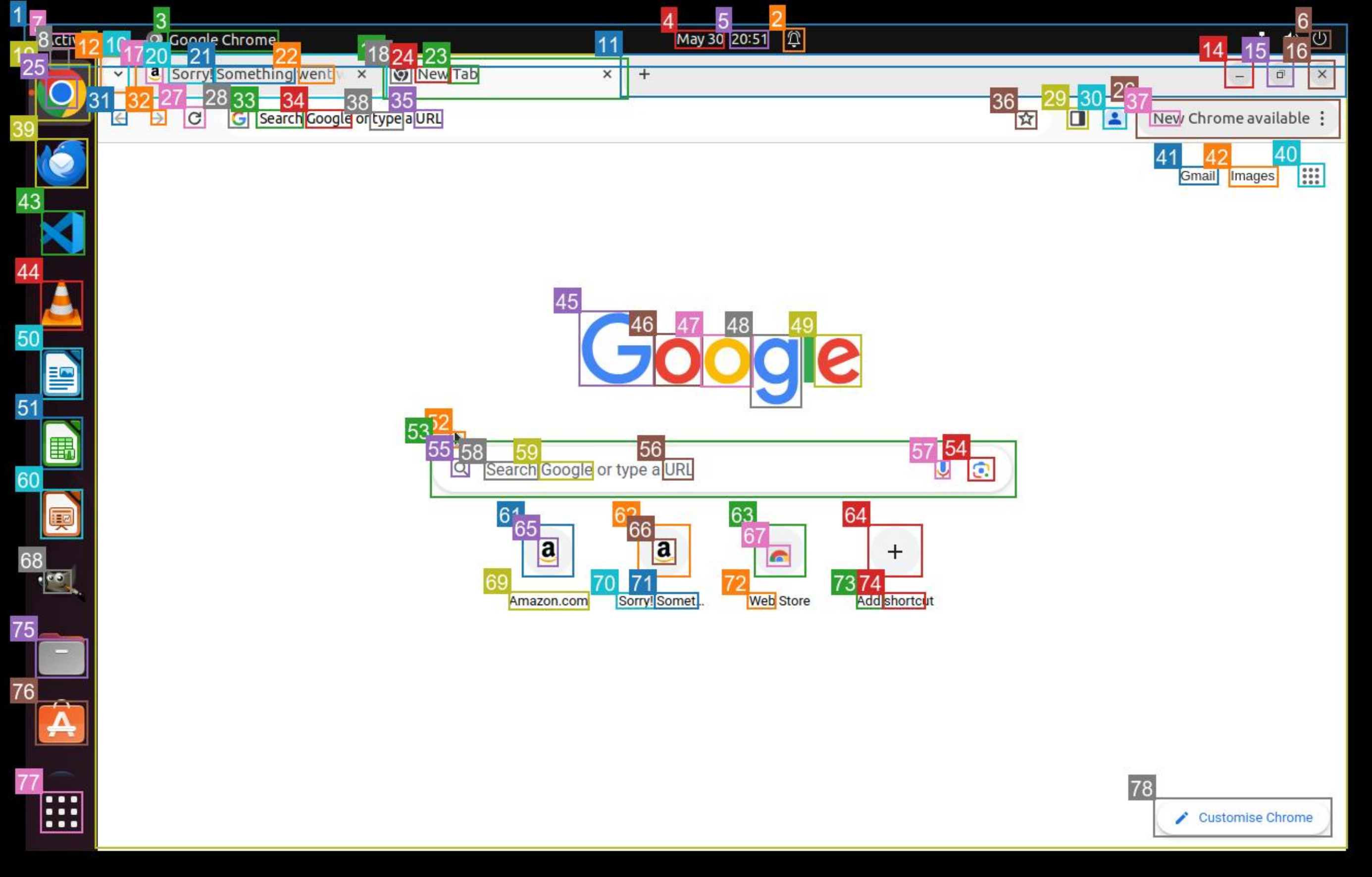}
      \caption{\projname SAM2SOM output} 
      \label{subfig:OSWorld_step9_sam2som}
    \end{subfigure}
        \begin{subfigure}{0.325\textwidth}
      \centering
      \includegraphics[width=\linewidth]{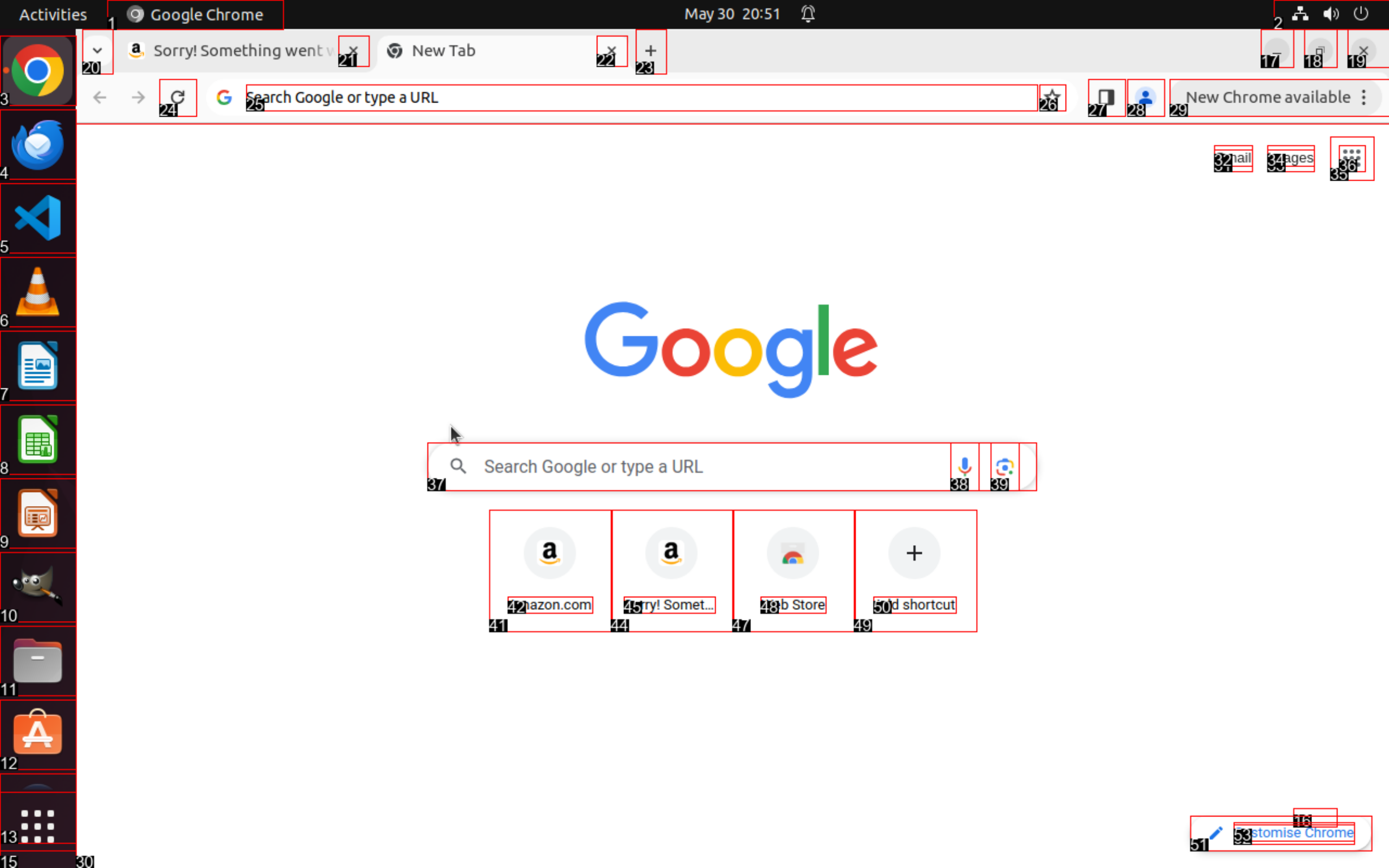}
      \caption{\emph{OSWorld} SOM image output} 
      \label{subfig:OSWorld_step9_som}
    \end{subfigure}

    \begin{subfigure}{0.325\textwidth}
      \centering
      \includegraphics[width=\linewidth]{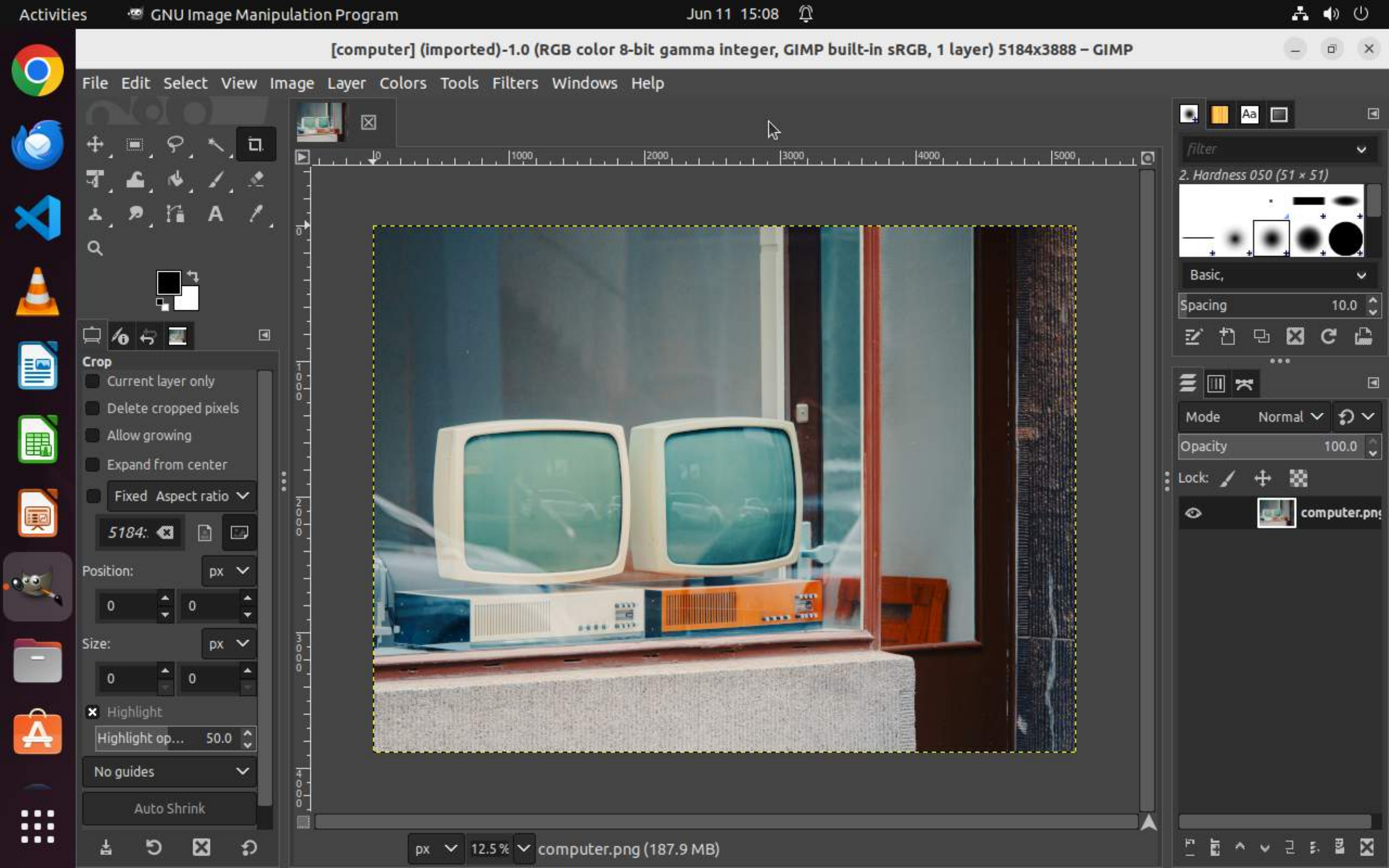}
      \caption{GIMP window} 
      \label{subfig:OSWorld_GIMP_base}
    \end{subfigure}
    \begin{subfigure}{0.325\textwidth}
      \centering
      \includegraphics[width=0.98\linewidth]{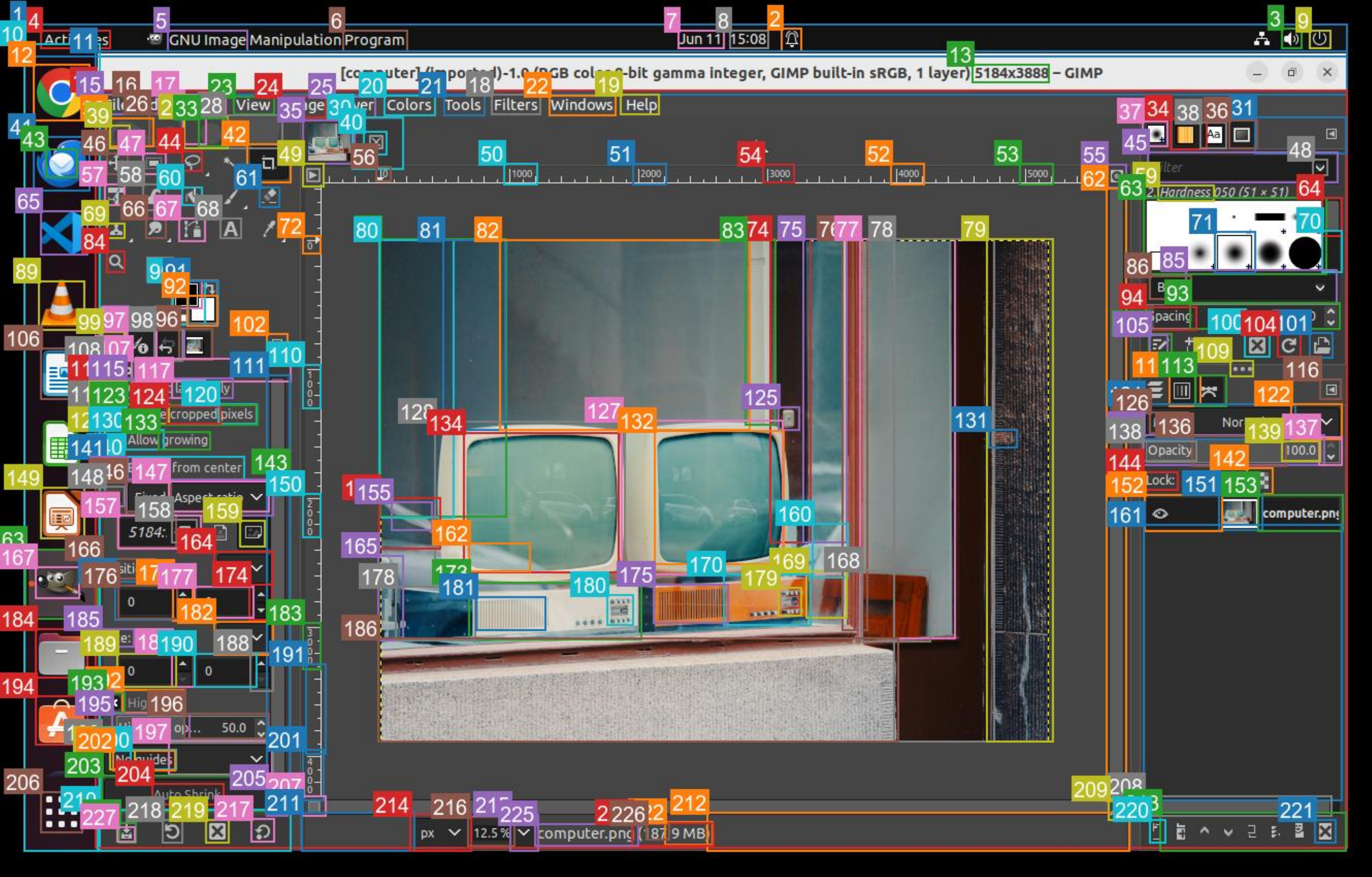}
      \caption{\projname SAM2SOM output} 
      \label{subfig:OSWorld_GIMP_sam2som}
    \end{subfigure}
    \begin{subfigure}{0.325\textwidth}
      \centering
      \includegraphics[width=\linewidth]{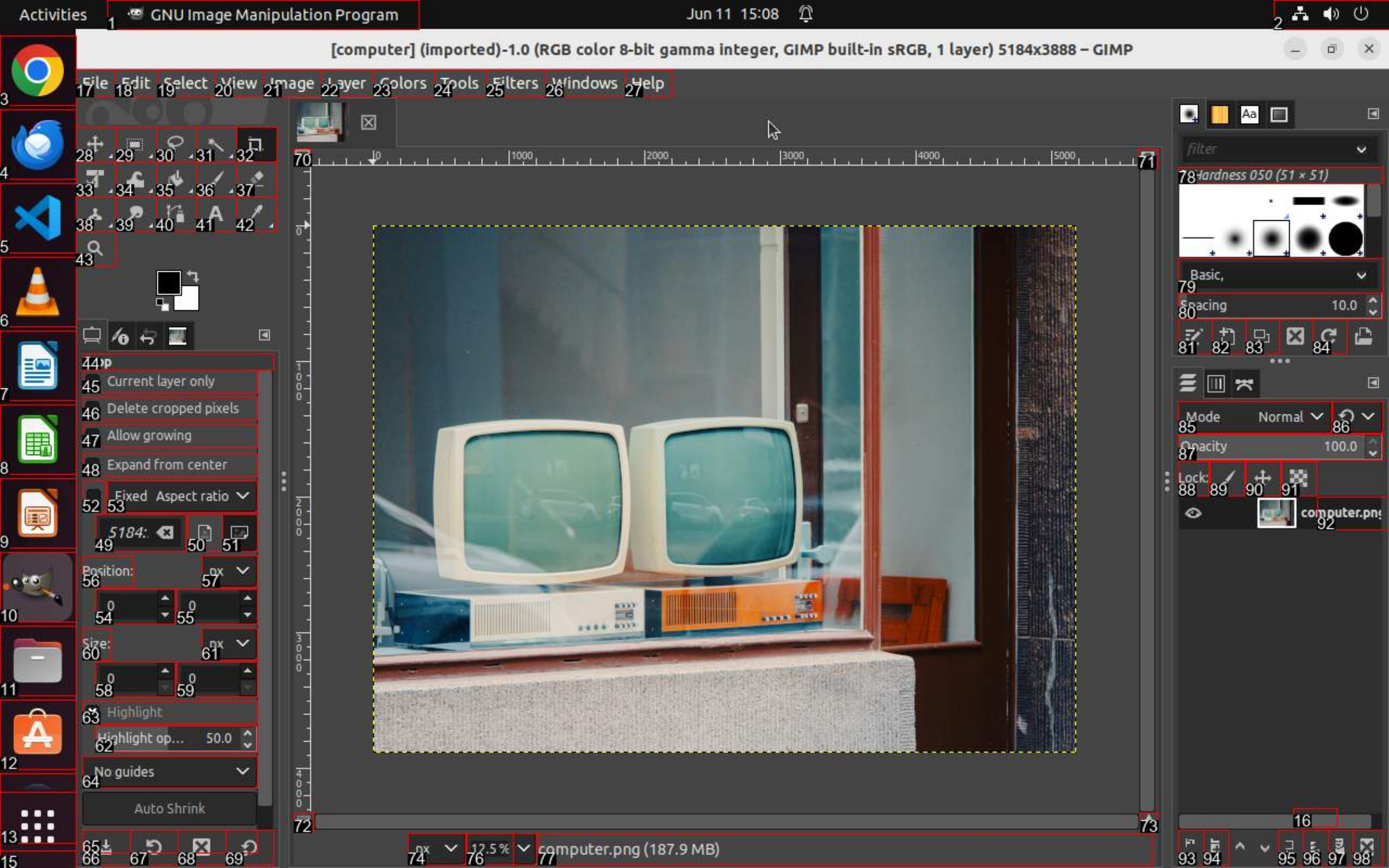}
      \caption{\emph{OSWorld} SOM image output} 
      \label{subfig:OSWorld_GIMP_som}
    \end{subfigure}
    
    \caption{Comparison of \projname's visual-only SAM2SOM and \emph{OSWorld}'s API-based SOM image results. Chrome: 78 vs. 53 bounding boxes; GIMP: 227 vs. 98 bounding boxes.}
    \label{fig:sam2som_comparison}
\end{figure}

To help mitigate such issues, we perform augmentation on the captured screenshots similarly to the Set-of-Mark (SoM) approach~\cite{yang2023setofmark}, by only utilizing SAM~\cite{kirillov2023segment} to generate potential UI items bounding boxes and assign them numerical tags. Our SoM-like augmentation \emph{differs} from recent agent-related work (\eg~\citep{zhang2024ufo, xie2024osworld}), which use OS-specific APIs to draw ground-truth bounding boxes for interactable elements (plus UI structure info, like types and element tree) to the results, while \projname relies only on image input and the segmentation output as augmentation. To make this distinction explicit, we call our augmentation approach SAM2SOM~\footnote{ We do not claim the method itself as a core contribution. SAM2SOM is used to illustrate a possible extra capability of the backend model, as mitigation for current spatial reasoning issues.}. Figure~\ref{fig:sam2som_comparison} illustrates the difference. While our approach produces many more potential bounding boxes, it is more general by relying only on a screenshot (or video frame).

\begin{figure}
    \centering
    \begin{subfigure}{0.495\textwidth}
      \centering
      \includegraphics[width=\linewidth]{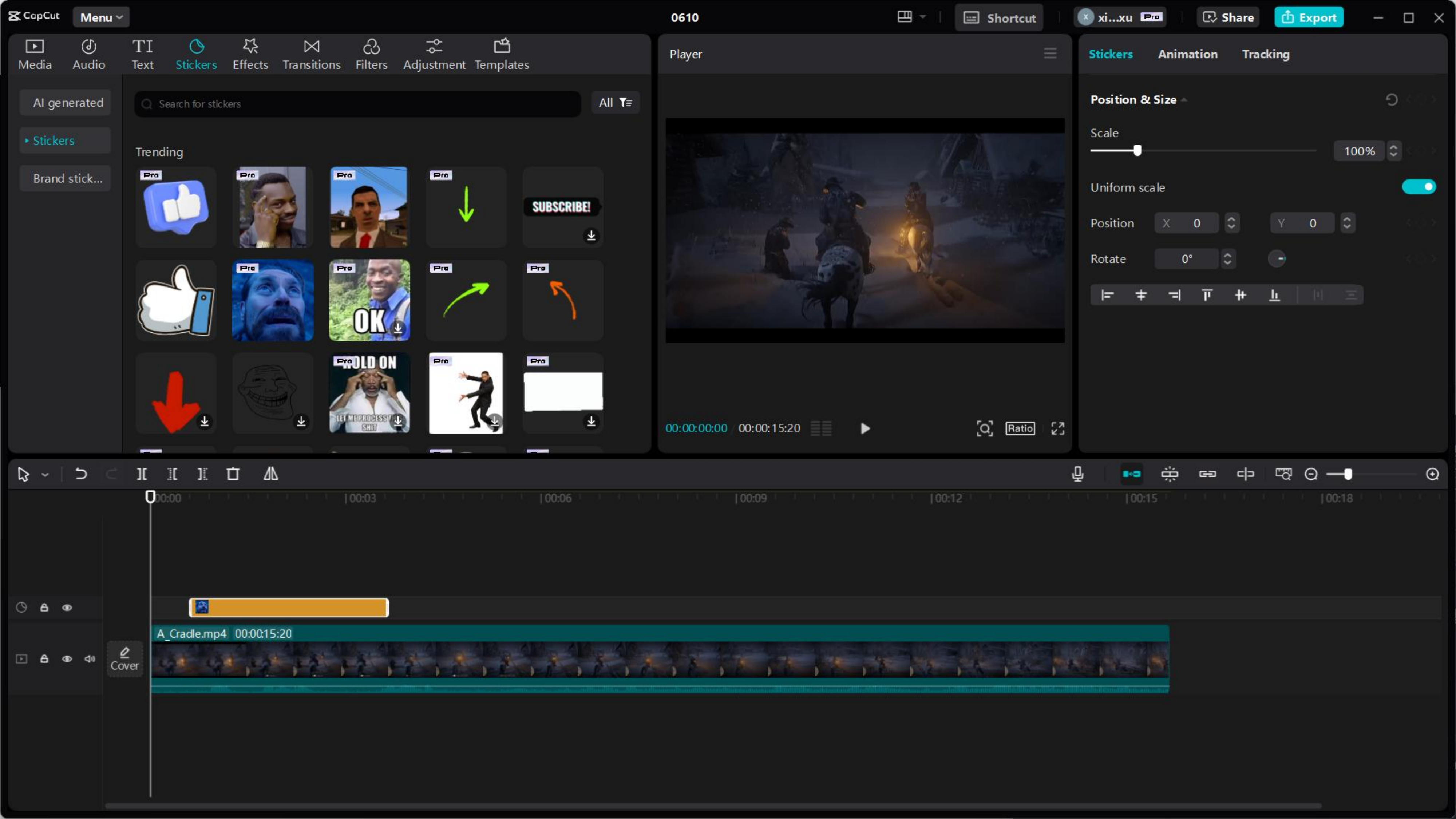}
      \caption{ \emph{CapCut}: Base image} \label{subfig:CapCut_base}
    \end{subfigure}
    \begin{subfigure}{0.495\textwidth}
      \centering
      \includegraphics[width=\linewidth]{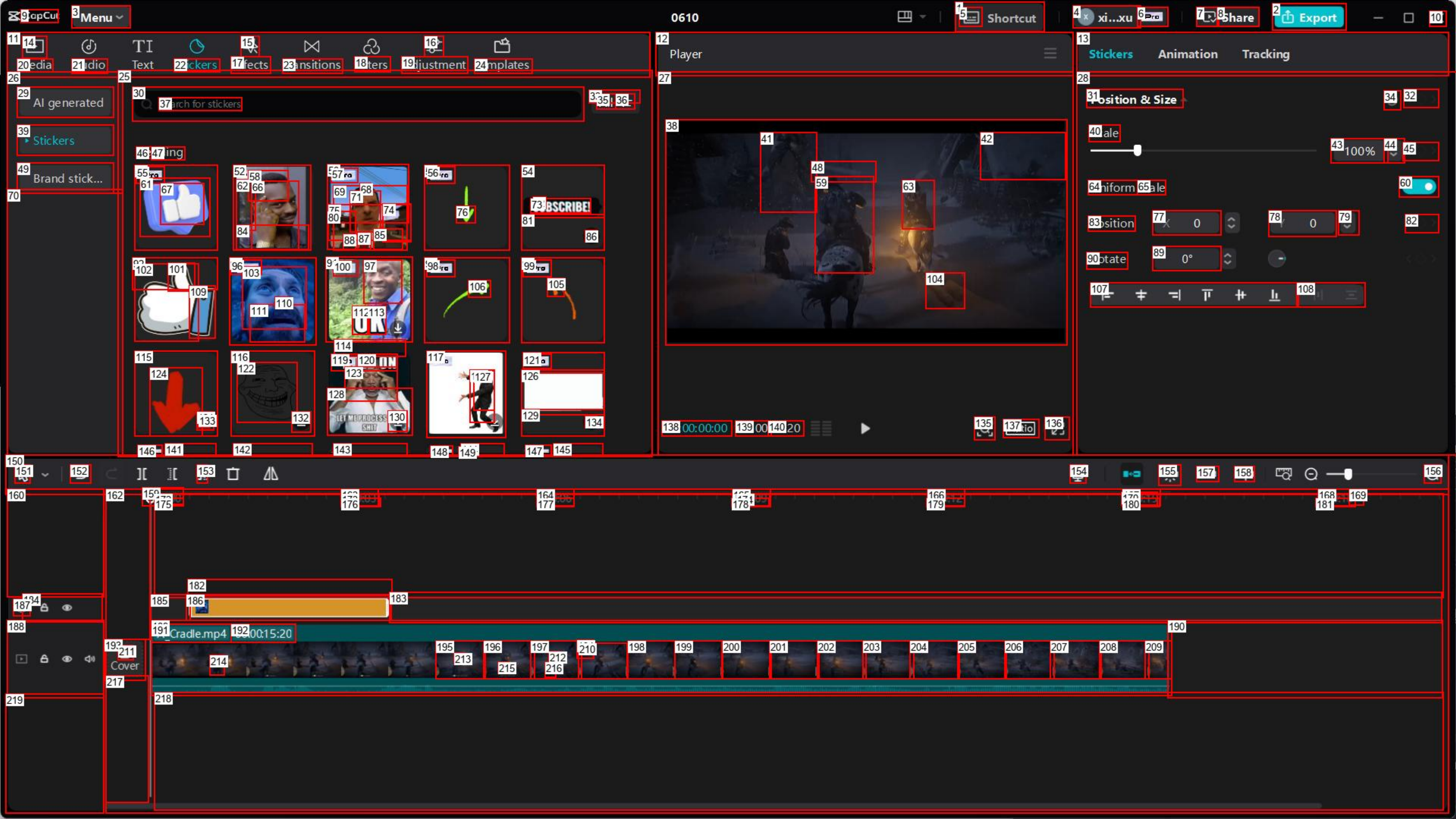}
      \caption{ \emph{CapCut}: Base image w/ SAM2SOM} \label{subfig:CapCut_som}
    \end{subfigure}

    \begin{subfigure}{0.495\textwidth}
      \centering
      \includegraphics[width=\linewidth]{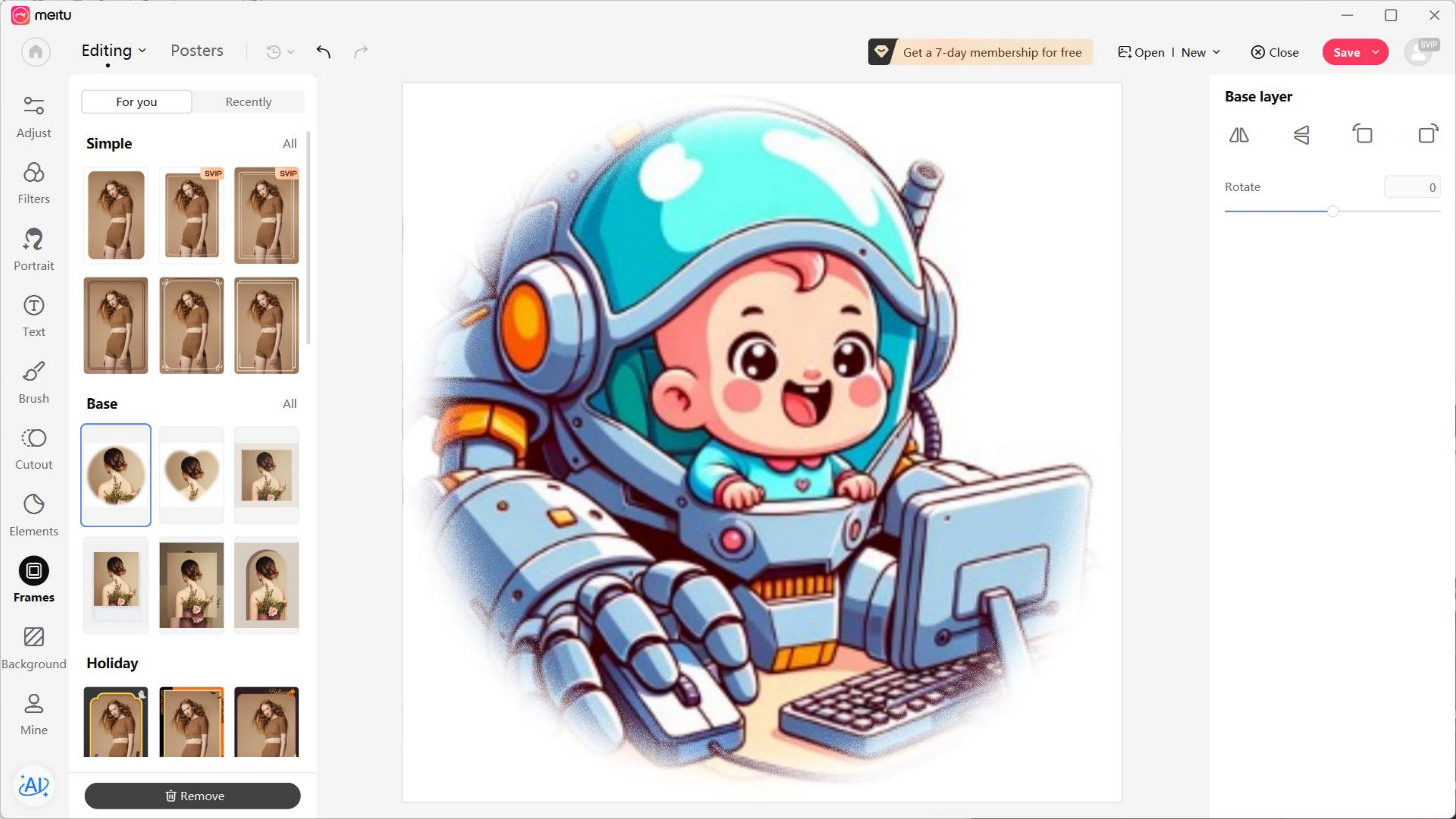}
      \caption{ \emph{Meitu}: Base image} \label{subfig:Xiuxiu_base}
    \end{subfigure}
    \begin{subfigure}{0.495\textwidth}
      \centering
      \includegraphics[width=\linewidth]{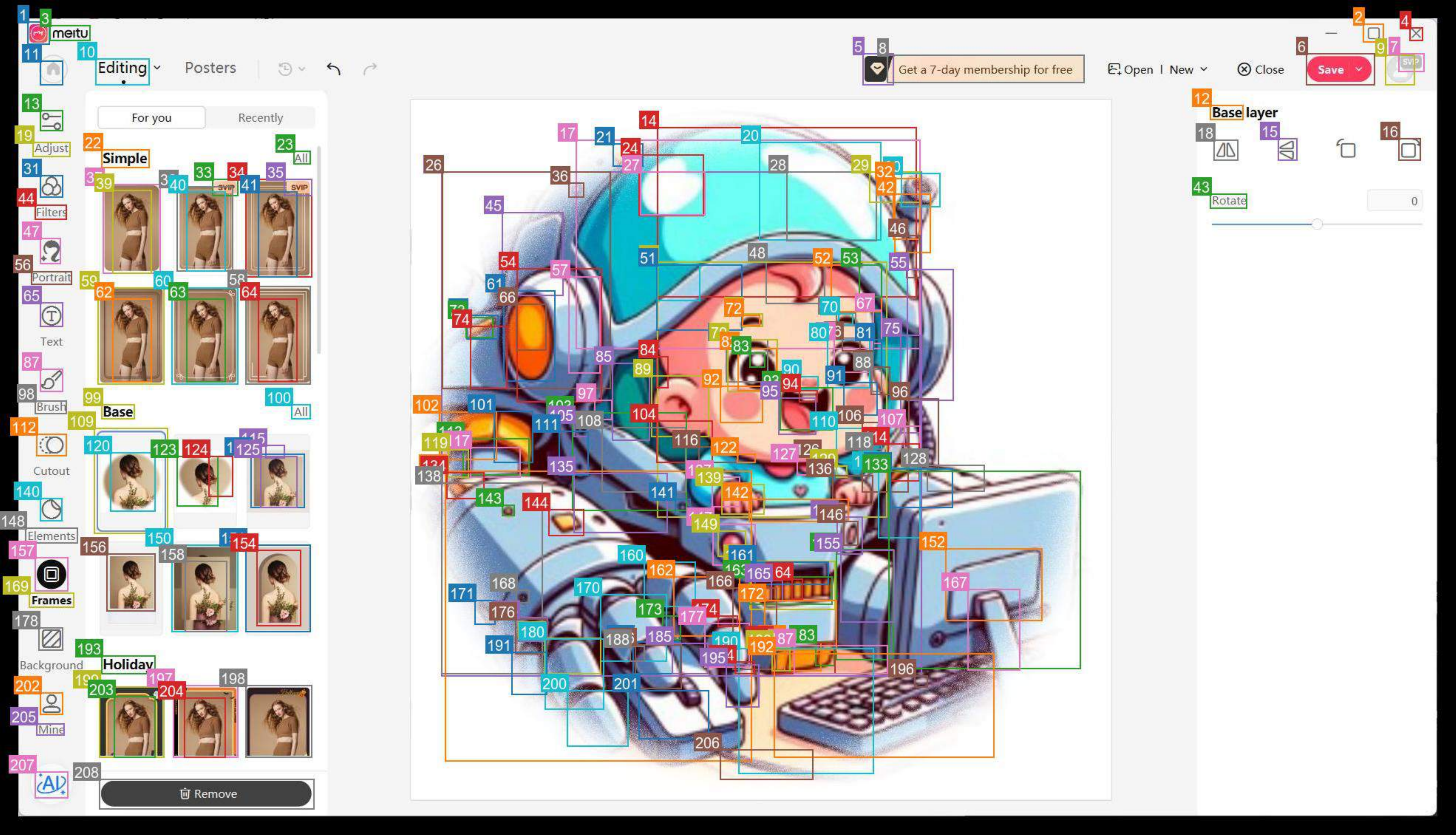}
      \caption{ \emph{Meitu}: Base image w/ SAM2SOM} 
      \label{subfig:Xiuxiu_som}
    \end{subfigure}
    
    \caption{Image examples of the two SAM2SOM augmentation styles. As CapCut's UI (top row) has very dark background, we utilize single-color borders with IDs in black text over white background, placed within the bounding box area. Other application software and OSWorld use the "standard" SAM2SOM multi-color style, as shown for Meitu (bottom row).}
    \label{fig:software_augmentation}
\end{figure}

To ensure all bounding box labels are consistently positioned, \projname's SAM2SOM implements two rendering styles, as shown in Figure~\ref{fig:software_augmentation} first and second rows. In the \emph{standard} style, we pad the SAM2SOM-enhanced image when showing the label IDs in the upper left corner of the bounding boxes (to prevent labels from hiding the contents of small areas), so no numerical label ID is drawn outside the image area). In the \emph{uniform} style, all bounding boxes utilize single-color borders with labels in black text over white background, placed within
the bounding box area (top left corner).

\begin{figure}
    \centering
    \begin{subfigure}{0.495\textwidth}
      \centering
      \includegraphics[width=\linewidth]{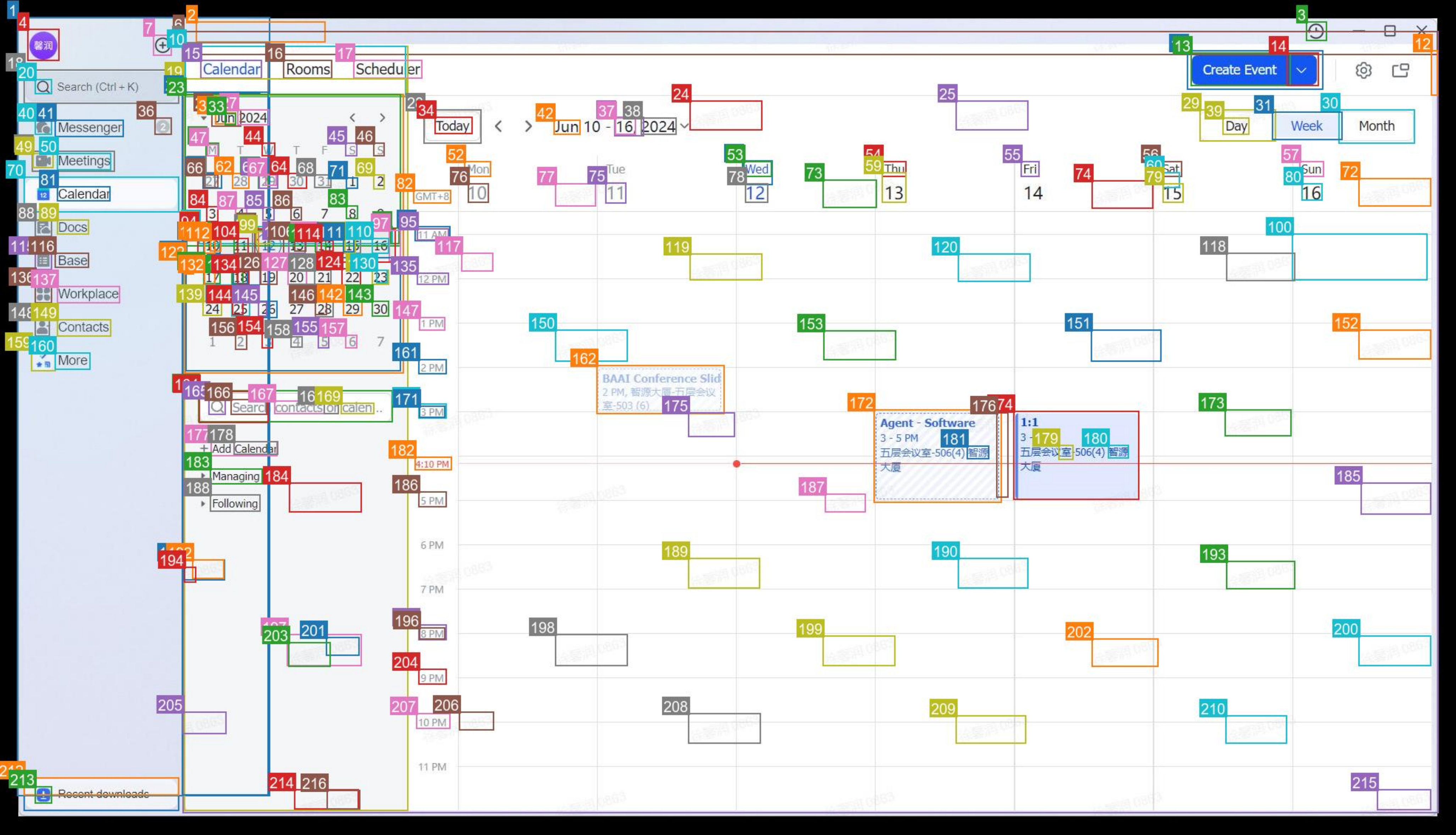}
      \caption{SAM2SOM image w/ watermarks}
      \label{subfig:Feishu_watermark}
    \end{subfigure}
    \begin{subfigure}{0.495\textwidth}
      \centering
      \includegraphics[width=\linewidth]{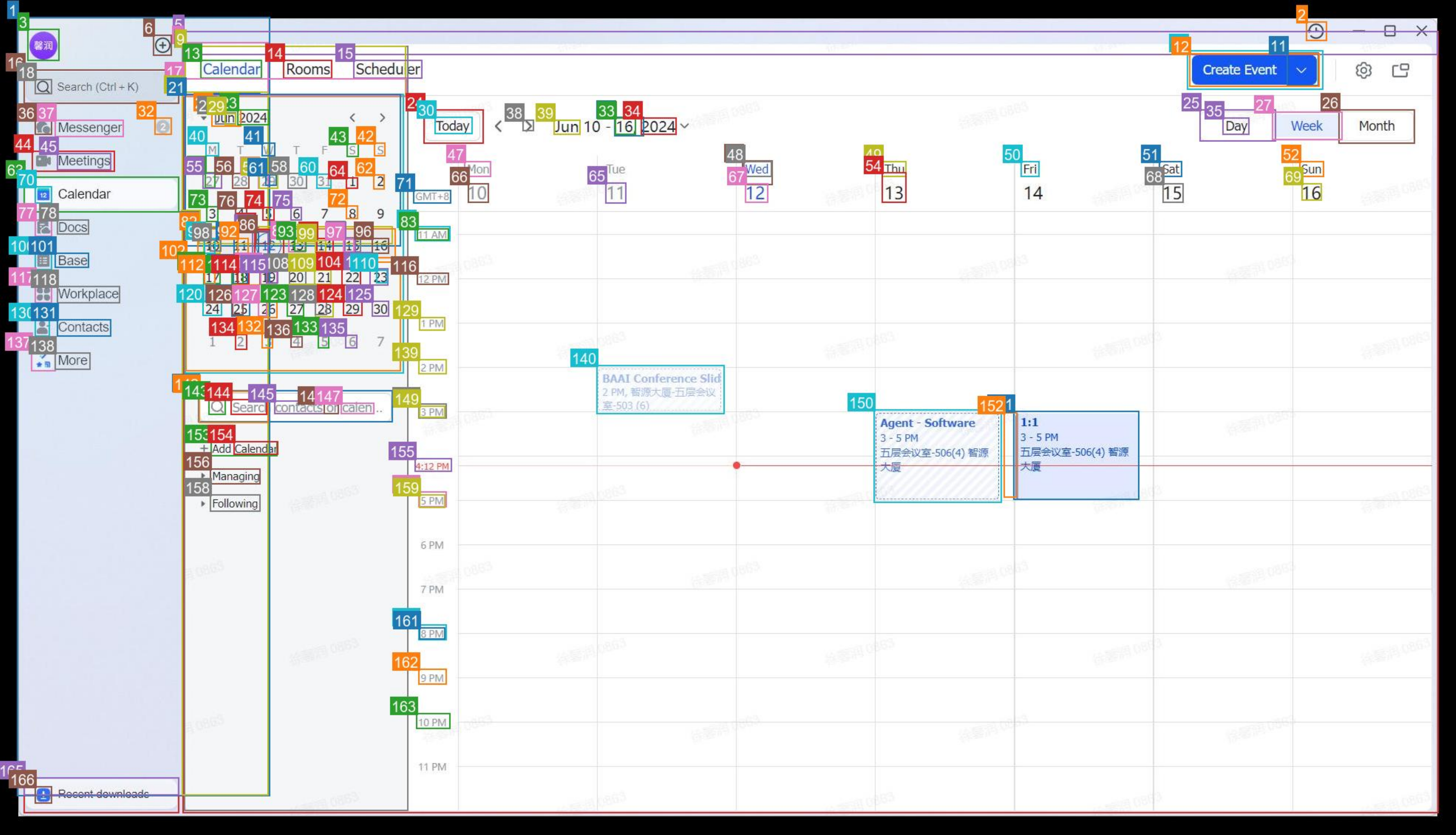}
      \caption{SAM2SOM image w/o watermarks} \label{subfig:Feishu_watermark_remove}
    \end{subfigure}
    
    \caption{Examples of filtering watermark in Feishu. The number of labels is greatly reduced from 216 to 166.}
    \label{fig:filtering_watermark}
\end{figure}

Moreover, in specific situations we may still need to refine SAM2SOM's output further. For example, in the Feishu case, we observe that watermarks generated by the software affect the segmentation negatively, complicating GPT-4o's selection of the correct bounding boxes to interact with. Therefore, we implement a simple filtering method for such watermarks. This filter is enabled only in the Feishu benchmark and, as shown in Figure \ref{fig:filtering_watermark}, can greatly reduce the number of unnecessary bounding boxes (from 216 to 166, in this example).

\begin{wrapfigure}{r}{0.5\textwidth}
\centering
\includegraphics[width=0.45\textwidth]{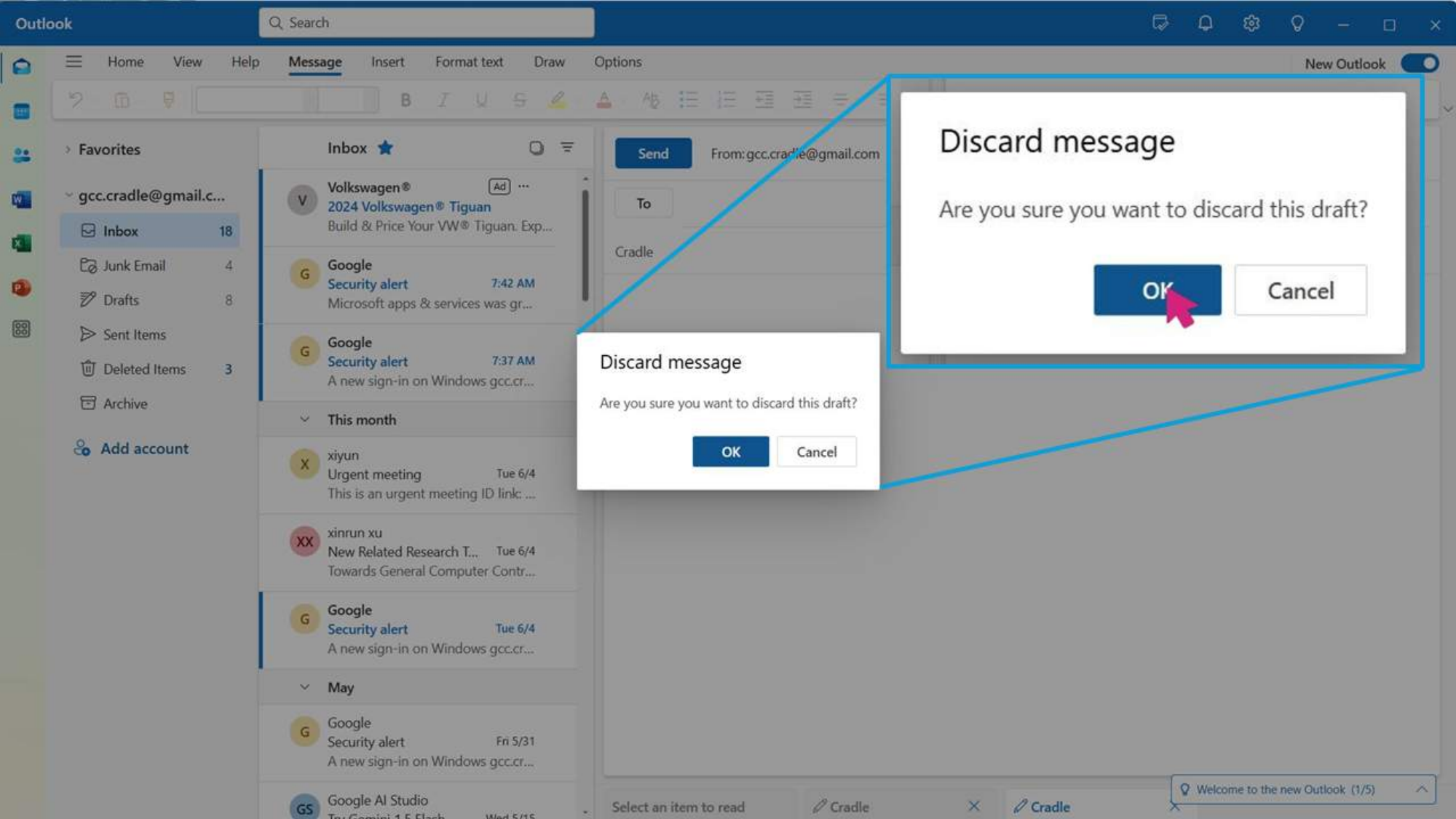}
\caption{Sample augmented image w/ drawn mouse pointer. Zoom overlay shows the image difference.}
\label{fig:outlook_with_mouse}
\end{wrapfigure}

In addition to using the SAM2SOM method for image augmentation, we also redraw the mouse pointer not present in captured screenshots in a more prominent magenta color based on its screen position, to emphasize both its presence and position for image understanding (\eg Figure~\ref{fig:outlook_with_mouse}). The augmentation process in Information Gathering can then result in four versions of a screenshot: a) base image, b) SAM2SOM image, c) base image with mouse pointer, and d) SAM2SOM image with mouse pointer.

\textbf{Self-Reflection.} As the applications in the software set are much less dynamic than complex games, there is no need to send multiple video frames to Self-Reflection. For the software applications, pre- and post-action screenshot usually suffice, \ie one image before and one image after an action is executed. Digital games often have continuous and dynamic environments that require multiple frames to properly capture the full context and thus help the backbone LMMs understand what happened. In contrast, software operations are typically more discrete and static, where the state before and after an action provides sufficient information for most analysis.

Nonetheless, we find that irrespective of images used, GPT-4o sometimes can have difficulty determining the success of certain tasks. For example, when downloading a file on Chrome, after either pressing `Ctrl + S', or using a `Save' menu, the agent must also press `Enter' or click the `Save' button to complete the task. However, GPT-4o often assumes the task is complete when the dialog opens and before this final step. Similar cases of incorrect conclusion happen when an action correctly closes a new panel or dialog. To address this category of issues, we add mandatory reasoning rules in the prompt for the Self-Reflection module to help mitigate such mistakes. If for specific applications this still remains an issue, we can use few-shot image examples to reinforce how the backend model should correctly judge success.

\clearpage

\begin{wrapfigure}{r}{0.5\textwidth}
\centering
\includegraphics[width=0.45\textwidth]{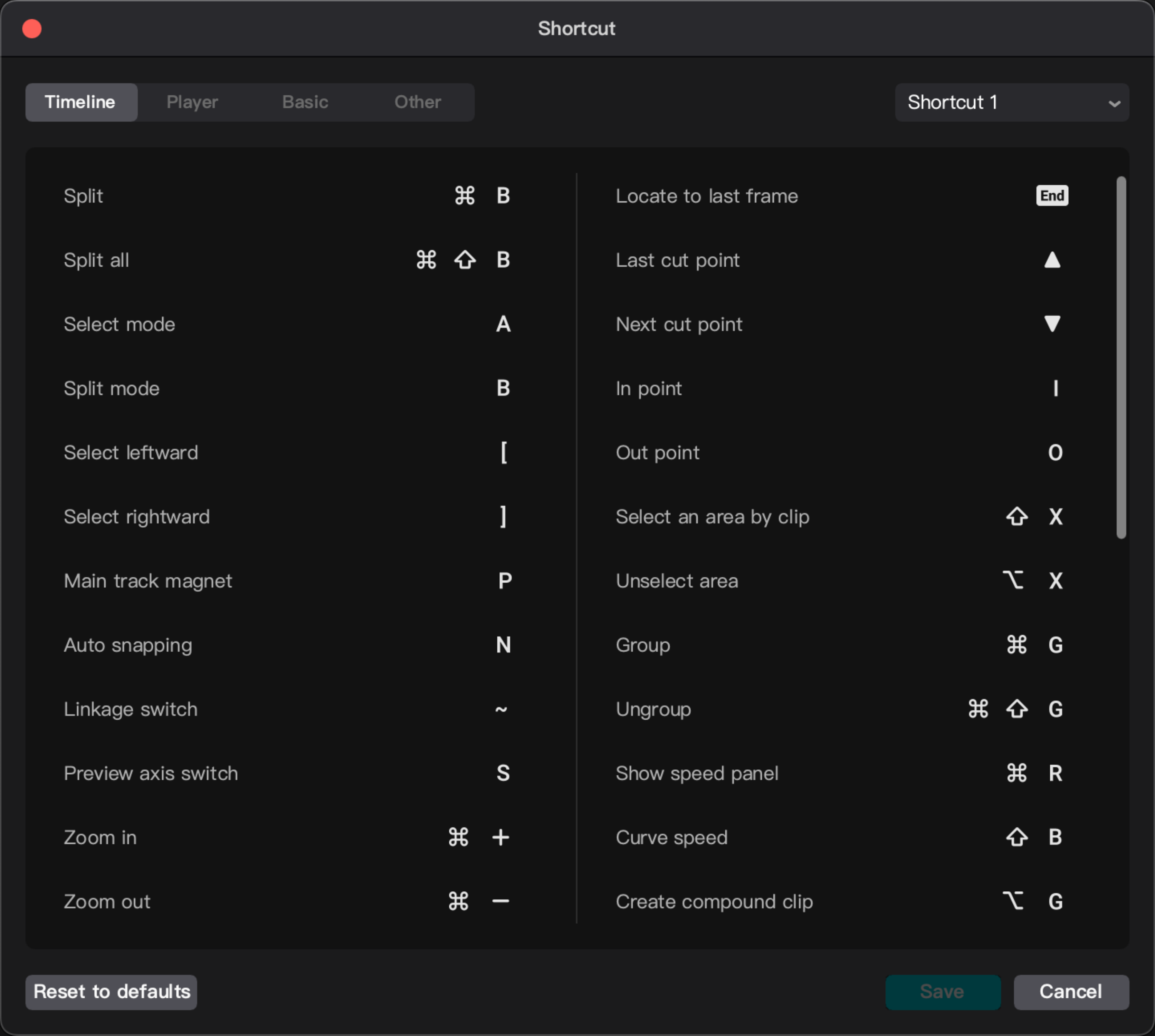}
\caption{Shortcuts screen in CapCut.}
\label{fig:capcut_shortcut}
\end{wrapfigure}

\textbf{Skill Curation.} In software tasks, direct skill generation was not necessary, as UI operations generally map closely to specific mouse or keyboard actions, making them more straightforward. In contrast, digital game environments involve continuous interactions and decision-making, raising new previously undiscovered information, and requiring the development of new skills to handle novel scenarios and adapt to changing contexts. 

However, we do add some additional pre-defined skills, on a per-application basis, for specific knowledge like less-widely known keyboard shortcuts which could be learnt from the application. For example, CapCut's shortcuts screen, shown in Figure~\ref{fig:capcut_shortcut}, or toolbar/icon processing output similarly to the process described for Cities: Skylines. Moreover, we also introduce pre-defined complex skills to demonstrate \projname's capability to leverage tools into novel functionality, such as using GPT-4o as a tool to extract information from a video to complete task 5 in CapCut.

When dealing with shortcuts, \eg as alternatives to mouse operations, it may be the case that specific shortcuts require "calibration". For example, using the keyboard to navigate the timeline in CapCut (as seen in the bottom area of Figure~\ref{subfig:CapCut_som}) requires mapping the keyboard shortcut (`Alt + arrow keys') to pixels or time, which we perform a priori and use the mapping in the pre-defined skill go\_to\_timestamp(seconds).

\textbf{Task Inference.} During the execution of an application task, we let GPT-4o decompose the execution strategy for the next step based on the overall task description and the subtask description. If the previous task decomposition is found to be unreasonable, a new decomposition plan should be proposed and this is evaluated at each iteration round.

\textbf{Action Planning.} To enable usage of SAM2SOM, for Action Planning, we insert new mouse skills, which mirror existing coordinates-based mouse skills (\ie that use x,y coordinates), but take a bounding box numerical label as an argument.

Furthermore, unlike in game playing, which focuses on performing one action per turn, when manipulating software \projname can be configured to perform two actions in sequence and thus lower interaction frequency requirements to the backend model. We find that GPT4-o can usually correctly output two-step compound actions. For example, when performing a search in the browser, it can typically output two consecutive action steps, \eg type\_text(text=`\{user\_query\}'), followed by the required press\_key(key=`enter').

\textbf{Action Execution.} While atomic and composite skills can involve complex operations, Action Execution happens over the regular \projname action space, as shown in Table \ref{tab:action_space}. For example, during Action Execution, a post-processing step converts the bounding box calls into regular mouse actions, using the centroid of a given bounding box as its coordinates for regular mouse operations.

Tool usage, like calling GPT-4o separately to analyze the contents of a media file, is not considered as an action, as tools do not operate on the environment, only as code steps inside a composite skill.

\subsection{Case Studies}
\label{appx:software:cases}

\subsubsection{Task Hardness}

It is well known that the difficulty of task completion can vary widely between humans and agents. The results in Table~\ref{tab:software_result_table} help illustrate some such cases. While many application operation issues may be attributed to UI variety or non-conformity, that is not necessarily the main source of task hardness (\ie how unexpectedly complex performing an operation is).

Here we use Outlook, a well-known e-mail client, as a case study to discuss how different factors affect \projname task completion in real-world application situations (the exact version used is listed in Table~\ref{tab:software_versions}). Taking task 1 ("Create a new e-mail to \{email\_address\} with the subject `Hello friend' and send it.") as an example, a success rate of 40\% and efficiency of 45.45\% may seem lower than expected.

Such a task could be reasonably broken down into steps like: a) Create new e-mail, b) Add recipient, c) Write title, and d) Send e-mail. And the Task Inference module performs such decomposition consistently. However, Action Planning needs to define specific actionable operations with mouse and keyboard to execute each step. 

Firstly, \projname needs to decide based on the knowledge and visual understanding capabilities available to it to either use a known keyboard shortcut (\eg `Ctrl + N') or to click at the "New mail" button. In our experiments, \projname tends to chose clicking on the button, which is then affected by the previously discussed issues that led to the integration of SAM2SOM into the framework. Issues in spatial reasoning issues or icon/image understanding may cause a few incorrect click attempts.

Adding the recipient to the e-mail requires typing an address at the appropriate location, \ie the typical "To" field. This can be accomplished in multiple ways, mainly by typing the address on the UI next to the "To" item or choosing a pre-existing contact. 

Clicking on the "To" button triggers the UI to search and select a pre-existing contact e-mail address (with no option of adding a new contact entry, which requires first accessing the "Contacts" menu, outside of "Mail"). Moreover, the UI interaction sequence to select an existing contact can be unintuitive even to experienced users, requiring a minimum of four steps, at each step offering multiple UI options that go away from contact selection. Attempting this flow usually leads \projname to exceed the maximum number of allowed step as it gets confused by the UI design.

\begin{figure}
    \centering
    
    \begin{subfigure}{0.32\textwidth}
      \centering
      \includegraphics[width=\linewidth]{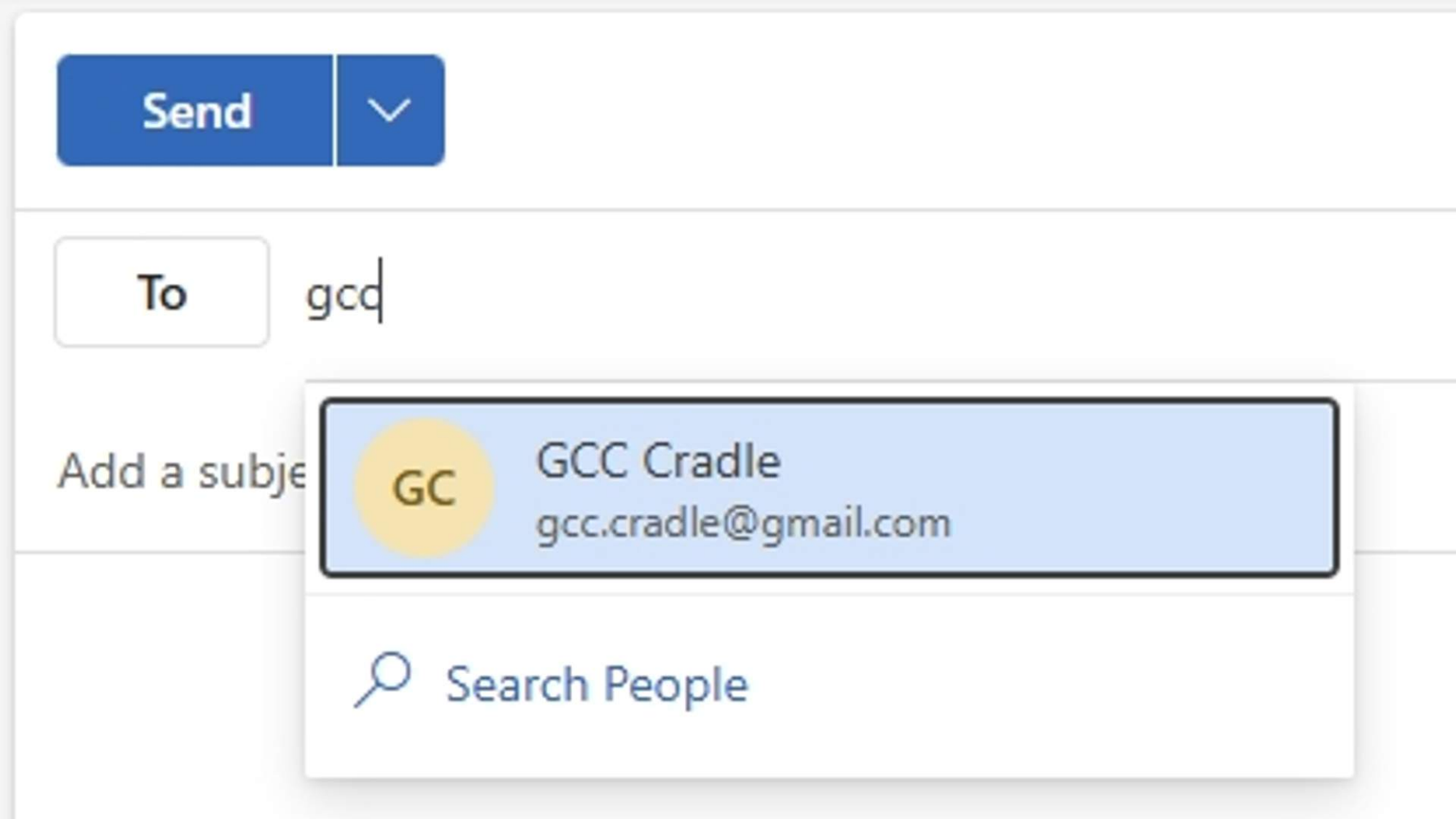}
      \caption{Pre-existing contact dropdown} 
      \label{subfig:Outlook_challenge_1}
    \end{subfigure}
    \begin{subfigure}{0.32\textwidth}
      \centering
      \includegraphics[width=\linewidth]{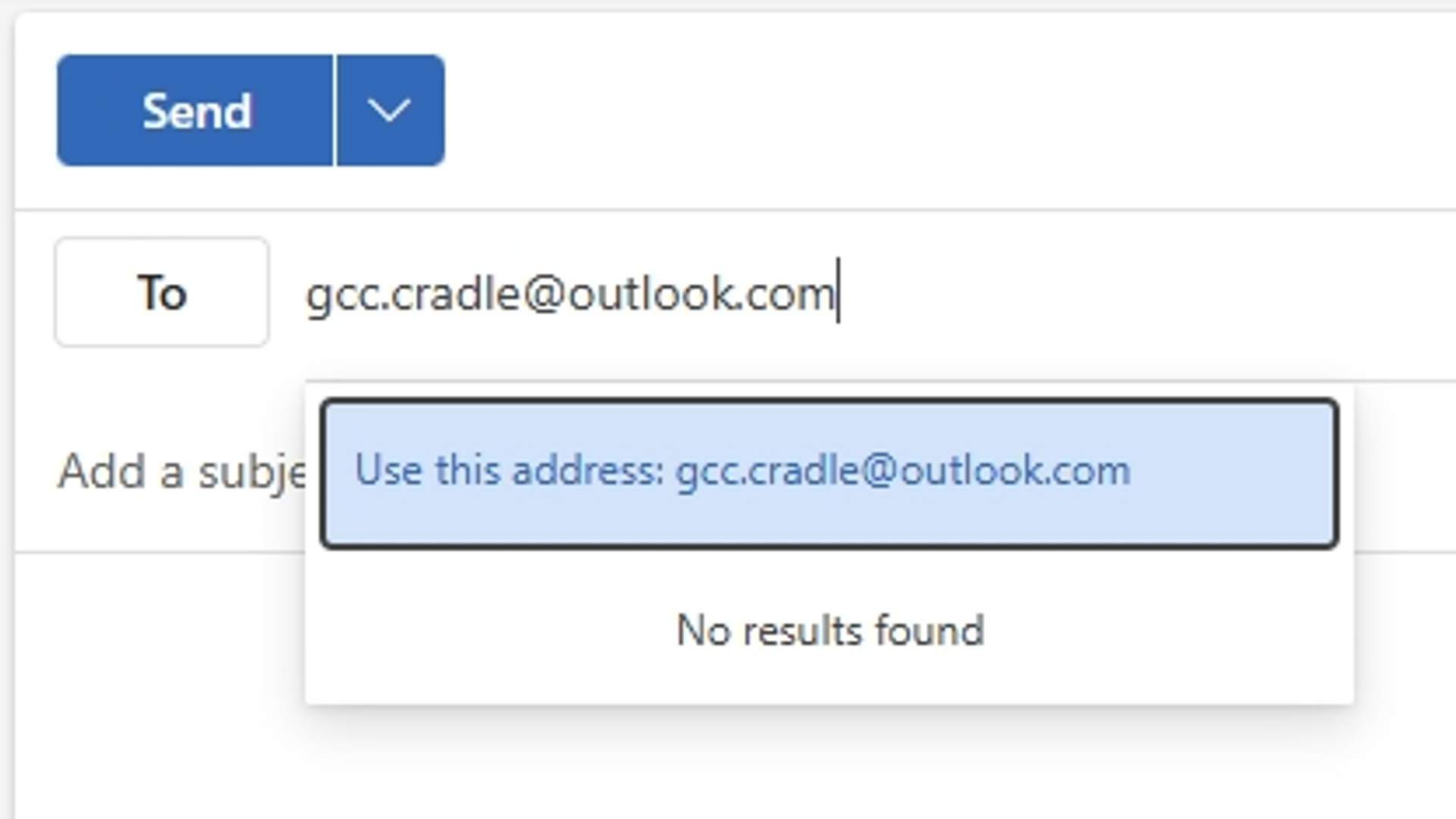}
      \caption{Contact search dropdown}
      \label{subfig:Outlook_challenge_2}
    \end{subfigure}
    \begin{subfigure}{0.32\textwidth}
      \centering
      \includegraphics[width=\linewidth]{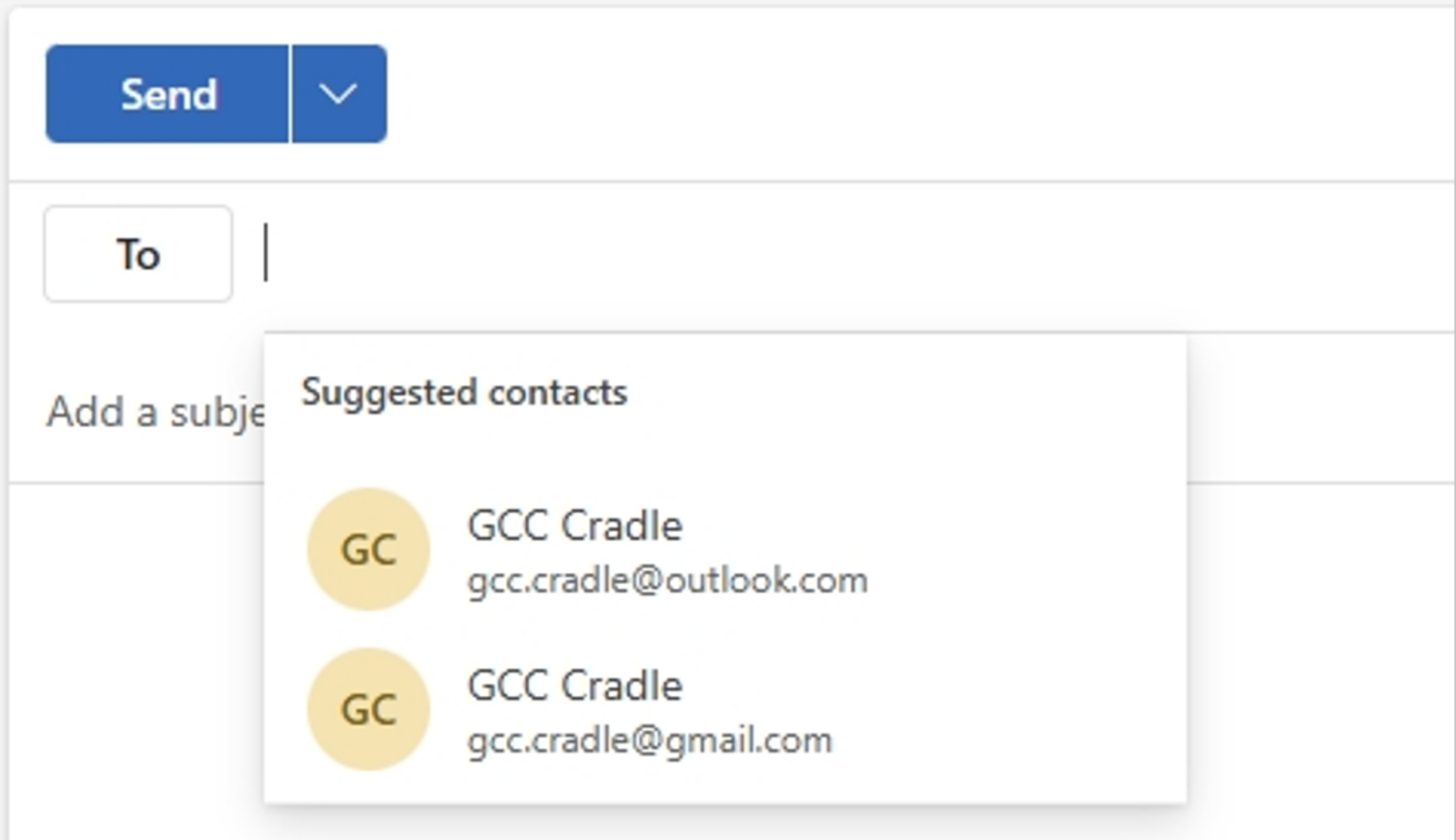}
      \caption{Contact suggestions}
      \label{subfig:Outlook_challenge_3}
    \end{subfigure}
    \caption{Visual behaviour in Outlook that may lead GPT-4o to visual understanding mistakes.}
    \label{fig:challenge_outlook}
\end{figure}

Nonetheless, choosing the simpler alternative of typing the e-mail address (assuming the correct text field is selected) triggers assistive UI pop-ups (as shown in Figure~\ref{fig:challenge_outlook}), which lead GPT-4o to falsely conclude the e-mail address is either already typed at the correct location or that it is duplicated and needs to be edited/removed. Furthermore, the pop-ups partially hide the subject area, making it harder for \projname to choose the next UI item to interact with for the next task step.

Similar issues with positioning and correctly identifying the typed subject text can also occur, but at a much smaller frequency.

Lastly, completing the task and sending the e-mail requires step similar to creating a new message. But determining send success requires additional attention/reflection as not all cases of the "Send mail" interface disappearing indicate a successful send (\eg clicking on an unrelated e-mail on the Inbox or closing the current window pop-up).

The Self-Reflection module plays a key role in moving task completion forward by detecting failed attempts
at executing each sub-task and providing rationale for failures, even if Information Gathering and Action Planning make repeated mistakes. Such feedback from Self-Reflection and allows Action Planning to tune its process and move ahead.

\subsubsection{Tool Use in CapCut}

Some general computer control tasks may require additional capabilities during execution preparation that can benefit from external tools to 
enhance agent abilities. 

When performing video editing, like in CapCut, a user may need to determine the precise frames to operate on based on video content. For such scenarios, \projname can easily leverage tool-using skills, like the LMM's ability to understand actions in a sequence of video frames, enabling it to comprehend video content and identify the exact frames for editing. 

\begin{figure}[ht]
    \centering
    \includegraphics[width=\textwidth]{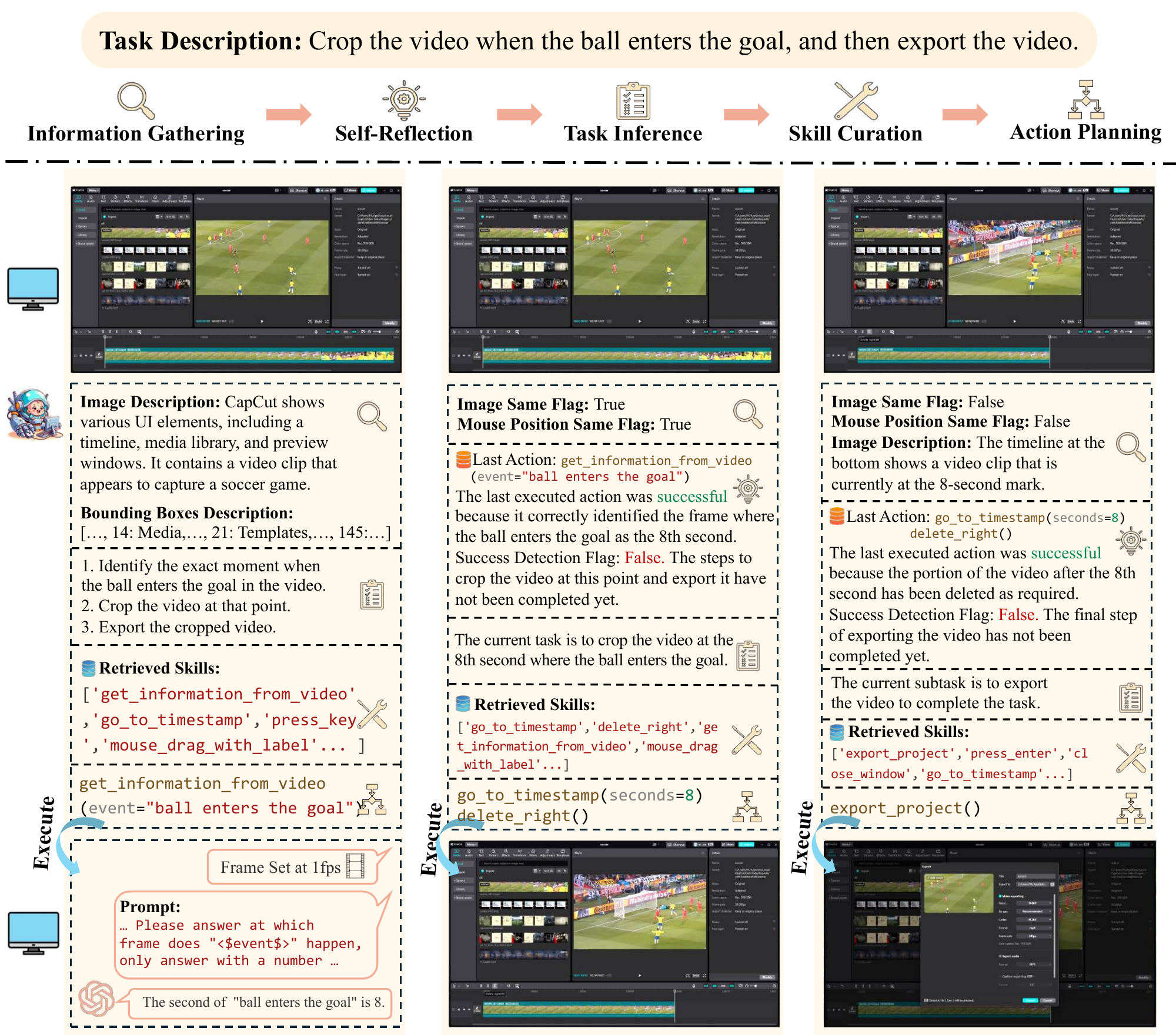}
    \caption{Showcase of Task 5 ("Crop the video when the ball enters the goal, and then export the new video") in CapCut.}
    \label{fig:Capcut_data_flow}
\end{figure}

We exemplify such tasks with task 5 ("Crop the video when the ball enters the goal, and then export the video") for CapCut, as illustrated in Figure \ref{fig:Capcut_data_flow}. This means our agent can effectively execute tool usage to find the specific frame where "the ball enters the goal". After the first round of Task Inference, \projname decomposes the task into three subtasks: 1. Identify the exact frame, 2. Crop the video, and 3. Export the video. 
Action Planning can then plan to execute `get\_information\_from\_video(event)' from our curated skills and generate "ball enters the goal" as its required argument for execution.

In this skill, we input a frame set of the video at 1 fps to identify the specific frame where the event occurs. The response is then recorded in Episodic Memory to ensure that subsequent operations can accurately utilize it and target the moment when the action occurs. Across subsequent iterations, \projname can then correctly plan and execute the remaining necessary actions for task completion: `go\_to\_timestamp(seconds=8)', `delete\_right()', and `export\_project()']. 

We have integrated few-shot learning into Self-Reflection to ensure \projname recognizes that following export\_project(), the expected screen is the CapCut application main window. This information allows it to verify the successful execution of the task, leading to success detection for the overall task.

\subsection{Limitations of GPT-4o}
\label{appx:software:limitations}

Besides the previously discussed limitations of GPT-4o, it is important to highlight a couple other GUI grounding issues.

\textbf{Non-standard UI and Noise.}

Non-standard UI, be it in visual style or in behaviour, can lead GPT-4o to misinterpret UI item functionality and application context state. The same applies to visual noise in the form of update pop-up, external contents (\eg ads), new e-mail/chat messages, etc. 

\begin{figure}
    \centering
    \includegraphics[width=\textwidth]{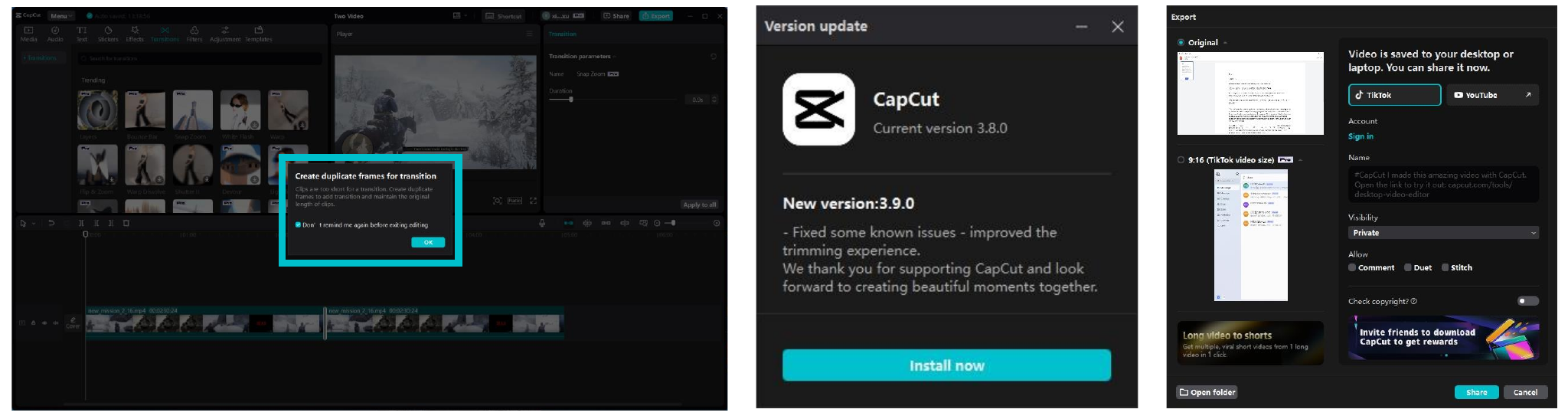}
    \caption{Different \emph{CapCut} pop-ups}
    \label{fig:pop_up_capcut}
\end{figure}

CapCut is affected by both factors, as further illustrated in Figure~\ref{fig:pop_up_capcut}. Moreover, its UI includes non-standard layouts involving precise positioning and drag/dropping. Lack of such prior knowledge by GPT4-o and differences in behaviour between similar functions, may also lead to mistakes in trying to decompose actions to perform. E.g., "Add an effect" requires very different UI-interaction depending on details.
Users can add effects in three different ways: i) dragging an effect to the timeline;
ii) click the plus sign in a given effect in the effects panel, which adds the effect to the current place on the timeline; and iii) drag an effect directly onto a video and apply the effect to the entire video.

\textbf{Visual Context Detail.}

GPT-4o still struggles with detailed visual understanding and over-relies on textual information or hallucinations, which results in insufficient  attention to visual context and leads to understanding and reasoning mistakes.

One such common example is GPT-4o declaring a dialog state to be ready to press a button like "Save", while ignoring no file name was provided, even if GPT-4o has been prompted to check for such situations. The same applies to it suggesting keyboard shortcuts to open menus that do not exist in the image being interpreted, \eg trying to press `Alt + F' to open the "File" menu on a screenshot that has no "File" menu.

Lastly, this lack of attention to context details can also affect understanding the outcome of operations over visual content, leading to incorrect estimation of operation success, \eg when retouching an image or deciding between a circle and a heart for a shape form.

%% file: appendix/osworld.tex
\subsection{Introduction to OSWorld}
\label{sec:OSWorld_intro}

OSWorld is a scalable, computer environment designed for multimodal agents. This platform provides a unified environment for assessing open-ended computer tasks involving various applications.

\subsection{OSWorld Tasks}
\label{sec:OSWorld_subtasks}

\begin{figure}[htp]
    \centering \includegraphics[width=0.6\linewidth]{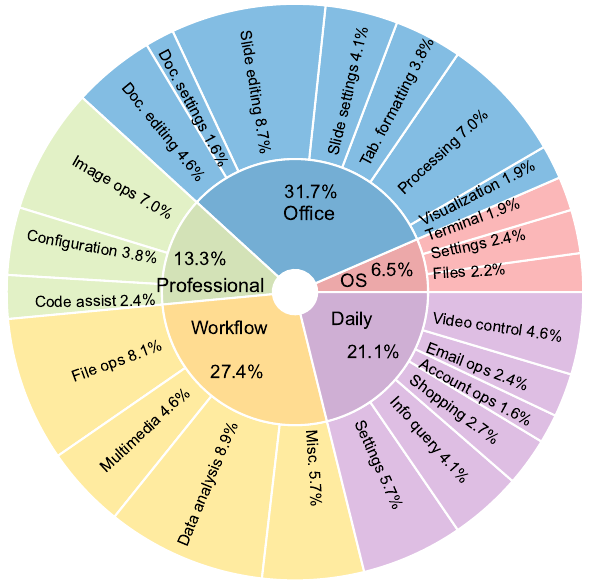}
    \caption{Task instructions distribution in OSWorld~\cite{xie2024osworld}}
    \label{fig:os_example}
\end{figure}

OSWorld is a benchmark suite of 369 real-world computer tasks (mostly on an Ubuntu Linux environment, but including a smaller set on Microsoft Windows) collected from authors and diverse sources such as forums, tutorials, guidelines. Each task is annotated with a natural language instruction and a manually crafted evaluation script for scoring.

\subsection{Implementation Details} 
\label{sec:implementation_details_OSWorld}

The OSWorld environment uses a virtual machine that takes in Python scripts based on PyAutoGUI for actions and provides screenshots and an accessibility tree for observations. We strictly follow the GCC settings. Our agent only uses the screenshot as input and outputs Python scripts using PyAutoGUI methods to control the keyboard and mouse (these operations are analogous to the regular action space for \projname). All 369 tasks use a same set of prompt templates.

We employ GPT-4o as the framework’s backbone model. We use the default experimental settings, as in OSWorld's baseline agent. The executable action space is the same as the OSWorld setting, the atomic skills are as follows:

\begin{itemize}

    \item \textbf{Mouse Actions}
    \begin{itemize}
        \item \emph{move\_mouse\_to\_position(x, y)}: Moves the mouse to a specified position on the screen.
        \item \emph{click\_at\_position(x, y)}: Performs a click at a specified position.
        \item \emph{mouse\_down(button)}: Presses the specified mouse button.
        \item \emph{mouse\_up(button)}: Releases the specified mouse button.
        \item \emph{right\_click(x, y)}: Right-clicks at the specified position.
        \item \emph{double\_click\_at\_position(x, y)}: Double-clicks at the specified position.
        \item \emph{mouse\_drag(x, y)}: Drags the cursor to the position.
        \item \emph{scroll(direction, amount)}: Scrolls the mouse wheel up or down by a specified amount.
    \end{itemize}

    \item \textbf{Keyboard Actions}
    \begin{itemize}
        \item \emph{type\_text(text)}: Types the specified text.
        \item \emph{press\_key(key)}: Presses and releases the specified key.
        \item \emph{key\_down(key)}: Holds a specified key.
        \item \emph{key\_up(key)}: Releases a specified key.
        \item \emph{press\_hotkey(keys)}: Presses a combination of keys and releases them in the opposite order (e.g., Ctrl+C), useful for shortcuts.
    \end{itemize}

    \item \textbf{Task Status}
    \begin{itemize}
        \item \emph{task\_is\_not\_feasible()}: Indicates that the task cannot be completed, providing feedback for scenarios where the agent encounters infeasible tasks.
    \end{itemize}

\end{itemize}

Many of these basic skills require GPT-4o to directly output an (x,y) position based on a screenshot. Given that the current GPT-4o is not able to achieve such precise control, we use a grounding tool to augment the screenshot. This way, GPT-4o only needs to choose an object ID. With the object ID and the bounding box of the object, we automatically convert it to the (x,y) position needed for skill execution. Instead of having GPT-4o directly choose the executable skills that require (x,y) position input, we provide several skills that only require a label ID as input for GPT-4o.

\begin{itemize}
    \item \textbf{Actions with Grounding Tools}
    \begin{itemize}
        \item \emph{click\_on\_label(label\_id)}: Clicks on a specified label in the grounding result.
        \item \emph{double\_click\_on\_label(label\_id)}: Double-clicks on a specified label in the grounding result.
        \item \emph{hover\_over\_label(label\_id)}: Moves the mouse to hover over a specified label in the grounding result.
        \item \emph{mouse\_drag\_to\_label(label\_id)}: Drags the mouse to a specified label in the grounding result.
    \end{itemize}
\end{itemize}

\textbf{Information Gathering.} Tasks in OSWorld require pixel-level mouse control. While GPT-4 exhibits grounding ability, using tools like SAM can further augment the screenshot with the grounding of icons in complex computer control tasks. The bounding box is helpful for GPT-4 to understand the occurrence of objects on the screen and can also be used to calculate the precise position for mouse control.

\begin{figure}[htp]
    \centering \includegraphics[width=\linewidth]{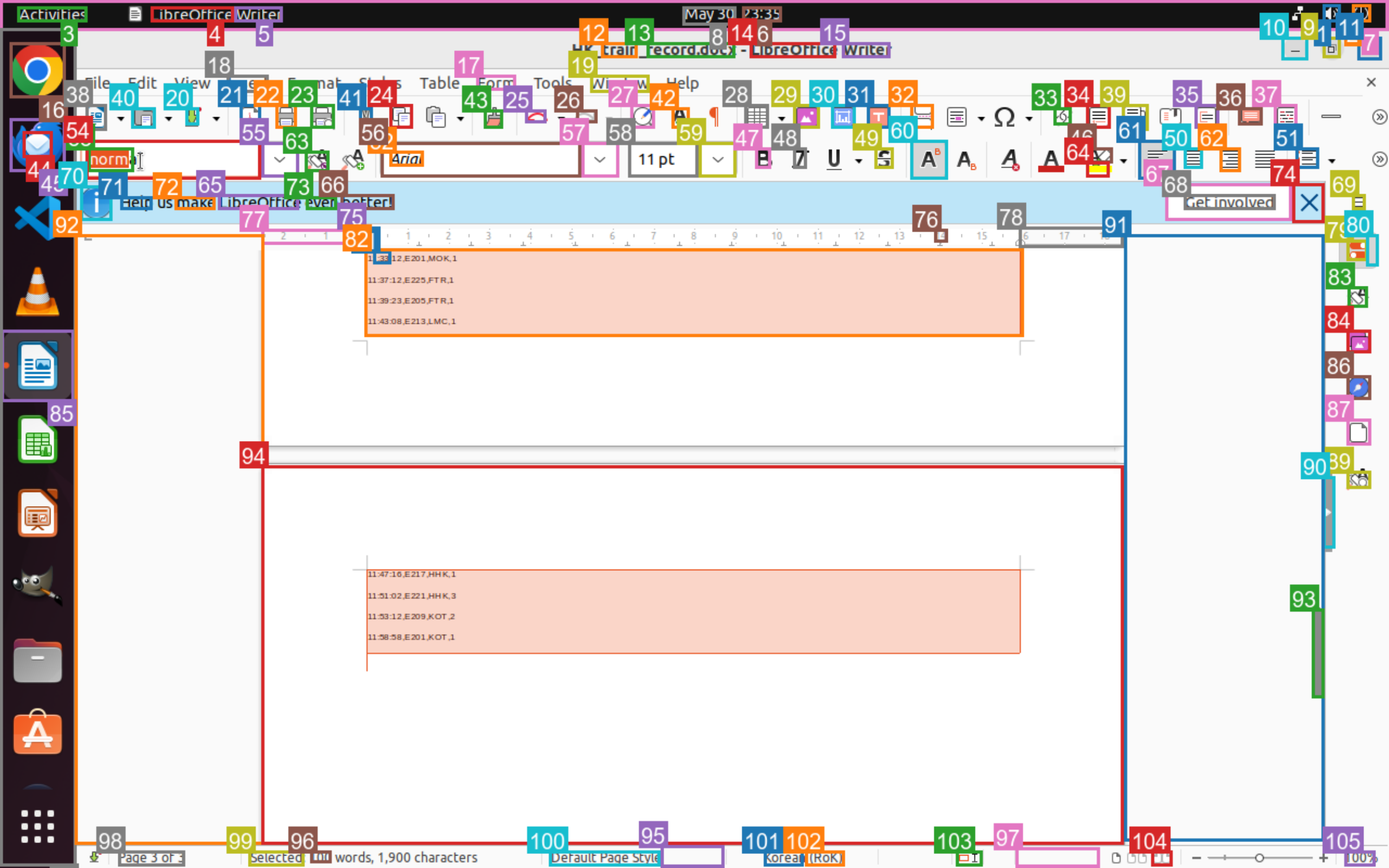}
    \caption{Augmented screenshot using \projname's SAM2SOM}
\end{figure}

\textbf{Self-Reflection.} The reflection module evaluates whether previous actions have been successfully executed and determines if the entire task was successful. The self-reflection module is important for tasks in OSWorld, which are sequential decision-making problems that require re-planning based on the current state and previous actions. The self-reflection module also helps to identify infeasible tasks.

\subsection{Application Target and Setting Challenges}
\label{sec:os_challenges}
Evaluations within OSWorld reveal notable challenges in agents' abilities, particularly in GUI understanding and operational knowledge~\cite{xie2024osworld}. To further complete tasks in OSWorld, the agent needs advanced visual capabilities and robust GUI interaction abilities. Furthermore, the agents face challenges in leveraging lengthy raw observation and action records. The next-level approach encompasses designing more effective agent architectures that augment the agents’ abilities to explore autonomously and synthesize their findings.

\subsection{Case Studies}

\subsubsection{Information Gathering}

\begin{figure}[htp]
    \centering \includegraphics[width=\linewidth]{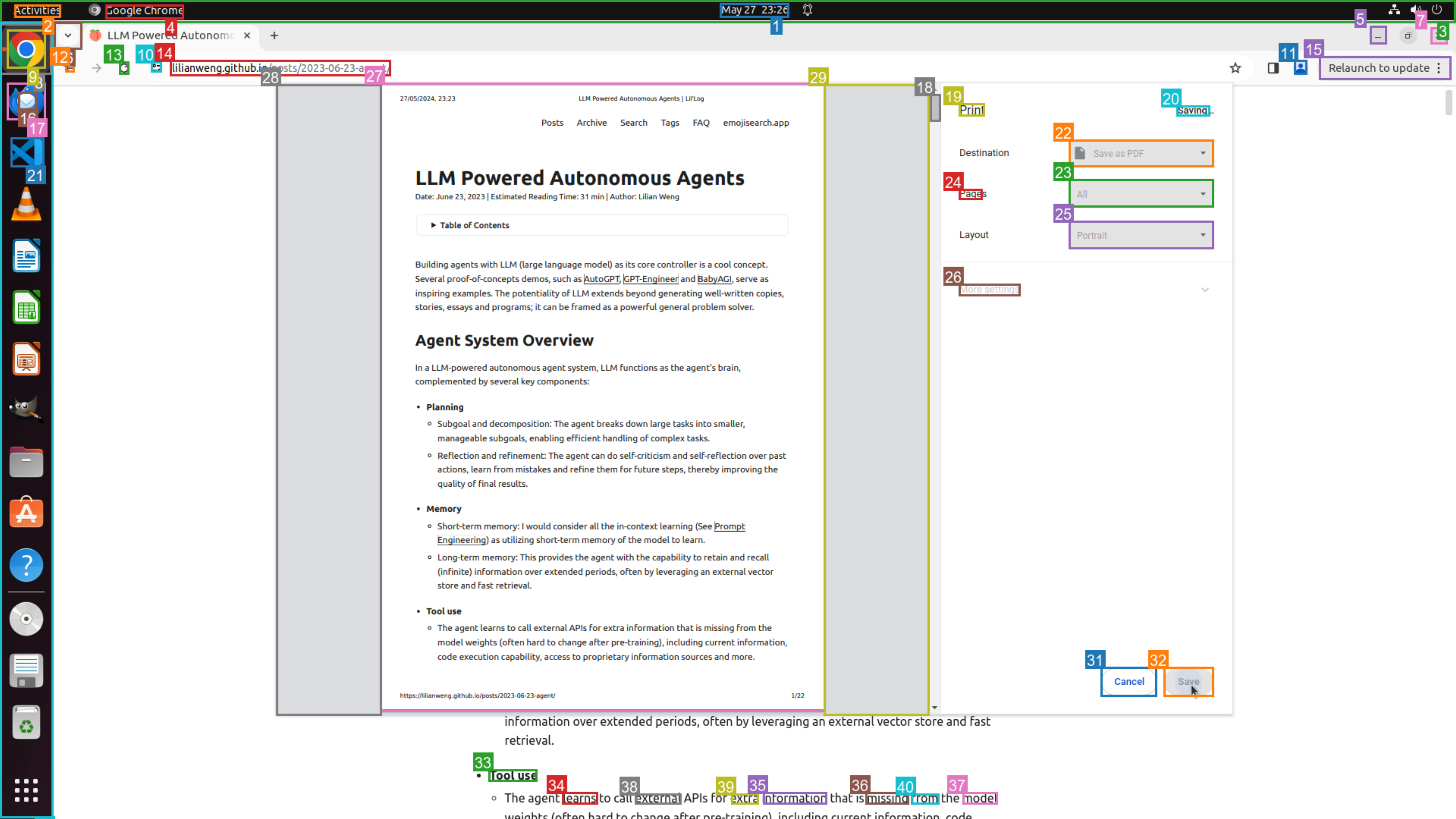}
    \caption{Case Study of robust and precise GUI interaction via information gathering}
    \label{fig:os_information_gather}
\end{figure}

With SAM as the grounding tool, we prompt the agent to identify the objects in each bounding box to determine the exact position of each object. As shown in Figure \ref{fig:os_information_gather}, the agent recognized the GUI element in box 32 as the Save button. In the planner, the agent chose to click on box 32 to save the PDF, resulting in success.

\subsubsection{Planning with Self-reflection}

We showcase how self-reflection combined with planning helps the agent complete a task by coming up with an alternative plan and validating its success.

\begin{figure}[htp]
    \centering \includegraphics[width=\linewidth]{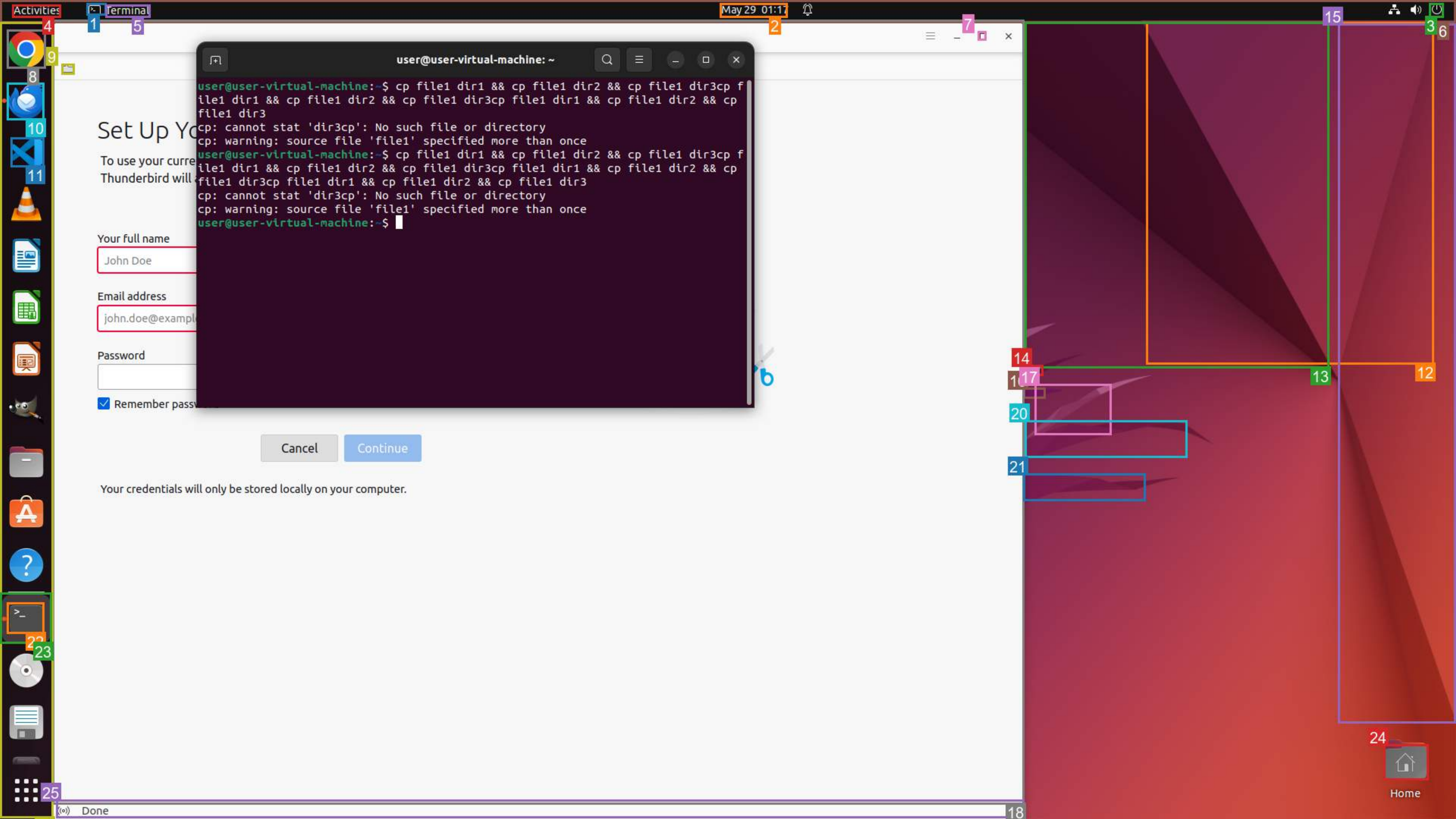}
    \caption{The agent fails to copy the files due to using incorrect commands}
    \label{fig:os_reflection_1}
\end{figure}

The current task instruction is "Copy the file `file1' to each of the directories `dir1', `dir2', `dir3'." As shown in Figure \ref{fig:os_reflection_1}, the agent made two attempts at implementing the command but encountered errors and warnings.

\begin{figure}[htp]
    \centering \includegraphics[width=\linewidth]{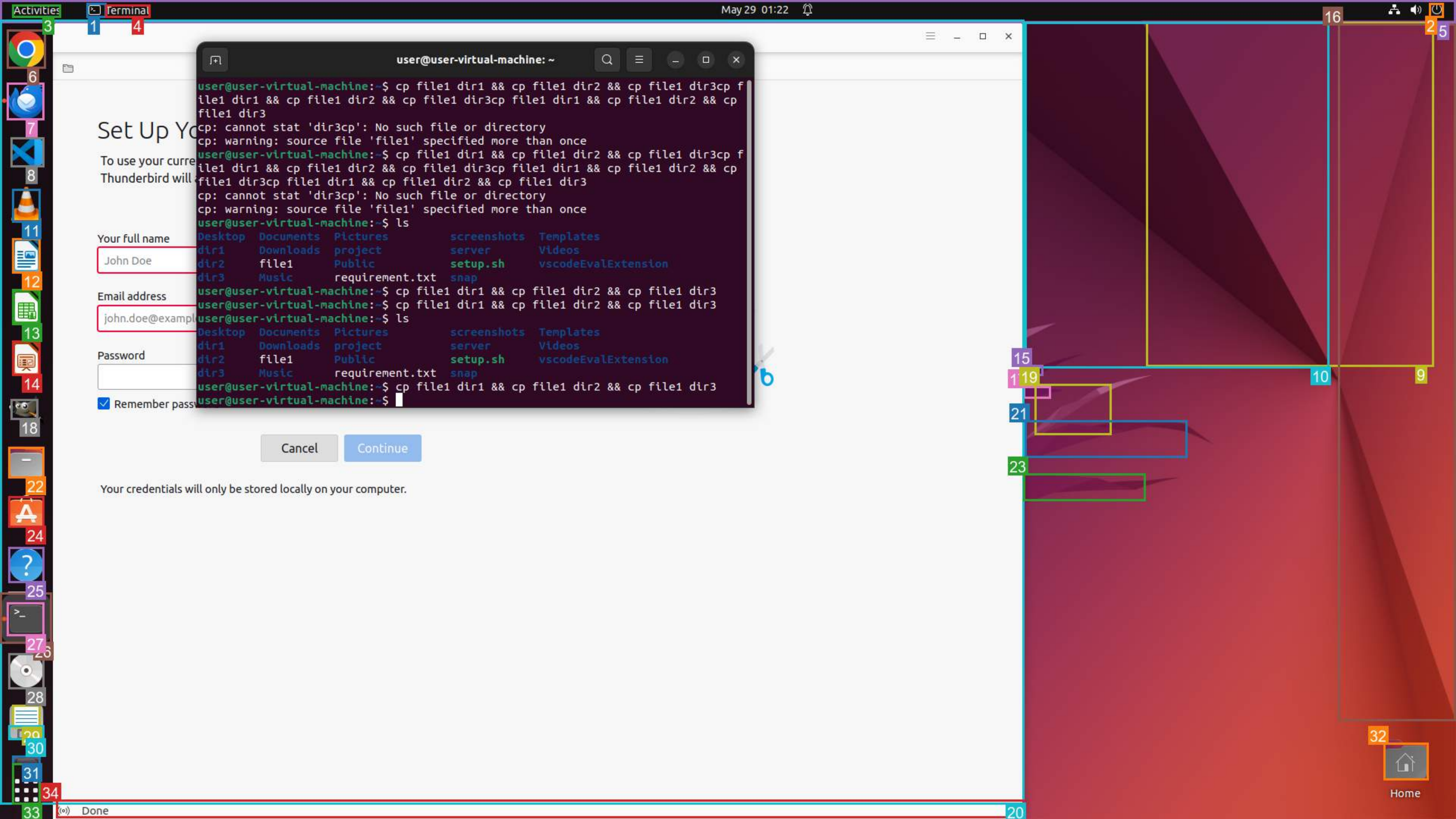}
    \caption{The agent reflects on the errors, checks the file structure and tries to debug}
    \label{fig:os_reflection_2}
\end{figure}

As shown in Figure \ref{fig:os_reflection_2}, after observing the errors and warnings in the previous steps, the agent checked the files in the directory to debug. After confirming the file structure, the agent tried different commands.

\begin{figure}[htp]
    \centering \includegraphics[width=\linewidth]{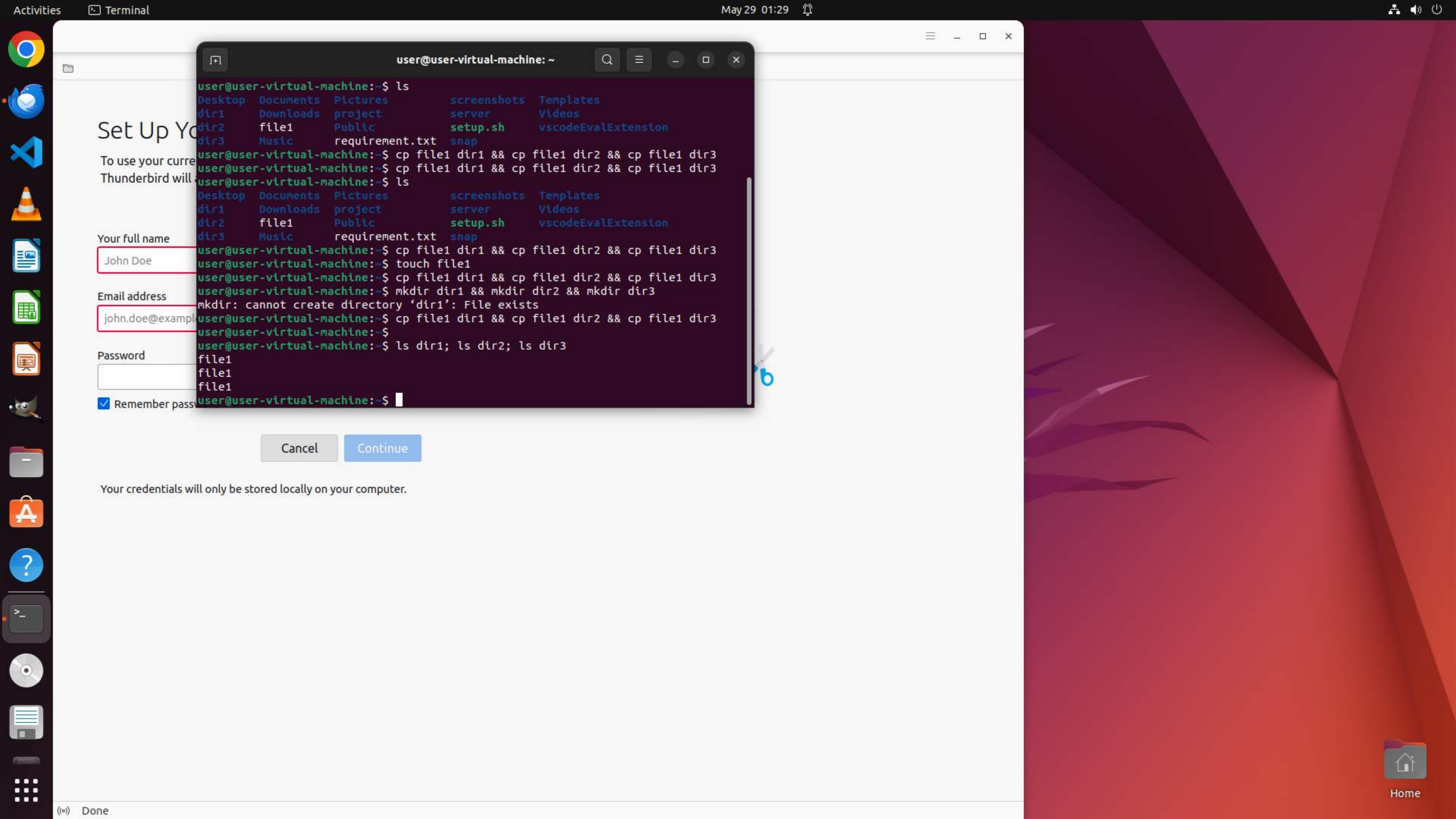}
    \caption{The agent checks if the files have already been copied}
    \label{fig:os_reflection_3}
\end{figure}

As shown in Figure \ref{fig:os_reflection_3}, after executing the new command without receiving an error message, the agent checks whether the files have been copied to the folders. After observing the result, it marks this task as a success.

\subsection{Quantitative Evaluation}

\begin{table}[]
\small
\centering
\caption{Detailed success rates divided by domains: OS, LibreOffice Calc, LibreOffice Impress, LibreOffice Writer, Chrome, VLC Player, Thunderbird, VS Code, GIMP, and Workflow (\ie involves multiple applications).}
\label{tab:exp_OSWorld_full}
    \begin{tabular}{c|ccccccccccc|c}
    \toprule
    Method 
    & \begin{tabular}[c]{@{}c@{}}OS \\(24) \end{tabular} 
    & \begin{tabular}[c]{@{}c@{}}Calc \\(47) \end{tabular} 
    & \begin{tabular}[c]{@{}c@{}}Impress \\ (47) \end{tabular} 
    & \begin{tabular}[c]{@{}c@{}}Writer \\(23) \end{tabular} 
    & \begin{tabular}[c]{@{}c@{}}VLC \\(17)\end{tabular}
    & \begin{tabular}[c]{@{}c@{}}TB \\(15) \end{tabular}
    & \begin{tabular}[c]{@{}c@{}}Chrome \\(46)\end{tabular}
    & \begin{tabular}[c]{@{}c@{}}VSC \\(23) \end{tabular}
    & \begin{tabular}[c]{@{}c@{}}GIMP \\(26)\end{tabular}
    & \begin{tabular}[c]{@{}c@{}}Workflow \\(101)\end{tabular}
    \\ 
    \midrule
    \begin{tabular}[c]{@{}c@{}} GPT-4o\end{tabular}    
   & 8.33 & 0.00 & 6.77 & 4.35 & 16.10 & 0.00 & 4.35 & 4.35 & 3.85 & 5.58 
     \\
    \begin{tabular}[c]{@{}c@{}} GPT-4o+SoM\end{tabular}    
& 20.83 &0.00 & 6.77 & 4.35 & 6.53 & 0.00 & 4.35 & 4.35 & 0.00 & 3.60  
     \\
    \begin{tabular}[c]{@{}c@{}}\projname\end{tabular}  
    & \begin{tabular}[c]{@{}c@{}} $16.67$\end{tabular} 
    & \begin{tabular}[c]{@{}c@{}} $0.00$\end{tabular} 
    & \begin{tabular}[c]{@{}c@{}} $4.65$\end{tabular} 
    & \begin{tabular}[c]{@{}c@{}} $8.70$ \end{tabular} 
    & \begin{tabular}[c]{@{}c@{}} $6.53$\end{tabular}
    & \begin{tabular}[c]{@{}c@{}} $0.00$\end{tabular}
    & \begin{tabular}[c]{@{}c@{}} $8.70$\end{tabular}
    & \begin{tabular}[c]{@{}c@{}} $0.00$\end{tabular}
    & \begin{tabular}[c]{@{}c@{}} $38.46$\end{tabular}
    & \begin{tabular}[c]{@{}c@{}} $5.48$\end{tabular}
     \\
    \bottomrule
    \end{tabular}
\end{table}

The detailed success rates for each application are listed in Table \ref{tab:exp_OSWorld_full}. We followed the same experimental settings as the OSWorld paper, running the experiment only once. Our results show that our agent performs better in the Chrome and GIMP domains. However, the difference in performance in the OS, Writer, and VSC domains is less statistically significant due to the smaller number of tasks. While improved information gathering and self-reflection empowered the agent in these domains, the complex pipeline and limitations of current grounding tools and GPT-4 hindered performance in domains like VLC and VSC. We identify these limitations as future directions for implementing the agent in real-world scenarios.